\documentclass{article}

\usepackage{arxiv}

\usepackage[utf8]{inputenc}
\usepackage[T1]{fontenc}
\usepackage{hyperref}       
\usepackage{url}            

\usepackage{booktabs}       
\usepackage{amsfonts}       
\usepackage{nicefrac}       
\usepackage{microtype}      
\usepackage{amsmath} 
\usepackage{cleveref}       
\usepackage{lipsum}         
\usepackage{graphicx}
\usepackage{doi}
\usepackage{amssymb}
\usepackage{wrapfig}
\usepackage{pifont}
\usepackage{multirow}
\usepackage{tcolorbox}
\usepackage{tikz}
\usepackage{fontawesome}
\usepackage{natbib}         
\usepackage{subfiles}
\usepackage{subfigure}
\usepackage{tabularx}
\usepackage{overpic}
\usepackage{xcolor}
\definecolor{mygray}{gray}{0.94}

\definecolor{MyGreen}{rgb}{0.02,0.7,0.02}
\definecolor{revise}{rgb}{0,0,0}
\definecolor{blue}{rgb}{0.21,0.49,0.74}
\definecolor{red}{rgb}{0.8, 0.2, 0.2}
\definecolor{green}{rgb}{0, 0.5, 0}
\definecolor{yellow}{RGB}{218, 160, 109}

\definecolor{cvprblue}{rgb}{0.21,0.49,0.74}
\usepackage{hyperref}
\usepackage{siunitx}
\hypersetup{
    colorlinks = true,
    linkcolor  = cvprblue,
    urlcolor   = magenta,
    citecolor  = teal,
}

\usepackage{cleveref}
\crefname{section}{Sec.}{Secs.}
\Crefname{section}{Section}{Sections}
\Crefname{table}{Table}{Tables}
\crefname{table}{Tab.}{Tabs.}
\usepackage{algorithm}
\usepackage[noend]{algpseudocode}
\usepackage{caption}
\usepackage{enumitem}
\usepackage{makecell}
\usepackage{subcaption}
\usepackage{soul}
\usepackage{appendix}
\usepackage[subfigure]{tocloft} 
\usepackage{circledsteps}
\usepackage{float}
\usepackage{subfloat}

\usepackage{tocloft}
\newcommand{\varitem}[3][black]{%
  \item[%
   \colorbox{#2}{\textcolor{#1}{\makebox(5.5,7){#3}}}%
  ]
}

\crefname{section}{Sec.}{Secs.}
\Crefname{section}{Section}{Sections}
\Crefname{table}{Table}{Tables}
\crefname{table}{Tab.}{Tabs.}
\crefname{figure}{Fig.}{Figs.}
\Crefname{figure}{Figure}{Figures}

\newcommand{\mpv}[1]{\textcolor{black}{#1}}

\title{Generative AI for Autonomous Driving: \\Frontiers and Opportunities}

\author{
\bf  Yuping Wang$^{1,3,19,*}$, Shuo Xing$^{1,*}$, Cui Can$^{4,*}$, Renjie Li$^{1,*}$, Hongyuan Hua$^{1,*}$, Kexin Tian$^{1,*}$, Zhaobin Mo$^{5,*}$,\\
\bf Xiangbo Gao$^{1,3,*}$, Keshu Wu$^{1}$, Sulong Zhou$^{1}$, Hengxu You$^6$, Juntong Peng$^4$, Junge Zhang$^2$, Zehao Wang$^2$,\\
\bf Rui Song$^7$, Mingxuan Yan$^2$,
Walter Zimmer$^{7}$, 
Xingcheng Zhou$^{7}$, Peiran Li$^{1,9}$, Fangzhou Lin$^{1}$, Peizheng Li$^{8}$,\\ 
\bf Zhaohan Lu$^{3}$, Chia-Ju Chen$^{10}$,
Yue Huang$^{11}$, Ryan A. Rossi$^{12}$, Lichao Sun$^{13}$, Hongkai Yu$^{14}$, Zhiwen Fan$^{1,10}$, \\
\bf Frank Hao Yang$^{14}$, Yuhao Kang$^10$,
Ross Greer$^{16}$, Chenxi Liu$^{17}$, Eun Hak Lee$^{18}$, Xuan Di$^{5}$, Xinyue Ye$^1$,\\
\bf Masayoshi Tomizuka$^{22}$, Liu Ren$^{20,21}$,
Alois Knoll$^{7}$, Xiaopeng Li$^9$,
Shuiwang Ji$^1$, Tianbao Yang$^1$, Jing Du$^6$,\\
\bf  Marco Pavone$^{23,24}$, Ming-Hsuan Yang$^{16}$,
Hua Wei$^{23}$, Ziran Wang$^4$, 
Yang Zhou$^1$, Jiachen Li$^2$,
Zhengzhong Tu$^{1,\dagger}$\\[2pt]
  $^1$Texas A\&M University, 
  $^2$Georgia Institute of Technology, 
  $^3$University of Michigan,
  $^4$Purdue University,\\
  $^5$Columbia University,
  $^6$University of Florida,
  $^7$Technische Universität München,
  $^8$University of Tuebingen,\\
  $^9$University of Wisconsin-Madison,
  $^{10}$University of Texas at Austin,
  $^{11}$University of Notre Dame,\\
  $^{11}$Adobe Research,
  $^{12}$Lehigh University,
  $^{13}$Cleveland State University,
  $^{15}$Johns Hopkins University,\\
  $^{16}$University of California, Merced,
  $^{17}$University of Utah,
  $^{18}$Texas A\&M Transportation Institute,\\
  $^{19}$Waymo LLC,
  $^{20}$Bosch Research North America,
  $^{21}$Bosch Center for Artificial Intelligence (BCAI),\\
  $^{22}$University of California, Berkeley,
  $^{23}$Stanford University, 
  $^{24}$NVIDIA, 
  $^{25}$Arizona State University,
}

\renewcommand{\shorttitle}{Generative AI for Autonomous Driving: Frontiers and Opportunities}

\hypersetup{
  pdftitle={Generative AI for Autonomous Driving: Frontiers and Opportunities},
  pdfsubject={cs.CV, cs.AI},
  pdfauthor={Yuping Wang, et al.},
  pdfkeywords={Scene Generation, Autonomous Driving, Generative Models},
}

\begin{document}
\maketitle

\vspace{-0.6cm}
\begin{abstract}
Generative Artificial Intelligence (GenAI) constitutes a transformative technological wave that reconfigures industries through its unparalleled capabilities for content creation, reasoning, planning, and multimodal understanding. This revolutionary force offers the most promising path yet toward solving one of engineering's grandest challenges: achieving reliable, fully autonomous driving, particularly the pursuit of Level 5 autonomy. This survey delivers a comprehensive and critical synthesis of the emerging role of GenAI across the autonomous driving stack. We begin by distilling the principles and trade-offs of modern generative modeling, encompassing VAEs, GANs, Diffusion Models, and Large Language Models (LLMs). We then map their frontier applications in image, LiDAR, trajectory, occupancy, and video generation as well as LLM-guided reasoning and decision making. We categorize practical applications, such as synthetic data workflows, end-to-end driving strategies, high-fidelity digital twin systems, smart transportation networks, and cross-domain transfer to embodied AI. We identify key obstacles and possibilities such as comprehensive generalization across rare cases, evaluation and safety checks, budget-limited implementation, regulatory compliance, ethical concerns, and environmental effects, while proposing research plans across theoretical assurances, trust metrics, transport integration, and socio-technical influence. By unifying these threads, the survey provides a forward-looking reference for researchers, engineers, and policymakers navigating the convergence of generative AI and advanced autonomous mobility. An actively maintained repository of cited works is available at 
\url{https://github.com/taco-group/GenAI4AD}.
\end{abstract}

\vspace{-0.4cm}

\vspace{-0.4cm}
\begin{figure}[h]
    \centering
\includegraphics[width=\textwidth]{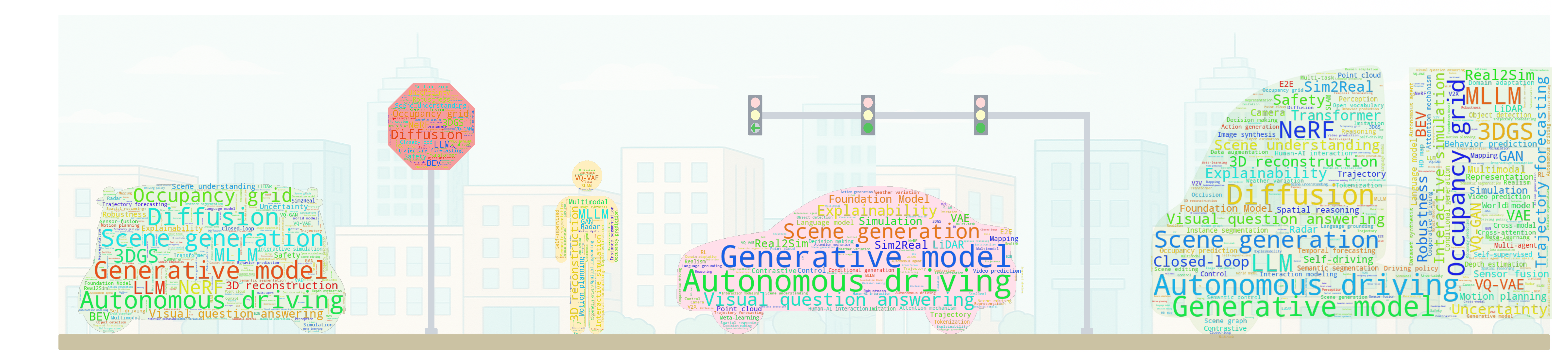}
\end{figure}

\thanks{$^*$Core Contributors. $^\dagger$Corresponding Author: Zhengzhong Tu (\texttt{tzz@tamu.edu}). $^\ddagger$Latest Update: \today.}

\clearpage
\tableofcontents
\clearpage


\section{Introduction}
\label{sec:introduction}

\begin{figure}[!ht]
    \centering
    \includegraphics[width=\linewidth]{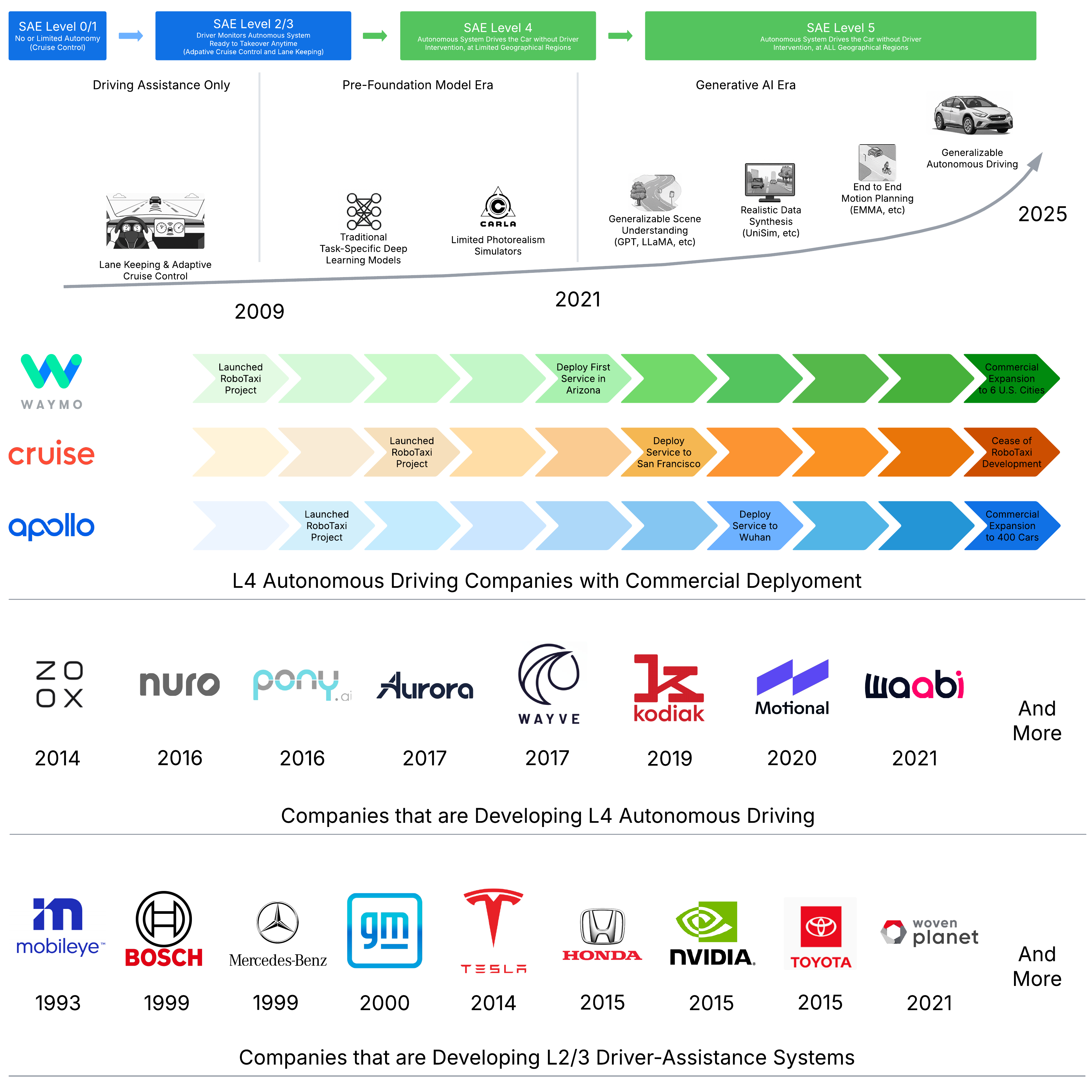}
    \caption{
    Historical overview of autonomous driving development, illustrating key technological periods (from early ADAS to the generative AI era) alongside timelines for major commercial L4 players (e.g., Waymo, Apollo), selected L4 startups, and established L2/3 ADAS companies.
    }    
    \label{fig:epoch}
\end{figure}

Autonomous driving has long been envisioned as a transformative technology that promises to revolutionize transportation by significantly impacting road safety, mobility, and logistics efficiency.
According to predictions by Goldman Sachs Research~\cite{goldsachs}, over 12\% of new car sales worldwide could reach SAE Level 3 automation (as shown in \cref{fig:epoch}) or higher by 2030, potentially launching a multibillion-dollar robotaxi market prior to achieving full autonomy.
This vision is steadily becoming an engineered reality, fueled by two decades of rapid progress in artificial intelligence (AI), computer vision, robotics, and intelligent transportation.
The development spans the entire stack, from massive data collection \cite{nuscenes2019, waymodataset}, self-supervised model training \cite{radford2021clip, he2022masked}, to large-scale validation \cite{bozga2021specification, schlichting2023savme, goyal2023automatic, li2023simulation} and efficient onboard deployment \cite{wang2024moving, Waymo_2024, waymosafe}, enabled by advances in high-performance computing devices (\textit{e.g.}, GPUs).

\begin{figure}[!ht]
    \centering
    \includegraphics[width=1.0\linewidth]{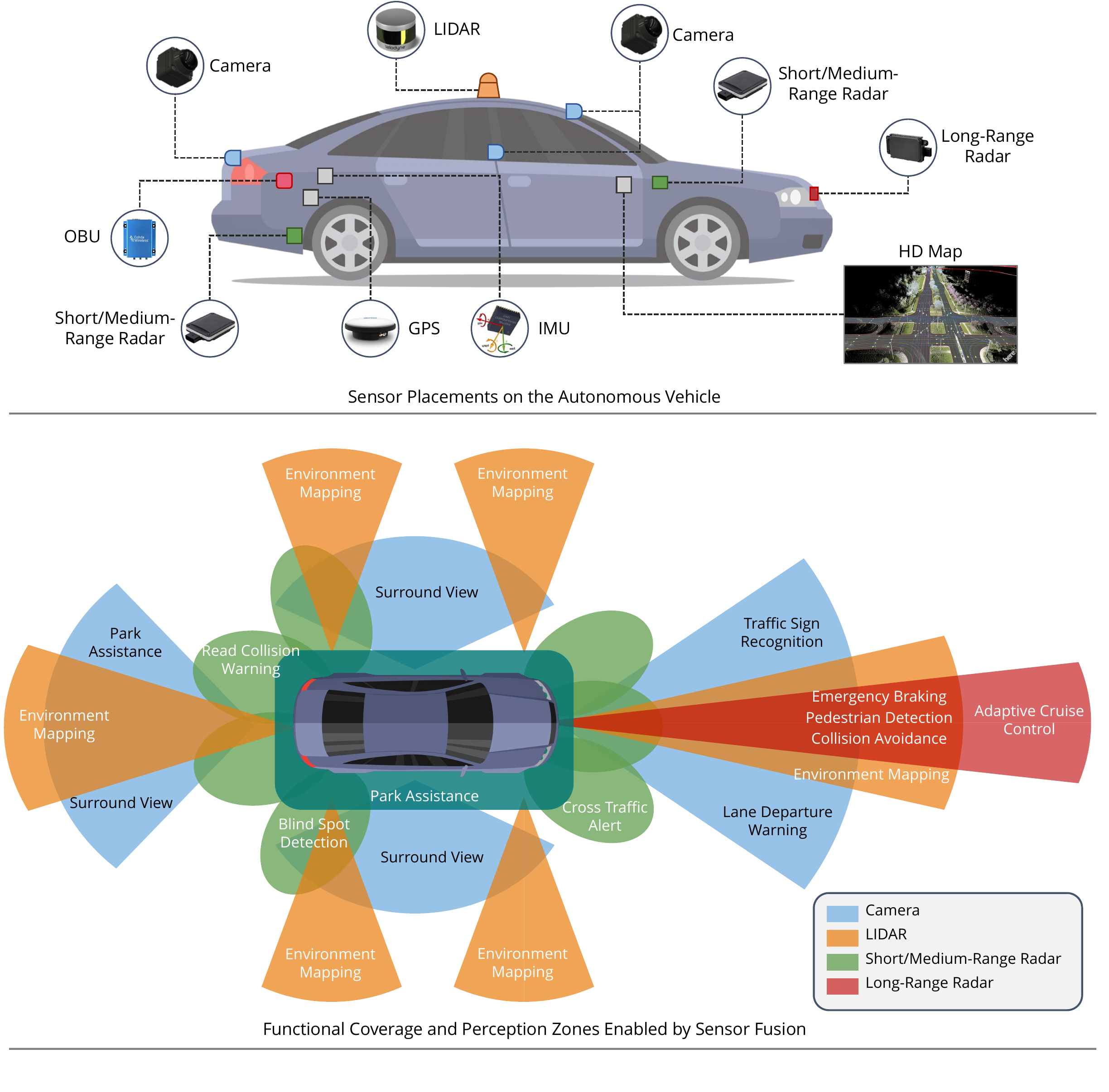}
    \caption{Overview of an autonomous vehicle's perception system, illustrating sensor fusion and coverage. \textbf{Top:} Typical placements for key sensors including cameras, LiDAR, and various radar types. \textbf{Bottom:} Functional coverage zones achieved through sensor fusion, showing areas responsible for tasks like collision avoidance, traffic sign recognition, and surround environmental mapping}
    \label{fig:sensor}
\end{figure}

Modern autonomous vehicles are usually outfitted with a suite of sensors like high-res cameras, LiDARs (both spinning and solid-state), radars, IMUs, and GNSS/GPS. These sensors offer various data about the dynamic surroundings (\cref{fig:sensor}).
Automotive-grade domain controllers instantly process and combine data using multi-core CPUs, efficient GPUs, high-bandwidth memory, and strong power-management circuits \cite{waymo2024gen6, waymo2020gen5, cnet2020baidulite, lidarnews2022cruise}.
These hardware components collectively enable key functions like real-time environmental perception, trajectory prediction, and motion planning, supporting different levels of driving automation.
Capabilities now extend from driver assistance systems (SAE Levels 2 and 3~\cite{sae2021sae}), needing human supervision for functions such as highway driving, to autonomous operation in defined environments (SAE Level 4~\cite{sae2021sae}). The ambitious target is Level 5 autonomy, which implies unrestricted operation.
The strong potential for safer roads, improved accessibility for various groups, and enhanced transportation efficiency drives extensive research and development in this area~\cite{usdot-ads}.

Academic research has laid a robust foundation by demonstrating the feasibility of autonomous driving and addressing critical challenges.
 A pivotal early moment was the 2005 DARPA Grand Challenge, won by Stanford University's vehicle, Stanley, demonstrating the potential for navigating complex environments autonomously~\cite{buehler20072005}.
Academic research has played a critical role in advancing key areas of autonomous driving, with techniques such as Simultaneous Localization and Mapping (SLAM)~\cite{cadena2016past} becoming fundamental to vehicle navigation and cartography in previously unexplored environments.
However, truly robust operation demanded breakthroughs in perception and decision-making, areas where traditional methods faced limitations.
 This set the stage for a paradigm shift driven by the rise of deep learning.
Advanced neural architectures such as ResNet~\cite{he2016deep} and Transformers~\cite{vaswani2017attention,dosovitskiy2020image} have emerged as effective mechanisms for extracting insights from extensive sensor data. These architectures have significantly enhanced machine perception, driving key developments in object detection~\cite{ren2015faster,carion2020end}, semantic segmentation~\cite{he2017mask,xie2021segformer}, and tracking~\cite{yilmaz2006object,wu2013online}, all crucial for intricate scene interpretation~\cite{zhao2017pyramid,cordts2016cityscapes}.
Building on this success, research expanded to apply machine learning techniques to higher-level behavior prediction~\cite{mozaffari2020deep}, motion planning~\cite{gonzalez2015review}, and even explore concepts of end-to-end driving systems that map sensor inputs directly to control outputs~\cite{xiao2020multimodal,hwang2024emma}.

However, as Clive Humby argued ~\cite{arthur2013tech}, ``\textbf{Data is the new oil}.''
This rapid progress of deep learning transformation was critically fostered by the availability of foundational vision datasets like ImageNet~\cite{deng2009imagenet}, MS COCO~\cite{lin2014microsoft}, YouTube-8M~\cite{abu2016youtube} built by industry leaders (\textit{e.g.}, Google, Microsoft, Meta), and equally importantly, bespoke autonomous driving datasets proving rich multimodal sensor data and annotations, such as KITTI~\cite{kitti}, nuScenes~\cite{nuscenes2019}, Waymo Open Dataset~\cite{waymodataset}, Argoverse~\cite{argoverse2019}, and BDD100K~\cite{yu2020bdd100k}.
Simulation environments such as CARLA~\cite{dosovitskiy2017carla}, AirSim~\cite{shah2018airsim}, SUMO~\cite{SUMO2018}, and Isaac Sim~\cite{liang2018gpu} are crucial for modern research. These resources deliver vital ground truth data for training and various platforms for validation. There is a reciprocal benefit: real-world data enhances simulator accuracy, and simulators produce synthetic data to supplement real datasets, especially for rare or dangerous situations.
Despite these powerful tools and algorithmic breakthroughs, most academic systems remain confined to research prototypes or controlled testing environments~\cite{xu2020mcity,dong2019mcity}, highlighting the substantial challenges in transitioning these technologies to robust, large-scale, real-world deployment and productization.

Bridging the gap from academic prototypes to productions, industry development has surged forward with ambitious commercialization efforts and large-scale deployments. 
Companies such as \textbf{Waymo} (with roots in Stanford's DARPA team mentioned above) and \textbf{Baidu Apollo Go} have emerged as leaders in the pursuit of Level 4 autonomy, operating driverless robotaxi services in constrained urban environments
Waymo, for instance, launched the first fully driverless service in Phoenix in 2020 and now operates across several major US cities, including San Francisco, Los Angeles, and Austin, while Baidu operates extensively in China, achieving fully driverless operations in over ten cities and accumulating over 10 million rides~\cite{baidu-apollo}.
Other players like \textbf{Zoox}, backed by Amazon, are pursuing a unique strategy with a purpose-built vehicle, actively testing in several US cities, planning public launches in Las Vegas and San Francisco later in 2025~\cite{zoox}.
However, translating L4 technology into widespread, profitable services faces immense technical, safety, and financial hurdles.
\textbf{Cruise}, once a major competitor backed by General Motors, faced significant setbacks following a safety incident in late 2023, leading GM to cease funding for its robotaxi operations in December 2024 and pivot towards developing advanced driver-assistance systems (ADAS) for personal vehicles~\cite{cruise}.
This strategic shift underscores the immense technical, safety, and financial challenges associated with scaling L4 robotaxis.
The market for production vehicles is predominantly controlled by SAE Level 2 and Level 3 Advanced Driver-Assistance Systems (ADAS), with leaders such as \textbf{Tesla}, known for its Autopilot/FSD Beta necessitating driver oversight~\cite{tesla}, and key suppliers like \textbf{Mobileye}~\cite{mobileye} delivering ADAS to various car manufacturers. This disparity underscores the persistent challenges in developing fully autonomous systems that operate beyond constrained environments. \mpv{\textbf{NVIDIA} has become a pivotal enabler in the autonomous driving ecosystem by providing scalable hardware-software platforms that power both development and deployment across the industry. Its DRIVE platform~\cite{nvidia-drive}, adopted by automakers such as Mercedes-Benz and Volvo, delivers end-to-end capabilities from perception to planning using AI-accelerated compute. In 2022, NVIDIA expanded its influence by launching the DRIVE Thor superchip, designed to unify ADAS and autonomous driving functions in next-generation production vehicles~\cite{nvidia-thor}.}

Despite significant advancements and investments, the autonomous driving industry confronts fundamental roadblocks impeding the transition towards truly autonomous Level 5 capability.
These span not only \textbf{technological hurdles} related to perception, prediction, and decision-making algorithms operating reliably in unpredictable environments, but also critical challenges in navigating an evolving \textbf{regulatory} and \textbf{legal} landscape, such as establishing clear liability frameworks for AV-involved accidents~\cite{taeihagh2019governing}.
\mpv{Achieving broad \textbf{public trust and acceptance} remains fragile for autonomous driving}, sometimes manifesting as ``AI anxiety'' evident in public reactions, triggering the recent rise of Waymo vandalism~\cite{waymo_rl_survey}---the acts of deliberate damage or destruction targeting Waymo's self-driving vehicles, ranging from throwing objects at the cars to slashing tires, tagging with graffiti, and even setting them on fire.
While acknowledging the importance of all these dimensions, this survey concentrates primarily on the core technical challenges, particularly those related to generalization, reliability, and system complexity. Key among these technical issues are:
\begin{enumerate}[noitemsep]
\varitem{red!40}{\textbf{1)}} \textbf{The Devil is in the ``Long Tails'': Robustness and Generalization}. Systems struggle to generalize reliably beyond training data, especially for rare but critical ``long tail'' events~\cite{Wayve}, encompassing diverse weather, lighting, and sensor noise, where real-world data collection is insufficient for complete coverage.
\varitem{blue!50}{\textbf{2)}} \textbf{Confidentially Confused: Reliability and Uncertainty.} Guaranteeing dependable real-time performance across millions of miles and diverse conditions, while effectively managing inherent AI model and environmental uncertainty, remains crucial but challenging.
\varitem{green!50}{\textbf{3)}} 
\textbf{An Arm and a LiDAR? Complexity and Scalability.}
The reliable scaling of these intricate systems is hindered by substantial computational requirements and economic expenses, especially due to costly sensor suites such as LiDAR, which impedes democratization and broad adoption.
\end{enumerate}
Despite notable successes, current autonomous driving paradigms face persistent technical barriers, suggesting they are approaching their inherent limits in achieving real-world generalization and robustness—necessitating a fundamental shift toward more powerful and adaptable AI architectures.

\begin{center}
\textbf{\textsf{``Quo Vadis, Autonomous Driving?''}}
\end{center}

The emergence of \textbf{DALL-E}~\cite{ramesh2021zero} from OpenAI in 2021 marked a pivotal turning point, triggering an unprecedented boom in \textbf{Generative AI (GenAI)} technologies. 
Quickly followed by platforms such as Midjourney~\cite{midjourney} and Stable Diffusion~\cite{rombach2021highresolution}, these innovations democratized, albeit to varying degrees, the accessibility of sophisticated AI-generated art ~\cite{ye2025generating}. Consequently, they set the stage for transformative impacts across diverse industries, including art, design, marketing, media, and entertainment~\cite{cao2023comprehensive, ye2025geodesign}.
Parallel to these advances in visual generative technologies, an even more profound revolution emerged within the realm of large language models (LLMs). Models such as ChatGPT~\cite{chatgpt} and GPT-4~\cite{openai2023gpt4} from OpenAI demonstrated unprecedented emergent capabilities in natural language processing, reasoning~\cite{wei2022cot}, and contextual understanding.
The landscape was further diversified by Meta's release of the open-source LLaMA model series \cite{touvron2023llama,llama3.2,roziere2023codellama}, fostering broader open research and development.
Furthermore, the integration of multimodal functionalities, particularly vision, into these powerful language architectures has opened new avenues for grounded visual understanding, vision-language reasoning, and more intuitive human-AI collaboration.
For the purposes of this survey, we define Generative AI models as a class of machine learning systems distinguished by their ability to learn underlying data distributions and subsequently synthesize novel data artifacts, such as images, videos, text, audio, code, or complex 3D environments. A critical characteristic is that these synthesized outputs exhibit statistical properties highly similar to the real-world data upon which the models were trained, enabling significant progress in applications demanding realistic, diverse, and scalable data representations.
%

This paragraph transitions from predictive models to advanced generative AI, revealing opportunities to overcome the constraints impeding Level 5 Autonomy.
For example, GenAI models directly confront the ``long tail'' challenge through high-fidelity synthesis of divers sensor data (\textit{e.g.}, LiDARs~\cite{wu2024text2lidar}, cameras~\cite{swerdlow2024street}, or trajectories~\cite{ivanovic2019trajectron}), and generation of complex driving scenarios~\cite{rowe2025scenario}, enabling the creation of rich datasets and simulation environments populated with rare but critical evens essential for robust generalization.
Moreover, it enhances system reliability by facilitating sophisticated modeling of multi-agent interactions and long-horizon prediction. bolstering situational awareness and planning under uncertainty.
Perhaps most transformative, multimodal foundation models like  LLaVA~\cite{liu2024visual} and DriveVLM~\cite{tian2024drivevlm} unify perception, prediction, and planning within a single language-centric architecture, carrying world knowledge in their pre-training, offer a path beyond brittle modular pipelines towards more scalable and adaptable systems.
Consequently, GenAI represents not merely an incremental tool but a potential paradigm shift for autonomous driving: a move towards unified, data-driven systems capable of deeper understanding, enhanced adaptability, and more effective generalization—key ingredients required to accelerate progress towards the ultimate goal of safe and reliable Level 5 autonomy.

Recognizing this pivotal moment and transformative potential, this survey undertakes a comprehensive review and synthesis, charting the course of how these GenAI technologies are actively reshaping the field of autonomous driving.
We aim to equip a diverse audience: engineers, researchers, practitioners, industry stakeholders, and crucially, policymakers, with the synergistic knowledge and critical perspective required to navigate the complex intersection of GenAI and autonomous driving. By fostering a deeper understanding of both the immense potential and the inherent responsibilities, we hope to accelerate the thoughtful development and conscientious deployment of intelligent, reliable, and humanity-centered autonomous systems on our shared planet~\cite{sanchez2025ethical}.

The outline of this survey is structured as follows:
\begin{itemize}[nolistsep]
    \item In Section \ref{sec:related_reviews}, we compare the scope of our survey with that of other related works on autonomous driving. Interested readers are encouraged to consult these surveys for complementary perspectives.
    \item In Section \ref{sec:popular_datasets}, we summarize the popular datasets used for autonomous driving research, classified by their target application domain. We compare their differences and provide a link to where to download them.
    \item In Section \ref{sec:foundations}, we systematically categorize the diverse landscape of generative models by their foundational architectures (\textit{e.g.}, VAEs, GANs, diffusion models, and autoregressive models).
    \item In Section \ref{sec:generation_methods}, we delve into the frontier GenAI models tailored for autonomous driving by application modality (\textit{e.g.}, image, video, LiDAR, trajectory) and core function (\textit{e.g.}, simulation, prediction, planning).
    \item In Section \ref{sec:applications}, we explore in detail the key applications of generative AI in autonomous driving, covering sensor generation, world modeling, multi-agent forecasting, scene understanding, and decision-making.
    \item In Section \ref{sec:embodied_ai}, we go beyond the scope of autonomous driving and discuss the research in the broader area of embodied AI.
    \item Lastly, in Section \ref{sec:future}, we move beyond model capabilities to critically examine current limitations and future challenges. This includes not only technical hurdles like data scarcity, theoretical gaps, evaluation methodologies, safety analysis, and simulation fidelity, but also extends to broader implications encompassing transportation planning, economic impacts, public health considerations, policy development, and vital ethical issues ~\cite{ye2025human,ye2023toward}. Here, we identify promising research directions aimed at building trustworthy, scalable, and ultimately beneficial generative systems for safe and equitable transportation for all.
\end{itemize}

\section{Related Surveys}
\label{sec:related_reviews}

In this section, we compare our survey with several recent works that focus on related aspects of autonomous driving and generative models. While these surveys provide valuable insights into specific subdomains, our work offers a broader, more integrative perspective on generative models for autonomous driving.

\textbf{A Survey on Data-Driven Scenario Generation for Automated Vehicle Testing}~\cite{cai2022survey} focuses on data-driven scenario generation for automated vehicle testing. It reviews various methodologies, such as reinforcement learning and accelerated evaluation, for creating critical driving scenarios. Unlike our work, which explores the role of generative models in multiple domains such as scene understanding and intelligent transportation, this survey is primarily concerned with generating test cases for evaluating autonomous vehicles.

\textbf{A Survey on Safety-Critical Driving Scenario Generation}~\cite{ding2023survey} categorizes scenario generation approaches into data-driven, adversarial, and knowledge-based methods. The survey highlights challenges in scenario fidelity, efficiency, and transferability. While this work provides a deep dive into safety-critical scenario generation, it focuses on conventional simulation-based scenario generation rather than those based on generative models.

\textbf{A Survey of World Models for Autonomous Driving}~\cite{feng2025survey} presents a structured review of world models that integrate perception, prediction, and planning. It proposes a taxonomy covering future physical world generation, behavior planning, and agent interactions. While this survey provides a brief introduction to some noteworthy world models and their application in autonomous driving, our survey covers related generative models more holistically and provides method comparison for each generation domain.

\textbf{Exploring the Interplay Between Video Generation and World Models}~\cite{fu2024exploring} investigates the synergy between video generation and world models in autonomous driving. The survey explores diffusion-based video generation methods, while our work also explores many other modalities.

\textbf{From Generation to Judgment: Opportunities and Challenges of LLM-as-a-Judge}~\cite{li2024generation} explores the use of large language models for AI evaluation tasks such as ranking, scoring, and selection. While this work is relevant to AI judgment and assessment, it does not address the role of generative models in synthetic data generation, simulation, or autonomous driving.

\textbf{Generative AI in Transportation Planning}~\cite{da2025generative} explores the integration of generative AI in transportation planning, such as optimal ways to move people and goods, and their focus is the safety and efficiency of these methods. Autonomous driving is an optional component in transportation planning. Our focus, on the other hand, is exclusively on autonomous driving. We will, however, discuss how it could benefit transportation planning.

\textbf{Vision-Language-Action Models: Concepts, Progress, Applications and Challenges}~\cite{sapkota2025vision} provides a broad survey of Vision-Language-Action (VLA) models across diverse embodied AI domains, including robotics and healthcare. Our survey on generative AI includes VLA models and many others, so that we can compare their differences and analyzer their unique challenges.

\paragraph{Additional Surveys on Autonomous Driving and Generative Models}
Beyond the surveys discussed above related to our, several other surveys delve into specific aspects of autonomous driving \cite{wei2021autonomous,chib2023recent, mao20233d, 10522953, Cui_2024_WACV, 10248946, gao2024survey, 10531702, 10604830, zhao2024autonomous, yang2023llm4drive, chen2024end} and generative modeling \cite{liu2024generative, 10230895, 10.1145/3626235, 10419041, 10.1145/3696415, de2023review, li2023comprehensive, 10521791, cao2023comprehensive, liu2024toward, 10081412, liu2024machine, 10419041, 10521640, liu2024generative}. These works provide in-depth insights into their respective fields and serve as valuable research resources.



\section{Datasets and Benchmarks for Autonomous Driving}
\label{sec:popular_datasets}

Generative modeling in autonomous driving requires diverse and large-scale datasets to train and evaluate models for perception, behavior prediction, and planning. Below we identify 30+ key datasets (both real-world and synthetic) used for data-driven simulation, trajectory forecasting, end-to-end driving, and collaborative driving. We organize them by single-vehicle vs. multi-agent focus and by primary task (perception, prediction, planning). Each dataset is compared across sensor modalities, 3D annotation availability, scale, HD map inclusion, and key applications.

\subsection{Real-World Single-Vehicle Perception Datasets}

Single-vehicle datasets typically provide sensor data collected from a single autonomous vehicle (\textit{i.e.}, ego-vehicle). They focus on perception tasks like object detection, tracking, and scene understanding. Many include \textbf{multimodal sensors} (cameras, LiDAR, etc.) and 3D annotations for training detection and segmentation models. These datasets enable \textit{end-to-end learning} and generative models that require real distributions. We list popular datasets in Table \ref{table:real_world_single_vehicle_perception_datasets}. These datasets vary in geographic coverage, sensor configurations, and annotation types, reflecting the diverse requirements of perception tasks. Datasets like KITTI  \cite{kitti} and Cityscapes \cite{cordts2016cityscapes} were among the pioneers, offering stereo camera data primarily for 2D and 3D object detection and segmentation tasks. As the field progressed, more comprehensive datasets such as nuScenes \cite{nuscenes2019} and Waymo Open Dataset \cite{waymodataset} emerged, incorporating a suite of sensors including LiDAR, radar, and multiple cameras, along with high-definition maps, to facilitate complex tasks like 3D object detection and tracking. Recent datasets like ONCE \cite{mao2021one} and PandaSet \cite{xiao2021pandaset} have pushed the boundaries further by providing extensive data volumes and diverse driving scenarios, although some lack certain sensor modalities like radar or HD maps. The inclusion of high-resolution sensors and detailed annotations in these datasets supports the development of robust perception algorithms capable of operating in varied environments.

\begin{table}[h]
\centering
\caption{Summary of Real-World Single-Vehicle Perception Datasets}
\resizebox{\textwidth}{!}{
\begin{tabular}{c|c|c|c|c|c|c|c}
\toprule
\textbf{Dataset} & \textbf{Data Source}  & \textbf{Sampling Rate} & \textbf{Camera Type} & \textbf{LiDAR} & \textbf{Radar} & \textbf{HD Map} & \textbf{Annotation Type} \\
\midrule
\href{http://www.cvlibs.net/datasets/kitti/}{KITTI (2012)} \cite{kitti} & Karlsruhe, Germany & \SI{10}{Hz} & Stereo (2 cameras) & \checkmark &   &   & 3D Bounding Boxes \\
\href{https://www.cityscapes-dataset.com/}{Cityscapes (2016)} \cite{cordts2016cityscapes} & 50 German Cities & N/A & Stereo (2 cameras) &   &   &   & 2D Segmentation \\
\href{http://apolloscape.auto/}{ApolloScape (2018)} \cite{huang2018apolloscape} & Various Cities in China & N/A & Stereo (2 cameras) & \checkmark &   & \checkmark & Semantic Segmentation \\
\href{https://usa.honda-ri.com/h3d}{Honda H3D (2019)} \cite{360LiDARTracking_ICRA_2019} & Bay Area, USA & N/A & Frontal View (1 camera) & \checkmark &  &  & 3D Bounding Boxes \\
\href{https://www.nuscenes.org/}{nuScenes (2019)} \cite{nuscenes2019} & Boston, Pittsburgh and Singapore & \SI{2}{Hz} & Surround View (6 cameras) & \checkmark & \checkmark &  & 3D Bounding Boxes \\
\href{https://waymo.com/open/}{Waymo Open Dataset (2019)} \cite{waymodataset} & Multiple US Cities & \SI{10}{Hz} & Frontal/Side View (5 cameras) & \checkmark &  & \checkmark & 3D Bounding Boxes \\
\href{https://www.argoverse.org/}{Argoverse (2019)} \cite{argoverse2019} & Miami and Pittsburgh & \SI{10}{Hz} & Surround View & \checkmark &  & \checkmark & 3D Bounding Boxes \\
\href{https://scale.com/open-datasets/pandaset}{PandaSet (2020)} \cite{xiao2021pandaset} & San Francisco & N/A & Surround View (7 cameras) & \checkmark &  &  & 3D Bounding Boxes, Segmentation  \\
\href{https://www.a2d2.audi/}{Audi A2D2 (2020)} \cite{geyer2020a2d2} & Various Cities in Germany & \SI{10}{Hz} & Surround View (6 cameras) & \checkmark &  &  & 3D Bounding Boxes \\
\href{https://once-for-auto-driving.github.io}{ONCE Dataset (2021)} \cite{mao2021one} & Various Cities in China & \SI{10}{Hz} & Surround View (7 cameras) & \checkmark &  &  & 3D Bounding Boxes \\
\bottomrule
\end{tabular}
}
\label{table:real_world_single_vehicle_perception_datasets}
\end{table}

\subsection{Real-World Multi-Agent Prediction \& Planning Datasets}

Multi-agent datasets capture the \textbf{behavior of multiple road users} (vehicles, pedestrians, etc.) over time, often in the form of tracks or trajectories in a shared scene. These datasets support \textbf{trajectory prediction, motion planning, and simulation of interactions} – areas where generative models (\textit{e.g.} CVAEs, GANs, world models) are actively applied to predict diverse futures or generate realistic driving scenarios. Many provide \textbf{bird’s-eye-view trajectories with HD maps} instead of raw sensor feeds, focusing on \textit{behavior modeling}.

\begin{table}[h]
\centering
\label{table:motion_forecasting_datasets}
\caption{Summary of Motion Forecasting and Cooperative Driving Datasets}
\resizebox{\textwidth}{!}{
\begin{tabular}{c|c|c|c|c|c|c}
\toprule
\textbf{Dataset} & \textbf{Data Source}  & \textbf{Sampling Rate} & \textbf{Camera Type} & \textbf{LiDAR} & \textbf{HD Map} & \textbf{Annotation Type} \\
\midrule
\href{https://levelxdata.com/highd-dataset/}{HighD (2018)} \cite{highDdataset} & German Highways & N/A & Drone (Bird's-eye View) &  &  & Agent 2D Bounding Boxes \\
\href{https://interaction-dataset.com/}{INTERACTION (2019)} \cite{interactiondataset} & US, China, EU Intersections & \SI{10}{Hz} & Drone and Fixed Cameras &  & \checkmark & Agent Trajectories \\
\href{https://github.com/aras62/PIE}{PIE (2019)} \cite{rasouli2019pie} & Toronto, Canada & \SI{30}{Hz} & Frontal View (1 camera) &  &  & Pedestrian Bounding Boxes, Intention Labels \\
\href{https://www.argoverse.org/}{Argoverse 1 \& 2 (2019, 2022)} \cite{argoverse2019, wilson2023argoverse} & Miami and Pittsburgh & \SI{10}{Hz} & Surround View & \checkmark & \checkmark & Agent Trajectories \\
\href{https://woven-planet.github.io/l5kit/}{Lyft Level 5 (2020)} \cite{houston2021one} & Palo Alto, USA & \SI{10}{Hz} & Surround View & \checkmark & \checkmark & Agent 3D Bounding Boxes \\
\href{https://levelxdata.com/round-dataset/}{rounD (2020)} \cite{rounDdataset} & German Roundabouts & N/A & Drone (Bird's-eye View) &  & & Vehicle 2D Bounding Boxes \\
\href{https://waymo.com/open/}{Waymo Open Motion (2021)} \cite{ettinger2021large} & Multiple US Cities & \SI{10}{Hz} & None & &  & Vehicle, Pedestrian, Cyclist Trajectories \\
\href{https://github.com/motional/nuplan-devkit}{nuPlan (2021)} \cite{karnchanachari2024towards} & Multiple US Cities & \SI{10}{Hz} & Surround View & \checkmark & \checkmark & Agent 3D Bounding Boxes \\

\href{https://thudair.baai.ac.cn/index}{LOKI (2021)} \cite{girase2021loki} & Japan Intersections & \SI{5}{Hz} & Vehicle Cameras & \checkmark & \checkmark & 3D Bounding Boxes, Intention Labels \\

\href{https://thudair.baai.ac.cn/index}{DAIR-V2X (2021)} \cite{yu2022dair} & China Intersections & N/A & Vehicle and Roadside Cameras & \checkmark &  & 3D Bounding Boxes \\
\href{https://levelxdata.com/exid-dataset/}{exiD (2022)} \cite{exiDdataset} & German Highway Exits & N/A & Drone (Bird's-eye View) &  &  & Vehicle 2D Bounding Boxes \\
\href{https://github.com/AIR-Act2Act/AIR-Act2Act}{V2X-Seq (2023)} \cite{v2x-seq} & Urban Intersections & \SI{10}{Hz} & Vehicle and Roadside Cameras & \checkmark & \checkmark & 3D Agent Bounding Boxes \\
\href{https://github.com/ai4r/V2V4Real}{V2V4Real (2023)} \cite{xu2023v2v4real} & Ohio, USA & \SI{10}{Hz} & Surround View & \checkmark &  & 3D Bounding Boxes \\
\href{https://github.com/autonomousvision/navsim}{NAVSIM (2024)} \cite{dauner2024navsim} & Multiple US Cities & \SI{10}{Hz} & Surround View & \checkmark & \checkmark & Agent 3D Bounding Boxes \\
\href{https://github.com/ai4r/UniOcc}{UniOcc (2025)} \cite{wang2025unioccunifiedbenchmarkoccupancy} & Various Cities in US & \SI{10}{Hz} & Surround View & \checkmark & & 3D Occupancy Grids \\
\bottomrule
\end{tabular}
}
\end{table}

\subsection{Synthetic and Simulation Datasets}
Synthetic datasets are generated via video games or simulators, such as CARLA \cite{dosovitskiy2017carla}, providing perfectly annotated data and controllable diversity. These are crucial for \textbf{data synthesis and domain adaptation} in generative modeling – for example, training image translation GANs or augmenting rare scenarios (rain, accidents) that are hard to capture in reality. Recent synthetic sets span single-vehicle tasks as well as multi-agent and V2X scenarios.

\begin{table}[h]
\label{table:simulation_datasets}
\centering
\caption{Summary of Simulation Datasets for Autonomous Driving}
\resizebox{\textwidth}{!}{
\begin{tabular}{c|c|c|c|c|c}
\toprule
\textbf{Dataset} & \textbf{Data Source} & \textbf{Camera Type} & \textbf{LiDAR}  & \textbf{HD Map} & \textbf{Simulation Task} \\
\midrule
\href{http://perso.lcpc.fr/tarel.jean-philippe/bdd/frida.html}{FRIDA/FRIDA2 (2010–2012)} \cite{frida, frida2} & MATLAB  & Monocular  &  &  & Foggy Images \\
\href{http://synthia-dataset.net/}{SYNTHIA (2016)} \cite{ros2016synthia} & Unity &  Multiple Views &  &   & Rain and Fog Images \\
\href{https://europe.naverlabs.com/research/computer-vision/virtual-kitti/}{Virtual KITTI (2016 \& 2019)} \cite{gaidon2016virtual} & KITTI, Unity & Monocular/Stereo &  &   & Real2Sim Transfer \\
\href{https://playing-for-benchmarks.org/}{Playing for Benchmarks (2018)} \cite{richter2017playing} & GTA-V Game Engine & Multiple Views & \checkmark &  & Interactive Driving Simulation \\
\href{https://people.ee.ethz.ch/~csakarid/SFSU_synthetic/}{Foggy Cityscapes (2018)} \cite{SDV18} & Cityscapes \cite{cityscapes} & Monocular &  &   & Foggy Images \\
\href{https://idda-dataset.github.io/}{IDDA (2020)} \cite{alberti2020idda} & CARLA Simulator & Fisheye &  &   & Semantic Segmentation \\
\href{https://github.com/xinshuoweng/AIODrive}{AIODrive (2021)} \cite{weng2021all} & CARLA & Multiple Views & \checkmark & \checkmark & Long Range Point Cloud \\
\href{https://github.com/OPV2V/OPV2V}{OPV2V (2021)} \cite{xu2022opv2v} & CARLA & Multiple Vehicles & \checkmark &  & Cooperative Perception \\
\href{https://github.com/SysCV/shift-detection-tta}{Shift (2022)} \cite{sun2022shift} & CARLA & Multiple Views & \checkmark  & \checkmark & Weather, Lighting Simulation \\
\href{https://deepaccident.github.io/}{DeepAccident (2023)} \cite{wang2024deepaccident} & CARLA  & Multiple Views & \checkmark & \checkmark  & Accident Scene Simulation \\
\href{https://warm-3d.github.io}{WARM-3D (2024)} \cite{zhou2024warm} & CARLA  & Monocular  &  & \checkmark & Sim2Real Transfer \\
\bottomrule
\end{tabular}
}
\end{table}

\subsection{Language Annotated Datasets}

In 2018, BDD-X \cite{kim2018bdd-x} pioneered the language-annotated datasets for autonomous driving. Influenced by the recent success in language models, several new multimodal datasets have been introduced since 2023 to integrate natural language understanding with autonomous driving perception. These datasets pair visual sensor data (camera views, and often LiDAR or maps) with textual annotations, ranging from question-answer (QA) pairs and free-form captions to instruction-like statements, specifically formatted for large language models or visual question answering tasks.

\begin{table}[h]
\label{table:language_based_datasets}
\centering
\caption{Summary of Language-Based Datasets for Autonomous Driving}
\resizebox{\textwidth}{!}{
\begin{tabular}{c|c|c|c|c}
\toprule
\textbf{Dataset} & \textbf{Data Source} & \textbf{Modality} & \textbf{QA Type} & \textbf{\# QA Pairs} \\
\midrule
\href{https://github.com/JinkyuKimUCB/BDD-X-dataset}{BDD-X (2018)} \cite{kim2018bdd-x} & Dashcam Recordings & Videos (40s clips) & Ego Intention, Scene Description & 7K \\

\href{https://usa.honda-ri.com/drama}{DRAMA (2023)} \cite{malla2023drama} & Japan Driving Videos & Video & Risk Object, Ego Intention, Ego Actions, Reasoning & 170K \\
\href{https://usa.honda-ri.com/rank2tell}{Rank2Tell (2024)} \cite{sachdeva2024rank2tell} & US Driving Videos & Video & Object Importance, Ego Intention, Ego Actions, Reasoning & 300K \\

\href{https://github.com/wayveai/LingoQA}{LingoQA (2024)} \cite{marcu2023lingoqa} & Driving Videos (4s clips) & Video & Scene Description, Recommended Actions, Reasoning & 419K \\
\href{https://github.com/qiantianwen/NuScenes-QA}{NuScenes-QA (2024)} \cite{qian2023nuscenesqa} & nuScenes & Same as nuScenes & Scene Description & 460K \\
\href{https://github.com/OpenDriveLab/DriveLM}{DriveLM (2024)} \cite{sima2024drivelm} & nuScenes, CARLA & Same as nuScenes & Multi-step Reasoning  & 360K \\
\href{https://github.com/sungyeonparkk/NuPlanQA}{NuPlanQA (2025)}\footnote{Not Released as of April 2025} \cite{park2025nuplanqalargescaledatasetbenchmark} & nuPlan & Same as nuPlan &  Perception, Spatial Reasoning, Ego Intentions & 1M \\
\href{https://github.com/xmed-lab/NuInstruct}{NuInstruct (2024)} \cite{ding2024holisticautonomousdrivingunderstanding} & nuScenes & Same as nuScenes & Instruction–Response Pairs Across 17 Task Types & 91K \\
\href{https://github.com/rossgreer/doscenes}{doScenes (2024)} \cite{roy2024doscenes} & nuScenes & Same as nuScenes & Free-Form Driving Instructions and Scene Reference Points & 4K \\
\href{https://github.com/LLVM-AD/MAPLM}{MAPLM (2024)} \cite{tencent2023maplm} & Chinese Cities & Image, LiDAR & Detailed Map Description (Lanes, Road, Signs) & 61K \\
\href{https://github.com/turingmotors/NuScenes-MQA}{NuScenes-MQA (2024)} \cite{inoue2023nuscenesmqaintegratedevaluationcaptions} & nuScenes & Same as nuScenes & Scene Captioning, Visual QA & 1.5M \\
\href{https://github.com/drive-bench/toolkit}{DriveBench (2025)} \cite{xie2025vlms} & nuScenes & Same as DriveLM & Visual QA & 20k \\
\bottomrule
\end{tabular}
}
\end{table}


\clearpage

\section{Fundamentals of Generative AI}
\label{sec:foundations}

\begin{figure}[!]
  \setkeys{Gin}{width=0.9\textwidth}
  \centering
    \includegraphics{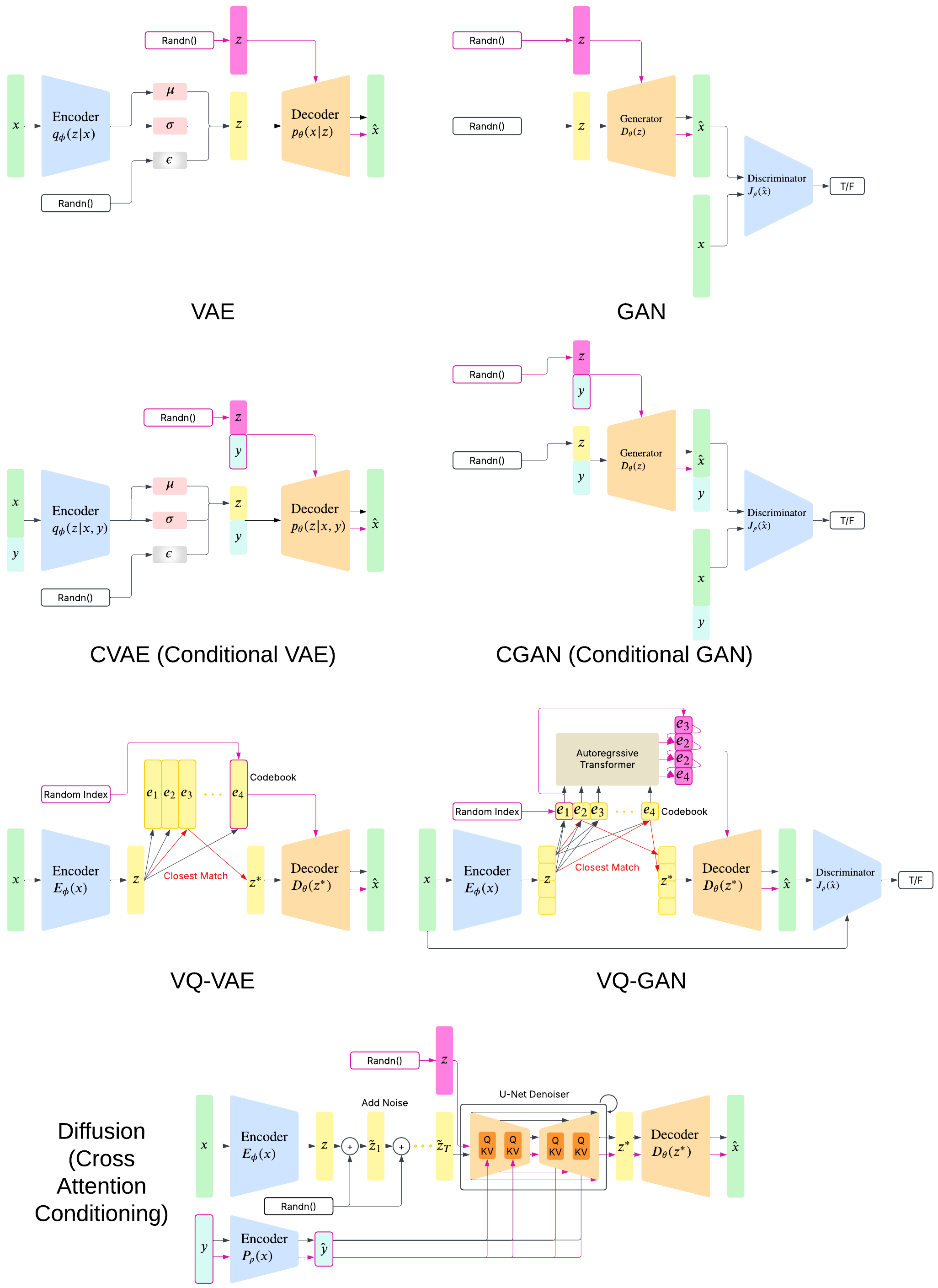}
\caption{Illustration of popular but easily confused generation models. The \textbf{black} arrows denote training time program flow, whereas the \textcolor{magenta}{purple} arrows denote inference time program flow.}
\label{fig:generative_foundation_comparison}
\end{figure}

This section provides the preliminaries on popular generative models used in autonomous driving. We list their architectures, training objectives, advantages, and limitations. In addition, we include a diagram to elucidate the commonly confusing model pipelines in Figure \ref{fig:generative_foundation_comparison}.

\subsection{Variational Autoencoder (VAE) and Variants}

\paragraph{Vanilla VAE}

Variational Autoencoder (VAE) is a probabilistic framework for generating data by learning a latent representation of the input distribution. Introduced by Kingma and Welling (2014) \cite{kingma2013vae}, VAE has gained popularity for its ability to produce diverse and semantically meaningful outputs, making it a natural choice for tasks like autonomous driving generation.

\textbf{Components:} A vanilla VAE consists of two key components: 
\begin{enumerate}
    \item $q_{\phi}(z|x)$: An encoder maps input data $x$ to a latent space representation $z$, parameterized by $\phi$. In practice, this encoder can be a simple multi-layer perceptron (MLP) that maps the input to a set of means and log variances.
    \item $p_{\theta}(x|z)$: A decoder maps the latent variable $z$ back to the data space, parameterized by $\theta$. This could also be a simple MLP to map sampled latents back to the input format.
\end{enumerate}

\textbf{Training Objective:} The latent variable $z$ is modeled as a random variable, typically sampled from a standard normal distribution $\mathcal{N}(0, I)$. At training, our goal is to maximize the Evidence Lower Bound (ELBO) of the marginal log-likelihood $\log p_{\theta}(x)$:
\begin{equation}
    \log p_{\theta}(x) \geq \mathbb{E}_{z \sim q_{\phi}(z|x)} \left[ \log p_{\theta}(x|z) \right] - \mathrm{KL} \left( q_{\phi}(z|x) || p(z) \right),
\end{equation}
where:
\begin{enumerate}
    \item $\mathbb{E}_{q_{\phi}(z|x)} \left[ \log p_{\theta}(x|z) \right]$ is the reconstruction loss, ensuring the decoded data $\hat{x}$ closely matches the original input $x$.
    \item $\mathrm{KL} \left( q_{\phi}(z|x) || p_{\theta}(z) \right)$ is the Kullback-Leibler (KL) divergence, regularizing the latent distribution $q_{\phi}(z|x)$ to approximate the prior $p_{\theta}(z)$.
\end{enumerate}

In terms of loss function, we seek to minimize:
\begin{equation}
    \mathcal{L}(\theta, \phi; x) = -\mathbb{E}_{z \sim q_{\phi}(z|x)} \left[ \log p_{\theta}(x|z) \right] + \mathrm{KL} \left( q_{\phi}(z|x) || p(z) \right).
\end{equation}

To enable gradient-based optimization, the latent variable $z$ is reparameterized as:
\begin{equation}
    z = \mu_{\phi}(x) + \sigma_{\phi}(x) \odot \epsilon, \quad \epsilon \sim \mathcal{N}(0, I),
\end{equation}
where $\mu_{\phi}(x)$ and $\sigma_{\phi}(x)$ are the mean and standard deviation of the approximate posterior $q_{\phi}(z|x)$, and $\odot$ denotes element-wise multiplication.

\textbf{Inference:} At inference time, we randomly sample $z \sim \mathcal{N}(0, I)$ and pass it through the decoder to acquire a generated image.

\begin{tcolorbox}[colback=gray!5!white, colframe=black!50!gray, parbox=false]
\textbf{Key Takeaways:}
While VAEs are versatile for many tasks \cite{xu2017variational, harvey2021conditional, zhang2021point}, particularly in the image domain, they often produce blurry outputs compared to adversarial methods (e.g., GANs). This limitation arises from the assumption of a Gaussian latent space and the pixel-wise reconstruction loss \cite{larsen2016autoencoding, van2017vqvae}.
\end{tcolorbox}

\paragraph{Conditional Variational Autoencoder (CVAE)}

A popular variant of Conditional Variational Autoencoder (CVAE) \cite{sohn2015CVAE} extends the standard VAE framework by introducing an additional conditioning variable $c$ that guides both the encoding and decoding processes. This is especially useful in tasks like autonomous driving, where the data generation process is context-dependent, such as generating future trajectories conditioned on past motion, or generating sensor data based on maps or driving commands.

\textbf{Components:} Given an input $x$ and condition $c$, CVAE consists of the following components:
\begin{enumerate}
    \item $q_{\phi}(z|x, c)$: An encoder network that maps the data $x$ and condition $c$ into a latent distribution over $z$.
    \item $p_{\theta}(x|z, c)$: A decoder that reconstructs or generates data $x$ from the latent code $z$ and the same conditioning variable $c$.
\end{enumerate}

\textbf{Training Objective:} The model now maximizes the conditional log-likelihood $\log p_{\theta}(x|c)$ through the Evidence Lower Bound (ELBO), similarly to VAE:
\begin{equation}
    \log p_{\theta}(x|c) \geq \mathbb{E}_{z \sim q_{\phi}(z|x,c)} \left[ \log p_{\theta}(x|z, c) \right] - \mathrm{KL} \left( q_{\phi}(z|x, c) || p(z|c) \right),
\end{equation}
where $p(z|c)$ is typically taken as a standard Gaussian $\mathcal{N}(0, I)$, independent of $c$, although it may be extended to a conditional prior in more expressive models.

The training objective becomes minimizing the following conditional loss:
\begin{equation}
    \mathcal{L}(\theta, \phi; x, c) = -\mathbb{E}_{z \sim q_{\phi}(z|x,c)} \left[ \log p_{\theta}(x|z, c) \right] + \mathrm{KL} \left( q_{\phi}(z|x, c) || p(z|c) \right).
\end{equation}

The reparameterization trick still applies, with the latent variable sampled as:
\begin{equation}
    z = \mu_{\phi}(x, c) + \sigma_{\phi}(x, c) \odot \epsilon, \quad \epsilon \sim \mathcal{N}(0, I),
\end{equation}
where $\mu_{\phi}(x, c)$ and $\sigma_{\phi}(x, c)$ are functions of both the input and condition.

\textbf{Inference:} At inference time, given only the condition $c$, we sample $z \sim \mathcal{N}(0, I)$ and generate data by:
\begin{equation}
    \hat{x} = p_{\theta}(x|z, c).
\end{equation}

\begin{tcolorbox}[colback=gray!5!white, colframe=black!50!gray, parbox=false]
\textbf{Key Takeaways:}
CVAE’s ability to capture uncertainty and multimodal futures makes it particularly suitable for stochastic domains like autonomous driving. They have shown success in LiDAR point cloud completion \cite{zhang2021point} and trajectory prediction \cite{lee2017desire, kim2021driving, ivanovic2019trajectron}. Compared to the vanilla VAE, CVAE offers finer control during inference through conditioning, enabling targeted scene or behavior synthesis based on structured inputs.
\end{tcolorbox}

\paragraph{Vector Quantized Variational Autoencoders (VQ-VAE)}

In the image generation domain, Vector Quantized Variational Autoencoders (VQ-VAE) \cite{van2017vqvae} were introduced to address certain limitations of vanilla VAE, such as blurry reconstructions caused by the assumption of a continuous Gaussian latent space. VQ-VAE replaces the continuous latent space of traditional VAEs with a discrete latent space, enabling it to learn more expressive representations for tasks like autonomous driving scene generation. Like a vanilla VAE, VQ-VAE consists of an encoder and a decoder. However, the “encoder” and “decoder” in VQ-VAE are \textbf{not} distributions because the latent mapping is done via a \textbf{discrete nearest-neighbor lookup}, rather than sampling from a continuous or parametric distribution as in a vanilla VAE. 

\textbf{Components:} Compared to VAE, VQ-VAE introduces a quantization step between these components:
\begin{enumerate}
    \item The encoder maps the input $x$ to a latent representation 
     \begin{equation}
     z_e(x) = E_{\phi}(x) \quad z_e(x) \in \mathbb{R}^d
    \end{equation}     
    \item Instead of directly using $z_e(x)$, VQ-VAE quantizes this latent vector to the nearest entry in a learned codebook $\mathcal{E} = \{e_i \in \mathbb{R}^d\}_{i=1}^K$ with $K$ discrete entries. The quantized representation is:
    \begin{equation}
        z_q(x) = e_k, \quad k = \arg\min_i \|z_e(x) - e_i\|^2.
    \end{equation}
    \item The decoder takes the quantized representation $z_q(x)$ and maps it back to the data space to reconstruct 
    \begin{equation}
        \hat{x} = D_{\theta}(z_q(x)).
    \end{equation}    
\end{enumerate}

\textbf{Training Objective:} The objective function for VQ-VAE consists of three terms:
\begin{equation}
\label{equation:vq_vae_loss}
    \mathcal{L}(\mathcal{E}, \theta, \phi; x) = \|x - \hat{x}\|^2 + \|z_e(x) - \text{sg}[z_q(x)]\|^2 + \beta \|z_q(x) - \text{sg}[z_e(x)]\|^2,
\end{equation}
where:
\begin{enumerate}
    \item The first term is the reconstruction loss, ensuring the decoded output $\hat{x}$ closely matches the input $x$. Compared to the reconstruction loss in VAE, VQ-VAE uses a discrete codebook and picks a \textbf{single} nearest vector for each input. The encoder outputs $z_e(x)$, and a quantization step maps that \textbf{deterministically} to one codebook entry, $z_q(x)$. In other words, there is no “sampling over multiple possible codes”; each $x$ is mapped to one code vector.
    \item The second term encourages the encoder output $z_e(x)$ to be close to its quantized version $z_q(x)$. The stop-gradient operator $\text{sg}[\cdot]$ prevents gradients from flowing through $z_q(x)$ during backpropagation.
    \item The third term (weighted by $\beta$) ensures that the codebook embeddings are updated to match the encoder outputs.
    \item Unlike the continuous VAE, there is no explicit KL term. The codebook and the discrete nature of the latent space act as a form of regularization.
\end{enumerate}

\textbf{Inference:} At inference time, we randomly sample $z \sim \mathcal{N}(0, I)$ and pass it through the decoder to acquire a generated image.

\begin{tcolorbox}[colback=gray!5!white, colframe=black!50!gray, parbox=false]
\textbf{Key Takeaways:}
VQ-VAE distinguishes itself from vanilla VAE by employing a discrete latent space, represented by a learned codebook $\mathcal{E}$, instead of a continuous variable $z$. This fundamental difference yields several key advantages. The discrete nature is often more expressive for capturing high-level semantic or categorical structures within data, such as object types or scene layouts. It helps mitigate the characteristic blurriness associated with traditional VAE reconstructions, which often stem from continuous latent assumptions and pixel-wise losses. Furthermore, this discrete representation facilitates effective data compression and integrates naturally with powerful autoregressive models (e.g., Pixel-CNN \cite{van2016conditional}) for learning priors over the latent codes. VQ-VAE's optimization mechanism, centered around a codebook loss, also offers an alternative approach that can avoid the posterior collapse problem sometimes observed in a vanilla VAE due to KL divergence balancing.

Despite its strengths, the VQ-VAE framework introduces specific challenges. The training process can be susceptible to issues like "codebook collapse," where only a subset of the discrete codes in the codebook ends up being actively used or updated, potentially limiting the model's representational capacity. Moreover, adequately capturing the nuances of highly complex datasets might require significantly large codebooks, which in turn increases the computational costs associated with training, storage, and inference. Therefore, utilizing VQ-VAE involves weighing its benefits in expressiveness and avoiding certain VAE pitfalls against these potential training instabilities and computational demands.
\end{tcolorbox}

\subsection{Generative Adversarial Network (GAN) and Variants}
\label{sec:gan}

\paragraph{Vanilla GAN}

Generative Adversarial Network (GAN) was first introduced by Goodfellow et al.\ (2014) \cite{goodfellow2014generative}. GAN frames the problem of generative modeling as a two-player minimax game between a \emph{generator} network $D_\theta$ and a \emph{discriminator} network $J_\rho$. The goal of the generator is to produce realistic samples that are indistinguishable from real data, while the discriminator attempts to distinguish generated samples from real ones.

\textbf{Components}: Let $x \sim p_{\text{data}}(x)$ denote samples from the real data distribution and $z \sim p_z(z)$ denote samples from a simple prior distribution (e.g., Gaussian or uniform). GAN has these components:
\begin{enumerate}
    \item A generator $D_{\theta}(z)$ maps latent variables $z$ to the data space.
    \item A discriminator $J_{\rho}(x)$ receives either real or generated data and outputs a scalar indicating the likelihood that the input is real.
\end{enumerate}

\textbf{Training Objective:} GAN's objective function is defined as:
\begin{equation}
    \min_\theta \max_\rho \;\mathbb{E}_{x \sim p_{\text{data}}(x)} \left[ \log J_\rho(x) \right] + \mathbb{E}_{z \sim p_z(z)} \left[ \log \left( 1 - J_\rho(D_{\theta}(z)) \right) \right].
\end{equation}

The discriminator $J_{\rho}$ aims to maximize this objective by outputting 1 for real data and 0 for fake data, while the generator $D_{\theta}$ aims to fool the discriminator by generating realistic samples.

GANs are trained in alternating steps:

\begin{enumerate}
    \item \textbf{Discriminator Update:} Fix $D_\theta$ and perform a gradient ascent step on $\psi$ to increase its ability to separate real and fake samples.
    \item \textbf{Generator Update:} Fix $J_{\rho}$ and perform a gradient descent step on $\theta$ to minimize the probability that $J_{\rho}$ correctly identifies $D_\theta(z)$ as fake.
\end{enumerate}

Note that the generator does not have access to real data directly; its learning signal comes solely through the discriminator's feedback.

\textbf{Alternative Losses:} In practice, the original GAN loss may lead to vanishing gradients when the discriminator becomes too strong. Several alternatives have been proposed:

\begin{itemize}
    \item \textbf{Non-saturating loss:} Replace $\log(1 - D(G(z)))$ with $-\log D(G(z))$ for the generator to provide stronger gradients.
    \item \textbf{Least-Squares GAN (LSGAN)} \cite{mao2017least}: Uses a least-squares loss to stabilize training.
    \item \textbf{Wasserstein GAN (WGAN)} \cite{arjovsky2017wasserstein}: Optimizes the Earth Mover (Wasserstein-1) distance, which provides more informative gradients.
\end{itemize}

\textbf{Inference:} At inference time, one samples a standard Gaussian as $z$ and pass it through the generator to get the output $\hat{x}$ (e.g., an image).

\begin{tcolorbox}[colback=gray!5!white, colframe=black!50!gray, parbox=false]
\textbf{Key Takeaways:}
GAN is widely recognized for its exceptional capability to synthesize highly realistic images rich in high-frequency details, often achieving superior visual fidelity compared to methods like VAE. Its strength stems from an adversarial training paradigm where a generator network learns to produce data that is indistinguishable from real data to a discriminator network, crucially without relying on direct reconstruction losses or assuming a specific output distribution. This allows GAN to capture complex data characteristics effectively, establishing it as a powerful tool for generating convincing visual content.

Despite its success in generation quality, GAN presents significant practical challenges. It is prone to "mode collapse," a phenomenon where the generator fails to capture the full diversity of the data distribution and instead produces only a limited variety of outputs. The adversarial training process itself is often unstable, demanding meticulous hyperparameter tuning and careful architectural design for convergence. The vanilla GAN also typically lacks an easily interpretable latent space, hindering controlled synthesis or meaningful interpolation between generated samples. Consequently, many advanced applications employ hybrid architectures, such as VQ-GAN \cite{esser2020vqgan} and StyleGAN \cite{karras2019style}, which combine GAN with other techniques like discrete latent spaces or style modulation to improve stability, control, and interpretability, reinforcing the foundational importance of GAN while mitigating their inherent difficulties.
\end{tcolorbox}




\paragraph{Conditional Generative Adversarial Network (CGAN)}
\label{sec:cgan}

Conditional GAN (CGAN or cGAN) \cite{mirza2014conditional} extends the vanilla GAN framework by introducing conditional inputs to both the generator and discriminator, allowing for the generation of data that adheres to specified attributes (e.g., class labels, semantic layouts, or textual descriptions). Just like CVAE vs. VAE, this extension is particularly valuable in autonomous driving, where scene composition, semantic control, and domain-specific constraints are essential for generating realistic and diverse outputs.

\textbf{Components:} The conditional GAN framework augments both generator and discriminator with side information $y$, such as class labels, bounding boxes, or other domain-specific conditions.  

\begin{itemize}
    \item The conditional generator $D_\theta(z, c)$ is the conditional generator that outputs an image $\hat{x}$,
    \item The conditional discriminator $J_\rho(x, c)$ is the conditional discriminator that evaluates whether the image $x$ matches the condition $y$.
\end{itemize}

\textbf{Training Objective:} The original CGAN loss function follows the standard GAN form, modified to incorporate conditioning:
\begin{equation}
    \min_\theta \max_\psi \;\mathbb{E}_{x,y \sim p_{\text{data}}(x, y)} \left[ \log J_\rho(x, y) \right] + \mathbb{E}_{z \sim p_z(z), y \sim p(y)} \left[ \log \left( 1 - J_\rho(D_\theta(z, y), y) \right) \right].
\end{equation}

\textbf{Inference:} Similar to GAN, one samples a gaussian vector $z$, but now also needs to concatenate it with a condition $y$, to feed into the generator.

\begin{tcolorbox}[colback=gray!5!white, colframe=black!50!gray, parbox=false]
\textbf{Key Takeaways:}
CGAN enhances the vanilla GAN framework by incorporating auxiliary information, or conditions $c$, into both the generator and discriminator, enabling explicit control over the generated output $x$. This allows for fine-grained manipulation of the synthesis process according to specific attributes like class labels, text descriptions, or semantic maps, making CGAN highly valuable for task-specific generation. While guided by the condition $c$, cGANs can still produce diverse outputs for the same condition by varying the input latent code $z$, and leveraging meaningful semantic or spatial priors as conditions often leads to more structurally coherent results than achievable with an unconditional GAN.

However, CGAN faces several challenges, inheriting issues like potential mode collapse from their GAN predecessors, where output diversity might be limited despite conditioning. A unique challenge is conditional overfitting, where the generator might overly rely on the condition $c$ while effectively ignoring the latent code $z$, leading to reduced variability. Furthermore, training a robust CGAN typically requires well-aligned paired data $(x,y)$, which can be costly or difficult to obtain. Significant research efforts have focused on overcoming these limitations, leading to advancements such as multimodal CGAN (e.g., BicycleGAN~\cite{zhu2017toward}) designed to improve output diversity, attention-based methods (e.g., SPADE~\cite{park2019semantic}) for better spatial feature alignment, and the emergence of powerful diffusion-based conditional models (e.g., ControlNet \cite{zhang2023adding}) offering alternative high-fidelity conditional generation capabilities.
\end{tcolorbox}




\paragraph{Vector Quantized Generative Adversarial Network (VQ-GAN)}

VQ-GAN \cite{esser2020vqgan} combines \textbf{Vector Quantization} (as in VQ-VAE) with a \textbf{GAN} (Generative Adversarial Network) \cite{goodfellow2014generative} objective. The motivation is to preserve the advantages of a discrete latent space (leading to high-quality reconstructions) while also leveraging an adversarial loss to further improve \emph{fidelity}, \emph{sharpness}, and \emph{perceptual quality} of generated images. 

\textbf{Components:} VQ-GAN has these components:

\begin{enumerate}
    \item \textbf{Encoder} $E_\phi(x)$: A convolutional neural network that transforms the input $x$ (e.g., an image) into a continuous latent representation $z_e$. In practice, $z_e$ may be a 2D feature map for image data:
    \begin{equation}
        z_e = E_\phi(x).
    \end{equation}
    
    \item \textbf{Vector Quantization} $\mathrm{Quantize}(z_e)$: Each local vector in $z_e$ is mapped to the nearest code vector from a learned codebook $\mathcal{E} = \{e_i \in \mathbb{R}^d\}_{i=1}^K$. Formally, for each vector $z_e(i)$ in $z_e$,
    \begin{equation}
        k^* \;=\; \arg\min_{k \in \{1,\dots,K\}} \bigl\|\,z_e(i) - e_k\bigr\|_2
        \quad \text{and} \quad
        z_q(i) \;=\; e_{k^*}.
    \end{equation}
    This yields a \emph{discrete} latent representation $z_q$.
    
    \item \textbf{Decoder} $D_\theta(z_q)$: Another neural network (the \emph{generator} in the GAN framework) that reconstructs an image $\hat{x}$ from the discrete codes $z_q$. Symbolically,
    \begin{equation}
        \hat{x} = D_\theta\bigl(z_q\bigr).
    \end{equation}
    
    \item \textbf{Discriminator} $J_\rho(\cdot)$: A separate neural network that learns to distinguish \emph{real} images $x$ from \emph{reconstructed/generated} images $\hat{x}$. The adversarial loss from $J_\rho$ encourages the generator to produce outputs with realistic textures and details.
\end{enumerate}

\textbf{Training Objectives:} Similar to GAN, the training process of VQ-GAN also has two alternating stages:

\begin{enumerate}
    \item \textbf{Discriminator Step:} We fix the generator components $(\phi, \theta, \mathcal{E})$ and only update $\psi$.
    \begin{equation}
        \mathcal{L}_{\mathrm{GAN}}(\psi) = [\text{log}D(x)+\text{log}(1-D(\hat{x}))]
    \end{equation}

    \item \textbf{Generator Step:} We fix the generator components $\psi$ and only update $(\phi, \theta, \mathcal{E})$, similar to equation \ref{equation:vq_vae_loss}.
    \begin{equation}
        \mathcal{L}(\mathcal{E}, \theta, \phi; x) = \|x - \hat{x}\|^2 + \|z_e(x) - \text{sg}[z_q(x)]\|^2 + \beta \|z_q(x) - \text{sg}[z_e(x)]\|^2,
    \end{equation}
    where $\mathrm{sg}[\cdot]$ is the \emph{stop-gradient} operator and $\beta$ is a hyperparameter (the \emph{commitment cost}).
\end{enumerate}

In short, the generator (encoder--decoder and codebook) is trained to \emph{minimize} these losses, while the discriminator is trained to \emph{maximize} its ability to classify real vs.\ generated images.

\textbf{Inference:} At inference time, we randomly sample an initial embedding from the codebook $\mathcal{E}$, and then use an autoregressive method to find the subsequent codes until their concatenated form can form the latent vector $z$. We then pass it to the decoder to generate an output.

\begin{tcolorbox}[colback=gray!5!white, colframe=black!50!gray, parbox=false]
\textbf{Key Takeaways:}
Compared to a vanilla GAN, VQ-GAN's intermediate discrete bottleneck allows it to perform compression-then-reconstruction, guided partly by reconstruction losses alongside the adversarial objective. This grounding in reconstruction often provides VQ-GAN with more stable training dynamics compared to many standard GANs and can help mitigate severe mode collapse, as the generator is anchored to reconstructing inputs rather than freely generating from noise.

Compared to VQ-VAE, VQ-GAN introduces a crucial addition: a patch-based discriminator network and an associated adversarial loss applied to the reconstructed images. While VQ-VAE relies solely on reconstruction fidelity (often measured by L1 or L2 loss) and codebook learning losses, VQ-GAN's adversarial component acts as a learned perceptual metric. This pushes the decoder (generator) to produce outputs that are not only accurate reconstructions but also possess higher perceptual quality, characterized by sharper details and more realistic textures, thereby addressing the potential blurriness sometimes observed in VQ-VAE results.

Despite effectively merging beneficial aspects, VQ-GAN is susceptible to challenges inherited from both its architectural parents. From VQ-VAE, it can suffer from issues like codebook collapse or dead embeddings, where portions of the learned discrete codebook become underutilized or inactive during training. From the GAN side, the inclusion of the adversarial loss introduces potential training instability and requires careful tuning of hyperparameters, especially the relative weighting between the reconstruction and adversarial loss terms. While potentially more stable than some pure GANs, it doesn't entirely eliminate the risk of mode collapse. Furthermore, the dual objectives create a potential tension: emphasizing the adversarial loss for perceptual sharpness and realism might lead the model to sacrifice fine-grained reconstruction accuracy or introduce visually plausible artifacts that deviate from the original data distribution.
\end{tcolorbox}





\subsection{Diffusion Models}

Diffusion models are a family of generative models that synthesize data by iteratively denoising random noise. Leveraging the power of stochastic processes, they have achieved state-of-the-art results in generating high-fidelity images, videos, and 3D scenes. Their robustness and flexibility make them particularly appealing for autonomous driving scene generation, where realism and diversity are critical. Unlike the aforementioned models, diffusion is more about a \textit{process} than a particular deep learning model.

\textbf{Processes:} Diffusion models consist of two main processes: a forward diffusion process and a reverse denoising process.

\begin{enumerate}
    \item \textbf{Forward Process:} The forward process gradually adds Gaussian noise to data \( x_0 \) over \( T \) timesteps to produce \( x_T \), effectively transforming the data into pure noise. This is modeled as:
    \begin{equation}
        q(x_t | x_{t-1}) = \mathcal{N}(x_t; \sqrt{1 - \beta_t} x_{t-1}, \beta_t \mathbf{I}),
    \end{equation}
    where \( \beta_t \) is a variance schedule controlling the noise level at each timestep.
    \item \textbf{Reverse Process:} The reverse process learns to denoise the noisy samples \( x_t \) step by step to reconstruct the original data \( x_0 \). This is parameterized as:
    \begin{equation}
        p_\theta(x_{t-1} | x_t) = \mathcal{N}(x_{t-1}; \mu_\theta(x_t, t), \Sigma_\theta(x_t, t)),
    \end{equation}
    where \( \mu_\theta \) and \( \Sigma_\theta \) are learned through a neural network.
\end{enumerate}

\textbf{Training Objective:} One common approach (DDPM) \cite{ho2020denoising} is to optimize a \emph{variational bound} on the log-likelihood:
\begin{equation}
    \log p_\theta(x_0)
    \;\;\ge\;\;
    \mathbb{E}_q\Bigl[\sum_{t=1}^T 
      -D_{\mathrm{KL}}\bigl(q(x_{t-1}\mid x_t, x_0) \,\|\, p_\theta(x_{t-1}\mid x_t)\bigr)
    \Bigr],
\end{equation}
where $q(x_{t-1}\mid x_t, x_0)$ is the \emph{true} posterior in the forward process. In practice, a simplified version of this objective is often used:
\begin{equation}
  \mathcal{L}(\theta; x)
  \;=\;
  \mathbb{E}_{x_0, \varepsilon, t}\bigl[
    \|\varepsilon - \varepsilon_\theta(x_t, t)\|^2
  \bigr],
\end{equation}
where $x_t = \sqrt{\bar{\alpha}_t}\, x_0 + \sqrt{1-\bar{\alpha}_t}\,\varepsilon$ (with $\varepsilon \sim \mathcal{N}(\mathbf{0},\mathbf{I})$), and $\varepsilon_\theta(\cdot)$ is the neural network estimating the noise.

\textbf{Inference:} After training, new samples are generated by starting from Gaussian noise $x_T$ and iteratively applying the learned reverse steps:
\begin{equation}
  x_{t-1} 
  \;\sim\;
  p_\theta\bigl(x_{t-1}\mid x_t\bigr),
  \quad
  t = T, T-1, \dots, 1.
\end{equation}
This denoising chain gradually removes noise and yields a final sample $x_0$ resembling the training distribution.

\paragraph{Conditioning the Denoiser}

In a diffusion model, the noise-prediction (or ``denoiser'') function can be viewed as
$$
    \varepsilon_\theta\bigl(x_t, t\bigr),
$$
which predicts the added noise in a noisy sample $x_t$ at timestep $t$. On top of it, the performance and flexibility of diffusion models can be significantly enhanced by conditioning the generation process on additional information, such as text, images, or semantic maps. When adding a conditioning signal $c$ (e.g., a CLIP \cite{radford2021clip} embedding or a Transformer \cite{vaswani2017attention} encoding), we typically modify this function to
$$
    \varepsilon_\theta\bigl(x_t, t, c\bigr).
$$
Below are two common ways to \emph{fuse} or incorporate $c$ into the denoiser:
\begin{enumerate}
        \item \textbf{Cross Attention:} A widespread approach (e.g., in Stable Diffusion \cite{rombach2021highresolution}) is to feed $c$ into a learned encoder $E_{\mathrm{cond}}(\cdot)$ and then inject those embeddings into each U-Net block of the denoiser via \emph{cross-attention}:
    \begin{equation}
        h_{\ell+1}
        \;=\;
        \mathrm{Block}_\ell\Bigl(
           h_\ell,\; t,\; \underbrace{\mathrm{CrossAttn}\bigl(\mathrm{LN}(h_\ell),\; E_{\mathrm{cond}}(c)\bigr)}_{\text{conditioning}}
        \Bigr),
    \end{equation}
    where $h_\ell$ is the latent feature map at level $\ell$, $\mathrm{Block}_\ell$ is a U-Net residual block, and
    \(\mathrm{CrossAttn}(Q,K)\) typically involves
    \begin{equation}
        Q = W_Q\,h_\ell,\quad
        K = W_K\,E_{\mathrm{cond}}(c),\quad
        V = W_V\,E_{\mathrm{cond}}(c),
    \end{equation}
    \begin{equation}
        \mathrm{CrossAttn}(Q,K,V)
        \;=\;
        \mathrm{softmax}\!\Bigl(\tfrac{Q\,K^\top}{\sqrt{d}}\Bigr) \,V.
    \end{equation}
    Here, the time step $t$ is also provided to each block (e.g., via positional embeddings). In essence, the \emph{denoiser network} is a U-Net that repeatedly attends to the conditioning vector (or sequence) $E_{\mathrm{cond}}(c)$, ensuring that each denoising step is guided by the condition.
    
    \item \textbf{Classifier-Free Guidance (Sampling-Time Technique) \cite{ho2022classifier}:} Another common formulation adds (or subtracts) a guidance term \emph{during sampling} rather than inside the architecture. One widely used variant is \textbf{classifier-free guidance}, where the model can be run \emph{with} or \emph{without} the condition $c$. Denoting these two calls as
    \[
        \varepsilon_\theta\bigl(x_t, t, \varnothing\bigr)
        \quad\text{and}\quad
        \varepsilon_\theta\bigl(x_t, t, c\bigr),
    \]
    the final noise prediction can be fused as
    \begin{equation}
        \tilde{\varepsilon}_\theta\bigl(x_t, t, c\bigr)
        \;=\;
        \varepsilon_\theta\bigl(x_t, t, \varnothing\bigr)
        \;+\;
        w \,\Bigl[
          \varepsilon_\theta\bigl(x_t, t, c\bigr)
          - 
          \varepsilon_\theta\bigl(x_t, t, \varnothing\bigr)
        \Bigr],
    \end{equation}
    where $w$ (the ``guidance scale'') controls how strongly the condition $c$ influences the denoising step. This effectively \emph{pushes} the model’s prediction toward the conditional result in a post-hoc way, even if the architecture itself uses only a single network $\varepsilon_\theta$. 
\end{enumerate}

In either case, the \emph{loss function} for diffusion (e.g., the simplified objective)
\begin{equation}
    \mathcal{L}(\theta; x) 
    \;=\;
    \mathbb{E}_{x_0,\varepsilon,t}\bigl[
        \|\varepsilon - \varepsilon_\theta(x_t, t, c)\|^2
    \bigr]    
\end{equation}
is updated to reflect that $\varepsilon_\theta$ \emph{now depends on} the condition $c$. In a cross-attention architecture, $c$ is integrated directly into the forward pass of $\varepsilon_\theta$. In classifier-free guidance, one trains both the conditioned denoiser and unconditioned denoiser, then mix the predictions \emph{during sampling} to control how strongly the model aligns with $c$. 

Diffusion model itself is a vast topic. We refer interested readers to a dedicated survey by Yang et al. \cite{yang2022diffusion}.

\begin{tcolorbox}[colback=gray!5!white, colframe=black!50!gray, parbox=false]
\textbf{Key Takeaways:}
Diffusion models have gained prominence for their remarkable ability to generate high-fidelity and diverse samples, often achieving state-of-the-art results that rival or surpass GANs, particularly in capturing the full range of data variation with reduced risk of mode collapse. Their training process is notably more stable and conceptually simpler than the adversarial dynamics of GANs, typically relying on straightforward objectives like predicting the noise added during a forward diffusion process. This iterative denoising mechanism also offers significant flexibility, enabling effective conditioning through techniques like classifier-free guidance for controlled generation, and allowing users to adjust the number of sampling steps to manage the trade-off between generation speed and final output quality.

Despite their strengths in sample quality and training stability, diffusion models face significant practical challenges primarily related to computational cost. The core iterative nature of the generation process, requiring numerous sequential steps (often hundreds or thousands) to denoise a sample, leads to substantially slower inference times compared to single-pass generative models like VAEs or GANs. This multi-step requirement also translates into higher computational demands for both training and inference. Furthermore, managing the intermediate representations across many iterations can necessitate considerable memory resources, particularly when dealing with high-dimensional data. These factors concerning speed and computational resources represent the main trade-offs against the high generation quality offered by diffusion models.
\end{tcolorbox}

\subsection{Neural Radiance Fields (NeRF)}
VAE, GAN, and diffusion models typically excel at the 2D image space generation. However, a growing trend in generative tasks is to harness the power of complete 3D representations. In 3D understanding and representation, Neural Radiance Field (NeRF) \cite{zhang2023nerf} and 3D Gaussian Splatting \cite{kerbl20233dgaussian} are the most popular. Introduced by Mildenhall et al. (2020)  \cite{mildenhall2020nerf}, NeRF models a scene with probabilistic rays originating from a view perspective, enabling high-quality rendering of complex 3D structures from a sparse set of 2D images. Its capability to generate photorealistic images and implicitly represent geometry makes it a powerful tool for autonomous driving scene generation.

\textbf{Components:} NeRF represents a scene as a set of tracing rays parameterized by a neural network:
\begin{equation}
 (\mathbf{c}, \sigma) = F_{\theta}(\mathbf{x}, \mathbf{d}),
\end{equation}
where:
\begin{itemize}
    \item \( \mathbf{x} \in \mathbb{R}^3 \): 3D spatial coordinates.
    \item \( \mathbf{d} \in \mathbb{R}^3 \): Viewing direction.
    \item \( \mathbf{c} \in \mathbb{R}^3 \): Emitted RGB color.
    \item \( \sigma \in \mathbb{R} \): Volumetric density.
    \item \( F_{\theta} \): A multilayer perceptron (MLP) parameterized by \( \theta \).
\end{itemize}

To render a pixel color, a ray \( \mathbf{r}(t) = \mathbf{o} + t\mathbf{d} \) is cast from the camera through the pixel, and the integral of the color along the ray is approximated as:
\begin{equation}
C(\mathbf{r}) = \int_{t_n}^{t_f} T(t) \sigma(\mathbf{r}(t)) \mathbf{c}(\mathbf{r}(t), \mathbf{d}) \, dt,
\end{equation}
where:
\begin{itemize}
    \item \( T(t) = \exp\left(-\int_{t_n}^t \sigma(\mathbf{r}(s)) \, ds \right) \): Transmittance, representing the probability of the ray reaching \( t \) without occlusion.
    \item \( t_n, t_f \): Near and far bounds of the ray.
\end{itemize}

The integral is approximated using stratified sampling:
\begin{equation}
C(\mathbf{r}) \approx \sum_{i=1}^N T_i \alpha_i \mathbf{c}_i,
\end{equation}
where \( \alpha_i = 1 - \exp(-\sigma_i \Delta t_i) \) is the opacity of the \( i \)-th sample, and \( T_i \) is the accumulated transmittance up to the \( i \)-th sample.

\textbf{Training Objective:} NeRF is trained to minimize the difference between the rendered pixel colors and the ground truth image:
\begin{equation}
\mathcal{L} = \frac{1}{P} \sum_{p=1}^P \|C_{\text{rendered}}(p) - C_{\text{target}}(p)\|^2,
\end{equation}
where:
\begin{itemize}
    \item \( C_{\text{rendered}}(p) \): Rendered pixel color for pixel \( p \).
    \item \( C_{\text{target}}(p) \): Ground truth color for pixel \( p \).
    \item \( P \): Total number of pixels in the image.
\end{itemize}

To improve efficiency and fidelity, NeRF uses a two-pass hierarchical sampling strategy:
\begin{itemize}
    \item \textbf{Coarse Model:} A coarse network samples the scene uniformly along each ray.
    \item \textbf{Fine Model:} A fine network focuses on regions with higher density, guided by the output of the coarse model.
\end{itemize}

The combined loss for both models is:
\begin{equation}
\mathcal{L}_{\text{total}} = \mathcal{L}_{\text{coarse}} + \mathcal{L}_{\text{fine}}.
\end{equation}

The neural network parameters \( \theta \) are optimized using gradient descent, with gradients computed through the differentiable volume rendering process. This end-to-end optimization allows NeRF to learn both color and density distributions that reconstruct the input views accurately.

\textbf{Inference:} Once trained, NeRF can synthesize novel views of a 3D scene from arbitrary camera poses. At inference time, the network is queried to render a new image by casting rays through each pixel of the target view. For each ray \( \mathbf{r}(t) = \mathbf{o} + t\mathbf{d} \), NeRF samples a set of 3D points \( \{\mathbf{x}_i = \mathbf{r}(t_i)\}_{i=1}^{N} \) along the ray and queries the MLP to obtain the corresponding densities \( \sigma_i \) and colors \( \mathbf{c}_i \):
\begin{equation}
(\mathbf{c}_i, \sigma_i) = F_\theta(\mathbf{x}_i, \mathbf{d}).
\end{equation}

These values are then aggregated using volumetric rendering, where the final pixel color is computed via the weighted sum:
\begin{equation}
C(\mathbf{r}) \approx \sum_{i=1}^{N} T_i \alpha_i \mathbf{c}_i,
\end{equation}
with opacity \( \alpha_i = 1 - \exp(-\sigma_i \Delta t_i) \) and accumulated transmittance
\begin{equation}
T_i = \exp\left(-\sum_{j=1}^{i-1} \sigma_j \Delta t_j \right).
\end{equation}

This process is repeated for each pixel in the target image, yielding a photorealistic rendering from the desired viewpoint. The rendering procedure is entirely differentiable and does not require any retraining or finetuning at inference; it relies solely on evaluating the learned MLP \( F_\theta \) across sampled points along camera rays.

\begin{tcolorbox}[colback=gray!5!white, colframe=black!50!gray, parbox=false]
\textbf{Key Takeaways:}
NeRF can capture the physical characteristics of an object, ~\cite{zhang2021nerfactor}, and can be used to generate realistic images of objects in different lighting conditions~\cite{xu2023renerf}. However, NeRF has the limitation of assuming the scene is static, unless extended with techniques like dynamic NeRF \cite{pumarola2021d} and deformable NeRF~\cite{park2020nerfies}. In addition, NeRF's reliance on viewing direction for color prediction can lead to artifacts when extrapolating to unseen viewpoints~\cite{yang2023freenerf}. Furthermore, compared to 3DGS, training and rendering of NeRF require significantly more computational resources due to its high-dimensional sampling and integration process.
\end{tcolorbox}


\subsection{3D Gaussian Splatting (3DGS)}


Unlike NeRF, which describes the scene with density and color volume, 3D Gaussian Splatting (3DGS) describes the scene using 3D Gaussians with attributes. 3DGS \cite{kerbl20233dgaussian} is a novel approach for generating and representing scenes by utilizing Gaussian primitives in a 3D space. Unlike traditional methods that rely on mesh-based representations or dense voxel grids, Gaussian splatting represents scenes as a set of parameterized 3D Gaussian distributions, enabling efficient and continuous scene generation. This method has gained traction due to its ability to balance fidelity and computational efficiency, making it a promising technique for applications like autonomous driving scene generation. A core difference between \emph{3D Gaussian Splatting (3DGS)} \cite{kerbl20233dgaussian} and neural-field methods like NeRF or generative autoencoders is \emph{how} the scene representation is parameterized and optimized. In many neural-field approaches (e.g., NeRF), the scene is encoded by a \emph{neural network} (often an MLP) that outputs color and density given continuous 3D coordinates. In contrast, \emph{Gaussian splatting} replaces the neural network with a \emph{direct} collection of 3D Gaussian primitives and \emph{optimizes their parameters explicitly}. 

\textbf{Components:} Gaussian splatting represents a scene as a collection of 3D Gaussians, each defined by its position, orientation, covariance matrix, and color/intensity parameters:

\begin{equation}
G_i = \{\mu_i, \Sigma_i, c_i\}, \quad i = 1, \dots, M,
\end{equation}
where:
\begin{itemize}
    \item \( \mu_i \in \mathbb{R}^3 \): Center of the Gaussian.
    \item \( \Sigma_i \in \mathbb{R}^{3 \times 3} \): Covariance matrix defining shape and orientation.
    \item \( c_i \in \mathbb{R}^3 \): Color attributes (e.g., RGB values).
\end{itemize}

The rendering process projects the 3D Gaussians into 2D image space and blends their contributions using splatting. The pixel value at position \( p \) is computed as:
\begin{equation}
I_{\text{rendered}}(p) = \sum_{i=1}^M w_i(p) c_i,
\end{equation}
where:
\begin{itemize}
    \item \( w_i(p) \): Weight of Gaussian \( i \) at pixel \( p \), computed from the projected covariance and intensity.
    \item \( c_i \): Color or intensity of Gaussian \( i \).
\end{itemize}

\textbf{Training Objective:} The key to training is that the \emph{rendering function} (the splatting process) is made \emph{differentiable}, so gradients flow back from the rendered image to each Gaussian’s parameters. Although this backpropagation is conceptually similar to training a neural network, there is \emph{no multi-layer perceptron} or CNN in the pipeline. Instead, each Gaussian’s position, covariance, and color are directly updated based on the reconstruction and regularization losses. Intuitively, each Gaussian is effectively a ``mini'' radial basis with its own shape. Hence, \emph{the entire scene} is a set of parameters $ \{\mu_i,\Sigma_i,c_i\}$ rather than a neural network function.

The total training loss is a combination of reconstruction, sparsity, and regularization terms:
\begin{equation}
\mathcal{L} = \mathcal{L}_{\text{recon}} + \lambda_{\text{sparsity}} \mathcal{L}_{\text{sparsity}} + \lambda_{\text{shape}} \mathcal{L}_{\text{shape}},
\end{equation}
where the \textbf{reconstruction Loss}:
    \begin{equation}
    \mathcal{L}_{\text{recon}} = \frac{1}{N} \sum_{p} \|I_{\text{rendered}}(p) - I_{\text{target}}(p)\|^2,
    \end{equation}
minimizes the pixel-wise difference between the rendered image \( I_{\text{rendered}} \) and the target image \( I_{\text{target}} \).

The \textbf{Sparsity Regularization}:
    \begin{equation}
    \mathcal{L}_{\text{sparsity}} = \sum_{i=1}^M \|w_i\|,
    \end{equation}
penalizes redundant or unnecessary Gaussians to promote efficiency.

The \textbf{Shape Regularization}:
    \begin{equation}
    \mathcal{L}_{\text{shape}} = \sum_{i=1}^M \|\Sigma_i - \Sigma_{\text{target}}\|^2,
    \end{equation}
ensures stable and well-formed Gaussian shapes.

\textbf{Inference:} Once trained, 3D Gaussian Splatting can render novel views by projecting the learned 3D Gaussian primitives into a new camera frame. Given a desired viewpoint, each Gaussian \( G_i = \{\mu_i, \Sigma_i, c_i\} \) is transformed into image space based on the camera's intrinsic and extrinsic parameters. The projection of each Gaussian onto the 2D image plane is computed by warping its 3D mean \( \mu_i \) and covariance \( \Sigma_i \) into the screen space. Each projected Gaussian contributes a weighted color \( w_i(p)c_i \) to the surrounding pixels, where \( w_i(p) \) is computed based on the 2D projected shape, depth ordering, and visibility. The final color at each pixel is obtained by alpha compositing the weighted contributions of all visible Gaussians:
\begin{equation}
I_{\text{rendered}}(p) = \sum_{i=1}^M w_i(p) c_i.
\end{equation}

\begin{tcolorbox}[colback=gray!5!white, colframe=black!50!gray, parbox=false]
\textbf{Key Takeaways:}
This rasterization-based approach enables real-time rendering, as it avoids the need for volumetric sampling and ray integration and allows for efficient parallelization of the rendering process. Unlike NeRFs, which must evaluate a radiance field (might be trained from a neural network or tensor factorization) at many sampled points per ray, 3DGS directly renders from its optimized primitive set without neural network inference. Thus, 3DGS achieves significantly faster rendering while preserving visual fidelity, making it highly suitable for downstream applications such as real-time simulation or view synthesis in autonomous driving scenarios.

The disadvantages of 3DGS lie in the difficulty of parameter tuning and limited fidelity: optimizing the parameters of each Gaussian (e.g., covariance and intensity) might be challenging, particularly for complex scenes with many objects; representing fine details, such as textures or sharp edges, may require a large number of Gaussians, reducing efficiency. In addition, while 3DGS is more efficient than dense representations, rendering a large number of Gaussians can still be computationally expensive.
\end{tcolorbox}


\subsection{Skinned Multi-Person Linear (SMPL) Model}
\label{sec:smpl} 

Crucial for realistic pedestrian modeling in autonomous driving perception and simulation, the Skinned Multi-Person Linear (SMPL) model~\cite{loper2015smpl} is a parametric representation of the human body that integrates a template mesh with linear blend skinning (LBS) to control body shape and pose articulation. Central to SMPL is a template mesh $M_h = (V, F)$, defined in a canonical rest pose, which consists of $n_v$ vertices $V \in \mathbb{R}^{n_v \times 3}$ and faces $F$. This template mesh can be deformed according to shape parameters $\beta$ and pose parameters $\theta$. The vertex locations in the shaped, rest-posed space, denoted as $V_S$, are computed via $V_S = V + B_S(\beta) + B_P(\theta)$. Here, $B_S(\beta) \in \mathbb{R}^{n_v \times 3}$ and $B_P(\theta) \in \mathbb{R}^{n_v \times 3}$ are functions representing the vertex offsets induced by shape and pose blend shapes, respectively. 

To transform the vertices $V_S$ into the target pose configuration defined by $\theta$, SMPL utilizes LBS. The final position $\mathbf{v}'_i$ of each vertex $\mathbf{v}_{S,i}$ (the $i$-th vertex in $V_S$) is determined using pre-defined LBS weights $W \in \mathbb{R}^{n_k \times n_v}$ and joint transformation matrices $G_k$:
$$ \mathbf{v}'_i = \sum_{k=1}^{n_k} W_{k,i} G_k \mathbf{v}_{S,i} $$
In this equation, $n_k$ represents the number of joints, $W_{k,i}$ denotes the skinning weight associating vertex $i$ with joint $k$, and $G_k \in SE(3)$ is the rigid transformation for joint $k$. These joint transformations $G_k$ are derived from the pose parameters $\theta$ and, implicitly, the shape parameters $\beta$, as $\beta$ influences the initial joint locations. The pose parameters $\theta$ typically comprise a body pose component $\theta_b \in \mathbb{R}^{23 \times 3 \times 3}$ (representing rotation matrices for 23 body joints) and a global orientation component $\theta_g \in \mathbb{R}^{3 \times 3}$. The shape variations are controlled by shape parameters $\beta \in \mathbb{R}^{10}$. For a more detailed exposition of the SMPL model, readers are directed to the original publication by Loper et al.~\cite{loper2015smpl}. For non-rigid actors such as pedestrians, one can incorporate the SMPL model to enable joint-level control using dynamic Gaussians~\cite{chen2024omnire}; SMPL provides both a prior template geometry for 3D Gaussian Splatting (3DGS) initialization and explicit control for modeling desired human behaviors, which is advantageous for downstream simulation applications.

\begin{tcolorbox}[colback=gray!5!white, colframe=black!50!gray, parbox=false]
\textbf{Key Takeaways:}
The SMPL model provides a widely-used parametric framework for representing human bodies, controllably deforming a template mesh using shape ($\beta$) and pose ($\theta$) parameters via LBS. Its explicit parametrization allows for realistic articulation and body shape variations, making it fundamental for human modeling tasks. Within autonomous driving, SMPL is crucial for generating diverse and controllable pedestrian simulations, understanding human behavior, and initializing dynamic 3D representations like Gaussian Splatting.
\end{tcolorbox}

\subsection{Autoregressive Models and Language Models}

Autoregressive Models refer to the models in which the output is formulated as a sequence, and the same trained model is run repeatedly to successively generate the new element in the sequence, conditioned on all previous elements. These models represent a foundational family of generative approaches that decompose high-dimensional data generation into a sequence of conditional predictions. 

For example, in the image generation domain, for an image $x = [x_1, x_2, \dots, x_N]$ consisting of $N$ pixels (or tokens), an autoregressive model defines the joint probability as:
\begin{equation}
    P(x) = \prod_{i=1}^{N} P(x_i | x_{<i}),
\end{equation}
where each token $x_i$ is generated based on all previously generated tokens $x_{<i}$. This property allows autoregressive models to capture complex spatial and temporal dependencies in data, making them suitable for generating coherent and high-fidelity sequences of pixels, frames, or trajectory elements.

At inference time, autoregressive models operate by sampling tokens sequentially. For image or video generation, a token is sampled from $P(x_i | x_{<i})$, appended to the context, and used to generate the next token until the full image or sequence is complete. Though slow due to their sequential nature, techniques such as parallel sampling, caching of key/value states, and token grouping can significantly accelerate inference. Below, we note the popular autoregressive generative models.

\textbf{PixelCNN and PixelRNN:} Early autoregressive models for image generation include PixelCNN~\cite{van2016conditional} and PixelRNN~\cite{van2016pixel}, which model pixel intensities one at a time using convolutional or recurrent networks. While effective at capturing local pixel dependencies, these models suffer from slow inference due to their strictly sequential nature and limited receptive field in deeper layers.

\textbf{Transformer-based Models:} Inspired by the success of language modeling, autoregressive Transformers such as GPT-like architectures have been adopted for vision tasks. Taming Transformers~\cite{haghighi2024taming} leverages a VQ-VAE encoder to discretize images into a grid of tokens, followed by an autoregressive transformer trained to predict the next token in the sequence. This decoupling of encoding and generation improves scalability and enables high-resolution image synthesis with better global coherence.

\textbf{Magvit and Magvit-v2:} Magvit~\cite{yu2023magvit} and its successor Magvit-v2~\cite{yu2023language} extend the autoregressive transformer paradigm to video generation. These models encode video frames into discrete tokens using a multi-scale VQ-VAE, and generate future tokens across space and time via 3D transformers. This allows them to model complex spatiotemporal dependencies in a fully autoregressive manner, enabling video prediction, completion, and interpolation tasks—essential for simulating autonomous driving scenes.

\textbf{VideoPoet:} VideoPoet~\cite{kondratyuk2023videopoet} is a generalist autoregressive model trained on a mixture of vision, audio, and text modalities. It uses a unified tokenization scheme and autoregressive transformer decoder to generate outputs conditioned on multimodal inputs. This generalist nature makes it applicable for closed-loop video simulation, planning, and multimodal driving assistance tasks, aligning with recent advances in MLLMs.

\textbf{Vector Autoregressive Transformers (VAR):} VAR~\cite{tian2024visual} proposes a token-based approach where latent feature vectors of driving scenes are modeled as sequentially dependent entities. Each latent token is generated conditioned on previously sampled tokens, enabling fine-grained generation of structured driving environments. This formulation allows trajectory or map elements to be autoregressively modeled, with applications in closed-loop policy learning and scene simulation.


\paragraph{Large Language Model (LLM)}

Apart from the above, the most popular type of autoregressive model is the Large Language Model (LLM). These models are based on the Transformer architecture \cite{vaswani2017attention} trained on vast corpora of text to model the probability distribution of natural language. These models, such as GPT \cite{gpt}, BERT \cite{devlin2018bert}, PaLM \cite{driess2023palm}, and LLaMA \cite{touvron2023llama}, have demonstrated remarkable performance on a wide range of language tasks, including question answering, summarization, translation, reasoning, and dialogue. The core idea behind LLMs is to learn contextual representations of text through self-attention mechanisms, allowing the model to generate coherent and contextually relevant outputs by predicting each token conditioned on previous tokens. Modern LLMs are often trained with hundreds of billions of parameters and utilize techniques such as masked language modeling (BERT) or causal language modeling (GPT-style) to capture syntactic and semantic patterns. 

\textbf{Components:} Large Language Models (LLMs) are typically composed of the following core components built on the Transformer architecture \cite{vaswani2017attention}:
\begin{itemize}
    \item \textbf{Tokenizer:} Converts raw input text into a sequence of discrete tokens, typically using subword tokenization methods such as Byte Pair Encoding (BPE) \cite{sennrich2015neural} or SentencePiece \cite{kudo2018sentencepiece}.
    \item \textbf{Embedding Layer:} Maps tokens into dense vector representations, forming the input to the model. Positional encodings are added to retain sequence order information.
    \item \textbf{Transformer Blocks:} A stack of multi-head self-attention layers and feed-forward neural networks. Each layer computes contextualized token representations by attending to all positions in the sequence.
    \item \textbf{Output Head:} A linear projection followed by a softmax function that outputs a probability distribution over the vocabulary for the next token prediction.
\end{itemize}

LLMs may also include additional architectural features such as layer normalization, residual connections, rotary embeddings \cite{su2024roformer}, or parallel attention/MLP layers in more recent designs (e.g., GPT-4 or LLaMA-2). These components enable LLMs to scale effectively and capture long-range dependencies within text, making them versatile tools for a wide array of natural language processing tasks.

\textbf{Training Objective:} LLMs are typically trained using an autoregressive language modeling objective, where the model learns to predict the next token in a sequence given all previous tokens. Formally, given a sequence of tokens $ x = (x_1, x_2, \dots, x_T) $, the training loss is defined as the negative log-likelihood:
\begin{equation}
    \mathcal{L} = -\sum_{t=1}^T \log P(x_t | x_{<t}).
\end{equation}
During training, \textit{teacher forcing} is employed: the ground-truth tokens $ x_{<t} $ are used as inputs to predict $ x_t $, rather than the model’s own previous predictions. This technique stabilizes training and accelerates convergence by preventing error accumulation during sequence generation. In contrast, during inference, models generate tokens autoregressively using their own previous outputs as inputs. Alternative objectives—such as masked language modeling (e.g., BERT~\cite{devlin2018bert}) or instruction tuning—have been used in other pretraining paradigms, but most generative LLMs like GPT-3~\cite{brown2020language} and LLaMA~\cite{touvron2023llama} rely on teacher-forced autoregressive training.

\textbf{Inference:} At inference time, LLMs generate text by sampling tokens autoregressively, \textit{i.e.}, one token at a time.

\begin{tcolorbox}[colback=gray!5!white, colframe=black!50!gray, parbox=false]
\textbf{Key Takeaways:}
Despite their impressive capabilities, LLMs face several significant challenges. First, they are highly \emph{data-hungry} and require massive corpora of high-quality, curated text to achieve competitive performance. Second, their \emph{computational demands}, both during training and inference, pose scalability and deployment concerns, particularly for real-time or resource-constrained applications. Third, LLMs suffer from \emph{hallucination}, \textit{i.e.} the tendency to generate fluent but factually incorrect or misleading content, undermining their reliability in safety-critical domains. Fourth, they lack explicit \emph{grounding} in external sensory input or real-world context, which can limit their reasoning in embodied settings. Lastly, aligning their outputs with human values and avoiding harmful or biased generations remains an open problem, necessitating ongoing research in \emph{alignment}, \emph{robustness}, and \emph{interpretability}.
\end{tcolorbox}

\paragraph{Multimodal Large Language Models (MLLM)}

MLLM is an extension of traditional Large Language Model (LLM) that are built with transformers. By incorporating inputs and outputs from multiple modalities, such as text, images, video, and audio, MLLM enables a richer understanding of the world and supports a wider range of applications, including autonomous driving, robotics, and creative tasks.

\textbf{Components:} The architecture of MLLMs builds upon the foundation of Transformer-based LLMs, augmented to handle multiple modalities. This is achieved by introducing modality-specific encoders and fusion strategies.

\textbf{Modality-Specific Encoders:} Separate encoders are used to preprocess inputs from different modalities:
    \begin{itemize}
        \item \textbf{Text:} Standard LLM encoders like GPT \cite{gpt} or BERT \cite{devlin2018bert} to process textual input.
        \item \textbf{Images:} Convolutional Neural Networks (CNNs) or Vision Transformers (ViT) to encode visual features.
        \item \textbf{Audio:} Spectrogram-based encoders or WaveNet-like \cite{oord2016wavenet} architectures to process audio signals.
    \end{itemize}

\textbf{Fusion Strategies:} Merge multimodal embeddings through:
\begin{itemize}
    \item \textbf{Latent Concatenation:} Direct combination of modality-specific embeddings.
    \item \textbf{Cross-Attention:} Dynamic alignment using transformer layers:
    \begin{equation}
        \text{Attention}(Q, K, V) = \text{Softmax}\left(\frac{QK^T}{\sqrt{d_k}}\right)V,
    \end{equation}
    where \( Q \) (queries) and \( K, V \) (keys/values) originate from different modalities.
\end{itemize}

For some MLLMs, all modalities are converted into a unified token space, allowing a single Transformer model to process multimodal inputs. Examples include encoding images into discrete visual tokens using Vector Quantization.

\textbf{Training Objective:} The training of MLLMs involves tasks that align multimodal inputs and outputs. Common objectives include:
\begin{itemize}
    \item \textbf{Cross-Modal Contrastive Learning:} Align paired data (e.g., image-text) via similarity maximization:
    \begin{equation}
        \mathcal{L}_{\text{align}} = -\log \frac{\exp(\text{sim}(z_{\text{text}}, z_{\text{image}}))}{\sum_{i,j} \exp(\text{sim}(z_{\text{text}}^i, z_{\text{image}}^j))},
    \end{equation}
    where \( \text{sim}(\cdot, \cdot) \) is a similarity metric such as cosine similarity.
    \item \textbf{Cross-Modal Joint Generation:} Condition outputs across modalities (e.g., text generation from images):
    \begin{equation}
        \mathcal{L}_{\text{gen}} = -\sum_t \log P(x_t | x_{<t}, z_{\text{image}}).
    \end{equation}
    \item \textbf{Contrastive Learning:} Learning modality-invariant embeddings via contrastive loss.
\end{itemize}

\textbf{Inference:} At inference time, Multimodal Large Language Models (MLLMs) operate by processing inputs from one or more modalities—such as text, images, or videos—through their respective modality-specific encoders. These encoded features are then fused into a shared representation using either latent concatenation or cross-attention mechanisms. In models like BLIP-2\cite{li2023blip2}, LLaVA\cite{liu2024llava}, or Flamingo\cite{alayrac2022flamingo}, the visual features (e.g., from ViT or CLIP) are injected as contextual tokens into the language model, enabling visual-conditioned language generation. For example, given an image and a prompt like ``Describe the traffic situation,'' the image is encoded into a feature embedding \( z_{\text{image}} \), which is combined with the tokenized prompt \( x_{\text{text}} \), and the model autoregressively generates a response using the conditional probability:
\[
P(x_t | x_{<t}, z_{\text{image}}).
\]
This allows MLLMs to reason over complex visual scenes and produce informative textual outputs. Similarly, in vision-and-action models such as DriveVLM~\cite{tian2024drivevlm} or LMDrive~\cite{shao2024lmdrive}, the MLLM processes camera inputs, route descriptions, and past trajectories to predict future actions (e.g., trajectory waypoints or control commands), often formatted as a sequence of output tokens. Depending on the application, the output of the MLLM can be natural language (for explanation or VQA), structured tokens (e.g., waypoints), or control signals. In summary, inference with MLLMs involves modality encoding, fusion, and conditional decoding—all handled within a unified transformer architecture, enabling generalizable reasoning across diverse autonomous driving scenarios.

\begin{tcolorbox}[colback=gray!5!white, colframe=black!50!gray, parbox=false]
\textbf{Key Takeaways:}
While MLLMs are known for their adaptability and simplicity, just like LLMs, MLLMs have the drawbacks of high computational demands, a huge need for training data, difficulty of alignment, and slow inference speed.
\end{tcolorbox}



\clearpage

\begin{figure}[!]
    \centering
    \includegraphics[width=0.85\linewidth]{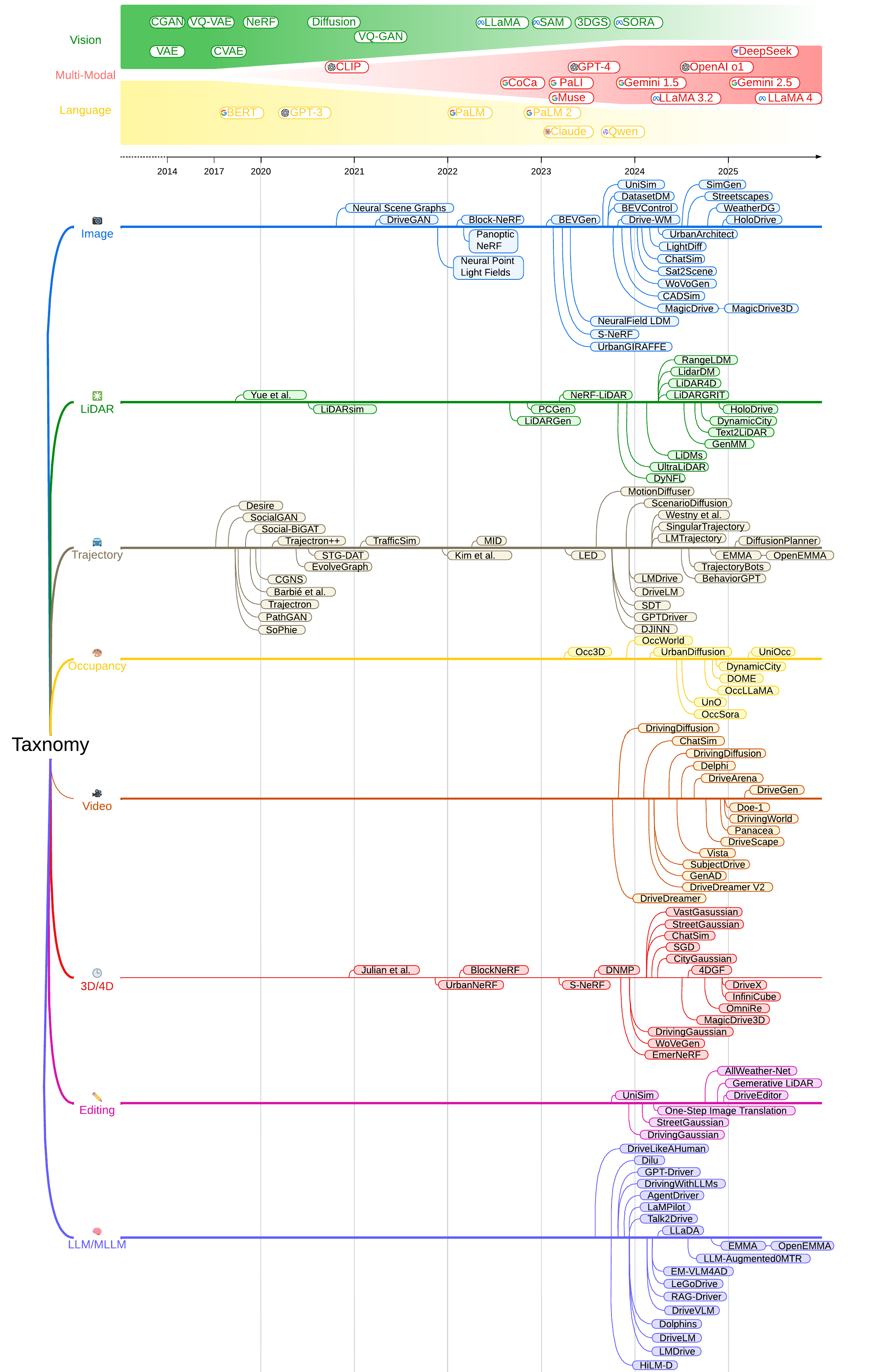}
    \caption{Emergence of various generative AI methods and the evolution of generative AI in autonomous driving.}
    \label{fig:generative_timeline}
\end{figure}

\clearpage

\section{Frontiers of Generative AI Models for Autonomous Driving}
\label{sec:generation_methods}


This section delves deep into the autonomous driving research that utilizes generative AI. \mpv{We divide the sections by the task modality: image in Section~\ref{sec:image_gen}, LiDAR in Section~\ref{sec:lidar_gen}, trajectory in Section~\ref{sec:traj_gen}, occupancy in in Section~\ref{sec:occupancy_gen}, video in Section~\ref{sec:video_gen}, editing in Section~\ref{sec:editing}, large language models in Section~\ref{sec:llm_gen} and multimodal large language models in Section~\ref{sec:mllm_gen}.} In each subsection, we cover the works in the recent years till 2025. We discuss their commonalities and differences, and highlight the challenges in each modality. \mpv{In Section \ref{sec:applications}, we'll further examine how these methods can be used in practice.}. We also include a table with hyperlinks to their code repository (if available). We list a timeline of these works in Figure \ref{fig:generative_timeline}.



\subsection{Image Generation}
\label{sec:image_gen}

Image generation is a significant branch of generative AI. Methods prominent in general image generation, such as NeRF, diffusion models, GANs, and VAEs, remain influential and have achieved remarkable accomplishments. However, the specific field of image generation for autonomous driving presents unique requirements. First, autonomous driving evaluation imposes higher demands on the safety assessment of self-driving algorithms. This necessitates that the generated images are not only photorealistic but also physically plausible. Second, driving scenarios involve multiple heterogeneous participants and are hierarchically complex, comprising static backgrounds and dynamic agents. Third, real driving scenarios exhibit a long-tailed distribution, where safety-critical scenarios are rare minorities. These specific challenges require image generation algorithms to incorporate unique designs in their pipelines.

Building on this, we will explore how existing methods address these challenges through specific modifications or extensions to general image generation approaches. These approaches can be broadly categorized into three levels, focusing on:  
1) \textbf{Controllable generation}: How different works leverage contextual information to inform the generation process, achieving diversified and physically-grounded outputs.  
2) \textbf{Decompositional generation}: How scenarios are generated in a decompositional manner to handle structurally complex driving scenarios. 

\begin{table}[h!]
\centering
\caption{Image Generation Methods. The upper part is controllable generation, and the lower part is decompositional generation. We also report the Fréchet Inception Distance (FID) whenever available.}
\label{table:imagegeneration_extended}
\resizebox{\textwidth}{!}{
\begin{tabular}{lllllll}
\toprule
\textbf{Method} & \textbf{Venue} & \textbf{Dataset (FID$\downarrow$)} & \textbf{Modeling Type} & \textbf{Backbone} & \textbf{Control Variables} & \textbf{Code} \\
\midrule
BEVGen \cite{swerdlow2024street} & IEEE RA-L'24 & Argoverse 2, nuScenes (25.5)  & VQ-VAE & Transformer & BEV Map, Object Box, Text & \href{https://github.com/alexanderswerdlow/BEVGen}{GitHub} \\
BEVControl \cite{yang2023bevcontrol} & arXiv'23 & nuScenes (24.9) & VAE & CNN, Transformer, CLIP & BEV Sketch, Text  & N/A \\
MagicDrive \cite{gao2023magicdrive} & ICLR'24 & nuScenes (16.2) & Diffusion, VAE & U-Net & Road Map, Object Box, Camera Pose & \href{https://github.com/cure-lab/MagicDrive}{GitHub} \\
Drive-WM \cite{wang2024driving} & CVPR'24 & nuScenes (15.8) & Diffusion, VAE & U-Net & Map, Text  & \href{https://github.com/BraveGroup/Drive-WM}{GitHub} \\
SimGen \cite{zhou2024simgen} & NeurIPS'24 & nuScenes (15.6) & Diffusion, SDEdit & U-Net & BEV, Text  & \href{https://github.com/metadriverse/SimGen}{GitHub} \\
DatasetDM \cite{wu2023datasetdm} & NeurIPS'23 & - (all models are pretrained) & Diffusion, LLM, VAE & U-Net, ControlNet & Text  & \href{https://github.com/showlab/DatasetDM}{GitHub} \\
DriveGAN \cite{kim2021drivegan} & CVPR'21 & RWD & GAN, VAE & CNN, LSTM, MLP & Steering, Speed, Scene Features  & \href{https://github.com/nv-tlabs/DriveGAN_code}{GitHub} \\
LightDiff \cite{li2024light} & CVPR'24 & nuScenes & VAE, Diffusion & U-Net & Lighting conditions  & \href{https://github.com/jinlong17/LightDiff}{GitHub} \\
Streetscapes \cite{deng2024streetscapes} & SIGGRAPH'24 & Google Street View (35.5) & Diffusion  & ControlNet & Road Map, Height Map, Camera Pose  & N/A \\
Wovogen \cite{lu2024wovogen} & ECCV'24 & nuScenes (27.6) & Diffusion, AutoEncoder & CNN, CLIP & Text, World Volumes, Ego Actions  & \href{https://github.com/fudan-zvg/WoVoGen}{GitHub} \\
\midrule
ChatSim \cite{wei2024editable} & CVPR'24 & Waymo Open Dataset & LLM, NeRF & MLP, Transformer & 3D Assets  & \href{https://github.com/yifanlu0227/ChatSim}{GitHub} \\
UrbanGIRAFFE \cite{yang2023urbangiraffe} & ICCV'23 & KITTI-360 (36.9), CLEVR-W (25.7) & NeRF & MLP & Camera Pose, Panoptic Prior  & \href{https://github.com/freemty/UrbanGIRAFFE}{GitHub} \\
Sat2Scene \cite{li2024sat2scene} & CVPR'24 & HoliCity (72.0) & NeRF & MLP & Satellite Images, Layout, 3D Constraints  & \href{https://github.com/lizuoyue/sat2scene}{GitHub} \\
Block-NeRF \cite{tancik2022block} & CVPR'22 & Block-NeRF Dataset & NeRF & MLP & Spatial Block Layout, 3D Constraints  & \href{https://github.com/freemty/UrbanGIRAFFE}{GitHub}\\
S-NeRF \cite{xie2023s} & CVPR'23 & nuScenes, Waymo Open Dataset & NeRF & MLP & Camera Path, 3D Constraints  & \href{https://github.com/fudan-zvg/S-NeRF}{GitHub} \\
NF-LDM \cite{kim2023neuralfield} & CVPR'23 & VizDoom, Replica, AVD & Diffusion, NeRF & MLP & Scene Embedding, 3D Constraints  & N/A \\
Panoptic NeRF \cite{kundu2022panoptic} & IEEE 3DIMPVT'22 & KITTI 360 & NeRF & MLP & Semantic Segmentation, 3D Constraints  & \href{https://github.com/fuxiao0719/panopticnerf}{GitHub} \\
Neural Point Light Field & CVPR'22 & Waymo Open Dataset & NeRF & MLP & Camera Pose, 3D Constraints  & \href{https://github.com/princeton-computational-imaging/neural-point-light-fields}{GitHub} \\
Neural Scene Graphs \cite{ost2021neural} & CVPR'21 & KITTI & NeRF & MLP & Object Graph Topology, 3D Constraints  & \href{https://github.com/princeton-computational-imaging/neural-scene-graphs}{GitHub} \\
UniSim \cite{yang2023unisim} & CVPR'23 & PandaSet & NeRF & MLP & Agent Profile, 3D Constraints  & N/A \\
CADSim \cite{wang2023cadsim} & CoRL'23 & MVMC, PandaSet & Differentiable CAD Rendering & MLP & CAD Geometry, 3D Constraints & N/A \\
\bottomrule
\end{tabular}
}
\end{table}

\paragraph{Controllable Generation} Controllable generation utilizes various domain input (\textit{e.g.}, layout) to enable conditional and context-aware generation, with the contextual elements modified in a controllable way. A critical and widely utilized contextual element is \emph{layout} information. BEVGen \cite{swerdlow2024street} utilizes Bird's-Eye View (BEV) layouts to generate realistic street-view images using a novel autoregressive framework with cross-view spatial attention. The model incorporates two autoencoders: one to encode the BEV layout into a discrete latent representation and another to process multi-view image tokens. These representations are then fed into a transformer with positional embeddings informed by camera intrinsics. BEVControl \cite{yang2023bevcontrol} improves upon BEVGen by introducing sketch-based inputs, enabling more flexible and precise editing of both foreground and background elements. Unlike BEVGen’s reliance on segmentation layouts, BEVControl uses a two-stage approach with a controller for geometry consistency and a Coordinator for appearance alignment. DrivingDiffusion \cite{li2025drivingdiffusion} utilizes 3D layouts to guide multi-view driving scene video generation with a latent diffusion model. By leveraging cross-view and temporal consistency modules, it ensures coherent spatial and temporal alignment across multi-camera views and video frames. MagicDrive \cite{gao2023magicdrive} and MagicDrive3D \cite{gao2024magicdrive3d} utilize  BEV information, including road maps and object boxes, for multi-view generation.

To enrich the data environment, \emph{text}-based conditions describing the target scene are widely used to guide the generation. For instance, DatasetDM \cite{wu2023datasetdm} leverages a text encoder to incorporate scene descriptions, while ChatSim \cite{wei2024editable} employs an LLM agent to inform the generation process. Among all text-based conditions, \emph{weather} conditions remain one of the most widely utilized contextual factors. MagicDrive3D \cite{gao2024magicdrive3d} and Drive-WM \cite{wang2024driving} both incorporate text-based weather inputs. In WeatherDG \cite{qian2024weatherdg}, LLMs are used to generate weather prompts based on weather hints, aiming to utilize their domain knowledge to enrich the details of weather-based prompts and guide a diffusion model toward the targeted scene.

Another widely used condition is \emph{urban architecture}. Streetscapes \cite{deng2024streetscapes} relies on satellite images to inform streetview generation, while SimGen \cite{zhou2024simgen} leverages road network architecture for the same purpose.

\begin{figure}[h!]
    \centering
    \includegraphics[width=0.8\textwidth]{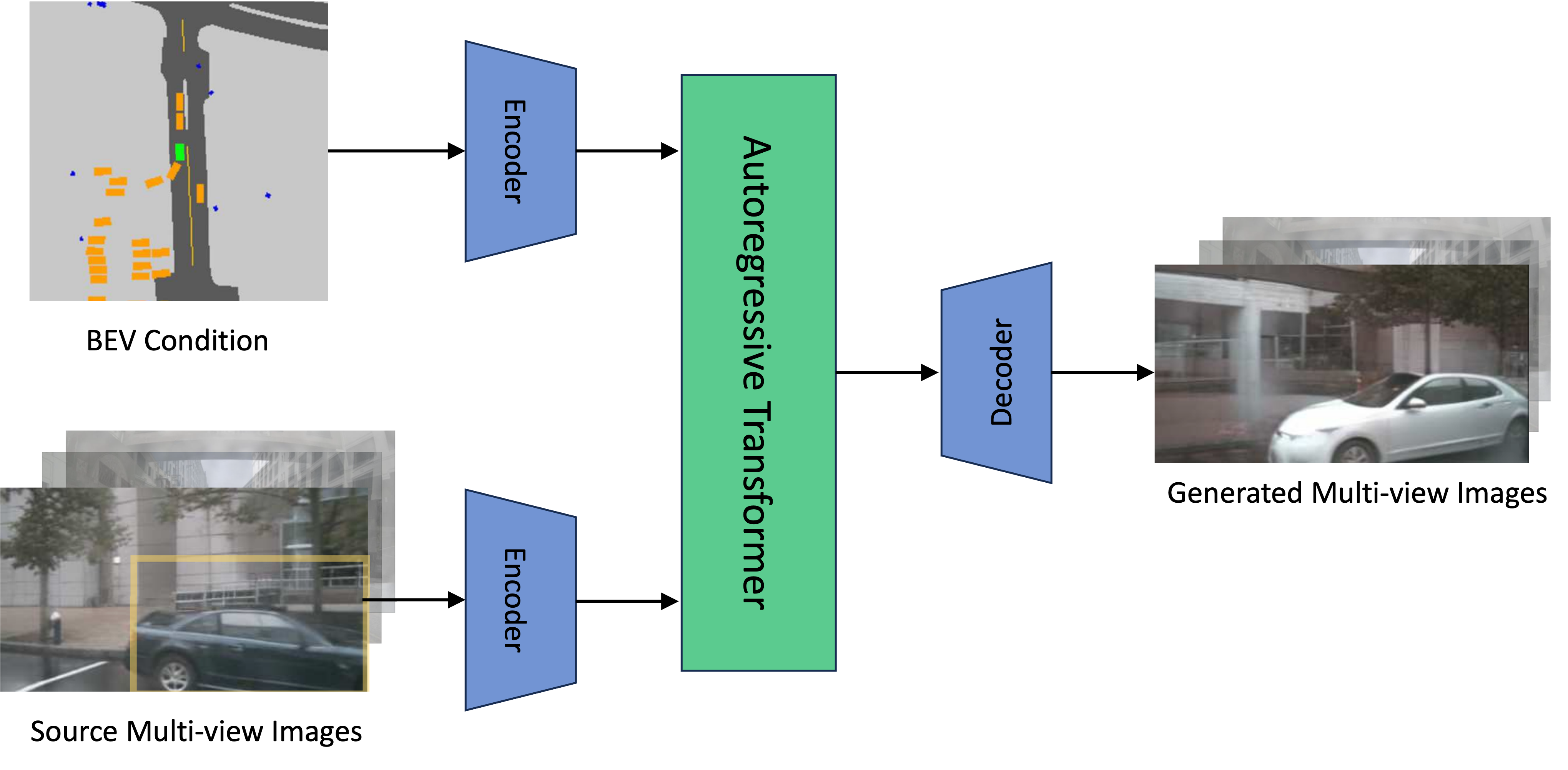}
    \caption{Illustration of controllable generation where BEV layout serves as the condition. The image is from \cite{swerdlow2024street}.}
    \label{fig:conditionGen}
\end{figure}

\begin{figure}[h!]
    \centering
    \subfigure{
        \includegraphics[width=0.8\textwidth]{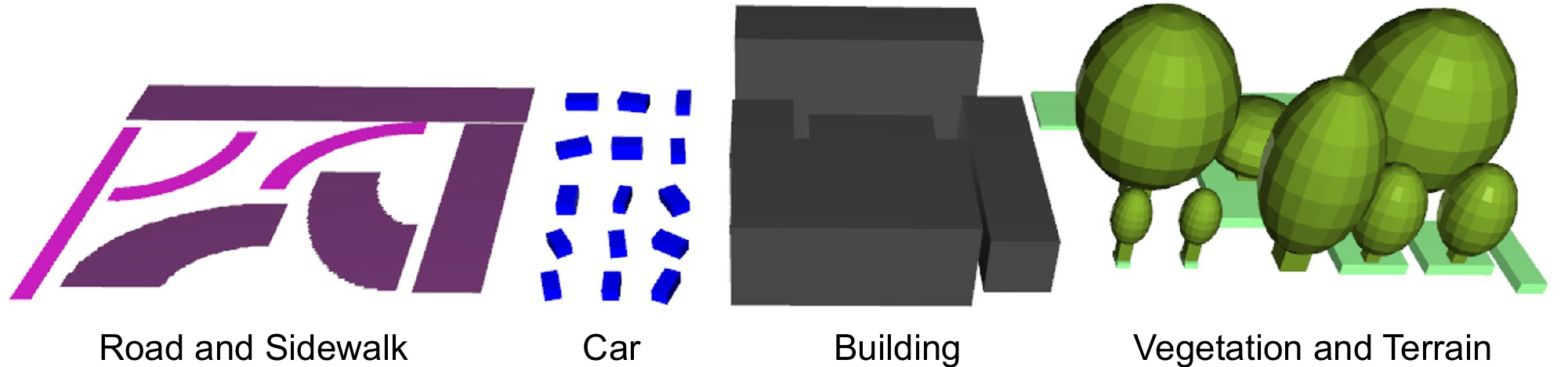} 
        \label{fig:urban_first}
    }
    \subfigure{
        \includegraphics[width=0.8\textwidth]{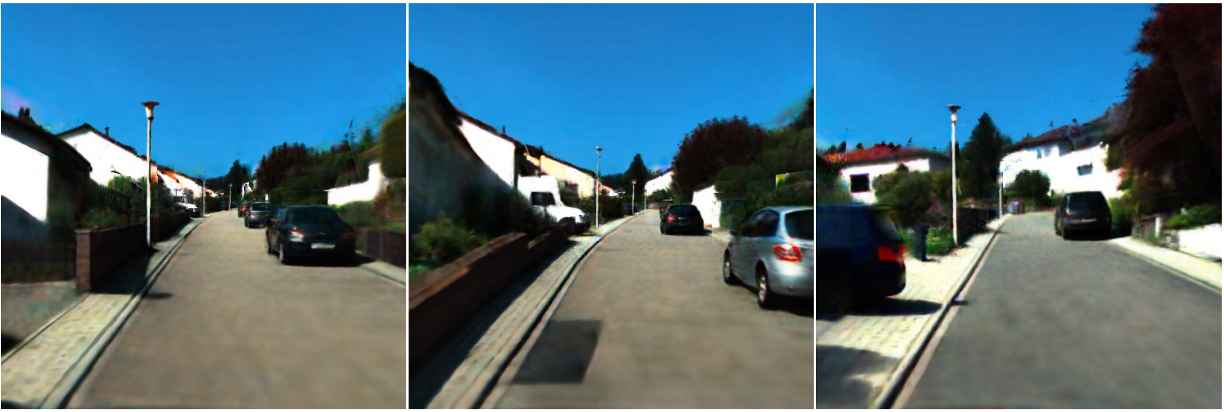} 
        \label{fig:urban_second}
    }
    \caption{An illustration of decompositional generation from \cite{lu2024urban}, where driving scenes in the bottom row are generated from separately constructed primitives in the top row.}
    \label{fig:compositionalGen}
\end{figure}

\paragraph{Decompositional Generation} In contrast to monolithic approaches that produce entire scenes in a single pass, \emph{decompositional generation} synthesizes environments in a structured, step-by-step manner. For instance, they generate the roads first and then populate the roads with vehicles. Panoptic Neural Fields \cite{kundu2022panoptic} exemplifies this paradigm by introducing a semantic, object-aware representation that factors a scene into distinct regions (\textit{e.g.}, roads, buildings, and vehicles). This segmentation not only yields a more interpretable model but also facilitates targeted edits, such as selectively modifying certain object classes.

Other recent works explore hierarchical or modular architectures to achieve similar goals. For instance, Block-NeRF \cite{tancik2022block} scales neural view synthesis to large scenes by dividing them into spatially coherent blocks, allowing for incremental and compositional building of complex cityscapes. NeuralField-LDM \cite{kim2023neuralfield} adopts a hierarchical latent diffusion process, breaking down the scene generation pipeline into coarser and finer stages. Urban Radiance Fields \cite{rematas2022urban} also embed city-scale structure within neural representations, capturing both global layout and local details in a compositional framework.

Beyond static scene generation, decompositional strategies are particularly useful for simulation and sensor modeling. UniSim \cite{yang2023unisim} and CADSim \cite{wang2023cadsim} each leverage neural fields to reconstruct and then manipulate large-scale urban scenarios. By disentangling various scene components—like geometry, appearance, and semantics—these simulators can independently update specific elements (\textit{e.g.}, changing vehicle positions, adjusting environmental factors) without regenerating the entire environment. This modular design promotes scalability, reusability, and real-time adaptability, underlining the importance of decompositional generation for complex 3D scene synthesis.

\subsection{LiDAR Generation}
\label{sec:lidar_gen}

\begin{figure}[!h]
  \setkeys{Gin}{width=\linewidth}
  \begin{tabularx}{\textwidth}{XX}
    \includegraphics{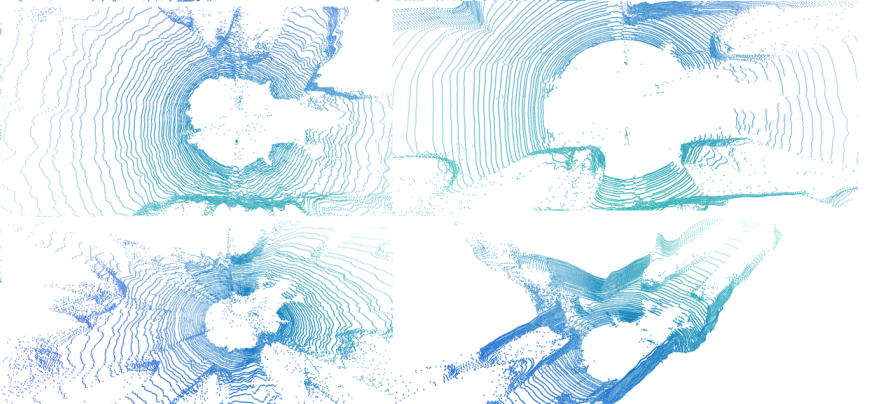} &
    \includegraphics{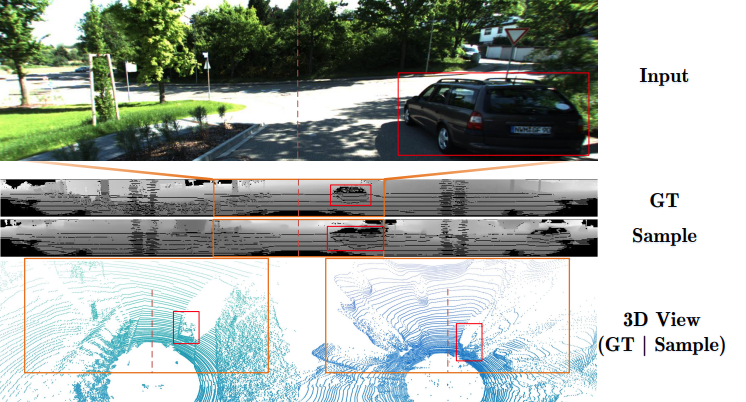} \\
  \end{tabularx}
\caption{Unconditional LiDAR generation (left) and conditional LiDAR generation guided by input data (right).}
\end{figure}

LiDAR point cloud generation is a crucial component in the advancement of generative AI for autonomous driving, as it directly supports critical tasks such as perception, simulation, and system validation. High-fidelity LiDAR data is essential for training robust perception models, testing planning algorithms, and simulating diverse driving scenarios. Unlike other modalities such as cameras, LiDAR provides precise 3D spatial information, making it indispensable for tasks like object detection, semantic segmentation, and trajectory prediction in L4/5 autonomous driving.

Despite its importance, LiDAR generation for autonomous driving faces significant challenges. One major obstacle is the limited availability of diverse real-world LiDAR scenes. Autonomous driving systems operate in highly complex environments, yet existing datasets often lack sufficient variability in conditions such as weather, traffic density, and road geometries. This scarcity hinders the robustness and generalization of models trained on such data. Additionally, accurately simulating LiDAR-specific characteristics, including sparse and non-uniform point clouds, raydrop effects, and interactions with reflective surfaces, poses unique difficulties. Another challenge lies in ensuring that synthetic data seamlessly integrates with real-world data. Generated LiDAR must exhibit geometric and semantic consistency across diverse scenarios while maintaining computational efficiency for large-scale simulation and testing. Furthermore, the lack of standard benchmarks for evaluating the fidelity and utility of generated LiDAR data adds complexity to the development and adoption of generative methods.

To address these challenges, recent advancements in generative AI have introduced novel approaches for LiDAR synthesis. These methods leverage powerful deep learning models to enhance the quality, adaptability, and controllability of generated point clouds. In particular, three major categories of LiDAR generation have emerged: diffusion models, which iteratively refine noise-based representations to create high-fidelity point clouds; neural radiance fields (NeRF) and neural implicit representations, which model LiDAR transmittance and scene geometry for realistic simulation; and transformer-based architectures, which enhance generative processes through structured representation learning and sequence modeling.

\paragraph{Early-Stage Physics-Based Generation}

Early LiDAR generation methods primarily relied on physics-based simulation, using ray casting and handcrafted sensor models to approximate real-world LiDAR characteristics. Yue et al. \cite{yue2018lidarpointcloudgenerator} introduced a game engine-based ray casting framework, enabling rapid dataset creation for segmentation and robustness testing. LiDARsim \cite{9157601} improved upon this by integrating a U-Net model to simulate raydrop effects, blending physics-based rendering with real-world data. PCGen \cite{10161226} continues this trend with First Peak Averaging (FPA) Raycasting, incorporating an MLP-based surrogate model to refine raydrop simulation and reduce the domain gap between simulated and real LiDAR. Transitioning towards generative approaches, LiDARGEN \cite{zyrianov2022learninggeneraterealisticlidar} employs a score-based diffusion model (SGM) to progressively denoise LiDAR point clouds while preserving physical consistency, supporting controllable synthesis without retraining. While these methods provided structured and interpretable LiDAR synthesis, their reliance on predefined sensor models limited adaptability, motivating the transition to diffusion models, NeRF-based approaches, and transformer-based architectures for more flexible and data-driven LiDAR generation.

\paragraph{Diffusion-based Generation}

Diffusion models have emerged as a powerful method for synthesizing LiDAR point clouds with strong realism, controllability, and computational efficiency. Building on this paradigm, LiDMs \cite{ran2024towards} present a latent diffusion framework for LiDAR scene generation, employing curve-wise compression, point-wise coordinate supervision, and patch-wise encoding to preserve structural and geometric details. Extending this approach further, RangeLDM \cite{hu2025rangeldm} introduces a latent diffusion framework that integrates Hough Voting for accurate range-view projection, VAE-based latent space compression, and a range-guided discriminator, enabling high-fidelity point cloud synthesis and downstream applications such as upsampling and inpainting. While these models focus on static LiDAR scene generation, LidarDM \cite{zyrianov2024lidardmgenerativelidarsimulation} extends diffusion to 4D LiDAR synthesis, first generating 3D scenes and dynamic actors before simulating temporally coherent sensor observations. Similarly, DynamicCity \cite{bian2024dynamiccitylargescalelidargeneration} employs a Diffusion Transformer (DiT) with VAE-based HexPlane encoding, optimizing large-scale LiDAR scene synthesis with a Projection Module and Pose Bedroll Strategy for more structured representation learning. Beyond scene generation, diffusion models are also applied to multimodal and controllable LiDAR synthesis. GenMM \cite{singh2024genmmgeometricallytemporallyconsistent} utilizes a diffusion framework to jointly edit RGB videos and LiDAR, ensuring temporal and geometric consistency when modifying existing 3D scenes. Meanwhile, Text2LiDAR \cite{wu2024text2lidar} introduces text-guided LiDAR synthesis, leveraging a Transformer-based equirectangular attention mechanism with a Control-Signal Embedding Injector (CEI) and Frequency Modulator (FM) to generate fine-grained, diverse point clouds based on textual descriptions. Collectively, these diffusion-based approaches advance LiDAR generation by improving computational efficiency, incorporating temporal and structural consistency, and enabling multimodal controllability, marking a shift from traditional handcrafted models to more flexible and scalable solutions.

\paragraph{NeRF-based Generation}

NeRF and neural implicit representations offer a continuous and structured approach to LiDAR generation, enhancing realism and adaptability in dynamic scenes. NeRF-LiDAR \cite{zhang2023nerf} proposes a NeRF-based framework that synthesizes realistic LiDAR point clouds with semantic labels from multi-view images and sparse LiDAR, using point- and feature-level alignment to enhance structural consistency. LiDAR4D \cite{10657876} introduces a differentiable LiDAR-only framework for novel space-time view synthesis, employing a 4D hybrid representation and geometric constraints to achieve geometry-aware and temporally consistent dynamic reconstruction. DyNFL \cite{Wu2023dynfl} introduces a compositional neural field framework that enables high-fidelity re-simulation of dynamic LiDAR scenes by separately reconstructing static backgrounds and dynamic objects, allowing for flexible scene editing and improved physical realism. These methods leverage implicit neural representations to provide high-fidelity, temporally consistent LiDAR data, bridging the gap between real and synthetic environments.

\paragraph{VQ-VAE-based Generation}

VQ-VAE-based methods have shown strong potential for structured and interpretable LiDAR generation. UltraLiDAR \cite{xiong2023learning} employs VQ-VAE to encode sparse LiDAR point clouds into compact discrete tokens, which are modeled using a transformer to enable sparse-to-dense completion and controllable scene generation. LidarGRIT \cite{haghighi2024taming} further builds on this design by combining an autoregressive transformer with VQ-VAE for progressive range image synthesis in the latent space, and explicitly models raydrop noise masks to improve geometric fidelity.
These approaches highlight the effectiveness of discrete representation learning for high-quality and semantically consistent LiDAR synthesis.

\begin{table}[h!]
\centering
\caption{LiDAR Generation Methods. We report the Fréchet Range Image Distance (FRID) whenever available and denote unconditional/conditional generation as u and c.}
\label{table:lidar_gen}
\resizebox{\textwidth}{!}{
\begin{tabular}{llllllll}
    \toprule
    \textbf{Method} & \textbf{Venue} & \textbf{Dataset (FRID$\downarrow)$} & \textbf{Modeling Type} & \textbf{Backbone} & \textbf{Control Mechanism} & \textbf{Generation Type} & \textbf{Code} \\
    \midrule
    LiDMs \cite{ran2024towards} & CVPR'24 & KITTI-360 (u 158.8, c 22.9) & Diffusion & CNN, U-Net & Multimodal conditions & Scene Generation & \href{https://github.com/hancyran/LiDAR-Diffusion}{GitHub} \\
    
    RangeLDM \cite{hu2025rangeldm} & ECCV'24 & KITTI-360 (u 1074.9) & Diffusion, VAE & CNN, U-Net & Partial Point Cloud & Scene Completion, Generation  & \href{https://github.com/WoodwindHu/RangeLDM}{GitHub} \\

    LidarDM \cite{zyrianov2024lidardmgenerativelidarsimulation} & ICRA'25 & KITTI-360, WOD & Diffusion, VAE & CNN & Semantic Map & LiDAR Simulation \& Raycasting & \href{https://github.com/vzyrianov/LidarDM}{GitHub} \\
    

    GenMM \cite{singh2024genmmgeometricallytemporallyconsistent} & arXiv'24 & BDD100K, WOD & Diffusion & U-Net, Transformer & 3D Bounding Boxes, Reference Image & Object-Level Manipulation & N/A \\
    
    Text2LiDAR \cite{wu2024text2lidar} & ECCV'24 & KITTI-360 (c 164.16), nuScenes & Diffusion & Transformer & Text & Full Scene Generation & 
    \href{https://github.com/wuyang98/Text2LiDAR}{GitHub} \\
    
    UltraLiDAR \cite{xiong2023learning} & CVPR'23 & PandaSet,KITTI & VQ-VAE & Transformer & Sparse Point Cloud & Scene Completion, Generation & N/A \\
    
    LidarGRIT \cite{haghighi2024taming} & CVPR-W'24 & KITTI-360, KITTI odometry & VQ-VAE & Transformer & Unconditional & Scene Generation & \href{https://github.com/hamedhaghighi/LidarGRIT}{GitHub} \\
    
    NeRF-LiDAR \cite{zhang2023nerf} & CVPR'24 & nuScenes & NeRF & U-Net, MLP & Camera Poses, Multi-view Images & LiDAR Simulation & \href{https://github.com/fudan-zvg/NeRF-LiDAR}{GitHub} \\

    LiDAR4D \cite{10657876} & CVPR'24 & KITTI, nuScenes & NeRF & U-Net, MLP & Camera Poses, Multi-view LiDAR Point Cloud & LiDAR Simulation & \href{https://github.com/ispc-lab/LiDAR4D}{GitHub} \\
    
    DyNFL \cite{Wu2023dynfl} & CVPR'24 &  WOD & Neural SDF & MLP & LiDAR Scans, 3D Bounding Boxes & LiDAR Simulation & \href{https://github.com/prs-eth/Dynamic-LiDAR-Resimulation}{GitHub} \\
    
    LiDARsim \cite{9157601} & CVPR'20 & LiDARsim Dataset & Physics-based Raycasting & Raycasting Engine, U-Net & 3D backgrounds, Dynamic Object Meshes & LiDAR Simulation & N/A \\
    
    PCGen \cite{10161226} & ICRA'23 & WOD & FPA Raycasting & Raycasting Engine, MLP & Reconstruced Scenario & LiDAR Simulation & N/A \\
    
    LiDARGEN \cite{zyrianov2022learninggeneraterealisticlidar} & ECCV'22 & KITTI-360 (2040.1) & Score-Based & U-Net & Sparse Point Cloud & Scene Generation & \href{https://github.com/vzyrianov/lidargen}{GitHub} \\
    
    Yue et al. \cite{yue2018lidarpointcloudgenerator} & ACM'18 & KITTI & Physics-based Raycasting & Raycasting Engine & Pre-defined In-game Scene Parameters & LiDAR Simulation & N/A \\
    \bottomrule
\end{tabular}
}
\end{table}



\subsection{Trajectory Generation}
\label{sec:traj_gen}

Trajectory generation, the task of synthesizing motion sequences for agents (\textit{e.g.}, vehicles, pedestrians, or other agents), is key to autonomous driving \cite{rudenko2020human,li2023pedestrian}. It enables applications from self-driving cars to mobile robot navigation \cite{yurtsever2020survey,li2024interactive,chen2024end,mavrogiannis2023core,yao2024sonic}. 
As autonomy advances into complex, dynamic environments, traditional methods like optimization-based motion synthesis and probabilistic graphical model-based methods \cite{li2018generic,zhan2018towards,li2020generic, wang2023eqdrive, wang2023equivariant} face limitations in handling uncertainty, multi-agent interactions, and real-time adaptability. 
This has spurred a paradigm shift toward generative AI, which leverages generative deep learning to model multimodal, context-aware trajectories while balancing safety, efficiency, and compliance with physical or social constraints. Recent breakthroughs in vision-language models (VLMs), diffusion models have further expanded the field, enabling systems to interpret multimodal inputs (\textit{e.g.}, LiDAR, maps, text instructions) and generate trajectories that align with human-like reasoning. This section systematically reviews these approaches, highlighting their strengths, limitations, and applications. We summarize a list of representative work in Table \ref{tab:trajectory_generation}.

\begin{figure}[h]
  \setkeys{Gin}{width=\linewidth}
  \begin{tabularx}{\textwidth}{XX}
    \includegraphics{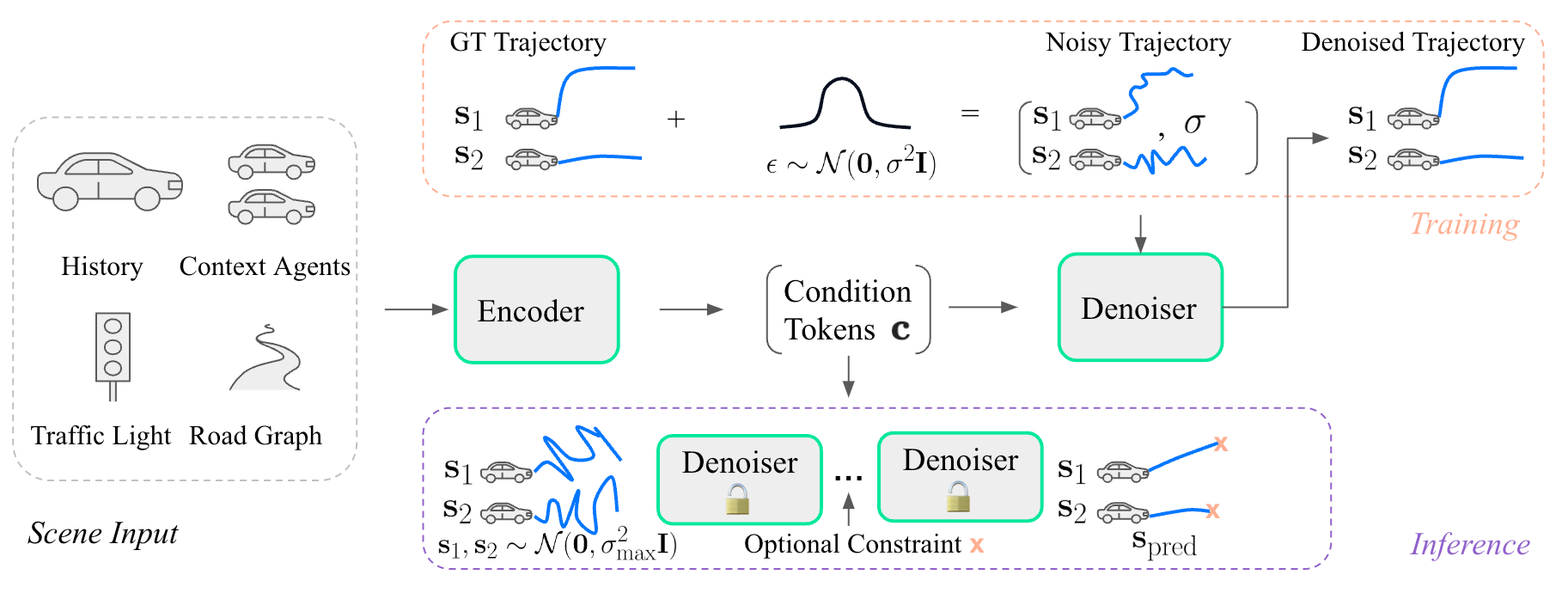} 
  \end{tabularx}
\caption{Overview of MotionDiffuser\cite{jiang2023motiondiffusercontrollablemultiagentmotion}, a diffusion-based trajectory generation framework that encodes scene context to condition the denoising processing during training and inference, enabling controllable multi-agent future trajectory generation with optional constraints.}
\end{figure}

\begin{table}[h!]
\centering
\caption{Trajectory Generation Methods. We report the full horizon Average Displacement Error (ADE) in meters, whenever available.}
\label{tab:trajectory_generation}
\resizebox{\textwidth}{!}{
\begin{tabular}{llllll}
\toprule
\textbf{Method} & \textbf{Venue} & \textbf{Dataset (ADE$\downarrow$)} & \textbf{Modeling Type} & \textbf{Backbone} & \textbf{Code} \\
\midrule
CGNS \cite{li2019conditional} & IROS'19 & ETH/UCY (0.62), SDD (15.6) & GAN & CNN & N/A  \\
EvolveGraph \cite{li2020evolvegraph} & NeurIPS'20 & ETH/UCY, SDD, H3D & Autoregressive & GNN & N/A  \\ 
STG-DAT \cite{li2021spatio} & T-ITS'21 & ETH/UCY (0.38), SDD (13.3) & CVAE & GNN & N/A  \\ 
PathGAN \cite{choi2021pathganlocalpathplanning} & ETRI'21 & iSUN & GAN & CNN & \href{https://github.com/d1024choi/pathgan_pytorch}{GitHub}  \\
MID \cite{gu2022stochastictrajectorypredictionmotion} & CVPR'22 & ETH/UCY (0.39), Stanford Drone (7.61) & Diffusion & Transformer & \href{https://github.com/Gutianpei/MID}{GitHub} \\
LED \cite{mao2023leapfrogdiffusionmodelstochastic} & CVPR'23 & ETH/UCY & Diffusion & Leapfrog & \href{https://github.com/MediaBrain-SJTU/LED}{GitHub} \\
SingularTrajectory \cite{bae2024singulartrajectoryuniversaltrajectorypredictor} & CVPR'24 & ETH (0.45) & Diffusion & SVD & \href{https://github.com/InhwanBae/SingularTrajectory}{GitHub} \\
Diffusion-Planner \cite{zheng2025diffusionbased} & ICLR'25 & nuPlan & Diffusion & Transformer & \href{https://github.com/ZhengYinan-AIR/Diffusion-Planner}{GitHub} \\
GPT-Driver \cite{mao2023gptdriver} & NeurIPS'23 & nuScenes (1.52) & LLM & Transformer & \href{https://github.com/PointsCoder/GPT-Driver}{GitHub} \\
DriveLM \cite{sima2024drivelm} & ECCV'24 & nuScenes & VLM & Transformer &  \href{https://github.com/OpenDriveLab/DriveLM}{GitHub}\\
LMDrive \cite{shao2024lmdrive} & CVPR'24 & CARLA & LLM & Transformer &  \href{https://github.com/opendilab/LMDrive}{GitHub} \\
OpenEMMA \cite{xing2025openemma} & WACV'25 & nuScenes (2.81) & VLM & Transformer &  \href{https://github.com/taco-group/OpenEMMA}{GitHub} \\
Desire \cite{lee2017desire} & CVPR'17 & KITTI (2.06), Stanford Drone & CVAE & RNN & \href{https://github.com/tdavchev/DESIRE}{GitHub} \\
Trajectron \cite{ivanovic2019trajectron} & ICCV'19 & ETH/UCY & CVAE & Graph RNN & \href{https://github.com/StanfordASL/Trajectron}{GitHub} \\
Trajectron++ \cite{salzmann2020trajectron++} & ECCV'20 & ETH/UCY, nuScenes & CVAE & Constrained Graph RNN & \href{https://github.com/StanfordASL/Trajectron-plus-plus}{GitHub} \\
Social GAN \cite{gupta2018socialgansociallyacceptable} & CVPR'18 & ETH/UCY (0.60) & GAN & RNN & \href{https://github.com/agrimgupta92/sgan}{GitHub}  \\
SoPhie \cite{sadeghian2018sophieattentiveganpredicting} & CVPR'19 & ETH/UCY (0.70) & GAN & Cross Attention & \href{https://github.com/coolsunxu/sophie}{GitHub} \\
Social-BiGAT \cite{kosaraju2019socialbigatmultimodaltrajectoryforecasting} & NeurIPS'19 & ETH/UCY (0.48) & Bicycle-GAN & Graph Attention Network & N/A \\
MotionDiffuser \cite{jiang2023motiondiffusercontrollablemultiagentmotion} & CVPR'23 & WOMD & Diffusion & Transformer & N/A \\
Westny et al. \cite{westny2024diffusionbasedenvironmentawaretrajectoryprediction} & arXiv'24 & rounD (0.21), highD (0.44) & Diffusion & GNN & N/A \\
LMTrajectory \cite{bae2024languagebeatnumericalregression} & CVPR'24 & ETH/UCY (0.41) & LLM & Transformer &  \href{https://github.com/InhwanBae/LMTrajectory}{GitHub} \\
TrafficSim \cite{Suo2021TrafficSim} & CVPR'21 & ATG4D\footnote{Private Dataset} & CVAE & GNN & N/A  \\
TrafficBots\cite{zhang2024trafficbots} & ICRA'23 & WOMD & CVAE & MLP & \href{https://github.com/zhejz/TrafficBotsV1.5}{GitHub}  \\
DJINN \cite{niedoba2023diffusion} & NeurIPS'23 & INTERACTION & Diffusion & Transformer & N/A \\
Scenario Diffusion \cite{pronovost2023scenario} & NeurIPS'23 & AV2 & Diffusion & UNet & N/A \\
BehaviorGPT \cite{Zhou2024BehaviorGPT} & NeurIPS'25 & WOMD & Autoregressive & Transformer &  N/A \\
\bottomrule
\end{tabular}
}
\end{table}

Traditional motion planning often seeks a single optimal trajectory based on criteria like safety and comfort, which can be insufficient under high uncertainty or complex interactions \cite{zhan2017safe,chai2019multipath,li2023game}. Generative models offer a paradigm shift, reframing trajectory generation as sampling from a learned distribution of plausible futures. This approach naturally handles uncertainty and multimodality, crucial for both predicting the behavior of external agents and planning the ego-vehicle's path, as well as for generating realistic, large-scale traffic simulations. We review key generative methodologies applied to these tasks below.

\paragraph{Variational Autoencoders (VAEs) and Generative Adversarial Networks (GANs)}

Early generative approaches leveraged VAEs and GANs to move beyond single-trajectory outputs \cite{li2019interaction,ma2019wasserstein,sun2022interaction}. 
\textbf{\textit{VAEs}} excel at capturing uncertainty and generating diverse possibilities by learning a latent distribution conditioned on input context (\textit{e.g.}, past trajectory, map, agent interactions). For single-agent prediction or planning, this allows sampling multiple potential futures conditioned on perceived intent or environmental factors \cite{kim2021driving, article}. In multi-agent settings, CVAEs were foundational. Influential work like DESIRE \cite{lee2017desire} generates diverse, socially plausible trajectories for multiple agents, employing ranking mechanisms to improve consistency. Building on this, Trajectron \cite{ivanovic2019trajectron} introduces dynamic graph structures to explicitly model interactions, significantly impacting the field. Its successor, Trajectron++ \cite{salzmann2020trajectron++}, further enhances realism by incorporating map information and vehicle dynamics constraints. 
S-CM \cite{choi2021shared} proposes a Cross-Modal Embedding framework that aims to benefit from the use of multiple input modalities under the CVAE framework.
\cite{ma2021continual} presents a multi-agent interaction behavior prediction framework with a graph-neural-network-based conditional generative memory system to mitigate catastrophic forgetting in continual learning for trajectory generation.
For simulation, CVAE-based architectures like the transformer-based model in TrafficBots \cite{zhang2024trafficbots} learn agent ``personalities" from data, enabling the generation of varied and interactive agent behaviors within simulated environments. 
STG-DAT \cite{li2021spatio} proposes a generic generative neural system for multi-agent trajectory prediction involving heterogeneous agents, taking a step forward to explicit interaction modeling by incorporating relational inductive biases.
DNRI \cite{dax2023disentangled} introduces novel disentanglement techniques to enhance the interpretability of CVAE for motion generation.
\textbf{\textit{GANs}} focus on generating highly realistic trajectories through an adversarial training process, where a generator tries to fool a discriminator trained to distinguish real data from generated samples. 
CGNS \cite{li2019conditional}, for instance, introduces a conditional generative neural system to generate multimodal future trajectories of vehicles.
PathGAN \cite{choi2021pathganlocalpathplanning}, for instance, generates realistic ego-vehicle paths directly from visual inputs and high-level intentions. In the multi-agent domain, GANs like Social GAN \cite{gupta2018socialgansociallyacceptable} introduced pooling mechanisms to aggregate social context, while SoPhie \cite{sadeghian2018sophieattentiveganpredicting} and Social-BiGAT \cite{kosaraju2019socialbigatmultimodaltrajectoryforecasting} employed attention mechanisms and graph representations to better capture interactions and generate socially compliant multi-agent forecasts.
CTPS \cite{li2019coordination} brings the ideas of Bayesian deep learning into deep generative models to generate diversified prediction hypotheses.

VAEs offer inherent diversity and probabilistic interpretation, making them suitable for exploring potential futures and representing uncertainty. GANs often achieve higher perceptual realism but can suffer from mode collapse (limited diversity) and training instability. Both paved the way for modeling complex interactions, moving beyond independent agent forecasting. TrafficSim \cite{Suo2021TrafficSim} also used implicit generative models (related to VAEs/GANs) to generate socially consistent simulation trajectories.

\paragraph{Diffusion Models}

More recently, diffusion models have emerged as a powerful class of generative models for trajectory tasks, offering high sample quality and stable training. They operate by learning to reverse a diffusion process that gradually adds noise to data, effectively learning to "denoise" random noise into structured trajectories conditioned on context.

Diffusion models naturally extend to complex multi-agent scenarios. MotionDiffuser \cite{jiang2023motiondiffusercontrollablemultiagentmotion} employed permutation-invariant transformers within the diffusion framework to generate controllable, collision-aware joint trajectories for multiple agents. SDT \cite{zhang2024multiagent} further scaled this combination. Integrating environmental context is key; environment-aware models \cite{westny2024diffusionbasedenvironmentawaretrajectoryprediction} condition the diffusion process on maps and dynamic interactions. The Residual Diffusion Model \cite{app142210350} focused on enforcing physical constraints during generation and improving efficiency.

Diffusion models are bridging the gap between prediction and planning. Diffusion-Planner \cite{zheng2025diffusionbased} integrates multi-agent prediction and ego-vehicle planning within a single diffusion framework, enabling interaction-aware driving without manually defined rules. For traffic simulation, diffusion models enable flexible, long-horizon, and controllable scenario generation. Examples include DJINN \cite{niedoba2023diffusion}, Scenario Diffusion \cite{pronovost2023scenario}, and SceneDiffuser \cite{Jiang2024SceneDiffuser}, which jointly model agent interactions and allow user-defined constraints. Conditional Traffic Diffusion (CTD) \cite{zhong2023guided} incorporates temporal logic constraints, RoAD \cite{liu2025rolling} enables closed-loop reactive simulation, and Scenario Dreamer \cite{rowe2025scenario} scales generation using latent diffusion. Hybrid approaches like SLEDGE \cite{chitta2024sledge} combine diffusion with transformers for efficient, high-fidelity simulation.

Diffusion models currently represent the state-of-the-art in generating realistic and diverse trajectories for both prediction and simulation. Their strength lies in capturing complex data distributions accurately. However, their iterative sampling process can be computationally expensive, posing challenges for real-time deployment, although significant progress is being made on acceleration techniques. Their ability to incorporate diverse conditioning information (maps, constraints, interactions) is a major advantage.

\paragraph{Sequence Models (Transformers and Large Language Models)}

Leveraging the success of sequence modeling in natural language processing, Transformers and MLLMs are increasingly applied to trajectory generation, treating trajectories as sequences of states or actions.

Beyond their use within diffusion models or CVAEs as conditioners, transformer architectures are used directly for sequence generation. BehaviorGPT \cite{Zhou2024BehaviorGPT} uses autoregressive next-patch prediction for lightweight, high-fidelity traffic simulation. TrafficGen \cite{feng2023trafficgen}, the engine behind ScenarioNet \cite{li2023scenarionet}, employs an interpretable autoregressive encoder-decoder structure to sequentially generate realistic traffic scenarios from logged data, proving effective for generating training data for reinforcement learning.

MLLMs offer the potential to incorporate broader world knowledge, reasoning capabilities, and multimodal inputs (text, images, LiDAR) into the trajectory generation process. Early work like GPT-Driver \cite{mao2023gpt} used GPT-3.5 on textual scene descriptions to generate trajectory tokens, demonstrating dynamic updates but limited by text-only input. Subsequent models integrated richer contexts. DriveLM \cite{sima2023drivelm} combined structured visual representations with LLM reasoning, LMDrive \cite{shao2024lmdrive} used a closed-loop approach mapping sensor data to control, and Waymo's EMMA \cite{hwang2024emma} (and its open-source counterpart OpenEMMA \cite{xing2025openemma}) directly mapped raw sensor data to driving actions using chain-of-thought reasoning. For multi-agent prediction, recent work explores encoding interactions as text-like tokens for LLM-based sequence generation \cite{bae2024languagebeatnumericalregression}, aiming to capture higher-level reasoning about group dynamics.

Transformer-based sequence models offer scalability and interpretability, particularly for simulation based on real data logs. MLLMs represent a rapidly evolving frontier, promising to handle complex instructions, reason about intricate scenarios, and integrate diverse data modalities. However, challenges include grounding language-based outputs in physical reality, ensuring safety and robustness, managing the immense data requirements, and the significant computational cost. Their application to complex, interactive multi-agent prediction and simulation is still relatively nascent compared to diffusion models or VAEs/GANs.

\subsection{Occupancy Generation}
\label{sec:occupancy_gen}


\begin{figure}[h]
  \setkeys{Gin}{width=\linewidth}
  \begin{tabularx}{\textwidth}{XX}
    \includegraphics{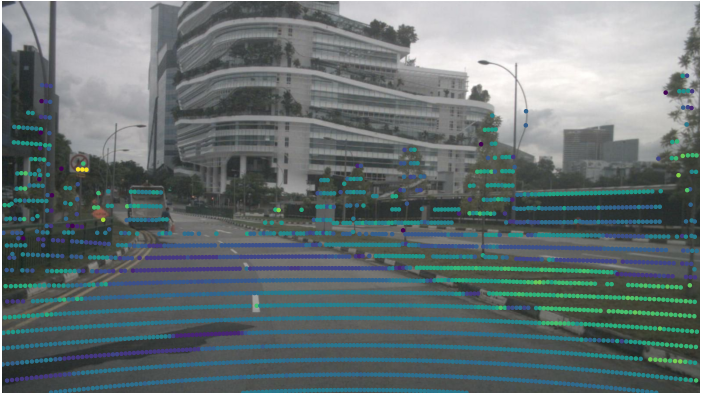} &
    \includegraphics{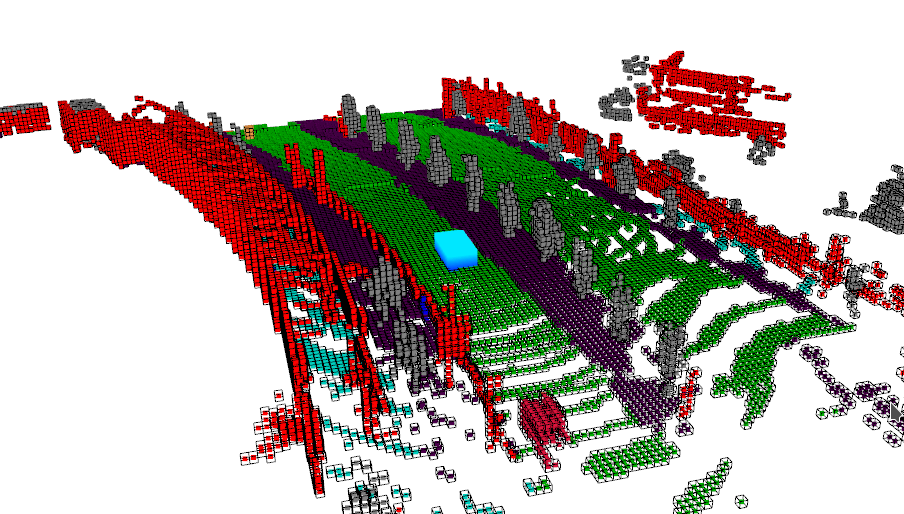} \\
  \end{tabularx}
\caption{On the right is a visualization of a 3D occupancy grid. It corresponds to the camera/LIDAR image on the left. The ego vehicle is shown as the center blue cube.}
\end{figure}

Unlike other modalities like images or LiDAR, 3D occupancy generation has the unique challenge that there exists no ground truth occupancy grid that researchers can use for training or quality measurement. Popular datasets, such as nuScenes \cite{nuscenes2019} and Waymo \cite{waymodataset}, do not include a 3D occupancy modality. Because of this, researchers in this area often need to develop their own methods to compute 3D occupancy grids for the scene at a particular timestep using camera and/or LIDAR inputs at that timestep and earlier. The works below present the approaches to acquire this ground truth. Works such as \cite{tian2024occ3d, li2024sscbench, tong2023scene, wang2023openoccupancy, openscene2023} introduce methodologies for generating 3D occupancy annotations. Occ3D \cite{tian2024occ3d} introduces a high-resolution occupancy benchmark using LiDAR and camera data, enabling highly detailed and diverse urban scene occupancies while capturing objects outside predefined ontologies. Occ3D is a foundational work because the results of this method are widely used as training data in many subsequent works for tasks in future occupancy grid prediction \cite{zheng2025occworld} and occupancy scene generation \cite{zhang2024urban, gu2024dome}. TPVFormer \cite{huang2023tpvformer}, on the other hand, approaches this problem from traditional Bird's-Eye-View (BEV) by adding side and front views to enhance 3D spatial representation. It uses a tri-perspective view (TPV) with transformer-based cross-attention to combine sparse LiDAR data with camera inputs for competitive performance even with limited 3D data. SurroundOcc \cite{wei2023surroundocc} addresses the sparseness in the occupancies by fusing multi-frame LiDAR scans with higher resolution camera images; it applies spatial attention and 3D convolutions that progressively upscale from low to high resolution, filling in occluded areas through Poisson reconstruction, which maintains scene consistency and depth accuracy. OccGen \cite{wang2024occgen} adopts a unique generative approach, framing occupancy computation as a “noise-to-occupancy” diffusion process that refines occupancy maps in steps, thus offering flexibility in balancing accuracy and computational cost while providing uncertainty estimates, making it particularly adept at handling occluded or ambiguous areas in the scene. Together, these methods reflect an evolution in occupancy grid computation, balancing dense and efficient 3D scene representation through advancements in data handling, generative modeling, and multi-view fusion. Recently, UniOcc \cite{wang2025unioccunifiedbenchmarkoccupancy} provides a unified benchmark and toolkit to acquire occupancy labels from popular driving datasets and simulations.

When it comes to synergy between generative AI and 3D occupancy, we see two dominant areas of research: \textbf{occupancy generation} and \textbf{occupancy forecasting}. The former refers to predicting future frame occupancy grids given historical ones, while the latter refers to generating current or future frame occupancy grids given a condition, like text or image.

\textbf{Occupancy generation} remains a nascent area of research, with relatively few works available at the time of writing. We highlight four promising methods that leverage generative techniques to achieve high-fidelity scene reconstruction and robust temporal forecasting, each introducing innovative strategies for realism and control in scene generation. UrbanDiffusion \cite{zhang2024urban} generates static 3D occupancy grids using a diffusion model conditioned on Bird’s-Eye View (BEV) maps. These BEV maps, editable via MetaDrive \cite{li2022metadrive}, guide the model’s semantic and geometric scene generation. The method uses a two-stage training process: first, a VQ-VAE \cite{van2017vqvae} embeds ground truth occupancy grids into a latent representation optimized for reconstruction loss; second, the latent embeddings are concatenated with BEV features and passed through a 3D U-Net \cite{ronneberger2015unet} to perform denoising. While this approach achieves structured, semantically consistent scenes, its output is limited to static, single-frame generation.

In contrast, OccSora \cite{wang2024occsora} extends generation to dynamic, multi-frame scenes conditioned on ego vehicle trajectories. Similar to UrbanDiffusion, OccSora uses a two-step training approach. First, a 4D occupancy tokenizer compresses historical occupancy grids into discrete tokens, reconstructing them via a 4D decoder. Next, a diffusion transformer \cite{peebles2022dit} generates future occupancy by denoising these tokens, conditioned on the encoded trajectory. This design enables OccSora to simulate temporally consistent scene evolution over multiple frames. Similarly, DOME \cite{gu2024dome} generates trajectory-conditioned dynamic occupancy sequences, employing a spatial-temporal diffusion transformer to predict long-term future frames. DOME begins by using a VAE \cite{kingma2013vae} to compress historical occupancy into latent embeddings optimized for reconstruction loss. These embeddings, combined with historical occupancy and ego trajectory inputs, guide the transformer to forecast future occupancy. Notably, DOME introduces a trajectory resampling method that enhances control over the generated scenes, supporting fine-grained, trajectory-aligned forecasting with high-resolution outputs. \cite{bian2025dynamiccity} employs a HexPlane \cite{cao2023hexplane} scene representation to conditionally generate 4D occupancies via DiT, which enables high-quality dynamic scene generation.

Finally, OccLLaMA \cite{wei2024occllama} uniquely integrates vision, language, and action modalities for occupancy generation. Like other methods, it uses a VQVAE to tokenize historical occupancy into discrete scene tokens. However, OccLLaMA also tokenizes ego motion as action tokens and language descriptions as text tokens. These three token types are processed by a pretrained LLaMA \cite{touvron2023llama}, fine-tuned to predict future scene, action, and text tokens. The predicted tokens are decoded into occupancy grids, motion plans, and textual outputs, enabling multimodal capabilities such as motion planning and visual question answering in addition to occupancy generation. This multimodal integration highlights OccLLaMA's potential as a versatile and extensible generative world model.

\textbf{Occupancy forecasting} generates future occupancy grids conditioned on the past occupancy grid in either 2D or 3D representations \cite{zhang2024vision,toyungyernsub2022dynamics,toyungyernsub2024predicting,lange2025self,zheng2025occworld,agro2024uno,lange2024scene}. In particular, OccWorld \cite{zheng2025occworld} leverages a generative model to predict the evolution of both the 3D environment and the ego vehicle’s trajectory. It encodes the current scene into discrete tokens using a variational autoencoder (VAE) and then predicts future scene tokens and vehicle states with a spatial-temporal transformer. This design allows OccWorld to capture fine-grained environmental changes over time without requiring instance-level annotations or pre-labeled object classes. In contrast, UNO (Unsupervised Occupancy) \cite{agro2024uno} forecasts 4D occupancy fields (spanning space and time) in a continuous fashion using self-supervised learning from LiDAR data. UNO constructs an occupancy map by generating pseudo-labels from LiDAR points, capturing both occupied and free space. It employs an implicit decoder to make occupancy predictions at any spatio-temporal point, making it adaptable for tasks like point cloud forecasting and semantic occupancy prediction in bird’s-eye-view (BEV) formats. UNO’s continuous occupancy representation achieves state-of-the-art performance across multiple datasets and is effective even with minimal supervision.

Together, these methods illustrate the evolution of conditioned 3D occupancy generation from single-frame scene reconstruction to complex multimodal, temporally-aware world models, each leveraging advanced generative architectures to capture intricate spatial and semantic information while accommodating diverse control inputs. A comparison of them is listed in Table \ref{table:occgeneration}.

\begin{table}
\caption{Comparison of 3D Occupancy Generation Methods. We report the mean average intersection-over-union whenever available.}
\label{table:occgeneration}
\resizebox{\textwidth}{!}{
\begin{tabular}{cccccccc}
    \toprule
    \textbf{Method} & \textbf{Venue} & \textbf{Dataset ($\text{mIoU}\downarrow$)} & \textbf{Modeling Type} & \textbf{Backbone} & \textbf{Control Mechanism} & \textbf{Generation Type} & \textbf{Code} \\
    \midrule
    UrbanDiffusion \cite{zhang2024urban} & arXiv'24 & nuScenes via Occ3D (0.80) & VQ-VAE & Diffusion & BEV Layout & Static Scene & Not Released \\
    
    DOME \cite{gu2024dome} & arXiv'24 & nuScenes via Occ3D (0.83) & VAE & DiT & Ego Trajectory & Scene and Agent Only & Not Released \\
    
    OccWorld \cite{zheng2025occworld} & ECCV'24 & nuScenes via Occ3D (0.66) & VQ-VAE & Transformer & Past Occupancy & Scene and Agent & \href{https://github.com/wzzheng/OccWorld}{GitHub} \\
    
    OccSORA \cite{wang2024occsora} & arXiv'24 & nuScenes via Occ3D (0.27) & VQ-VAE & DiT & Ego Trajectory, Past Occupancy & Scene and Agent & \href{https://github.com/wzzheng/OccSora}{GitHub}\footnote{As of July 2026, the released code contained redacted lines, thus the results are not reproducible.} \\
    
    OccLLaMA \cite{wei2024occllama} & arXiv'24 & nuScenes via Occ3D (0.75) & VQ-VAE & LLaMA & Language & Scene and Agent & Not Released \\
    
    UnO \cite{agro2024uno} & CVPR'24 & nuScenes, Argoverse2 & Not Specified & Transformer & Past Occupancy & Semantic LiDAR & Not Released \\
    
    DynamicCity \cite{bian2025dynamiccity} & ICLR'25 & CARLA (0.65) & VAE & DiT & Ego Trajectory & Scene and Agent & \href{https://github.com/3DTopia/DynamicCity}{GitHub} \\

    UniOcc \cite{wang2025unioccunifiedbenchmarkoccupancy} & ICCV'25 & UniOcc (0.51) & VQ-VAE & Transformer & Ego Trajectory & Scene and Agent & \href{https://github.com/tasl-lab/UniOcc}{GitHub} \\
    \bottomrule
\end{tabular}
}
\end{table}





\subsection{Video Generation}
\label{sec:video_gen}

Video is one of the mainstream representations of scenes from the ego view~\cite{Ji:TPAMI2012}. Although significant progress has been made in the field of video generation, ensuring temporal consistency over a long duration remains a challenge \cite{liu2024sora, kondratyuk2023videopoet, dai2023emu, zeng2024make}. Historically, algorithm-generated videos in driving scenarios are used to facilitate the learning of autonomous driving algorithms by providing more training data or forming a driving simulation platform that provides real-world-like sensor input for autonomous driving algorithms \cite{dosovitskiy2017carla}. Thus, the video generation approaches in driving scenarios should maintain the following features: (1) geometric and temporal consistency regarding the environment and the objects; (2) the movements of the ego and other vehicles should follow the real traffic flow; (3) the driving data is long-tailed thus the video generation procedure should be fine-grained controlled. 

Below, we review existing driving video generation approaches to investigate how they tackle these problems.

From a methodology perspective, existing approaches typically (1) achieve geometric and temporal consistency by developing spatial- or temporal- attentions \cite{wen2024panacea, ma2024unleashing, wu2024drivescape, zhao2024drivedreamer}; (2) ensure plausible traffic flow by utilizing structural information including BEV maps~\cite{wen2024panacea, ma2024unleashing, wang2023drivedreamer}, HD maps~\cite{wang2023drivedreamer, zhao2024drivedreamer, yang2024drivearena}, 3D layout~\cite{wu2024drivescape, ma2024unleashing, wen2024panacea, wang2023drivedreamer}; (3) condition on images~\cite{wen2024panacea, ma2024unleashing, wu2024drivescape, wang2023drivedreamer}, texts~\cite{wen2024panacea, ma2024unleashing, zhao2024drivedreamer, wang2023drivedreamer, yang2024drivearena}, depth~\cite{wen2024panacea}, camera pose~\cite{wen2024panacea, ma2024unleashing, wu2024drivescape, yang2024drivearena}, BEV~\cite{wen2024panacea, ma2024unleashing, wu2024drivescape, zhao2024drivedreamer, yang2024drivearena}, HD maps~\cite{zhao2024drivedreamer, wang2023drivedreamer}, 3D layout~\cite{ma2024unleashing, wu2024drivescape, zhao2024drivedreamer, wang2023drivedreamer, yang2024drivearena}, and driving actions~\cite{wang2023drivedreamer}.

From a task perspective, these works target different downstream use cases: \textbf{Driving video generation}, \textbf{closed-loop simulation}, \textbf{language-explainable video generation}. We review each case separately. 

\paragraph{Driving Video Generation} Panacea~\cite{wen2024panacea} generates panoramic multi-view driving videos from multiple control signals, including image, text, and BEV sequence. It utilized the latent diffusion model as a generative prior and employed the ControlNet~\cite{zhang2023adding} structure to control the generation of the videos. It optimizes the video generation to fit the provided BEV sequence, to ensure the videos adhere to the real traffic flows. It developed intra-view and cross-view attentions and cross-frame attentions to ensure the geometric and temporal consistency of the generated videos. However, it typically generates 8 frames of video at \SI{2}{Hz}, which is relatively short and sparse for training autonomous driving algorithms.
Delphi~\cite{ma2024unleashing} further achieves longer video generation of 40 frames by developing techniques including noise decomposition and reinitialization and feature-aligned temporal consistency. However, due to the dependency on 3D bounding box annotation, its temporal resolution is restricted to \SI{2}{Hz}.
DriveScape~\cite{wu2024drivescape} further overcomes this problem and achieves a temporal resolution up to 10 Hz by developing a Bi-Directional Modulated Transformer technique to ensure precise alignment of 3D structural information under sparse conditioning control.

\paragraph{Video Generation in Closed-loop Autonomous Driving}
Unlike general-purpose video generation tasks, video generation for closed-loop simulation requires the accurate modeling of the interaction between the ego car and the environment. Many works show an apparent generate-and-act paradigm. DriveDreamer~\cite{wang2023drivedreamer}, GenAD~\cite{yang2024generalized}, and Vista~\cite{gao2024vista} can jointly model drivers' actions by utilizing a two-stage training. In the first stage, a video generation model is trained. In the second stage, the driver actions are provided, and the model is asked to predict future frames of the video. Thus, these models can react to the driving policies. DrivingWorld~\cite{hu2024drivingworld} models video generation jointly with vehicle position to predict future frames using a GPT-style structure and end-to-end one-stage training. Doe-1~\cite{zheng2024doe} proposes an autonomous driving system with a perception-planning-prediction paradigm in which the video generation is conditioned on agent predictions.

\paragraph{MLLM Assisted Video Generation}
with the development of multimodality large language models, their strong reasoning abilities are widely used in many fields. In autonomous driving, besides using them as a general QA solution in driving scenes~\cite{ma2024dolphins, sima2023drivelm}, many works have specifically utilized MLLMs in driving video generation. DriveDreamer2~\cite{zhao2024drivedreamer} leverages MLLMs to generate a plausible BEV trajectory as a conditioning signal. ChatSim~\cite{wei2024editable} built an agent system using MLLMs to enable interactive and spatially consistent video editing. Doe-1~\cite{zheng2024doe} leveraged VQA as a scene description generator in the perception-planning-generation loop.


\begin{table}[h!]
\caption{Comparison of Video-based Scene Generation Methods. I = image, T = text, E = BEV, B = bounding boxes or layout, D = depth, C = camera, M = maps, A = driver action, O = optical flow, J = trajectory, S = subject, H = high-level instructions like command and goal point. Conditions in brackets are optional. Fréchet Video Distance (FVD) is tested using nuScenes \cite{nuscenes2019}.}
\label{table:video}
\resizebox{\textwidth}{!}{
\begin{tabular}{lllllllllll}
    \toprule
    \textbf{Method} & \textbf{Venue} & \textbf{Modeling} & \textbf{Backbone} & \textbf{Frames} & \textbf{FPS} & \textbf{FVD} & \textbf{Condition} & \textbf{Closed-loop} & \textbf{LLMs} & \textbf{Code}\\
    \midrule
    Panacea~\cite{wen2024panacea} 
    & CVPR'24 & Diffusion & ControlNet & 8 & 2 & 139.0 & ITEBDCM &  &  & \href{https://github.com/wenyuqing/panacea}{GitHub}\\
    
    Delphi~\cite{ma2024unleashing} 
    & CoRR'24 & Diffusion & U-Net & 40 & 2 & 113.5 & TEBC & \checkmark &  & N/A \\
    
    DriveDreamer~\cite{wang2023drivedreamer} 
    & ECCV'24 & Diffusion & U-Net,Transformer & 32 & 12 & 340.8 & ITMBA &  &  & \href{https://github.com/JeffWang987/DriveDreamer}{GitHub}\\
    
    DriveScape~\cite{wu2024drivescape} 
    & arXiv'24 & Diffusion & U-Net & 30 & 2--10 & 76.39 & IMEB &  &  & N/A \\
    
    DriveArena~\cite{yang2024drivearena} 
    & CoRR'24 & Diffusion,AR & U-Net & N/A & 12 & N/A & TBCM & \checkmark &  & \href{https://github.com/PJLab-ADG/DriveArena}{GitHub}\\
    
    DriveGen~\cite{lin2025drivegen} 
    & arXiv'24 & Diffusion & U-Net & - & - & 125.88 & ITB &  &  & \href{https://github.com/Hongbin98/DriveGEN}{GitHub}\\
    
    DrivingDiffusion~\cite{li2025drivingdiffusion} 
    & ECCV'24 & Diffusion & U-Net & - & - & 346 & ITBO &  &  & \href{https://github.com/shalfun/DrivingDiffusion}{GitHub}\\
    
    Vista~\cite{gao2024vista} 
    & CoRR'24 & Diffusion,AR & U-Net & 25 & 10 & 89.4 & I(AHJ) &  &  & \href{https://github.com/OpenDriveLab/Vista}{GitHub}\\
    
    SubjectDrive~\cite{huang2024subjectdrive} 
    & CoRR'24 & Diffusion & ControlNet & 8 & 2 & 124 & ITSB &  &  & N/A\\
    
    GenAD~\cite{yang2024generalized} 
    & CVPR'24 & Diffusion & Transformer & 8 & 2 & N/A & ITAJ & \checkmark &  & N/A\\
    
    DrivingWorld~\cite{hu2024drivingworld} 
    & arXiv'24 & AR & Transformer,GPT & 400 & 10 & 90.9 & IJ & \checkmark &  & \href{https://github.com/YvanYin/DrivingWorld}{GitHub}\\
    
    Doe-1~\cite{zheng2024doe} 
    & arXiv'24 & N/A & N/A & - & 2 & N/A & ITJ & \checkmark & \checkmark & \href{https://github.com/wzzheng/Doe}{GitHub}\\
    
    ChatSim~\cite{wei2024chatsim}  
    & CVPR'24 & Agent & N/A & 40 & 10 & N/A & IT &  & \checkmark & \href{https://github.com/yifanlu0227/ChatSim}{GitHub}\\
    
    \bottomrule
\end{tabular}
}
\end{table}



\subsection{3D/4D Reconstruction and Generation}
\label{sec:3d4d_gen}

3D Geometric correctness is crucial in driving scenes. Instead of learning indirect 3D structure in image or video generation, many scene generation approaches in driving scenes introduce direct 3D structure learning utilizing 3D representations such as point cloud, NeRF, and 3D Gaussian.

\paragraph{3D Representations}
3D representation refers to implicit or explicit approaches to store and manipulate static or dynamic geometry (and appearance). There are many 3D representations like voxels, implicit surfaces, or parametric models; however, in this section, we focus on the representations mainly used in driving scene generation. Specifically, we will focus on 3D point cloud, neural radiance field (NeRF), and 3D Gaussian splatting (3DGS). 3D representations allow us to primarily perform two tasks, namely \textbf{Scene Reconstruction} and \textbf{Scene Generation}. Below, we discuss both separately.

\begin{tcolorbox}[colback=gray!5!white, colframe=black!50!gray, parbox=false]
\textbf{Note:} Although multi-view-stereo (MVS) images are one of the common representations that are used in driving scenarios, we discuss them already in video~\ref{sec:video_gen} and image~\ref{sec:image_gen} generation and thus we will not discuss them here. 
\end{tcolorbox}





\paragraph{3D Driving Scene Reconstruction}
Reconstruction or representation is a fundamental component of the type of generative models that assume a fixed prior on the structure of the input data. There are several approaches to representing and reconstructing static or dynamic street scenes. Classified by the representation, many driving dataset~\cite{geiger2013visionkitti,sun2020scalabilitywaymo,maddern20171robotcar, caesar2020nuscenes} provide point cloud of the scene by fusing the calibrated LiDAR scans, serves as benchmarks for point cloud reconstruction on street scene using local LiDAR frames~\cite{shan2020lio, zhang2014loam, shan2018lego} or multi-view images~\cite{schoenberger2016mvs, schoenberger2016sfm}. 


Neural Radiance Field~\cite{mildenhall2020nerf} utilizes learnable density and appearance fields to represent 3D scenes implicitly. While it achieves high-performance rendering quality in novel view synthesis, adopting it to driving scenarios faces challenges like large-scale scenes, low-overlap views, unbounded space, etc.
Significant progress has been made in adopting NeRFs to driving scenes. Block NeRF~\cite{tancik2022block} develops a splitting-and-merging schema to reconstruct very large scenes; Urban NeRF~\cite{rematas2022urban} and DNMP~\cite{lu2023urban} leverage additional LiDAR input, which is usually available on autonomous vehicles, to enhance geometry information. S-NeRF~\cite{xie2023s} specifically addresses the low-overlap problem by disentangling background and foreground, utilizing LiDAR supervision, and specially designed camera parameterization for the ego car. 
UniSim~\cite{yang2023unisim} leverages NeRF and performs closed loop evaluation.
EmerNeRF~\cite{yang2023emernerf} and Julian et al.~\cite{ost2021neural} decompose dynamic and static scenes to achieve comprehensive reconstruction for dynamic street scenes.

3DGS explicitly represents scenes as a set of attributed 3D Gaussians. It achieves real-time differentiable novel view synthesis (NVS) by developing a GPU-friendly rasterization procedure. Many works have achieved modeling dynamic street scene~\cite{yan2024street, fischer2024dynamic, zhou2024drivinggaussian, chen2023periodic}. OmniRe~\cite{chen2024omnire} notably achieves panoptic reconstruction of 4D street scenes, including background environment, vehicles, pedestrians, and other deformable dynamic objects with 3DGS and SMPL. SGD~\cite{yu2024sgd} tackled the sparse view problem of driving frames by utilizing diffusion prior and sparse LiDAR scans. VastGaussian~\cite{lin2024vastgaussian} and CityGaussian~\cite{liu2024citygaussian} specifically improve large-scale scene reconstruction by divide-and-conquer techniques. 

Another line of work explores feed-forward Transformer-based models that map multi-view images directly to pixel-aligned point maps, as demonstrated by DUSt3R~\cite{wang2024dust3r} and its metric-scale extension MASt3R~\cite{leroy2024grounding}. In the context of autonomous driving, STORM~\cite{yang2024storm} and DrivingForward~\cite{tian2025drivingforward} unify diverse driving datasets to train Transformers capable of inferring dynamic 3D scenes from sparse input images, enabling high-fidelity rendering and near real-time reconstruction. This feed-forward process can be further enhanced by incorporating additional sensor inputs, such as LiDAR, as shown in StreetCrafter~\cite{yan2024streetcrafter}.


\begin{figure}[ht]
    \centering
    \includegraphics[width=0.99\linewidth]{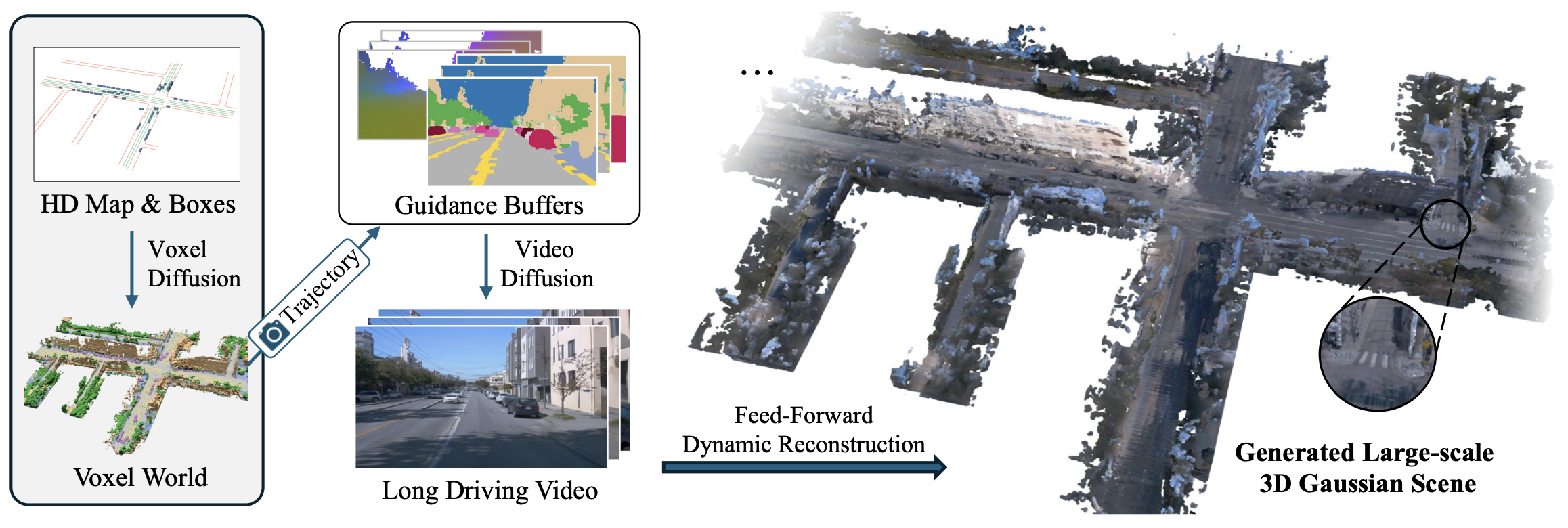}
    \caption{InfiniCube~\cite{lu2024infinicube} generates large-scale dynamic 3D driving scenes from the HD map, 3D bounding boxes and text
controls.}
    \label{fig:3d4d-infini}
\end{figure}

\paragraph{3D Driving Scene Generation}
In this section, we divide 3D/4D driving scene generation approaches into two types: (1) Approaches that generate a 3D representation directly. (2) Approaches that utilize 3D representations as intermediate geometry constraints and generate non-3D outputs (images, videos, etc).

\textbf{3D as Generated Representation}
These approaches aim to generate dynamic or static 3D representations for driving scenes. As illustrated in~\ref{fig:3d4d-infini}, InfinCube~\cite{lu2024infinicube} achieves infinite scale dynamic driving scene generation in an outpaint-and-fuse manner, with the dynamic driving scene extended by a fine-tuned controllable video diffusion model and dual-branch 3D reconstruction module. MagicDrive3D~\cite{gao2024magicdrive3d} utilized a diffusion prior of video generative models and aligned the monocular depth prediction across frames to generate the 3DGS representation for the static driving scene. DreamDrive \cite{mao2024dreamdrive} similarly relies on 3DGS to model driving scenes in 3D, then applies diffusion to generate future scenes.
DiST-4D~\cite{guo2025dist} predicts metric depth by leveraging prior multi-view RGB sequences and performing novel-view synthesis at existing camera positions, thereby optimizing a generalizable spatiotemporal diffusion model.

\textbf{3D as Intermediate Representation}
These approaches aim to generate driving video or images with the assistance of 3D representations like voxels, point clouds, etc. Different from the above category, these methods do not generate 3D representations that support novel view rendering at any given trajectory once the scene is generated. WoVoGen~\cite{lu2025wovogen} generates future videos conditioned on past observations and driving actions. Employ diffusion models as a generative prior and develop a world volume representation to enhance geometric consistency. Similarly, Stag-1~\cite{wang2024stag} generated multi-view driving videos of dynamic 3D scenes, by using a project-and-outpaint paradigm utilizing diffusion models as a generative prior and point clouds as geometry constrain. ChatSim~\cite{wei2024chatsim} developed a multi-agent system to generate photo-realistic videos for dynamic 3D scenes, where NeRFs are employed as background 3D representation and 3D assets to represent vehicles. 

\begin{table}[h!]
\centering
\caption{\textbf{Comparison of 3D/4D Generation Methods.} In the condition column, M stands for maps, I for images or videos, B for 3d bounding boxes or layout, J for trajectory, T for text, O for opacity, C for camera, and A for driving action. $^*$ means not presented in the original paper but further supported afterward. $^\dagger$ means reconstruction models with a generative prior.}
\label{table:3d4d_gen}
\resizebox{\textwidth}{!}{
\begin{tabular}{llllllll}
    \toprule
    \textbf{Method} & \textbf{Venue} & \textbf{Task} & \textbf{Modeling Type} & \textbf{Backbone}  & \textbf{Condition} & \textbf{Output} & \textbf{Code}\\
    \midrule
    InfiniCube~\cite{lu2024infinicube} & ArXiv'24 & 4D Gen. & 3DGS, Diffusion & 3D U-Net, ControlNet, DiT & MBJT & Video, 3DGS & N/A \\
    WoVoGen~\cite{lu2025wovogen}  & ECCV'24 & 4D Gen. & Diffusion & 3D U-Net, Transformer &  MOTA & Video & \href{https://github.com/fudan-zvg/WoVoGen}{Github}\\
     DriveX~\cite{yang2024driving}  & ArXiv'24 & 4D Gen. & Diffusion & U-Net & MOTA & Video, 3DGS & \href{https://github.com/fudan-zvg/DriveX}{\color{red} Github}\\
     ChatSim~\cite{wei2024chatsim}   & CVPR'24 & 4D Gen. & NeRF, 3DGS$^*$ & Transformer & IT & Video & \href{https://github.com/yifanlu0227/ChatSim}{Github}\\ 
     MagicDrive3D~\cite{gao2024magicdrive3d}   & CORR'24 & 4D Gen. & 3DGS & MLP &  TEBJ & Video, 3DGS & \href{https://github.com/flymin/MagicDrive3D}{\color{red} Github}\\     
     DreamDrive~\cite{mao2024dreamdrive}   & Arxiv'24 & 4D Gen. & 3DGS , Diffusion & MLP &  IJ & Video, 3DGS & N/A\\
     \midrule
      OmniRe~\cite{chen2024omnire}   & ICLR'25 & 4D Rec. & 3DGS, Graph & N/A  & I(CD) & 3DGS, SMPL & \href{https://github.com/ziyc/drivestudio}{Github}\\ 
    CoDa-4DGS~\cite{song2025coda}   & ArXiv'25 & 4D Rec. & 3DGS & MLP  & ICD & 3DGS & N/A\\ 
     4DGF~\cite{fischer2024dynamic}   & NeurIPS'24 & 4D Rec. & 3DGS, Graph & N/A  & IC(D) & 3DGS & \href{https://github.com/tobiasfshr/map4d}{Github}\\ 

     StreetGaussian~\cite{yan2024street}  & ECCV'24 & 4D Rec. & 3DGS & N/A  & ICD & 3DGS & \href{https://github.com/zju3dv/street_gaussians}{Github}\\ 
     DrivingGaussian~\cite{zhou2024drivinggaussian}  & CVPR'24 & 4D Rec. & 3DGS & N/A , Graph & ICD & 3DGS & N/A \\ 
     SGD~\cite{yu2024sgd}  & CORR'24 & 4D Rec.$^\dagger$ & 3DGS & U-Net, ControlNet  & ITCD & 3DGS & N/A \\
     EmerNeRF~\cite{yang2023emernerf}  & ICLR'24 & 4D Rec. & NeRF & MLP  & ICD & NeRF & \href{https://github.com/NVlabs/EmerNeRF}{Github}\\ 
  VastGaussian~\cite{lin2024vastgaussian} & CVPR'24 & 3D Rec. & 3DGS & CNN  & IC & 3DGS & N/A\\
   CityGaussian~\cite{liu2024citygaussian} & ECCV'24 & 3D Rec. & 3DGS & N/A  & IC & 3DGS & \href{https://github.com/Linketic/CityGaussian}{Github}\\ 
    DNMP~\cite{lu2023urban}  & ICCV'23 & 3D Rec. & Voxel, Mesh& MLP  & ICD & Voxel,Mesh & \href{https://github.com/DNMP/DNMP}{Github}\\ 
     S-NeRF~\cite{xie2023s}  & ICLR'23 & 3D Rec.& NeRF & MLP  & ICD & NeRF & \href{https://github.com/fudan-zvg/S-NeRF}{Github}\\
     
    BlockNeRF~\cite{tancik2022block}  & CVPR'22 & 3D Rec.  & NeRF& MLP & IC & NeRF & N/A \\
    UrbanNeRF~\cite{rematas2022urban}  & CVPR'22 & 3D Rec.  & NeRF & MLP& ICD & NeRF & N/A \\

     Julian et al.~\cite{ost2021neural} & CVPR'21 & 4D Rec. & NeRF, Graph & MLP & IC & NeRF & \href{https://github.com/princeton-computational-imaging/neural-scene-graphs}{Github}\\ 
  STORM~\cite{yang2025storm} & ICLR'25 & 4D Rec. & 3DGS & Transformer & IC & 3DGS & \href{https://github.com/NVlabs/GaussianSTORM}{Github}\footnote{Not Yet Released as of Apr.2025}\\ 
  \bottomrule
\end{tabular}
}
\end{table}

\subsection{Editing}
\label{sec:editing}

Scene editing is an emerging yet relatively underexplored area of autonomous driving. The are two main task directions: \textbf{image editing} and \textbf{3D editing}. Both directions have the goal to manipulate raw sensor data by seamlessly adding, removing, or modifying objects within the scene. This capability allows for the creation of diverse, contextually accurate scenarios, which are critical for training and evaluating autonomous systems. By enabling precise scene alterations, scene editing addresses key challenges such as generating rare edge cases and improving data diversity for robust perception models.

\paragraph{Image Editing}

\begin{figure}[!h]
  \setkeys{Gin}{width=\linewidth}
    \includegraphics{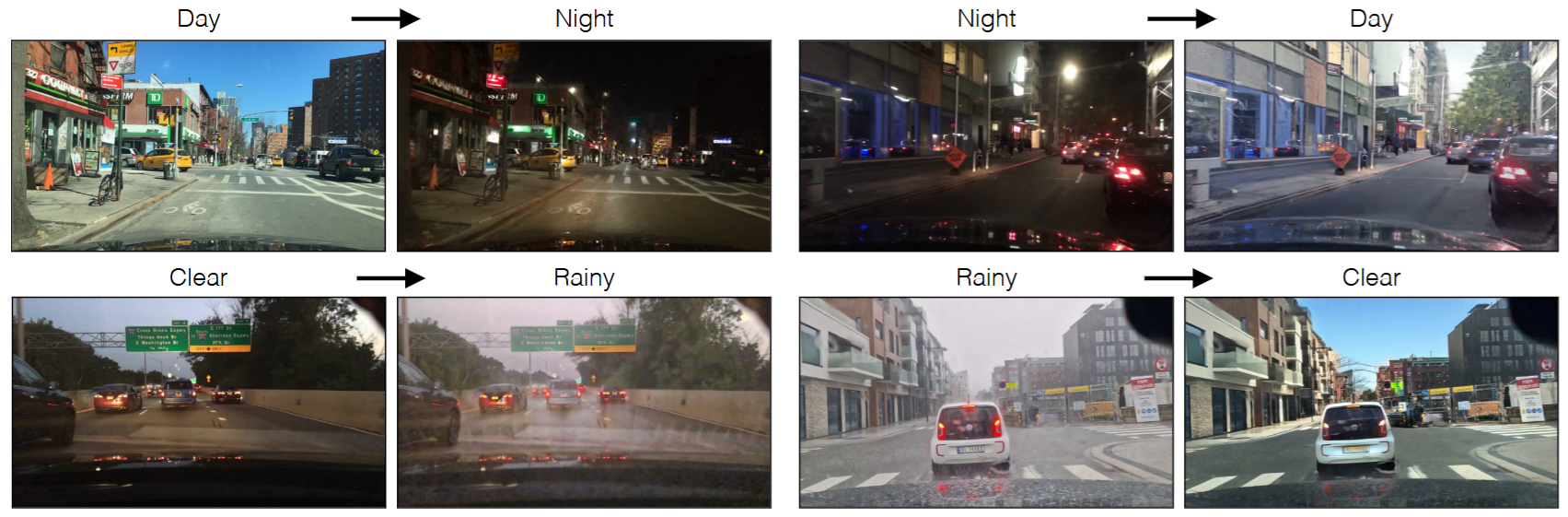}
\caption{Example of image editing on weather by One-Step Image Translation \cite{parmar2024one}.}
\end{figure}

Image editing aims to edit a camera image directly, without the need to comprehend the driving scene in 3D. Historically in the wider computer vision domain, direct image editing relies on paired image data to map an image from one domain to another, such as Pix2Pix \cite{isola2017image}, SPADE \cite{park2019semantic}, Scribbler \cite{sangkloy2017scribbler}, SEAN \cite{zhu2020sean}, SpaText \cite{avrahami2023spatext}, InstructPix2Pix \cite{brooks2023instructpix2pix}, GLIGEN \cite{li2023gligen}, Palette \cite{saharia2022palette}, and Re-Imagen \cite{zhang2023adding}. In the autonomous driving domain, although there are similar works \cite{qian2025allweather,zhu2024mwformer,li2024v2x, hetang2023novel} that performs derain, dehaze, de-snow, or other restoration~\cite{wu2024one,mei2024codi,tu2022maxim} tasks on driving images, they require the paired image data that is not as available as in general vision tasks (for example, it is difficult to collect the exact same driving scene in both clear day and rain). This necessitates a more data-friendly method that does not rely on paired data. One-Step Image Translation \cite{parmar2024one} introduces a text-conditioned, diffusion-based image translation method. Specifically, the authors apply a CLIP \cite{radford2021learning} text conditioner on top of a pretrained StableDiffusion \cite{rombach2021highresolution} model to enable weather, lighting changes on a driving scene without the need for a paired ground truth. 

\paragraph{3D Editing}

\begin{figure}[!h]
  \setkeys{Gin}{width=\linewidth}
    \includegraphics{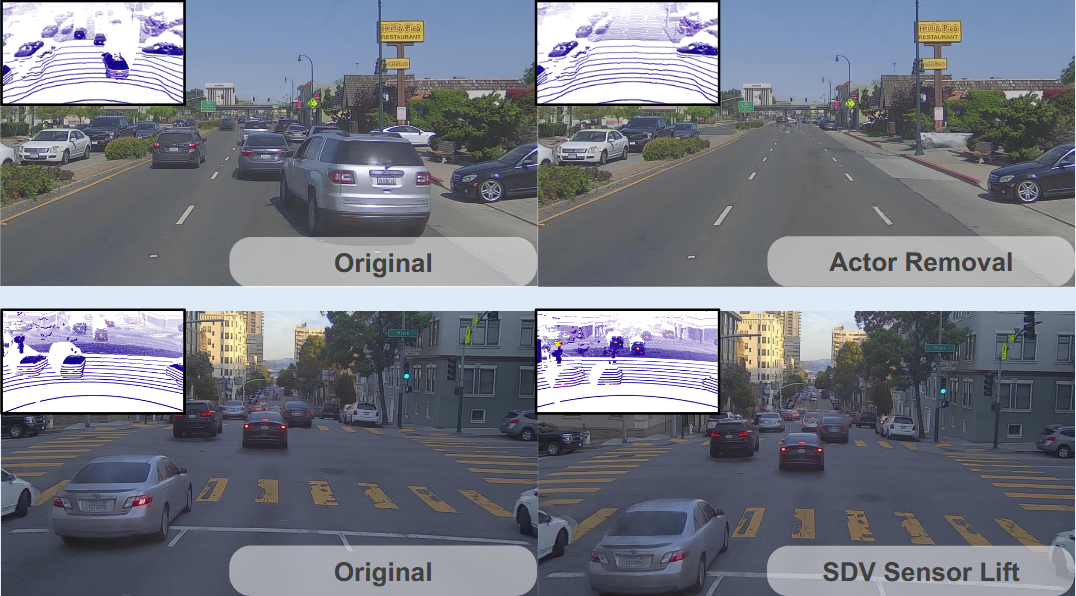}
\caption{Example of vehicle removal and manipulation in UniSim~\cite{yang2023unisim}.}
\end{figure}

It is important to note that many 3D reconstruction works are also suitable for 3D editing. These works include SimGen \cite{zhou2024simgen, tonderski2024neurad, fischer2024dynamic, chen2024omnire}. Please refer to the details in \ref{sec:3d4d_gen}. In this section, we focus on works that solely aim for 3D editing. 3D editing edits a driving scene after comprehending the scene. One dominant paradigm is to employ two sequential steps, where it first learns the representation of the static background and then learns that of the moving objects. UniSim \cite{yang2023unisim} adopts a NeRF-based backbone, using multi-resolution voxel grids to represent static backgrounds and neural shape priors to model dynamic actors. DrivingGaussian \cite{zhou2024drivinggaussian} leverages Gaussian splatting to decompose scenes into incremental Static 3D Gaussians and dynamic Gaussian graph components. StreetGaussian \cite{yan2024streetgaussian} applies 3DGS to both the background and the objects, but the objects are additionally appended with their historical poses. CoDa-4DGS \cite{song2025coda} employs deformable Gaussians and context feature distillation to achieve dynamic scene rendering and semantic-based scene editing in 4D autonomous driving environments. Generative LiDAR Editing \cite{ho2024generativelidar} approach focuses on LiDAR point clouds and uses background inpainting on spherical voxelization to extract objects from a static background. DriveEditor \cite{liang2024driveeditor} relies on a pretrained SegmentAnything \cite{kirillov2023segment} to infer the 3D bounding box of the agent of interest from camera images.

Once the representations are learned, the editing and rendering processes are tailored to each method's backbone and scene requirements. UniSim \cite{yang2023unisim} employs a NeRF-based framework where edits, such as adding or removing actors and modifying their trajectories, are applied directly to the neural feature fields. The edited representations are rendered using a voxel-based approach with neural feature interpolation, ensuring photorealism and consistency across both static and dynamic elements. DrivingGaussian \cite{zhou2024drivinggaussian} utilizes Gaussian splatting, where edits are applied by modifying the Gaussian primitives representing dynamic objects or background elements. The rendering process aggregates Gaussian contributions to produce smooth and realistic results, integrating seamlessly into multi-camera views and LiDAR data. StreetGaussian \cite{yan2024streetgaussian} extends Gaussian splatting to urban-scale scenes, where edits such as introducing vehicles or pedestrians are applied to the multi-resolution Gaussian representations. The rendering process consolidates these changes into the static background while maintaining high photorealism, even in dense urban settings. CoDa-4DGS \cite{song2025coda} integrates LSeg \cite{li2022languagedriven} to perform feature distillation on 2D images. By rasterizing these semantic features into the Gaussians and jointly deforming them over time, it enables semantic-driven operations, such as moving, removing, or adding Gaussians, during scene editing, thereby synthesizing new scenes. These methods demonstrate flexibility in both editing and rendering, ensuring that changes are coherent and realistic for autonomous driving scenarios. Generative LiDAR Editing  \cite{ho2024generativelidar} uses generative inpainting to edit the point cloud representations of dynamic objects, enabling object additions, removals, and spatial rearrangements. The rendering step voxelizes the edited LiDAR data and integrates the changes into the static background, ensuring robust spatial consistency and high fidelity. DriveEditor \cite{liang2024driveeditor} takes out the agent of interest as an image patch, fills its pixels with a mask, and uses CLIP \cite{radford2021clip} conditioned video diffusion SV3D \cite{voleti2024sv3d} to simultaneously fill the mask and reposition/insert the agent in a designated location.

\begin{table}[h!]
\centering
\caption{Summary of 3D Scene Editing Methods. Here we note their supported operations and output format.}
\resizebox{\textwidth}{!}{
\begin{tabular}{llllllll}
\toprule
\textbf{Method} & \textbf{Modeling Type} & \textbf{Insertion} & \textbf{Removal}& \textbf{Manipulation} & \textbf{Camera} & \textbf{LiDAR} & \textbf{Code} \\ \midrule
UniSim \cite{yang2023unisim}  & NeRF & \checkmark & \checkmark & \checkmark & \checkmark & \checkmark & N/A \\
DrivingGaussian \cite{zhou2024drivinggaussian} & 3DGS & \checkmark & & & \checkmark & & \href{https://github.com/VDIGPKU/DrivingGaussian}{Github} \\ 
StreetGaussian \cite{yan2024streetgaussian} & 3DGS & \checkmark & \checkmark & \checkmark & \checkmark &  & N/A\\ 
CoDa-4DGS \cite{song2025coda} & 3DGS, LSeg & \checkmark & \checkmark & \checkmark & \checkmark &  & N/A\\
Generative LiDAR \cite{ho2024generativelidar} & Generative Inpainting & \checkmark & \checkmark & \checkmark & \checkmark & \checkmark & N/A\\ 
DriveEditor \cite{liang2024driveeditor} & SAM, Video Diffusion & \checkmark & \checkmark & \checkmark &  & \checkmark & N/A\\ 
\bottomrule
\end{tabular}
}
\label{tab:scene_editing_summary}
\end{table}





\subsection{Large Language Models}
\label{sec:llm_gen}
The emergence of powerful LLMs~\citep{devlin2018bert, radford2019gpt2, brown2020gpt3, team2023gemini, roziere2023codellama, touvron2023llama, touvron2023llama2, raffel2020t5, qwen2, qwen2.5} has revolutionized the fields of multi-round chat~\citep{openai2023gpt4,touvron2023llama2}, instruction following~\citep{ouyang2022instructgpt} and reasoning~\citep{roziere2023codellama,azerbayev2023llemma,plum}. Building on their capacity to interpret and synthesize complex information, LLMs hold significant promise for autonomous driving. By processing diverse data from real-time sensors and intricate driving scenarios, LLMs can enhance decision-making, bolster safety, and expedite the adoption of advanced driver-assistance systems. 

\begin{table}[!]
\centering
\caption{\textbf{Comparison of LLM-based Autonomous Driving Systems.} In the condition column, QA stands for question answering, PL stands for planning, DM for decision making, ED for environment description, SU for scene understanding, and DC for driving context.}
\label{table:llm}
\resizebox{\textwidth}{!}{
\begin{tabular}{llllllllllllll}
    \toprule
    \textbf{Method} & \textbf{Venue} & \textbf{Interaction} & \textbf{Task} & \textbf{Scenario} & \textbf{Backbone} & \textbf{Strategy} & \textbf{Input} & \textbf{Output} & \textbf{Code} \\
    \midrule
    Dilu~\cite{wen2023dilu} & ArXiv'23 & Prompting & QA & DM & GPT-4~\citep{gpt4} & ReAct~\citep{yao2023react} & ED & Action & \href{https://github.com/PJLab-ADG/DiLu}{Github} \\ 
    Drive-Like-A-Human~\cite{fu2024drive}  & WACV'24 & Prompting & QA & DM & GPT-3.5~\citep{chatgpt} & ReAct~\citep{yao2023react} & ED & Action & \href{https://github.com/PJLab-ADG/DriveLikeAHuman}{Github}\\
    Driving-with-LLMs~\citep{chen2024driving}  & ICRA'24 & Fine-tuning & QA & SU & LLaMA-7b~\citep{touvron2023llama} & None & Question & Answer & \href{https://github.com/wayveai/Driving-with-LLMs}{Github}\\ 
    LaMPilot~\cite{ma2024lampilot}  & CVPR'24 & Prompting & QA  & SU & General LLMs & PoT~\citep{pot} & Instruction, DC & Code  & \href{https://github.com/PurdueDigitalTwin/LaMPilot}{Github}\\ 
    LLaDA~\citep{li2024driving} & CVPR'24 & Prompting & QA & DM & GPT-4~\citep{gpt4} & CoT~\citep{wei2022cot} & Intended Command & Action & \href{https://github.com/Boyiliee/LLaDA-AV}{Github} \\
    \midrule
    GPT-driver~\citep{mao2023gpt} & NeurIPS'23 & Fine-tuning & PL & E2E & GPT-3.5~\citep{chatgpt} & CoT~\citep{wei2022cot} & Instruction, DC & Object, Action, Trajectory & \href{https://github.com/PointsCoder/GPT-Driver}{Github} \\
    Talk2Drive~\citep{cui2023large} & ITSC'24 & Prompting & PL & E2E & GPT-4~\citep{gpt4} & CoT~\citep{wei2022cot} & Instruction, DC &  Executable Controls & \href{https://github.com/PurdueDigitalTwin/Talk2Drive}{Github} \\
    Agent-Driver~\citep{mao2023language} & COLM'24 & Prompting & PL & E2E & GPT-3.5~\citep{chatgpt} & ReAct~\citep{yao2023react} & Observation & Object, Action, Trajectory & \href{https://github.com/USC-GVL/Agent-Driver}{Github} \\
  \bottomrule
\end{tabular}
}
\end{table}

\paragraph{Question Answering}
Early applications of LLMs in autonomous driving primarily leverage their basic capabilities, using them as expert chatbots to assist with perception, prediction, and planning tasks, ultimately supporting the decision-making process. Typically, key cues, such as the status of the ego vehicle or details about the driving scenario, are first translated into natural language and then processed by the LLM. The LLM is then prompted to generate responses based on this linguistic input in a question-answering format. In general, this type of LLM application in autonomous driving centers around question-answering tasks \cite{huang2024drivlme, ma2022dorothie}. Dilu~\cite{wen2023dilu} introduces a knowledge-driven decision-making framework that leverages LLMs with few-shot prompting. This framework comprises four core modules—Environment, Reasoning, Reflection, and Memory—to collaboratively guide the decision-making process. Building on this approach, Drive-Like-A-Human~\cite{fu2024drive} proposes a closed-loop autonomous driving system for meta decision-making using GPT-3.5, demonstrating human-like driving capabilities. \citep{chen2024driving} proposes a multimodal architecture that fuses an object-level vectorized numeric modality with LLaMA-7b~\citep{touvron2023llama}, demonstrating the model’s capabilities through fine-tuning. LaMPilot~\cite{ma2024lampilot} leverages Program-of-Thought prompting techniques to bolster the instruction-following capabilities of large language models integrated into autonomous driving systems during perception, prediction, and planning. LLaDA \citep{li2024driving} introduces a training-free mechanism to assist human drivers and adapt autonomous driving policies to unfamiliar environments. \cite{xing2025can} pioneers VLM-based navigation on human-readable maps, such as Google Maps. By leveraging the zero-shot generalizability of LLMs, LLaDA can be integrated into any autonomous driving stack, improving performance in locations with different traffic rules—without requiring additional training. 

\paragraph{Motion Planning}
In contrast to question-answering tasks, LLM-based autonomous driving systems for planning tasks treat autonomous driving as a holistic sequence modeling problem, where raw or structured language-based representations are directly mapped to driving actions. The final output typically consists of either executable controls or future ego-vehicle status (\textit{e.g.}, speed, curvature, and waypoints), enabling the vehicle to navigate dynamically and adaptively in complex environments. GPT-driver~\citep{mao2023gpt} builds on the powerful reasoning and generalization capabilities of GPT-3.5 to develop an end-to-end autonomous driving motion planning system by formatting both inputs and outputs into language tokens. Talk2Drive~\citep{cui2023large} presents an LLM-based framework that translates natural verbal commands into executable controls in a human-in-the-loop manner, enabling the system to learn and adapt to individual preferences. To further enhance the capabilities of LLM-based autonomous driving systems, Agent-Driver~\citep{mao2023language} is the first to introduce an agentic framework that transforms the traditional autonomous driving pipeline, which integrates a versatile tool library, a cognitive memory, and a reasoning engine. 

Noted that, despite these advances, the LLM-based autonomous driving systems still face a fundamental limitation: the absence of a native visual perception module. As a result, they are unable to process the full perception-to-action pipeline within a unified framework.

\subsection{Multimodal Large Language Models}
\label{sec:mllm_gen}

By integrating vision encoders such as CLIP~\citep{radford2021clip}, the powerful capabilities of LLMs can be extended to the visual domain~\citep{li2022blip, li2023blip2, liu2024llava, llavanext, llama3.2, Qwen-VL, Qwen2VL}. These encoders convert image patches into tokens and align them with text token embeddings, enabling unified multimodal understanding. Consequently, MLLMs can seamlessly process and reason over both textual and visual inputs, supporting tasks like visual question answering (VQA)~\citep{antol2015vqa, hudson2019gqa, gurari2018vizwiz, chen2025interleaved, singh2019textvqa, bao2024autobench} and image captioning~\citep{chen2015microsoftcoco, agrawal2019nocaps}. This unified architecture also lays the foundation for developing end-to-end autonomous driving systems capable of effectively handling the entire perception-to-action pipeline.

\begin{table}[!]
\centering
\caption{\textbf{Comparison of MLLM-based Autonomous Driving Systems.} In the condition column, VQA stands for visual question answering, PL stands for planning, SU for scene understanding, DS for driving scene,  MVF for multi-view frame, and TC for transportation context.}
\label{table:mllm}
\resizebox{\textwidth}{!}{
\begin{tabular}{llllllllllllll}
    \toprule
    \textbf{Method} & \textbf{Venue} & \textbf{Interaction} & \textbf{Task} & \textbf{Scenario} & \textbf{Backbone} & \textbf{Strategy} & \textbf{Input} & \textbf{Output} & \textbf{Code} \\
    \midrule
    HiLM-D~\citep{ding2023hilm} & ArXiv'23 & Prompting & VQA & SU &  MiniGPT-4~\citep{zhu2023minigpt} & None & Question, DS (Video)  & Answer & N/A \\
    DriveLM~\citep{sima2023drivelm}  & ECCV'24 & Fine-tuning & VQA & SU & BLIP-2~\citep{li2023blip2} & CoT~\citep{wei2022cot} & Question, DS (Image)n & Answer & \href{https://github.com/OpenDriveLab/DriveLM}{Github}\\
    Dolphins~\citep{ma2024dolphins} & ECCV'24 & Fine-tuning & VQA & SU & OpenFlamingo~\citep{awadalla2023openflamingo} & CoT~\citep{wei2022cot} & Question, DS (Video) & Answer & \href{https://vlm-driver.github.io/}{Github}\\
    EM-VLM4AD~\citep{em-vlm4d} & CVPR'24 & Fine-tuning & VQA & SU & T5/T5-Large~\citep{raffel2020t5} & None & Question, DS (MVF) & Answer & \href{https://github.com/akshaygopalkr/EM-VLM4AD}{Github}\\ 
    LLM-Augmented-MTR~\citep{zheng2024large} & IROS'24 & Prompting & VQA  & SU & GPT-4V~\citep{gpt4v} & CoT~\citep{wei2022cot} & Instruction, TC-Map & Context Understanding  & \href{https://github.com/SEU-zxj/LLM-Augmented-MTR}{Github}\\ 
    \midrule
    LMDrive~\citep{shao2024lmdrive} & CVPR'24 & Fine-tuning & PL & E2E & LLaVA-v1.5~\citep{liu2024llava} & CoT~\citep{wei2022cot} & Instruction, DS (MVF), LiDAR & Control Signal & \href{https://github.com/opendilab/LMDrive}{Github} \\
    LeGo-Drive~\citep{paul2024lego} & IROS'24 & Fine-tuning & PL & E2E &  CLIP~\citep{radford2021clip} & None & Instruction, DS (Image) & Trajectory & \href{https://github.com/reachpranjal/lego-drive}{Github} \\
    RAG-Driver~\citep{yuan2024rag} & ArXiv'24 & Fine-tuning & PL & E2E &  ViT-B/32~\citep{vit}, Vicuna-1.5~\citep{vicuna1.5} & RAG & Instruction, DS (Video) & Action, Trajectory & \href{https://github.com/YuanJianhao508/RAG-Driver}{Github} \\
    DriveVLM~\citep{tian2024drivevlm} & CoRL'24 & Fine-tuning & PL & E2E &  Qwen-VL~\citep{Qwen-VL} & CoT~\citep{wei2022cot} & Instruction, DS (Video) & Action, Trajectory & N/A\\
    EMMA~\citep{hwang2024emma} & ArXiv'24 & Fine-tuning & PL & E2E &  Gemini 1.0 Nano-1~\citep{team2023gemini} & CoT~\citep{wei2022cot} & Instruction, DS (MVF) & Object, Action, Trajectory & N/A\\    OpenDriveVLA~\citep{zhou2025opendrivevla} & ArXiv'25  & Fine-tuning & PL & E2E &  Qwen2.5 & None & Instruction, DS (MVF) & Action, Trajectory & \href{https://github.com/DriveVLA/OpenDriveVLA}{Github}\\
    OpenEMMA~\citep{xing2025openemma} & WACV'25  & Prompting & PL & E2E &  General MLLMs & CoT~\citep{wei2022cot} & Instruction, DS (Image) & Object, Action, Trajectory & \href{hhttps://github.com/taco-group/OpenEMMA}{Github}\\
    
  \bottomrule
\end{tabular}
}
\end{table}

\paragraph{Perception and Prediction} With the incorporation of a vision/video encoder, MLLMs can directly process visual information from driving scenarios. By leveraging the pretrained knowledge of large language models, MLLMs are capable of understanding complex driving scenes, identifying key objects and events, and performing high-level reasoning and analysis to support decision-making in autonomous driving systems. HiLM-D~\citep{ding2023hilm} leverages MLLMs to process driving scene videos and generate natural language that simultaneously identifies and interprets risk objects, understands ego-vehicle intentions, and provides motion suggestions—eliminating the need for task-specific architectures. DriveLM~\citep{sima2023drivelm} introduced Graph VQA to model graph-structured reasoning for perception, prediction, and planning in question-answer pairs. Based on this, it further leverages the approach proposed in RT-2~\citep{brohan2023rt} to develop an end-to-end DriveVLM by converting the action to the trajectory. Dolphins~\citep{ma2024dolphins}, building on OpenFlamingo~\citep{awadalla2023openflamingo}, enhances fine-grained visual reasoning by leveraging large-scale public VQA datasets. To adapt these capabilities to the autonomous driving domain, Dolphins is further trained on a custom VQA dataset constructed from the BDD-X dataset~\citep{kim2018bdd-x}. The model is capable of processing rich multimodal inputs—including video or image sequences, textual instructions, and historical vehicle control signals—to generate contextually grounded and instruction-aware responses. Building upon the DriveLM dataset~\citep{sima2023drivelm}, EM-VLM4AD~\citep{em-vlm4d} proposes an efficient and lightweight multi-frame vision-language model tailored for the visual question answering (VQA) task in autonomous driving scenarios. DriveVLM~\citep{tian2024drivevlm} introduces a MLLM-based framework that incorporates the Chain-of-Thought (CoT) reasoning paradigm~\citep{wei2022cot} that enables more sophisticated scene understanding and decision-making in complex driving environments. LLM-Augmented-MTR~\citep{zheng2024large} leverages GPT-4V to interpret driving scenarios from visualized images using carefully crafted prompts. The model generates rich transportation context information, which augments traditional motion prediction algorithms and enhances their performance in complex environments.

\begin{figure}[!h]
  \setkeys{Gin}{width=\linewidth}
    \includegraphics{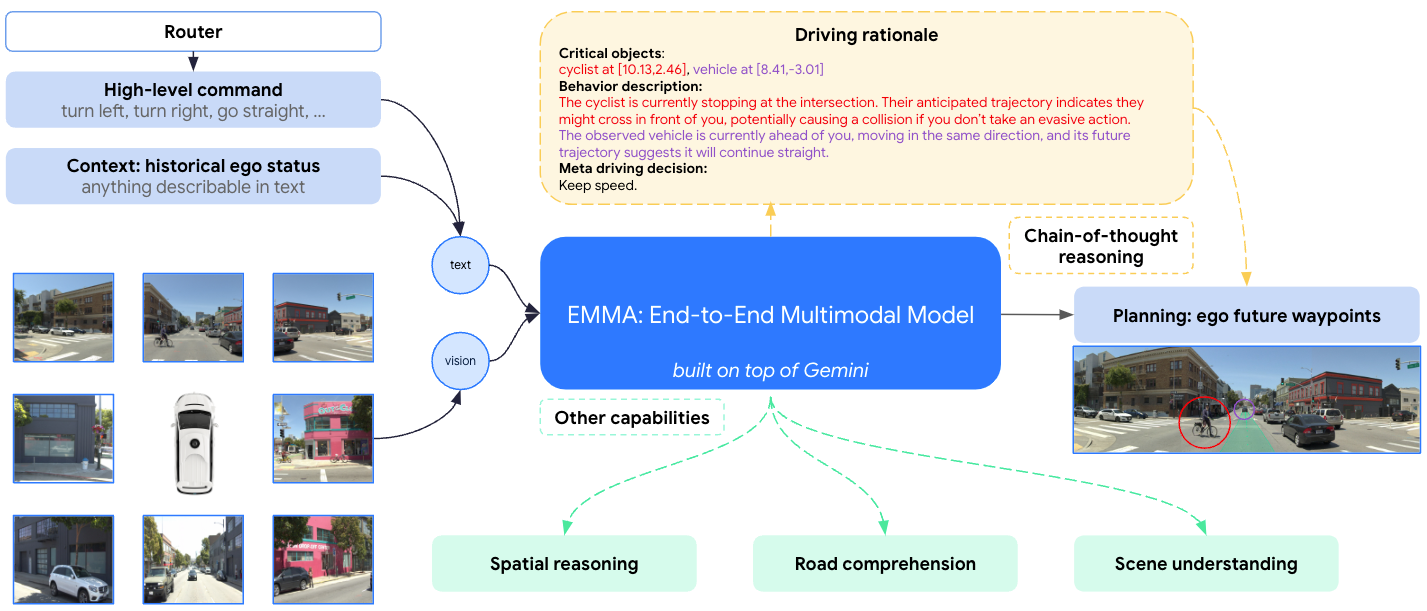}
\caption{Example of MLLM-based end-to-end autonomous driving system in EMMA~\cite{hwang2024emma}.}
\end{figure}

\paragraph{End-to-End Systems}
The powerful reasoning capabilities and generalization abilities of MLLMs have enabled the development of language-guided, end-to-end autonomous driving systems. These systems integrate perception, planning, and control within a unified framework. By leveraging MLLMs, such systems can interpret complex commands, adapt to novel scenarios, and provide interpretable decision-making processes, marking a significant step toward more flexible and human-aligned autonomous vehicles. LMDrive~\citep{shao2024lmdrive} presents a novel language-guided, end-to-end, closed-loop autonomous driving framework that interacts with dynamic environments using multimodal, multi-view sensor data and natural language instructions based on LLaVA-v1.5. Instead of producing long-term trajectory waypoints, LeGo-Drive~\citep{paul2024lego} proposes an approach to estimate a goal location based on the given language command as an intermediate representation in an end-to-end setting. While RAG-Driver~\citep{yuan2024rag} introduces a novel retrieval-augmented in-context learning framework for MLLM-based, generalizable, and explainable end-to-end autonomous driving, it demonstrates exceptional zero-shot generalization to previously unseen environments—achieving this without the need for additional training or fine-tuning. Building upon DriveVLM~\citep{tian2024drivevlm}, the authors introduce DriveVLM-Dual, a hybrid autonomous driving system that integrates the vision-language capabilities of DriveVLM with a traditional modular pipeline. This dual framework enhances spatial reasoning and significantly improves real-time decision-making and planning performance in complex driving scenarios. EMMA~\citep{hwang2024emma} proposes a unified MLLM-based framework that encodes all non-sensor inputs—such as navigation instructions and ego vehicle status—and outputs—such as planned trajectories and 3D object locations—into natural language representations. This language-centric design enables seamless integration of diverse modalities and tasks within a single, general-purpose model architecture. Furthermore, OpenEMMA~\citep{xing2025openemma} leverages powerful pre-trained MLLMs, such as LLaVA-1.6~\citep{llavanext}, Qwen2-VL~\citep{Qwen2VL}, and GPT-4o~\citep{gpt4o}, to build an end-to-end AD system that incorporates CoT~\citep{wei2022cot} reasoning. Rather than directly generating the future ego-vehicle trajectory, OpenEMMA first predicts the future ego-vehicle's speed and curvature, which are then integrated to derive the future waypoints. This design allows the system to achieve competitive performance in a zero-shot, training-free manner.


\clearpage

\section{Real World Applications}
\label{sec:applications}

Generative AI has significantly reshaped autonomous driving. In this section, we cover the applications where we see the widest adoption of generative AI.

\subsection{Synthetic Data Generation}
\label{app:data_gen}

Synthetic data generation has become a pivotal component in developing and validating autonomous driving systems. By creating artificial yet realistic datasets, developers can simulate various driving scenarios, including rare or hazardous situations that are challenging to capture in real-world data collection. 

The first synthetic dataset specifically for autonomous driving tasks, FRIDA \cite{frida}, was introduced in 2010 to address depth estimation in foggy weather. This dataset, later expanded into FRIDA2 \cite{frida2}, featured synthetic road images with varying levels of fog to aid perception in adverse conditions.

A significant leap occurred in 2016 with the Flying Things and Driving datasets \cite{mayer2015large}, which leveraged 3D object models from the ShapeNet \cite{shapenet2015} database to create realistic urban street scenes. However, early synthetic datasets struggled with realism due to simplistic image splicing techniques. To address this, researchers turned to commercial video game engines. That same year, Virtual KITTI \cite{virtualkitti2016} was introduced, utilizing Unity to create synthetic versions of real-world KITTI \cite{kitti} dataset sequences. This development marked a shift toward using commercial game engines as powerful tools for synthetic dataset generation.

The first generative AI in synthetic data generation is the 2017 work Pix2Pix \cite{brooks2023instructpix2pix}, which was developed to generate semantic segmentation data from real-world images with GANs. This work is followed by Pix2PixHD \cite{wang2018pix2pixHD} in 2018, which enhanced detail generation. GAN-based models like GauGAN \cite{gaugan} and MaskGAN \cite{lee2020maskgan} further advanced scene synthesis, though challenges such as deformation artifacts remained.

With the generative AI becoming increasingly popular, data synthesis has now come to a whole new era. Below, we discuss the different data synthesis modalities, from the low-level sensor data generation to high-level scene generation.

\subsubsection{Sensor-Space Data Generation}

Sensor-space generation focuses on synthesizing raw sensor data, such as camera images or LiDAR point clouds. They can be used for training and testing perception models. Traditional simulators like CARLA produce sensor data via game engines (rendering photorealistic 3D scenes to images and simulating LiDAR through ray-casting) \cite{chen2025s}. However, purely graphics-based simulators require extensive manual asset creation and often exhibit a sim-to-real domain gap \cite{he2024neural}. Recent approaches leverage generative models to create photorealistic sensor data, reducing manual effort and narrowing the gap between virtual and real data. These techniques enable rapid generation of diverse training images and sequences, including rare corner cases, to improve model robustness.

\paragraph{Image and Video Synthesis} Camera data provides important information to the autonomous driving system that enables it to detect objects, construct road topology, recognize road signs, etc. However, real-world camera data collection involves thousands of hours of real driving and is costly \cite{nuscenes2019}. Furthermore, edge scenarios are difficult to collect. In contrast, the image generation methods described in Section \ref{sec:image_gen} offer cheap but powerful tools for synthesizing diverse and realistic driving scenes, providing critical data for training and evaluating autonomous systems. In terms of how to control image generation, methods such as BEVGen \cite{swerdlow2024street}, BEVControl \cite{yang2023bevcontrol}, MagicDrive \cite{gao2023magicdrive}, and DriveDiffusion \cite{li2025drivingdiffusion} condition the image output on interpretable priors like bird's-eye-view (BEV) layouts, road maps, or textual descriptions (\textit{e.g.}, weather, traffic conditions). These methods enable control over scene composition, making them ideal for systematically generating training samples with desired object configurations, viewpoints, and environmental conditions. Meanwhile, methods like UrbanGIRAFFE \cite{yang2023urbangiraffe}, Panoptic NeRF \cite{kundu2022panoptic}, and UrbanRF \cite{rematas2022urban} provide hierarchical and modular scene synthesis by separating static infrastructure (\textit{e.g.}, roads, buildings) from dynamic agents (\textit{e.g.}, vehicles, pedestrians). This disentanglement allows for scalable generation across large urban areas while also supporting structured editing or perturbation of individual scene elements—useful for simulating rare events or testing system robustness. These models serve as versatile engines for generating photorealistic, label-rich image datasets that capture the long-tail and multi-agent complexities inherent in real-world driving. 

Going beyond static image generation, video generation is also essential for autonomous driving. Video generation possesses the inherent difficulty of ensuring temporal consistency between images. Models like Panacea~\cite{wen2024panacea}, Delphi~\cite{ma2024unleashing}, and DriveScape~\cite{wu2024drivescape} generate high-quality driving videos conditioned on multimodal inputs such as images, BEV layouts, camera poses, and 3D structures, allowing for the creation of diverse driving situations with controllable environmental factors and agent behaviors. These approaches enhance temporal consistency via techniques such as cross-frame attention~\cite{wen2024panacea}, feature-aligned temporal modules~\cite{ma2024unleashing}, and bidirectional transformer structures~\cite{wu2024drivescape}, making them highly suitable for producing synthetic video datasets that simulate realistic motion and traffic dynamics. Meanwhile, methods like DriveDreamer\cite{wang2023drivedreamer}, Vista\cite{gao2024vista}, and DrivingWorld\cite{hu2024drivingworld} incorporate driver policy modeling into video generation through closed-loop simulation, enabling the synthetic scenes to reflect realistic ego-environment interactions and temporal progression. The integration of multimodal large language models (MLLMs), as seen in DriveDreamer2\cite{zhao2024drivedreamer}, GAIA-2 \cite{russell2025gaia}, and ChatSim~\cite{wei2024editable}, further facilitates fine-grained scene composition and interactive editing, enabling the generation of tailored, scenario-specific videos. These advancements collectively enable scalable and customizable video synthesis pipelines that contribute not only to dataset augmentation but also to simulation-based testing and policy learning in autonomous driving.

\paragraph{Evaluation of Visual Generative Models}
Evaluating the quality of generated imagery and videos is crucial, yet challenging.
Traditional perceptual quality metrics~\cite{mittal2012making,ke2021musiq,zhang2018unreasonable,tu2021rapique,tu2020comparative,zheng2024faver,he2024cover,wu2022fast,zheng2022completely,wu2023q} that are typically used to evaluate user-generated videos~\cite{tu2021ugc,tu2021rapique} often fail to fully capture how humans perceive this new type of content.
The evaluation of visual generation models requires a multi-axis protocol that accounts for photorealism, semantic fidelity, temporal coherence, and task utility:
\begin{enumerate}[noitemsep]
 \varitem{red!40}{\textbf{1)}}  \textbf{Distributional fidelity}. Fréchet-Inception Distance (FID) \cite{heusel2017gans}, Kernel-Inception Distance (KID) \cite{binkowski2018demystifying}, and their video analogue FVD \cite{unterthiner2018towards} remain the de-facto standards for image/video realism. Recent works introduce Spatial-FID (sFID) to penalize geometry distortions in BEV-conditioned images \cite{swerdlow2024street} and FlowEval \cite{wu2024drivescape}, which couples optical-flow consistency with FVD to capture motion plausibility.
\varitem{blue!40}{\textbf{2)}} \textbf{Text and image/video alignment}. Prompt-faithfulness is now routinely probed with CLIP-derived scores—CLIPScore \cite{hessel2021clipscore}, ALIGNScore \cite{zha2023alignscore}, and TIFA \cite{hu2023tifa}—while BEV-conditioned methods report Chamfer-IoU or vector-lane precision between generated images and the supervising layout \cite{yang2023bevcontrol}. Hierarchical scene generators (\textit{e.g.}, UrbanGIRAFFE) further measure static/dynamic disentanglement error by swapping infrastructures or agents and computing LPIPS \cite{zhang2018unreasonable} change.
\varitem{green!40}{\textbf{3)}}  \textbf{Temporal consistency and physical realism.}
    Metrics such as temporal stability~\cite{lai2018learning}, wrapping error~\cite{yeh2024diffir2vr}, and the newly proposed Action-FVD \cite{wu2024drivescape} explicitly penalize flicker and physically implausible dynamics. For long-horizon simulations, closed-loop Driving Score—the success rate of an RL or imitation-learning policy trained purely on the generated videos—acts as an end-to-end litmus test \cite{wang2023drivedreamer,gao2024vista}.
\varitem{purple!40}{\textbf{4)}}  \textbf{Downstream transferability} Ultimately, synthetic data must boost in-distribution and out-of-distribution performance of perception stacks~\cite{xing2024autotrust}. A common protocol trains a detector or segmentation network on varying synthetic-to-real ratios and reports mAP/mIoU on real benchmarks~\cite{lee2022fifo} (\textit{e.g.}, nuScenes \cite{nuscenes2019}, Waymo \cite{waymodataset}). Gap-to-real reduction~\cite{xu2023bridging}, rather than absolute detector accuracy alone, is the key metric.
\varitem{black!40}{\textbf{5)}}  \textbf{Human preference and safety vetting.} Paired A/B preference tests and Likert-scale surveys remain the de facto gold standard for assessing perceptual realism—especially in long-tail scenarios where automated metrics plateau. Recent toolkits~\cite{huang2023t2i,jiang2024genai} streamline this process by fusing crowd-sourced judgments with ELO-style rating schemes, yielding scalable, statistically robust quality estimates while flagging safety-critical artefacts for manual review.
\end{enumerate}

No single metric can fully capture the multifaceted notion of ``\textbf{quality}.'' Practitioners therefore favor a score-card paradigm that aggregates distributional, alignment, temporal-consistency, and task-level measures. Creating a unified, open-source evaluation suite for bespoke generative models in autonomous driving remains a pivotal open challenge.

\paragraph{3D Synthesis} To meet the geometric fidelity required for autonomous driving, recent generative methods have moved beyond 2D representations to directly model scenes in 3D space. Rather than inferring geometry implicitly from images or videos, these approaches leverage explicit representations, such as point clouds, NeRFs, and 3D Gaussian Splatting (3DGS), to reconstruct or synthesize spatially accurate, dynamic environments. Works like Block-NeRF\cite{tancik2022block} and UrbanNeRF\cite{lu2024urban} extend NeRF to large-scale urban scenes using modularity and LiDAR supervision. Similarly, 3DGS methods such as StreetGaussian\cite{yan2024street} and OmniRe\cite{chen2024omnire} explicitly model scene elements—including vehicles and pedestrians. Complementary to these, occupancy-based generation offers a compact, voxelized encoding of space, supporting robust downstream planning. Methods like UrbanDiffusion\cite{zhang2024urban} generate the 3D voxels for the static environment, such as roads, trees, and buildings. 

Direct LiDAR generation helps test the onboard LiDAR processing system. Unlike image synthesis, generating realistic LiDAR scenes poses unique challenges due to the sparse, non-uniform nature of point clouds and their strong dependence on physical sensor dynamics. Early efforts in this space were primarily physics-based, simulating LiDAR rays via handcrafted models and raycasting engines~\cite{yue2018lidarpointcloudgenerator,9157601,10161226}. While interpretable, these approaches often suffered from limited generalizability and realism. More recent advances adopt generative frameworks, particularly diffusion models, NeRF-based neural implicit representations, and transformer architectures, to improve fidelity, control, and scalability. Diffusion-based models like LiDMs~\cite{ran2024towards}, RangeLDM~\cite{hu2025rangeldm}, and LidarDM~\cite{zyrianov2024lidardmgenerativelidarsimulation} employ latent-space denoising strategies to synthesize static and dynamic point clouds, with applications ranging from full-scene generation to conditional editing and text-to-point-cloud synthesis~\cite{wu2024text2lidar}. NeRF-based approaches~\cite{zhang2023nerf,10657876,Wu2023dynfl} leverage volumetric rendering and scene transmittance modeling to generate structurally consistent LiDAR data, even enabling dynamic scene editing and sensor adaptation. In parallel, transformer-based models such as UltraLiDAR~\cite{xiong2023learning} and LidarGRIT~\cite{haghighi2024taming} utilize vector quantization and auto-regressive token prediction for completion and inpainting tasks, introducing structured generative capabilities with greater interpretability. These methods collectively demonstrate a shift toward learning-based LiDAR synthesis pipelines that offer high-quality, controllable, and temporally coherent 3D scene generation, opening new possibilities for training, validating, and simulating autonomous driving systems at scale.

\subsubsection{Trajectory Generation}

While sensor space data generation allows the training and testing for perception systems or end-to-end systems, traffic and trajectory generation enables modulized training and testing of the motion planning components \cite{cao2021spectral,xu2024matrix,arief2024importance}. Established traffic simulators provide a backbone for multi-agent simulation, and they can be enhanced with learned generative behaviors. SUMO (\textbf{S}imulation of \textbf{U}rban \textbf{MO}bility) \cite{SUMO2018}, for example, is an open-source microscopic traffic simulator that can model thousands of vehicles with configurable routes, traffic lights, and detailed dynamics. While SUMO by default uses rule-based or randomized driver models, one can inject learned agent policies (from reinforcement learning or imitation learning) to make the traffic respond more naturally \cite{ma2021reinforcement,lee2023robust}. 
Another framework, ScenarioNet \cite{li2023scenarionet}, builds a large repository of real-world traffic scenarios extracted from logs (Waymo, nuScenes, etc.) and provides a platform to replay or modify these scenarios in simulation.

While traditionally dominated by rule-based or optimization-based models, the field has seen a transformative shift toward generative paradigms, motivated by the need to model uncertainty, social interactions, and long-term multi-agent dynamics. Recent advances leverage deep generative models, including VAEs, diffusion models, and transformer-based architectures, to learn complex distributions over agent motion, enabling diverse and physically plausible future trajectories. In ego-centric settings, conditional VAEs and diffusion-based planners~\cite{kim2021driving,zheng2025diffusionbased, gu2022stochastictrajectorypredictionmotion} allow the ego vehicle to plan its own trajectory under uncertainty. Some works allow incorporating contextual cues such as intentions, sensor observations, or even natural language instructions~\cite{mao2023gpt, sima2023drivelm}. In scene-centric multi-agent scenarios, models like Trajectron++ \cite{ivanovic2019trajectron} and MotionDiffuser \cite{jiang2023motiondiffusercontrollablemultiagentmotion} capture social interactions through graph-based or permutation-invariant architectures, enabling coordinated, socially consistent trajectory sampling. Diffusion models, in particular, have shown strong performance in modeling complex joint behaviors and integrating environmental or agent-level constraints~\cite{zhang2024multiagent, westny2024diffusionbasedenvironmentawaretrajectoryprediction, you2024followgen}. The performance of diffusion models can be further enhanced by integrating physics models to reduce parameter size, increase convergency rate, and reduce input data needs \cite{long2025physical}. 
Beyond predictive use, generative methods have also revolutionized traffic simulation, replacing rule-based systems with data-driven, multi-agent simulators that support long-horizon, controllable rollouts~\cite{Suo2021TrafficSim, niedoba2023diffusion, Zhou2024BehaviorGPT, tan2024promptable}. These simulators leverage generative policies or autoregressive transformers to synthesize realistic traffic behavior, supporting scenario-based testing and closed-loop validation of driving systems. Additionally, vision-language models have emerged as powerful tools to interface with human-like reasoning, incorporating scene understanding and high-level instructions into trajectory generation~\cite{shao2024lmdrive, hwang2024emma, xing2025openemma}. Despite these advances, challenges remain in balancing fidelity and efficiency, improving controllability under constraints, and integrating explainability into stochastic generative processes—key frontiers for enabling safe, interpretable, and generalizable autonomous behavior.

\subsubsection{Traffic State Generation}

For autonomous vehicles, understanding and anticipating traffic conditions at the network level is critical for safe, efficient, and adaptive decision-making. Segment- and lane-level traffic attributes—such as average speed, density, and flow—serve as fundamental inputs for route selection, dynamic rerouting, behavior prediction, and motion planning~\cite{min2023deep}. These variables are particularly important in complex urban environments where traffic patterns fluctuate rapidly and where real-time responsiveness is essential~\cite{lee2024congestion}. Traffic state information also plays a vital role in evaluating the suitability of road segments and in defining or updating operational design domain (ODD) boundaries. As autonomous systems move toward broader deployment in mixed traffic environments, their ability to adapt to evolving traffic states becomes increasingly crucial. In this context, the capacity to generate, simulate, and forecast traffic conditions (even in the absence of real-time data) forms a critical foundation for robust and context-aware navigation.

Traffic state generation refers to the learning, estimation, and synthesis of macroscopic traffic variables across large-scale road networks. Unlike trajectory-level generation, which focuses on modeling the behavior of individual vehicles, traffic state generation provides a mesoscopic perspective that captures the collective dynamics of vehicle flows across space and time. This process typically begins with data collected from heterogeneous sources, including roadside sensors (\textit{e.g.}, loop detectors, cameras), in-vehicle sensors, and intersection control systems, which are then aggregated and processed at the network level. These data streams offer spatially and temporally rich insights into how traffic evolves across segments, corridors, and zones. By learning from such data, generative models can simulate traffic behavior across wide areas without requiring complete or continuous sensing coverage~\cite{zhang2024generative, zou2023novel}. The resulting traffic state data support a variety of applications, including digital twin environments, large-scale simulation platforms, and infrastructure-vehicle interaction systems such as adaptive signal control or route guidance.

Among its diverse applications, traffic state generation has been used to enable both short-term forecasting and long-term scenario planning. Deep learning models such as temporal convolutional networks, graph-based attention models, and recurrent neural networks have demonstrated the ability to capture spatiotemporal dependencies in traffic data and predict how speed, volume, or congestion levels will evolve in the near future~\cite{lee2023traffic, lee2021travel}. These forecasts are vital for autonomous vehicle decision-making, particularly in scenarios that involve anticipatory braking, speed smoothing, or rerouting based on downstream traffic conditions.

Another important use case involves the reconstruction of traffic conditions in sensor-limited or uninstrumented areas of the road network. By learning the spatial correlations between upstream and downstream nodes, generative models can infer missing traffic states with reasonable accuracy~\cite{min2023deep}. This allows traffic management systems and autonomous vehicles to operate effectively even in environments where continuous data coverage is unavailable. For autonomous vehicles, in particular, the ability to reason about unobserved road segments, based on contextual traffic flow patterns, enhances situational awareness and supports safer planning in partially observable domains.

Traffic state generation also contributes to simulation-based evaluation and stress testing of autonomous systems. By creating realistic yet diverse traffic scenarios, including rare but high-impact conditions such as abrupt congestion, lane-blocking events, or large-scale detours, researchers can assess autonomous vehicle behavior under uncertainty and edge cases. This is particularly important in mixed traffic environments where autonomous and human-driven vehicles interact under dynamic and sometimes unpredictable conditions. Generative traffic scenarios enable the testing of autonomous vehicle responses to behaviors such as aggressive lane changes, inefficient gap acceptance, or noncompliance with traffic rules, which are common in real-world settings. These simulations help evaluate operational safety, ensure cooperative adaptability, and identify potential conflict points between autonomous vehicles and conventional drivers. They also support the refinement of decision policies, the definition of ODD limits, and the validation of fallback strategies in controlled yet challenging environments, ultimately contributing to more resilient and socially compatible autonomous driving systems.

\subsection{End-to-End Autonomous Driving}

End-to-end driving is an emerging deep learning method for autonomous driving that generates the planning trajectory and/or low-level control actions directly from sensory data and ego vehicle status~\cite{chen2024end, wu_trajectory-guided_nodate, li_exploring_2024, chen_ppad_2025}. This new technology has gained attention from both the industry~\cite{driveirl,hu2023imitation,pmlr-v164-scheel22a} and academia \cite{hu_planning-oriented_2023}. Unlike the conventional paradigm that stitches a series of components (perception, prediction, planning, etc.) in serial, the end-to-end autonomous driving system is fully differentiable, allowing more holistic data flow in the model, thereby reducing the compounding errors that arise in the conventional paradigm. Generative AI plays a significant role in this emerging field.

\paragraph{Action Generation with Generative Models} 

Inspired by the success of LLMs, methods utilizing generative and autoregressive approaches are introduced in end-to-end autonomous driving. GenAD~\cite{zheng_genad_2024} proposed a novel generative framework by turning autonomous driving into a generative modeling problem. The framework contains an instance-centric tokenizer to transform surrounding elements into instance tokens, and a VAE to learn a latent space for trajectory modeling. Previously separated motion prediction and planning tasks are simultaneously performed by sampling from the latent space conditioned on instance tokens. DriveGPT~\cite{huang2024drivegpt} then proposes a GPT-style model to predict encoded future scenes as tokens.

DiffusionDrive~\cite{liao_diffusiondrive_2024} utilizes a truncated diffusion policy with an efficient cascade diffusion decoder to generate diverse and plausible planning trajectories at real-time speed. DiffAD \cite{wang_diffad_2025} formulates the autonomous driving task as an image generation task and uses a diffusion model to perform joint perception and planning tasks, thereby reducing the overall system complexity. GoalFlow \cite{xing_goalflow_2025} proposed a goal-guided diffusion model to solve the trajectory divergence problem inherent in diffusion-based methods.

\paragraph{Multimodal Foundation Model in End-to-end Driving} As the capabilities of multimodal foundation models continue to improve, their applications in end-to-end autonomous driving are also thriving. LMDrive~\cite{shao_lmdrive_2023} proposes the first VLM-based method for closed-loop end-to-end driving. This model takes in navigation instructions, multimodal sensor data, and possible notice instructions as input, and generates control signals directly. DriveLM~\cite{wang_drivemlm_2023} integrates Visual Question Answering (VQA) into autonomous driving, providing a way to instruct an autonomous system to produce vehicle actions given verbal commands and camera inputs. 

To utilize the reasoning ability of foundation models, DriveGPT4~\cite{xu_drivegpt4_2024} proposed a question-answering framework for autonomous driving. Also aiming to improve the reasoning capability, DriveCoT~\cite{wang_drivecot_2024} collects a dataset for chain-of-thought (CoT) reasoning, providing labels for the reasoning process and final decisions. It also provides a simple baseline that feeds multiple task-specific learnable queries into the CoT process. EMMA\cite{hwang2024emma} and its open source counterpart OpenEMMA\cite{xing2025openemma} are the first comprehensive frameworks that produces multimodal outputs (drivable trajectories, object bounding boxes, etc.) directly from multimodal sensor inputs (cameras, LiDARs, etc.), demonstrating effectiveness, generalizability, and robustness across a variety of challenging driving scenarios. 

MLLM can also be used together with other end-to-end autonomous driving models, facilitating the model's capability on scene understanding and solving unseen scenarios. Wang et al.~\cite{wang_drive_2024} uses the visual and textual understanding ability from a multimodal foundation model to enhance the robustness and adaptability of the autonomous driving system. The multimodal representations also enable language-augmented latent space editing and simulation, giving the system the potential to better generalize. Mei et al.~\cite{mei_continuously_2024} proposed a dual-process decision-making framework, combining the strong but slow analytic process using a large language model and a fast and empirical process using a smaller language model. This process imitates the human cognitive process and successfully utilizes the logical reasoning ability to continuously improve performance in varied environments with low inference overhead. Dong et al.~\cite{dong2024generalizing} uses a zero-shot LLM with CoT prompt in the training phase to provide instructions to the standalone end-to-end model, showing its feasibility in real-world deployment. SENNA \cite{jiang2024senna} introduces an integrated framework for VLM's usage in end-to-end autonomous driving. In this framework, VLM can use the perceptual information to provide meta-action guidance to the end-to-end driving model. VDT \cite{guo2025vdt} uses VLM for scene understanding to assist the diffusion-based path generation. Orion \cite{fu2025orion} further connects the generative planner to the upstream VLM to enable the end-to-end gradient backpropagation at training time.

\subsection{Personalized Autonomous Driving}
The autonomous driving industry is experiencing an evolution towards human-centric systems~\cite{calvert2020human,yang_human-centric_2024}, where vehicle automation extends beyond considering only traditional safety and efficiency metrics but also provides personalized driving experiences. This trend reflects a growing recognition that a successful adoption of the autonomous vehicle requires not just technically sound driving capabilities, but also the ability to provide human-like driving experiences that align with individual preferences and expectations~\cite{cui_drive_2024,cui_receive_2024}. Recent advances in human-centric autonomous driving have been significantly influenced by generative AI, especially LLMs and VLMs. These models demonstrate strong capabilities in continuous adaptation and learning. In this section, we will explore personalized and human-centric designs in autonomous driving enabled by generative AI.

\paragraph{Large Language Models} Li et al.~\cite{li_large_2024} reviews how LLMs enable more human-like autonomous driving by bridging the gap between data-driven approaches and human-like decision making. They identify how LLMs enhance context understanding, scenario reasoning, and decision interpretability through their language understanding and commonsense knowledge capabilities. Another study by Li et al.~\cite{li_personal_2024} explores personal LLM agents as AI assistants customized through personal data and device integration, emphasizing the importance of adapting AI systems to individual user contexts.

The use of LLMs in personalized natural language interaction systems has gained significant popularity in recent years. Yang et al.~\cite{yang_human-centric_2024} explore using LLMs as an intermediary to understand and reason about natural language commands from human users in autonomous vehicles, demonstrating how generative AI can enable more intuitive human-centric vehicle interactions. Cui et al.~\cite{cui_drive_2024} develop an LLM framework using chain-of-thought prompting for continuous feedback, while their subsequent work~\cite{cui_receive_2024} focuses on translating human intent into safe vehicle actions. Roque et al.~\cite{roque2025automateddatacurationusing} develop an automated data engine to generate such data which relates intent or instruction to vehicle actions, utilizing GPS application data as a language stream. Han et al.'s Words2Wheels framework~\cite{han_words_2024} introduces a novel approach by combining LLMs with reward function generation, utilizing a driving style database and statistical evaluation for policy alignment.  Martinez-Baselga et al.~\cite{martinez-baselga_hey_2024} present a method for personalizing navigation using LLMs to interpret natural language commands and automatically generate/tune cost functions for Model Predictive Control (MPC).

Some researchers explore using LLMs in trajectory and behavior prediction. TrajLLM~\cite{lan_traj-llm_2024} leverages LLMs to predict future motion trajectories from past observations and scene semantics, enhanced by lane-aware probabilistic learning and a multimodal Laplace decoder for scene-compliant predictions, effectively emulating human-like lane focus. Duan et al.~\cite{duan_prompting_2024} propose to integrate LLMs into autonomous driving systems by using multimodal prompts (combining visual and LiDAR data) and letting LLMs help correct driving mistakes. This enhances human-like and personalized autonomous driving by leveraging language models' ability to understand and reason about complex driving scenarios in a more natural, semantic way. Li et al.~\cite{li_ai-empowered_2024} explores the development of AI-empowered personalized co-pilot systems for autonomous driving, which covers personalized trajectory prediction for ramp-merging scenarios and the potential of LLMs for enhancing perception and decision-making in autonomous driving.

Extensive research focuses on integrating LLMs into personalized agents. Chen et al.~\cite{chen_edge-cloud_2024} propose EC-Drive, an edge-cloud collaborative framework for autonomous driving that uses smaller language models (\textit{e.g.} LLaMA) on edge devices for routine driving decisions while selectively offloading complex scenarios to larger cloud-based models (\textit{e.g.} GPT-4) to enhance personalization and adaptability in open-world scenarios. Shi et al.~\cite{shi_maximizing_2024} presents LLMOps (Large Language Model Operations) as a methodology for enhancing personalized recommendation systems that can be applied to vehicle infotainment systems. CockpitGemini~\cite{ren_cockpitgemini_2024} is a novel framework that leverages generative AI models (LLMs), multi-agent systems, and human digital twins to enable highly personalized smart vehicle cockpit experiences, demonstrating the potential through key aspects like personalized product design, interactive interfaces, user state monitoring, and driving strategy recommendations based on individual preferences and real-time status. Ma et al.~\cite{ma_lampilot_2024} leveraged RAG, which is an approach that enhances model capabilities by retrieving relevant historical information to augment the LLM generation process to learn from human feedback and achieve human-like driving, while Sun et al.~\cite{sun_optimizing_2024} combine RLHF with LLMs for physiological feedback processing.  Wong et al.~\cite{wong_autonomous_2023} examine ChatGPT's potential to transform autonomous travel decision-making, showing that ChatGPT can act as a personalized travel assistant.

\paragraph{Vision Language Model} Vision-language integration is an important topic in personalized driving. Guo et al.~\cite{guo_vlm-auto_2024} propose VLM-Auto, a VLM-based autonomous driving system designed to enhance human-like driving behavior by leveraging advanced road scene understanding. pFedLVM~\cite{kou_pfedlvm_2024}, a framework that integrates Large Vision Models with federated learning for autonomous driving to address the challenges of model under-fitting as training data grows. The key innovation is to keep VLMs only on central servers while having vehicles exchange learned features rather than full model parameters - this preserves personalized driving characteristics for each vehicle while enabling shared knowledge across the fleet. Cui et al.~\cite{cui2024onboardvisionlanguagemodelspersonalized} develop an on-board VLM system to provide a personalized control strategy for MPC and PID controllers on a real vehicle. Long et al. \cite{long2024vlm} utilized the reasoning capacity of VLM to adjust MPC parameters for safe and informed decision-making. You et al. \cite{you2024v2x}  extended the previous work to autonomous vehicle control with V2X-enabled cooperative perception. 

In addition, some of the researchers leverage generative AI for human-like multimodal networks and communication. Zhang et al.~\cite{zhang_generative_2024} propose using generative AI (LLMs and diffusion models) to enhance vehicular networks through a semantic-aware framework that leverages multimodal inputs (text and images) for more reliable vehicle-to-vehicle communication. Liang et al.~\cite{liang_generative_2024} examine how generative AI can enhance human-like semantic communication networks by proposing a novel framework that incorporates multimodal AI models, semantic encoding/decoding, and knowledge management capabilities. Zhang et al.'s survey paper~\cite{zhang_personalization_2024} on LLM personalization offers valuable insights into how personalization techniques can be applied across different granularities (user-level, persona-level, and global preferences) with potential applications for adapting autonomous driving systems to individual driver preferences and characteristics. 

\paragraph{Generative Deep Learning Models} Bao et al.~\cite{bao_prediction_2021} propose a probabilistic deep generative model (CVAE and LTSM) for predicting personalized driving behaviors like velocity, acceleration, and steering angle that considers individual driving styles and surrounding vehicle interactions. p-BEAM~\cite{collabor19} is a personalized driving behavior modeling, which trains a generative adversarial recurrent neural network (GARNN) model in the cloud that adapts to dynamic changes in normal driving, and then transfers and fine-tunes a personalized model on the vehicle's edge device using CGARNN-Edge. Xu et al.~\cite{xu_generative_2023} propose integrating generative AI technologies (like diffusion models and LLMs) into vehicular digital twins and simulation systems to enhance autonomous driving, introducing a multi-task offloading framework optimized through distributed deep reinforcement learning. Shan et al.~\cite{shan_generative_2024} propose using diffusion models and LoRA to generate personalized car front-end designs by controlling personality descriptors and emotional tags.

\subsection{Digital Twins}

As mentioned in Section \ref{app:data_gen}, autonomous driving requires massive amounts of diverse, high-quality data for training and validation. In addition to that, recent works in reinforcement learning (RL) have sparked interest in interactive, closed-loop simulation \cite{yang2024drivearena, zhang2022rethinking, gao2025rad, han2024autoreward, lu2023imitation, tan2024promptable}, and justify further research in Sim2Real and Real2Sim transfer frameworks.

\paragraph{Real2Sim: Building Realistic Digital Twins}
Real2Sim refers to the process of converting real-world data into a virtual counterpart, forming the foundation of digital twins—simulated replicas of physical environments that enable scalable, controllable, and reproducible testing. Modern Real2Sim pipelines involve reconstructing driving environments using multimodal sensor inputs (LiDAR, cameras, HD maps) and populating them with dynamic agents and semantics. Works like UrbanDiffusion~\cite{zhang2024urban}, OccSora~\cite{wang2024occsora}, and DOME~\cite{gu2024dome} generate high-resolution 3D occupancy scenes conditioned on trajectory or layout, effectively capturing geometric and semantic realism from real data. Similarly, NeRF-based reconstructions (\textit{e.g.}, BlockNeRF~\cite{tancik2022block}, UrbanNeRF~\cite{lu2024urban}) and 3D-GS models (\textit{e.g.}, OmniRe~\cite{chen2024omnire}, DrivingGaussian~\cite{zhou2024drivinggaussian}) reconstruct dense 3D/4D environments, supporting novel view synthesis and agent rendering.

Real2Sim for behavior is achieved by mining agent trajectories from logs and learning generative models (\textit{e.g.}, CVAEs~\cite{lee2017desire}, Trajectron++ \cite{salzmann2020trajectron++}, diffusion models like MotionDiffuser~\cite{jiang2023motiondiffusercontrollablemultiagentmotion}) that replicate plausible interactions. Language-conditioned frameworks like DriveLM~\cite{sima2023drivelm}, LMDrive~\cite{shao2024lmdrive}, and GPT-driver~\cite{mao2023gpt} further abstract scene context into linguistic inputs, enabling high-level understanding and flexible simulation via multimodal large language models (MLLMs). This fusion of spatial, behavioral, and semantic fidelity enables the creation of semantically rich, interactive digital twins.

\paragraph{Sim2Real: Bridging the Domain Gap}
Sim2Real aims to transfer models trained in simulation to real-world deployment. However, simulators often suffer from the reality gap due to simplified physics, low visual realism, or mismatched agent behavior. Image and video generation models like BEVGen~\cite{swerdlow2024street}, MagicDrive3D~\cite{gao2024magicdrive3d}, and DriveDiffusion~\cite{li2025drivingdiffusion} improve realism via controllable generation using BEV layouts, text prompts, or weather conditions. Diffusion models and 3D-GS rendering allow these systems to generate diverse and photorealistic scenes while preserving geometry and temporal consistency. When combined with NeRFs or point clouds, models like Stag-1~\cite{wang2024stag} and ChatSim~\cite{wei2024chatsim} produce semantically consistent sequences that can simulate various driving conditions and environments, narrowing the sim-real visual gap.

For LiDAR, diffusion-based models like RangeLDM~\cite{hu2025rangeldm} and LiDMs~\cite{ran2024towards} offer high-fidelity simulation of point clouds, modeling sensor-specific artifacts like ray-drop and sparsity. NeRF-LiDAR~\cite{zhang2023nerf} and DyNFL~\cite{Wu2023dynfl} generate physically plausible LiDAR signals from implicit 3D scene models, contributing to sensor-level realism.

At the behavioral level, Sim2Real requires not only realistic agent dynamics but also socially-aware interactions. Models like TrafficSim~\cite{Suo2021TrafficSim} and DJINN~\cite{niedoba2023diffusion} simulate multi-agent traffic behaviors learned from real data, offering stochastic yet realistic traffic flows. Newer approaches like BehaviorGPT~\cite{Zhou2024BehaviorGPT} leverage transformers to autoregressively simulate behavior, capturing long-horizon dependencies and contextual reasoning. \cite{dong2019mcity}

\paragraph{Digital Twins as a Sim-Real Bridge}

The integration of Real2Sim and Sim2Real techniques has given rise to Digital Twins: high-fidelity, interactive simulations that mirror real-world conditions and behavior. These twins enable controlled testing of edge cases (\textit{e.g.}, occlusions, rare maneuvers, collisions) and provide a closed-loop environment for training autonomous policies. The MCitiy digital twin (2025) \cite{dong2019mcity, mcitiy_2025} is the pioneer research in providing a complete digitalized replica of a real-world autonomous driving testing facility. It serves both as rich synthetic environments (training in sim, testing in real) and as mirrored digital layers over real environments (simulation informed by real data), creating a bidirectional feedback loop where models can continually improve.

\subsection{Scene Understanding}
Scene understanding with multimodal Large Language Models (MLLMs) has emerged as a promising frontier in autonomous driving, offering a unified framework to interpret complex traffic scenarios through multimodal reasoning. Unlike traditional perception modules, which rely on task-specific architectures for object detection, segmentation, or intent estimation, MLLMs integrate powerful visual encoders (\textit{e.g.}, CLIP \cite{radford2021clip}, BLIP-2 \cite{li2023blip2}, LLaVA \cite{liu2024llava}) with pretrained large language models to enable open-ended understanding from both visual and linguistic inputs. These models can process raw driving scene images or video sequences alongside natural language queries or instructions, making them suitable for a wide array of tasks such as risk identification, behavior prediction, and semantic understanding. HiLM-D~\cite{ding2023hilm} showcases the ability of MLLMs to perform unified perception and high-level reasoning by generating natural language descriptions of risks, intentions, and suggested motions from driving videos. Similarly, DriveLM~\cite{sima2023drivelm} introduces Graph VQA for structured reasoning and is further extended in DriveVLM~\cite{tian2024drivevlm} to incorporate Chain-of-Thought prompting for spatial understanding and trajectory forecasting. Dolphins~\cite{ma2024dolphins} and TUMTraffic-VideoQA \cite{zhou2025tumtraffic} adapt general VQA datasets and domain-specific driving data to enhance fine-grained visual reasoning, while EM-VLM4AD~\cite{em-vlm4d} introduces a lightweight MLLM optimized for efficiency in multi-frame analysis. These models excel at interpreting dynamic scenes and generating context-aware predictions or plans, effectively bridging perception and decision-making. Beyond reasoning, MLLMs have also been integrated into end-to-end driving stacks, as demonstrated by LMDrive~\cite{shao2024lmdrive}, which fuses sensor inputs and language commands in a closed-loop control system. LeGo-Drive~\cite{paul2024lego} further proposes to predict semantic goal locations from language as an intermediate planning target, enabling flexible goal-directed navigation. Cube-LLM \cite{cho2024language} is the first method to allow open-vocabulary 3D grounding (provide 3D bounding box for items in a given image). Collectively, these MLLM-driven approaches represent a step toward cognitively-informed scene understanding, offering interpretability, flexibility, and generalization in complex driving environments.

\subsection{Intelligent Transportation Systems}

\begin{figure}[h!]
    \centering
    \includegraphics[width=0.6\linewidth]{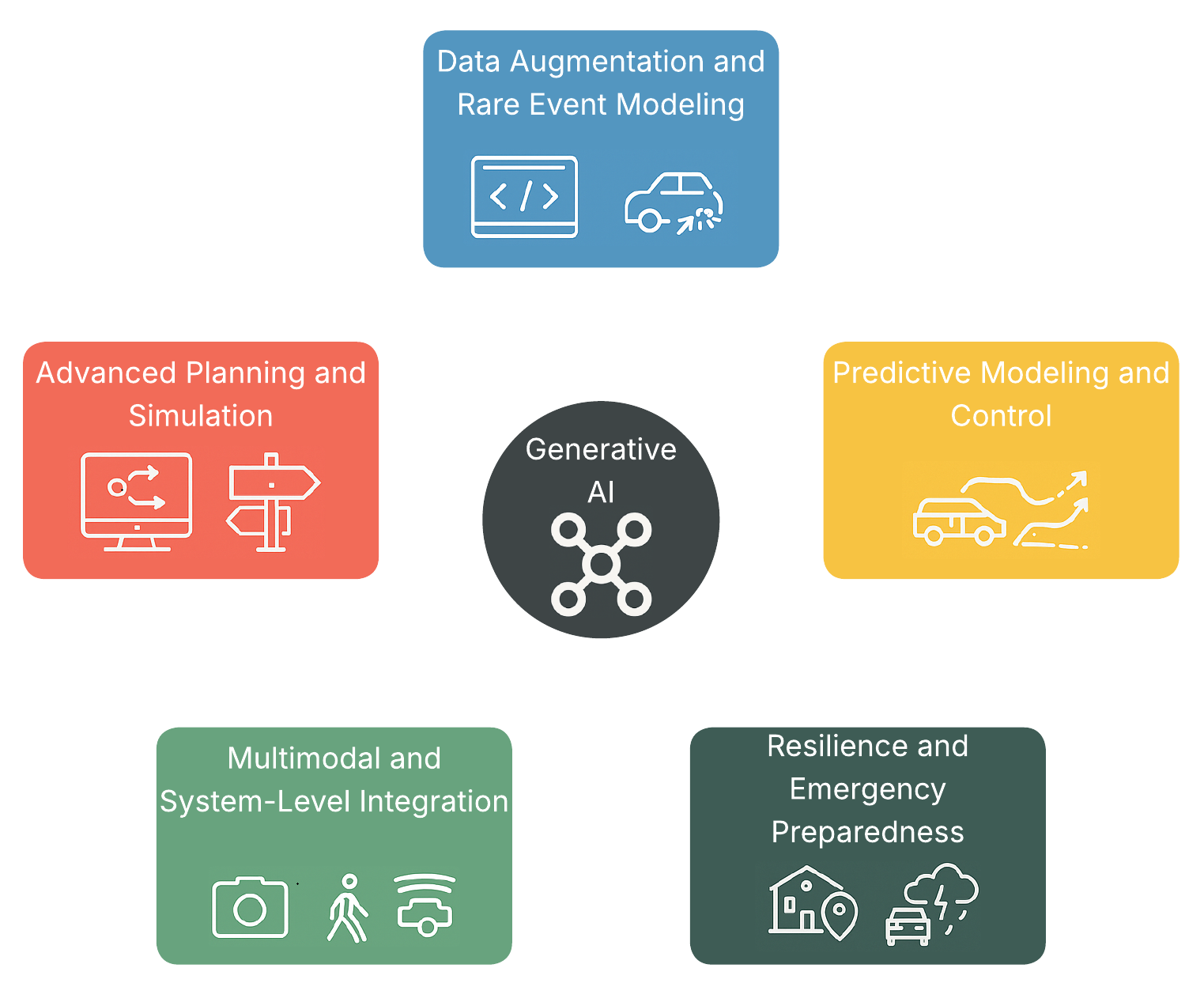}
    \caption{Generative AI helps solve the challenges in Intelligent Transportation Systems.}
\end{figure}

It is also important to contextualize how advancements in generative AI, intersect with the broader domain of Intelligent Transportation Systems (ITS). ITS generally refers to the integration of advanced information, communication, and AI technologies within transportation infrastructure and vehicles to enhance safety, efficiency, and sustainability \cite{liu2024potentials, pu2024frontiers}. Unlike the vehicle-centric perspective typical of autonomous driving, ITS adopts a more comprehensive approach, emphasizing system-level or network-level coordination and traffic management \cite{song2024first}. ITS frameworks facilitate communication among vehicles through Roadside Units (RSUs) or cloud servers, enabling the sharing of global information to optimize the movement of people and goods across multimodal transportation networks.

Recent research highlights the transformative role of generative AI in ITS. More specifically, GenAI offers new tools for data augmentation \cite{luo2025senserag}, demand forecasting \cite{yang2025independent}, scenario generation, multimodal optimization, and simulation of complex traffic systems under dynamic and uncertain conditions \cite{da2025generative, wang2024traffic}. This complements, and often enhances, the core objectives of autonomous driving: safe, efficient, and context-aware mobility. The integration of GenAI into ITS enables the transportation agents (\textit{e.g.}, transportation planner, policy maker, engineers, and managers) This includes:

\begin{enumerate}[noitemsep]
 \varitem{red!40}{\textbf{1)}} \textbf{Advanced Planning and Simulation}: Generative AI significantly enhances the ability of Intelligent Transportation Systems (ITS) to plan and simulate complex traffic environments \cite{xu_generative_2023}. By generating diverse and realistic traffic scenarios, including rare and extreme conditions, planners and engineers can better understand how systems respond to different situations. This capability is particularly valuable for virtual testing of traffic control strategies, infrastructure design, and mobility policies without costly or disruptive real-world trials. In urban planning, generative models allow the simulation of future transportation demand and the impact of new infrastructure developments, supporting more informed and proactive decision-making.

 \varitem{blue!40}{\textbf{2)}} \textbf{Data Augmentation and Rare Event Modeling}: A persistent challenge in ITS is the imbalance in datasets—especially the scarcity of critical but rare events such as traffic crashes, near misses, or infrastructure failures. Generative AI can address this by synthesizing high-quality, realistic data that augments existing datasets \cite{luo2025senserag}. This is particularly useful for training machine learning models, which often underperform in underrepresented scenarios. By generating rare event data, generative models help improve system robustness, model generalization, and the fairness of AI-based traffic applications, ensuring that safety systems are reliable even in low-frequency but high-impact situations. 

 \varitem{yellow!40}{\textbf{3)}} \textbf{Predictive Modeling and Control}: Generative AI contributes to more accurate and adaptive predictive modeling within ITS by learning the underlying distributions of traffic dynamics. It can forecast future traffic states, generate synthetic traffic flow patterns, and model traveler demand across different times and locations \cite{yang2025independent}. These predictions are crucial for enabling proactive traffic management strategies such as dynamic routing, congestion pricing, and adaptive signal control. Furthermore, by generating plausible control actions under uncertainty, generative AI supports real-time decision-making and system-level optimization for both human-operated and autonomous transportation systems.

 \varitem{green!40}{\textbf{4)}} \textbf{Multimodal and System-Level Integration}: Modern ITS must integrate and coordinate across multiple transportation modes—vehicles, transit, micromobility, and pedestrians. Generative AI excels in modeling the interactions and dependencies across these modes, allowing simulation of complex, multimodal environments \cite{balog2025user}. For example, it can generate realistic interactions between connected automated vehicles (CAVs), human-driven vehicles, and vulnerable road users. This helps in developing and testing cooperative driving policies, shared space designs, and traffic control systems that ensure seamless, efficient, and safe operation across the entire network. System-level simulations enabled by generative AI are crucial for managing traffic holistically rather than in isolated subsystems.

 \varitem{black!40}{\textbf{5)}} \textbf{Resilience and Emergency Preparedness}: Generative AI plays a vital role in preparing transportation systems for emergencies and enhancing infrastructure resilience \cite{rane2024artificial}. It can simulate large-scale disruptions such as natural disasters, infrastructure failures, or cyberattacks, helping agencies evaluate the performance and vulnerabilities of existing systems. By generating stress-test scenarios that exceed historical extremes, planners can identify weak points, test evacuation strategies, and design more robust emergency response systems. These capabilities contribute to building resilient ITS frameworks that can adapt to and recover from unexpected events, safeguarding both mobility and public safety.
    
\end{enumerate}

These capabilities align with goals like system-level optimization, resilience, and equity, which are vital at the city or regional level but typically fall outside the narrow purview of a single autonomous vehicle system.

Although autonomous vehicles primarily navigate and plan at the local level, there is growing interest in linking autonomous vehicle behavior with system-wide transportation objectives. For instance, vehicle trajectory generation (as discussed in previous sections) can be aligned with goals set by centralized intelligent systems (\textit{e.g.}, reducing congestion in real-time, adhering to dynamic tolling policies, or optimizing curbside usage). Recently, cooperative autonomous driving is a rising topic that lies in the intersection between autonomous driving and intelligent transportation. Works like CoBEVT \cite{xu2022cobevt} and CMP \cite{wang2025cmp} explore the improvement in safety and quality brought by inter-vehicle cooperation. The use of GenAI in these areas could be a promising future direction.

Moreover, many techniques developed for autonomous driving, such as diffusion models for trajectory synthesis, MLLMs for perception, and generative world modeling, can directly support transportation planning tasks \cite{rodrigue2020geography, xing2025can}.


\clearpage

\section{Generative AI in the Broader Area of Embodied Robotics}
\label{sec:embodied_ai}
Generative AI for autonomous driving is increasingly intersecting with the broader field of embodied AI, AI that interacts with the physical world (as in robotics)~\cite{liu2024aligning}. The boundary between a self-driving car and a robot is largely semantic (a car is essentially a robot that drives). Recent trends show a convergence of techniques: vision-language-action (VLA) models and large multimodal models developed for robotics are being adapted to driving, and vice versa. This opens the door for foundation models that can operate across different platforms (cars, drones, home robots) by leveraging common representations. For example, 2023 saw the release of OpenVLA (Open Vision-Language-Action) \cite{kim2024openvla}, a 7-billion-parameter model trained on 970k real robot demonstrations. OpenVLA and related models (like Google DeepMind’s RT-2 \cite{brohan2023rt}) combine language understanding with visual perception and action outputs. While OpenVLA was aimed at robotic manipulation, the architecture is generalizable: indeed, EMMA and OpenEMMA \cite{xing2025openemma, hwang2024emma} can be seen as a driving-specific VLA model that links vision, language, and action for an AV. Similarly, diffusion models that generate actions in driving, such as in MotionDiffuser \cite{jiang2023motiondiffusercontrollablemultiagentmotion}, are also applied to the action generation for embodied agents \cite{black2024pi_0, bjorck2025gr00t}. 
This section dives into how generative modeling techniques benefit the perception, action generation, and sim-to-real transferability of embodied agents.

\subsection{Related Surveys in Embodied AI with Foundation Models}

A growing body of recent surveys has systematically examined the convergence of foundation models and embodied AI, which demonstrates how advances in generative modeling and large-scale multimodal learning are reshaping robotics and autonomous systems. These surveys can be broadly grouped into three thematic directions: integration of foundation models into robotics, development of Vision-Language-Action (VLA) frameworks, and practical applications in industry.

As robotics moves toward open-world deployment, integrating foundation models across the autonomy stack has become increasingly essential for achieving generalizable and scalable behaviors. Several surveys have analyzed how foundation models enhance robotic perception, planning, and control. Firoozi et al. \cite{firoozi2023foundation} highlight how internet-scale pretraining supports generalization and zero-shot capabilities. Xu et al. \cite{xu2024survey} provide a detailed breakdown of vision-language and large language models applied across high- and low-level robotic control. Hu et al. \cite{hu2023toward} emphasize the need for robotics-specific foundation models that integrate multimodal sensing, embodied reasoning, and safe action generation. Across these works, challenges such as data scarcity, hallucination risks, and evaluation standardization are consistently identified.

Given the growing need for embodied agents to interpret sensory environments and execute language-conditioned actions, there has been increasing focus on Vision-Language-Action (VLA) models. Ma et al. \cite{ma2024survey} propose a taxonomy of VLA architectures spanning visual representations, dynamics models, and language-conditioned policies, while Xu et al. \cite{xu2024survey} further analyze world models, Chain-of-Thought (CoT) reasoning, and large-scale datasets supporting VLA development. Awais et al. \cite{awais2025foundation} discuss how recent vision-language foundation models are enabling agents to reason and act across diverse environments, reflecting a shift toward fully integrated multimodal perception-to-action pipelines.

Additionally, industrial deployment and agent-based applications highlight the need for foundation models capable of robust real-world operation. Ren et al. \cite{ren2024embodied} introduce an "ABC model" coordinating multimodal perception and planning for manufacturing intelligence. Fan et al. \cite{fan2025embodied} demonstrate how large language models like GPT-4 can autonomously generate manufacturing plans and robot programs. Durante et al. \cite{durante2024agent} propose the broader concept of Agent AI, unifying perception, cognition, and action in physical and virtual environments. Together, these efforts illustrate how embodied intelligence is advancing from theoretical frameworks toward impactful real-world deployment.

\subsection{Generative Modalities and Their Integration in Embodied AI}

In embodied AI systems, different sensory modalities contribute to complementary aspects of perception and interaction. Visual modalities, such as RGB imagery, depth sensing, and 3D scene reconstruction, serve as high-level observation methods that provide detailed global representations of the environment. These visual inputs are crucial in enabling agents to understand spatial layouts, identify objects, and perform high-level planning in complex and dynamic surroundings. In the meantime, an essential perspective of embodied AI extends beyond passive observation to active interaction with the environment. In particular, object manipulation and contact-rich physical operations require fine-grained, localized sensory information that visual modalities alone cannot fully provide. Tactile sensing addresses this need by capturing surface properties, contact geometry, and interaction forces, offering detailed feedback critical for dexterous operation and manipulation tasks. Based on these complementary roles, generative modalities in embodied AI can be broadly categorized into visual modalities and contact modalities. 

\textbf{Generative Modeling for Visual-Based Modalities:} Visual sensing provides embodied agents with rich observations of their environments, supporting high-level perception, planning, and action selection. Recent advances in generative modeling have enabled synthetic augmentation and predictive simulation across various visual modalities, ranging from RGB images to 3D point clouds and depth scans.

Generative image models, such as GANs, diffusion models, and video prediction networks, enhance an embodied agent’s ability to perceive and anticipate changes in its surroundings. These models can synthesize novel views or predict future frames, allowing the agent to visualize potential outcomes and handle partial observations. For instance, diffusion models have been employed to generate goal images or hallucinate intermediate object states, providing “common sense” geometric reasoning that guides policy learning and decision-making \cite{yu2023scaling}. Video-generation-based world models similarly allow agents to simulate future sensory experiences based on candidate actions, a technique widely adopted in model-based reinforcement learning. In robotics, such generative foresight has been applied to tasks like multi-step manipulation planning and navigation, enabling agents to predict how scenes may evolve over time and select actions that lead to favorable imagined outcomes \cite{driess2023palm, yang2023learning}.

Beyond 2D visual prediction, embodied agents equipped with depth sensors also benefit from advances in generative 3D scene modeling. Neural Radiance Fields (NeRFs) and related neural field representations allow robots to generate realistic 3D and 4D scenes, reconstructing unseen viewpoints or creating synthetic depth signals that closely match real-world environments. NeRF-based models have been used to anticipate obstacles, plan collision-free paths, and simulate dynamic changes in the environment \cite{wang2024nerf}. CLIP-Fields \cite{shafiullah2022clip} further enrich the representation by embedding 3D structures into vision-language spaces, supporting semantic navigation tasks where an agent reasons about objects described in natural language. In addition to neural fields, purely synthetic LiDAR data generation has emerged through models like LidarDM, a diffusion-based approach that produces physically plausible sequential 4D LiDAR point clouds conditioned on dynamic scenes \cite{zyrianov2024lidardmgenerativelidarsimulation}. These generative sensor models can be directly integrated into simulation pipelines or agent perception systems, providing diverse and lifelike training data that improves robustness without requiring extensive real-world data collection.

\textbf{Generative Modeling for Tactile-Based Contact Modalities:} While visual modalities provide global environmental understanding, tactile sensing offers fine-grained, local feedback that is essential for contact-rich tasks such as manipulation, insertion, and object recognition \cite{yumimictouch}. In embodied AI, tactile modalities play a critical role in capturing interaction dynamics that visual observations alone cannot resolve, particularly in scenarios involving soft, deformable, or occluded objects. As vision-based tactile sensors (\textit{e.g.}, DIGIT \cite{lambeta2020digit}, GelSight \cite{yuan2017gelsight}) become more accessible and robust, a growing body of work has begun to explore tactile-driven generative modeling. These models learn to reconstruct contact maps, simulate missing tactile input, align touch with vision and language, or generate reactive action plans based on tactile feedback.

Several works have focused on generative modeling of contact geometry from tactile inputs, enabling agents to reason about where and how contact occurs beyond direct sensor measurements. NCF-v2 \cite{higuera2023perceiving}, for example, predicts extrinsic contact fields between manipulated objects and their environments using only vision-based tactile inputs from a robot gripper. By combining a variational autoencoder with a transformer-based contact regressor, it generates probabilistic contact distributions over object surfaces and improves downstream insertion policies in real-world tasks such as mug-in-cupholder or dish-in-rack assembly. Similarly, TacMAE \cite{cao2023learn} applies masked autoencoding to reconstruct missing tactile regions from partial GelSight images, enabling robust representation learning in scenarios with weak or incomplete contact.\cite{lambeta2020digit} propose a compact and low-cost high-resolution tactile sensor that supports generative contact modeling by compressing tactile observations into keypoint-based latent features and learning a forward dynamics model for Model Predictive Control, allowing for closed-loop dexterity in in-hand manipulation. In a more spatially grounded setting, \cite{dou2024tactile} introduces TaRF that integrates touch with 3D vision by conditioning a latent diffusion model on NeRF-rendered RGB-D inputs to generate dense tactile distributions across unobserved surface regions. This cross-modal completion allows agents to predict tactile feedback in physically inaccessible or previously unexplored areas of the scene.

Recent research explores how tactile signals can be semantically grounded through generative alignment with language and vision. Sparsh \cite{higuerasparsh} and UniTouch \cite{yang2024binding} apply contrastive and masked autoencoding methods to vision-based tactile data, creating unified latent representations transferable across semantic tasks like force prediction, slip detection, textile classification, and bidirectional generation, such as touch-to-image or touch-to-text. Similarly, OCTOPI \cite{yu2024octopi} aligns tactile videos with language using Vicuna-based LLMs, enabling commonsense reasoning about object properties (softness, stickiness, fragility) and excelling in tactile reasoning tasks such as ripeness prediction and object matching. Generative approaches also use tactile signals for direct action conditioning and policy learning. Reactive Diffusion Policy \cite{xue2025reactive} integrates a slow visual planner with a fast tactile feedback loop, leveraging diffusion models for trajectory planning and real-time tactile refinement for reactive control in complex manipulation scenarios. The Visuo-Tactile Transformer (VTT) \cite{chen2023visuo} and NeuralFeels \cite{suresh2024neuralfeels} employ transformer architectures for fusing tactile and visual modalities to generate embeddings and depth estimates, respectively, significantly enhancing manipulation and incremental visuo-tactile object reconstruction tasks. Self-supervised tactile representation learning frameworks like T-DEX \cite{guzey2023dexterity} utilize BYOL-trained encoders for dexterous task imitation, while MViTac \cite{dave2024multimodal} employs cross-modal contrastive learning, achieving robust material classification and grasp prediction even under limited supervision. Broader transfer across tactile sensors and manipulation tasks is demonstrated by T3 \cite{zhaotransferable} and AnyTouch \cite{fenganytouch}, using masked autoencoding and semantic alignment for generalizable tactile representations across diverse sensors and applications.

\subsection{LLMs and Multimodal Models for Perception-to-Action Translation}

The pursuit of general-purpose robotic agents has increasingly drawn from advances in foundation models, particularly large language models (LLMs), vision-language models (VLMs), and diffusion-based action generators \cite{moroncelli2024integrating, li2024cogact}. These models aim to unify perception, reasoning, and control into a common framework, allowing robots to map diverse sensory observations directly into meaningful actions across a wide variety of manipulation tasks and embodiments. Recent works reflect a growing convergence between robotics and generalist AI, with embodied systems adopting sequence modeling, multimodal representation learning, and scalable dataset-driven training paradigms to bridge perception and action in complex environments.

A primary trend centers on constructing generalist policies through large-scale foundation models. OpenVLA \cite{kim2024openvla} and PaLM-E \cite{driess2023palm} illustrate how pretrained language architectures \cite{oquab2023dinov2,zhai2023sigmoid, touvron2023llama} can be extended with visual and sensor tokens to support manipulation, navigation, and high-level planning. OpenVLA emphasizes efficiency, demonstrating that even a 7B model trained on real-world demonstrations can rival or surpass more resource-intensive systems. On the other hand, Palm-E \cite{driess2023palm} integrates massive Internet-scale knowledge with embodied experience, enhancing embodied reasoning through positive transfer. Meanwhile, RT-2 \cite{brohan2023rt} and RT-X \cite{vuong2023open} pioneer web-scale training pipelines, where models jointly learn from Internet vision-language datasets and robot trajectories to acquire robust manipulation capabilities with emergent reasoning abilities. Pi-0 \cite{black2024pi_0} introduces continuous action generation via flow matching, further improving zero-shot generalization across manipulation domains. Complementary approaches like RoboFlamingo \cite{li2024vision} focus on practical deployment. The authors propose lightweight adaptation layers that enable flexible language-driven manipulation using open-source vision-language models. Similarly, VIMA \cite{jiang2022vima} frames manipulation as multimodal prompt-driven sequence prediction, achieving strong generalization to unseen tasks, objects, and environments through scalable transformer-based architectures.

Furthermore, advances in action representation and policy learning architectures have further expanded the horizons of generalization in manipulation. Diffusion Policy \cite{chi2023diffusion} models action sequences as conditional denoising processes, offering robust multimodal trajectory generation in high-dimensional spaces. Motion Planning Diffusion \cite{carvalho2023motion} extends the application of diffusion models from action sequence prediction to trajectory planning. The model learns a generative prior over entire motion plans and guides diffusion-based sampling toward satisfying task objectives such as collision avoidance and smoothness. ALOHA \cite{zhao2023learning} introduces transformer-based chunking strategies that predict temporally coherent action segments, which improves the stability and efficiency of real-world fine manipulation from minimal demonstrations. CrossFormer \cite{doshi2024scaling} proposes a shared policy architecture capable of controlling robots with widely varying embodiments and sensor configurations. Parallelly, large language models have been increasingly incorporated into the planning and reasoning layers of embodied agents. Frameworks such as SayCan\cite{saycan2022arxiv} leverage LLMs to translate natural language instructions into sequences of executable robot skills or policy code, demonstrating that language models can orchestrate flexible, context-aware decision-making across multiple abstraction levels. Recent efforts like PrefVLM \cite{ghosh2025preference} further integrate selective human feedback with vision-language models to enhance generalization while reducing annotation demands. RDD \cite{yan2026rdd} applies VLMs to align high-level demonstrations with low-level ones during hierarchical planning.

\subsection{Simulation-to-Reality Transfer with Generative Models}

A notorious challenge in robotics and embodied AI is the Sim2Real gap, where policies or perception models trained in simulation often fail when deployed in the real world due to discrepancies in visual appearance and physical behavior. Recent advances in generative modeling have begun to address this gap across multiple dimensions, including visual observation gap, dynamic mismatch, and task variability.

A primary line of work focuses on bridging the visual observation gap. To overcome the unrealistic textures and lighting in synthetic environments, Li et al. \cite{li2024aldm} propose a layout-to-image diffusion model (ALDM) that photorealistically renders simulation scenes to enhance the zero-shot transfer performance of grasping policies. More broadly, diffusion and GAN-based stylization methods have been used to translate simulated camera images into realistic ones \cite{zhao2024exploring}, learning style mappings that reduce visual discrepancies without extensive manual tuning. Beyond the RGB domain, the generative methods have also been tested on 3D visual observation. LidarDM \cite{zyrianov2024lidardmgenerativelidarsimulation} generates realistic LiDAR point clouds from scene layouts, enabling sim-to-real augmentation for planners relying on range data. These generative models also facilitate targeted domain randomization by inserting variations such as obstacles or weather effects, producing plausible yet diverse sensor observations that improve robustness \cite{zhang2024lidar}.

Another equally critical challenge lies in addressing the dynamic mismatch between simulation and reality. Simulators often idealize physical properties such as mass, friction, and compliance, while real-world interactions exhibit significant variability. To address this, generative models can be employed to synthesize diverse physical parameters or simulate plausible perturbations in object dynamics. Recent advances have explored this direction through learned representations and automated modeling. Le Cleac’h et al. \cite{le2023differentiable} propose Dynamics-Augmented Neural Objects (DANOs), where continuous density fields parameterized by deep networks capture both visual appearance and dynamic properties, enabling simulation environments to better reflect real-world interaction behavior. Semage et al. \cite{semage2023zero} introduce the Reverse Action Transformation (RAT) framework, which enhances zero-shot sim-to-real transfer by learning to adjust simulated policy outputs to account for real-world dynamics without requiring separate adaptation modules. Complementing these approaches, a Real2Sim pipeline \cite{pfaff2025scalable} automates the generation of simulation-ready assets by estimating visual geometry, collision models, and inertial parameters directly from robotic interaction data, minimizing manual asset tuning and improving simulation realism. Together, these methods demonstrate that generative modeling and data-driven system identification offer powerful tools to reduce the dynamics gap, leading to more robust embodied agents capable of generalizing to imperfect, variable physical environments.

Rare task variability also poses a persistent bottleneck for transfer learning in embodied AI. Simulation environments often fail to capture the long-tail distribution of real-world scenarios \cite{zhang2023deep}. Generative models offer a promising solution by enabling the targeted creation of challenging corner cases, such as scenes with rare clutter configurations, dynamic obstacles, or environmental disturbances. Building on this idea, several recent methods have proposed systematic frameworks for rare event generation. Diffusion Augmented Agents (DAAG) \cite{di2024diffusion} employ diffusion models to autonomously relabel and augment agents' past experiences, synthesizing successful outcomes from previously failed trajectories and thus exposing agents to rare but meaningful training examples without human intervention. Gen2Sim \cite{katara2024gen2sim} automates the entire simulation pipeline by combining image diffusion models and large language models (LLMs) to generate textured 3D assets, task decompositions, and reward structures, producing complex manipulation environments with diverse physical dynamics. GenSim \cite{wang2023gensim} focuses on scaling task diversity at the semantic level, using LLMs to create novel goal-directed and exploratory robotic tasks that significantly improve generalization to unseen scenarios. Zook et al. propose a real-to-sim pipeline that generates realistic simulation tasks from a single real-world image, leveraging vision-language models to match assets and iteratively refine task setups \cite{zook2024grs}. Together, these generative pipelines illustrate that leveraging diffusion models and LLMs provides a scalable and systematic strategy for synthesizing rare and complex training scenarios.

\subsection{Generative Tools for Training Augmentation, Reasoning, and Safety}
Generative AI is not only improving the models inside embodied agents but also revolutionizing how training data is synthesized and diversified. One major benefit is in scalable dataset augmentation for robot learning. Instead of collecting millions of real-world trials, generative pipelines synthesize new training examples, augmenting both diversity and realism. A representative example is ROSIE (Robot Learning with Semantically Imagined Experience) \cite{yu2023scaling}, which leverages diffusion models and LLMs to semantically expand robot datasets by inpainting new objects or distractors into existing scene images, resulting in photorealistic and physically consistent augmentations. Similarly, UniSim \cite{yang2023unisim} proposes a universal action-conditioned simulator that predicts future observations conditioned on agent actions, enabling fully simulated long-horizon training across manipulation and navigation tasks. Beyond pixel-level generation, multiple recent works emphasize layout-level synthesis to diversify embodied training scenarios. Context-aware methods such as Context-Aware Layout Generation and Geometry-to-Culture frameworks \cite{he2021context, asano2025geometry} generate 3D object arrangements by reasoning over semantic and cultural contexts, thus producing more realistic and varied environments compared to purely geometric placement. Further, LayoutVLM \cite{sun2024layoutvlm} and LayoutReasoning \cite{zhu2024automatic} exploit VLMs to synthesize physically plausible and semantically coherent scene layouts through optimization-based or reasoning-driven pipelines. On the 2D layout side, TextLap \cite{chen2024textlap} fine-tunes LLMs for text-to-layout planning, demonstrating that even from natural language descriptions alone, coherent spatial templates can be generated to bootstrap visual environment creation.

Generative models also contribute to multimodal reasoning for embodied AI, supporting agents in perception, planning, and action. Recent work has proposed closed-loop architectures in which large language models (LLMs) are used to generate high-level reasoning about goals and strategies, while generative models simulate the outcomes of proposed actions to verify their feasibility \cite{wang2024nerf, jin2023alphablock}. Recent works illustrate concrete realizations of this loop. In grasping and manipulation, Reasoning-Tuning \cite{xu2023reasoning} and Grasp Reasoning \cite{jin2024reasoning} integrate semantic affordance reasoning with low-level control adaptation to improve grasp success in dynamic scenes. Multimodality Grasping \cite{jin2024reasoning} extends this approach by interpreting implicit language instructions to guide part-based grasping behaviors. RoboMamba \cite{liu2024robomamba} further optimizes vision-language-action pipelines by using structured state space models for efficient $SE(3)$ pose prediction. In addition to execution-level reasoning, failure detection and recovery have been addressed through AHA \cite{duan2024aha}, which trains vision-language models to reason about manipulation failures and provide corrective feedback. In addition to individual manipulation tasks, human-robot collaboration and multi-agent coordination have also benefited from reasoning-based architectures. Vision-Language HRC \cite{liu2024vision} uses visual perception and LLM-driven reasoning to dynamically interpret and fulfill human assembly instructions. Temporal Subgraph Reasoning \cite{li2023self} addresses the problem of multi-human-multi-robot collaboration by dynamically constructing task graphs that adapt to changing team compositions and environmental contexts.  At a larger scale which focus on teamed agents, Dynamic Relational Reasoning \cite{li2024multi} models evolving social interactions among agents for socially compliant navigation, and Topological Reasoning \cite{wang2023coordination}  proposes decentralized path planning strategies by inferring topological constraints without explicit inter-agent communication. Across these domains, the integration of language-based reasoning and generative predictive modeling is enabling embodied agents to extend their capabilities from low-level control to high-level with flexible planning in complex environments.

Generative tools shine in environment synthesis and safety validation as well. Language-driven generative models can synthesize diverse layouts, modify object placements, and introduce environmental variations, providing a scalable means to evaluate the robustness of embodied agents. Traditional simulators such as Habitat \cite{szot2021habitat} and iGibson \cite{pmlr-v164-li22b} depend on static, hand-crafted environments, but generative approaches now enable dynamic creation and targeted alteration of virtual worlds. For example, a system may generate random house layouts populated with varied objects, or introduce scene modifications such as additional furniture, clutter, or lighting changes to systematically challenge an agent’s perception and planning capabilities. ChatScene \cite{zhang2024chatscene} illustrates this direction by using large language models to generate safety-critical driving scenarios from natural language prompts, translating them into executable test cases in CARLA. Although originally developed for autonomous driving, similar strategies apply to robotics: an LLM could generate hypothetical conditions such as a slippery floor or fragile objects, with generative simulation models bringing these scenarios to life without physical risk to the hardware. Building on this foundation, generative models also enable rare-event creation for safety validation. Systems like LidarDM \cite{zyrianov2024lidardmgenerativelidarsimulation} synthesize realistic LiDAR sequences with inserted unexpected obstacles or dynamic agents to evaluate how autonomous systems respond to rare or dangerous situations. These techniques allow what-if analysis, adversarial training, and systematic exploration of boundary conditions without the prohibitive cost of manually modeling rare events or waiting for them to occur in real deployments. 

In summary, the technologies first developed to simulate and predict the world for self-driving cars are catalyzing a revolution in embodied AI at large. Image and sensor generative models enrich an agent’s perception and ground its understanding in realistic inputs. Generative planners and trajectory models offer flexible, multimodal action generation that adapts to novel situations. LLM-based architectures give agents a powerful reasoning engine to translate raw perceptions into goal-directed behavior, using knowledge far beyond their direct experience. And generative data augmentation and simulation provide the fuel for training robust, generalizable agents while ensuring they are tested against the rare situations that matter for safety. By integrating these advances, we are moving toward general-purpose embodied agents that can learn, reason, and act across diverse tasks and environments – a leap enabled by the creative power of generative AI repurposed from autonomous driving to all of robotics and beyond.

\begin{figure}[h!]
    \centering
    \includegraphics[width=0.9\linewidth]{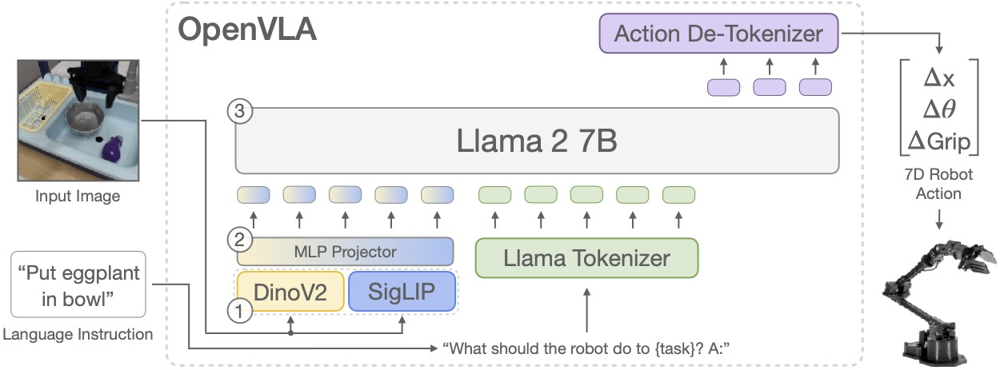}
    \caption{Architecture of OpenVLA \cite{kim2024openvla}. It builds on top of an open-source LLM, Llama \cite{touvron2023llama}, by adding custom action tokenizers.}
\end{figure}

\clearpage

\section{Discussions, Opportunities, and Future Directions}
\label{sec:future}

\subsection{Building More Diverse Scenarios, Datasets, and Benchmarks}

\begin{figure}[h!]
    \centering
    \includegraphics[width=0.99\linewidth]{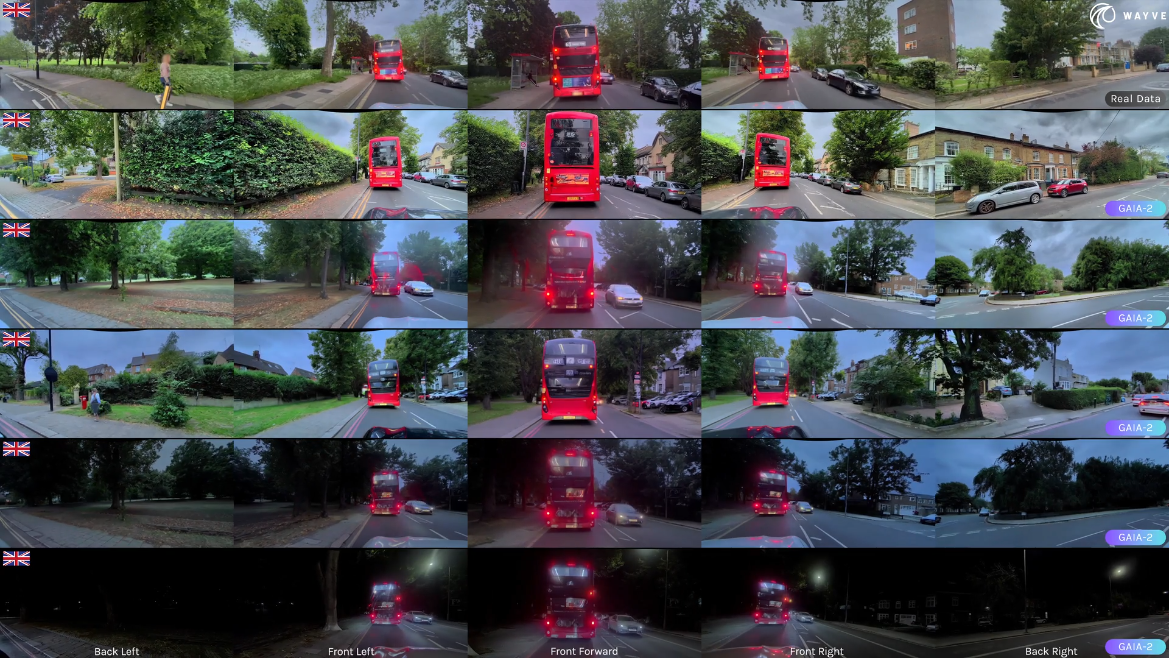}
    \caption{Demonstration from GAIA-2 \cite{russell2025gaia}. The top row shows an actual real-world scenario, while the bottom rows display several generated versions of that same scene.}
\end{figure}

\paragraph{Long-Tail Scenarios}
A key challenge for autonomous driving is handling long-tail scenarios – rare and safety-critical situations that are underrepresented in real-world data (faulty traffic lights~\cite{yang2024mallight}, unpredictable pedestrians, etc).
Generative AI offers a promising avenue to diversify driving scenarios and create synthetic datasets that expose autonomous vehicles (AVs) to a broader spectrum of conditions than traditional data collection. 
Recent works explicitly focus on generating adversarial or uncommon events to improve robustness. 
For example, Wayve's GAIA-2 \cite{russell2025gaia} is a generative world model designed specifically for producing high-fidelity driving video. 
It supports controllable video generation across diverse geographic locations and conditions, 
enabling simulation of both common and rare scenarios in a unified framework. 
By leveraging text, map, and action inputs, GAIA-2 and similar models help reduce reliance on expensive real-world data collection and can generate edge-case driving scenes on demand. 
While GAIA-2 focuses on perception realism, ScenarioDreamer \cite{rowe2025scenariodreamer} is another type of generative models that emphasize trajectory-wise realism. 
It uses latent diffusion models to generate lane graphs and agent trajectories.
These models enable semi-controllable driving scenario generation by manipulating the latent space of the generative model or providing conditions to the neural network.
However, these fully generative and data-driven frameworks often ignore or oversimplify the physical constraints of the driving environment. 
For example, loaded and unloaded trucks have different friction coefficients on the road and different momentum while driving at the same speed, leading to different driving dynamics.
Similarly, road surface and weather conditions significantly impact driving dynamics, which are difficult to capture with fully data-driven generative models.

\paragraph{Control the Generation}
To enable higher controllability and physical realism, applying physics-informed neural networks (PINNs) \cite{RAISSI2019686} to generative models is a promising direction.
PINNs integrate physical laws, typically expressed as partial differential equations (PDEs), directly into the training process of neural networks.
This approach enables models to honor known dynamics, such as vehicle mass, tire-road friction, and momentum, 
thereby enhancing generalization and physical plausibility, especially in scenarios with limited or noisy data.
In autonomous driving, PINNs have been applied to model complex vehicle dynamics such as trajectory prediction and uncertainty quantification \cite{mo2025physics}. 
They can also incorporate a "Physics Guard" layer to ensure that learned parameters remain within physically meaningful ranges, improving prediction accuracy and stability in high-speed scenarios.
Another approach is to incorporate generative models with driving simulators such as CARLA \cite{dosovitskiy2017carla}, MetaDrive \cite{li2021metadrive}, or NVIDIA Drive Sim.
TeraSim \cite{sun2025terasim} is a 2025 traffic simulation platform that uses generative models to construct diverse, 
high-fidelity traffic environments with statistically realistic behavior, 
and then amplifies rare but critical events to systematically expose autonomous vehicles to edge cases. 
Such generative simulation can uncover hidden failure modes by producing scenarios (\textit{e.g.}, near-crashes, sudden cut-ins) that conventional testing might miss. 
The pioneer works demonstrate the potential of combining generative models with simulation to create more diverse and realistic driving scenarios. 
However, existing Digital Twin approaches face challenges such as the efficiency of real-world cloning, the gap between real-world and simulated data, and the lack of a unified framework for scenario generation and evaluation.
This topic will be discussed in detail in Section \ref{diginal_twin}.

\paragraph{Generation for Self-Supervised Learning}
Generative AI has recently shown promise as a means for selecting novel non-synthetic data for training autonomous perception, prediction, and planning systems. When a generative model is trained at foundation scale toward a task, such as the generation of caption text describing a driving scene, both the model outputs and model latents become the criteria by which novel data can be mined. Such data mining supports operations in active and self- or semi-supervised learning, with examples of such techniques including AIDE \cite{liang2024aide} and VisLED \cite{greer2025language}, and even reactive control to anomalous events \cite{sinha2024real}.

\paragraph{Challenges}
A classic dilemma in GenAI is the evaluation of generated output with regard to a goal task. To build datasets of more diverse scenarios, generated representations of scene understanding are important as a tool by which data can be selected using a language-based human or machine query \cite{jiang2026navtrust,xing2025can, cao2024maplm, gopalkrishnanmulti, sachdeva2024rank2tell, keskar2025evaluating}. A variety of metrics may attempt to quantify the ability of the generative method to retrieve and present relevant information for such data selection when compared to human annotation. A future direction for creation of quantitative evaluation schemes may function in a more semi-supervised manner, using generative models themselves to reconstruct input and thus validate the information contained in the model output as sufficient or insufficient information for reconstruction (analogous to the information bottleneck of an autoencoder and the evaluation of a GAN discriminator) \cite{ross2021evaluating}, and thereby forming a basis for whether the low-dimensional representation of scene information is correct or incorrect. An example in autonomous driving comes from the proposed evaluation scheme of \cite{bossen2025can}, whereby a VLM's assessment of a nonverbal instruction in a driving scene is assessed based on the ability of a human or other prompt-generated model, such as \cite{lucas2022posegpt, guo2020action2motion, xu2023actformer}, to recreate a similar pose action sequence. As with all semi-supervised approaches, the ability of the foundation model to generate accurate reconstructions is a limiting factor, and in language-prompted cases, can be further restricted by linguistic prompt ambiguity.    

\begin{tcolorbox}[colback=gray!5!white, colframe=black!50!gray]
    \begin{itemize}[leftmargin=0em]
        \item \fontsize{9}{10}\selectfont
              \textbf{Opportunities:}
              \textit{In the near term, we anticipate new benchmarks that explicitly evaluate an autonomous vehicle's performance on generated long-tail scenarios. At the same time, ensuring realism in these synthetic scenarios remains an open problem: generative models must be constrained by physics and human-like behavior distributions. Future work may tie together standards for scenario description (such as OpenSCENARIO \cite{openscenario2022asam}) with generative techniques, so that any created scenario can be specified in a common format for evaluation and shared within the community. Lastly, validating the realism of generative scenarios is challenging. Developing a systematic framework to evaluate the realism and diversity of synthetic scenario data presents an opportunity for future research.}
    \end{itemize}
\end{tcolorbox}

\subsection{Theoretical and Algorithmic Foundations for End-to-End Autonomous Driving}


\begin{wrapfigure}{L}{0.50\textwidth}
    \includegraphics[width=\linewidth]{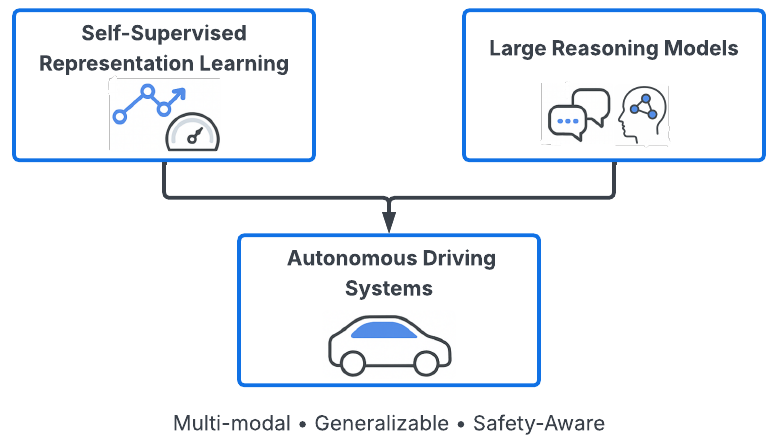}
    \caption{Self-Supervised Representation Learning and Large Reasoning Models enhance Autonomous Driving.}
\end{wrapfigure}

Advancing end-to-end autonomous driving requires strengthening two core pillars: \textbf{(1) robust visual representation models}, and \textbf{(2) strong multimodal reasoning models}. While substantial progress has been made in both areas, current methods often suffer from inefficiencies and limitations due to underdeveloped theoretical and algorithmic foundations, particularly in self-supervised representation learning (SSRL) and large reasoning model (LRM) training. This section reviews the underlying foundations and identifies recent breakthroughs and future directions.

\paragraph{Algorithmic Foundations of Self-Supervised Representation Learning (SSRL)}
Practical frameworks such as SimCLR~\cite{simclrv1}, MoCo~\cite{DBLP:conf/cvpr/He0WXG20}, SwAV~\cite{DBLP:journals/corr/abs-2006-09882}, DINO~\cite{oquab2024dinov}, CLIP~\cite{radford2021learning}, and SogCLR~\cite{yuan2022provable} have made SSRL highly effective for visual data. Many of these methods—, especially contrastive approaches, rely on mini-batch contrastive losses, which require either very large batch sizes or memory banks to approximate global similarities effectively. This leads to optimization inefficiencies due to gradient estimator variance. To address this, Yuan et al.~\cite{yuan2022provable} proposed a global contrastive loss formulation using a finite-sum coupled compositional optimization (FCCO) framework~\cite{wangsox}, revealing the theoretical limitations of mini-batch-based contrastive losses. They developed SogCLR, an algorithm that ensures convergence even with small batch sizes, outperforming SimCLR at equivalent performance with significantly fewer resources (e.g., batch size 256 vs. 8192). Building upon this, Qiu et al. introduced iSogCLR~\cite{qiu2023not}, linking global contrastive learning to distributionally robust optimization by incorporating individualized temperature tuning. Additionally, TempNet~\cite{qiu2024cool} learns a neural network to predict personalized temperature schedules for CLIP training, further improving training stability and adaptability. NexusFlow \cite{lin2026nexusflow} proposes a method to allow distinct task heads to provide complementary supervision during multi-task learning.

\paragraph{Theoretical Foundations of SSRL}
Several theoretical studies have attempted to explain the success of contrastive SSRL with respect to generalization, feature learning, and downstream transferability~\cite{DBLP:conf/icml/SaunshiPAKK19, haochen2021provable, DBLP:conf/icml/LeiYYZ23}. However, these works often face major drawbacks:
\ding{182} Unrealistic assumptions, such as conditional independence of augmentations given labels~\cite{DBLP:conf/icml/SaunshiPAKK19};
\ding{183} Over-simplified objectives that lack practical relevance~\cite{haochen2021provable};
\ding{184} Unsound theoretical claims, sometimes stemming from weak formulations or approximations.

To overcome these issues, a promising direction is to develop a principled statistical framework that unifies theory and practice. Wang et al.~\cite{wang2025on} proposed such a framework using discriminative probabilistic modeling in a continuous domain. Their work shows that both mini-batch-based and global contrastive losses are biased estimators of the desired learning objective, leading to non-vanishing generalization error. To address this, they introduced a multiple-importance-sampling estimator with non-parametric estimation of per-sample weights, paving the way for more statistically consistent self-supervised learning.

\paragraph{Foundations of Large Reasoning Models (LRMs)}
Current techniques for fine-tuning LRMs typically follow three paradigms: \ding{182} supervised fine-tuning (SFT) uses next-token prediction over labeled input-output data, \ding{183} reinforcement learning (RL) with synthetic data optimizes models based on rewards from rule-based or simulation environments, and \ding{184} preference optimization (PO) fine-tunes models using pairwise human feedback (e.g., "A is preferred over B"). While RL is still in early-stage exploration for large-scale systems, SFT and PO have become standard for instruction-tuning and alignment. Pioneering works in PO~\cite{christiano2017deep, ouyang2022training} used reward models trained from human-labeled preferences. More recently, Direct Preference Optimization (DPO)\cite{rafailov2023direct} replaced reward modeling with a direct objective based on preference data, prompting many variants:
R-DPO~\cite{Park2024DisentanglingLF}, CPO~\cite{xu2024contrastive}, IPO~\cite{ipo_2022}, SimPO\cite{meng2024simpo}, KTO~\cite{ethayarajhmodel}, ORPO\cite{hong2024orpo}, and DPO-p~\cite{pal2024smaug}, among others~\cite{zhao2023calibrating, jung2024binaryclassifieroptimizationlarge}. Despite empirical progress, theoretical understanding of these methods remains limited. To bridge this gap, Guo et al.~\cite{guo2025discriminativefinetuninggenerativelarge} proposed Discriminative Fine-Tuning (DFT), a principled alternative to SFT. DFT replaces the generative paradigm with a discriminative objective that maximizes the likelihood of preferred responses while downweighting less-preferred ones, shifting from next-token prediction to full-response classification. This aligns better with the nature of human preference data and provides a more theoretically grounded framework for fine-tuning LLMs.

\begin{tcolorbox}[colback=gray!5!white, colframe=black!50!gray]
    \begin{itemize}[leftmargin=0em]
        \item \fontsize{9}{10}\selectfont
              \textbf{Opportunities:} 
              \textit{Opportunities for future research in this area lie in establishing robust theoretical and algorithmic foundations that can unify and advance the development of end-to-end autonomous driving systems. Key directions include improving the statistical consistency and optimization efficiency of self-supervised representation learning through principled loss functions and adaptive mechanisms, as well as building more theoretically grounded frameworks for fine-tuning large reasoning models using human preferences. Bridging these advances with multimodal learning, robustness to rare events, and generalization across environments could enable scalable, reliable, and interpretable autonomous driving systems that are both data-efficient and safety-aware.} 
    \end{itemize}
\end{tcolorbox}

\subsection{Digital Twin and Real2Sim2Real Generalization}
\label{diginal_twin}


\begin{wrapfigure}{R}{0.50\textwidth}
    \includegraphics[width=\linewidth]{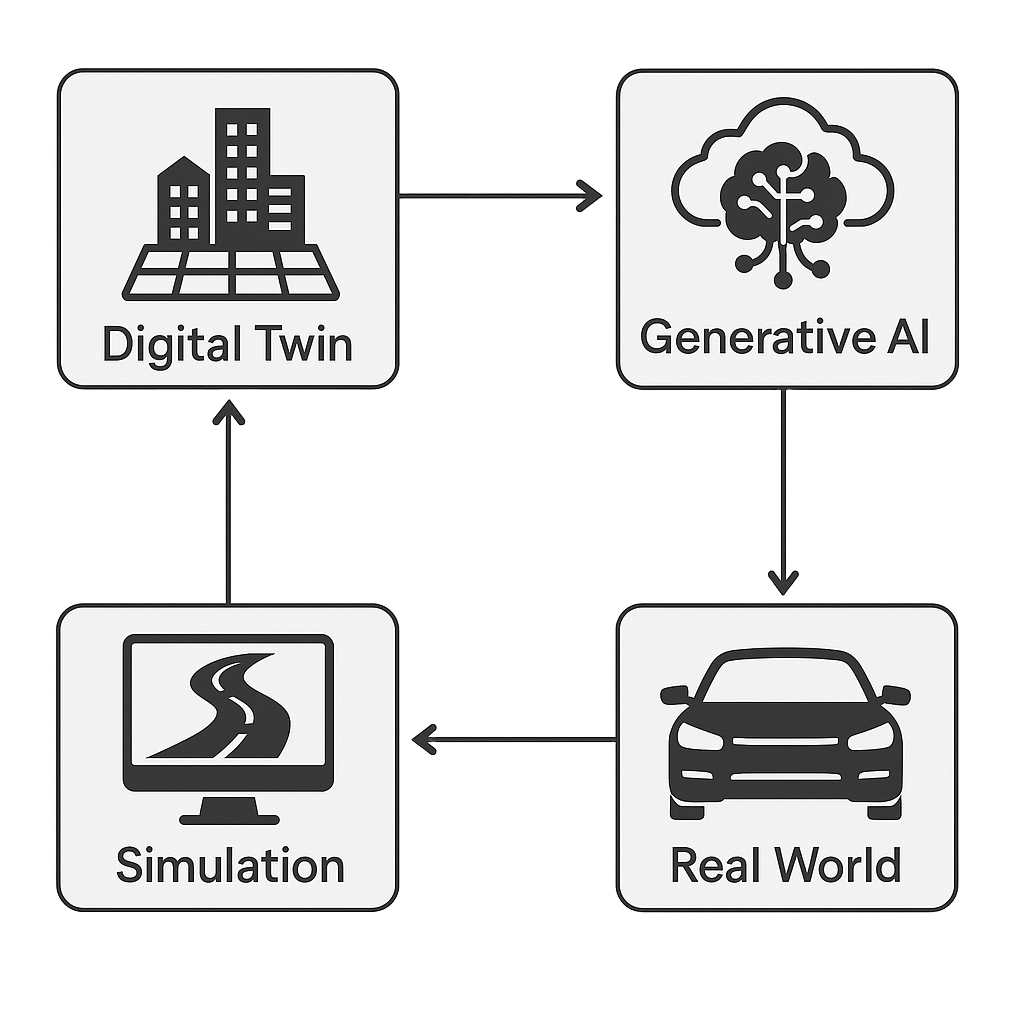}
    \caption{Closed-loop Real2Sim2Real framework powered by digital twins and generative AI. Real-world data informs the construction of high-fidelity digital twins, which are enhanced by generative models to create diverse and realistic simulation scenarios. These simulations are used to train and test autonomous systems, whose behaviors are validated in both virtual and physical environments, enabling continuous refinement and robust deployment.}
\end{wrapfigure}

Digital twins, high-fidelity virtual replicas of physical environments, are becoming indispensable in autonomous driving research and development. 
Generative AI significantly enhances the power of digital twins by enabling Real to Sim to Real transfer learning and closed-loop testing. 
A notable example is the newly released open-source digital twin of the Mcity Test Facility (2024),
which mirrors a real 32-acre proving ground in simulation. This digital twin incorporates detailed road geometries, signage, and even varied road surface materials, 
and it works in tandem with generative traffic simulation (via Mcity's TeraSim \cite{sun2025terasim}) to populate the twin with other vehicles, pedestrians, 
and even randomized adversarial events. The benefit is twofold: 
engineers can drive millions of miles virtually in a digital clone of a real environment before the autonomous vehicle ever touches the physical track, 
and then seamlessly test the same scenarios in reality. 
This closed-loop cycle greatly accelerates iteration. 
As Mcity researchers note, one can precisely control factors in the virtual world, 
such as orchestrating pedestrian movements or weather changes, which would be random or impossible to repeat consistently in real life. 
By focusing and speeding up such testing, digital twins serve as a bridge between simulation and road testing. 

Despite these advances, building and maintaining such realistic digital twins is very expensive. For example, the Mcity
Test Facility can cost around \$2,400 per day for U-M faculty, while using a vehicle from the Open CAV fleet might
cost around \$400 per vehicle \cite{mcity_cost}. Such a high cost makes it difficult to build a digital twin for large-scale
areas such as a city or a state. Generative AI can be a solution to create virtual replicas of real-world environments by
reconstructing 3D or 4D environments from 2D images or videos \cite{gao2025citygs} using 3D Gaussian. Vid2Sim \cite{xie2024vid2sim} further postprocesses the generated 3D environment into mesh structures to enable object interaction and safety assessment.
Generative AI is also tackling the longstanding sim-to-real gap by making simulated sensor data and agent behaviors more realistic, 
and by learning simulation models directly from real-world data (real-to-sim). 
For instance, DriveDreamer \cite{wang2023drivedreamer} is a world model derived entirely from real driving videos that uses a latent diffusion model to capture the complexity of real-world scenes. 
It can generate controllable driving video features conditioned on text prompts or planned trajectories, and even predict plausible future actions for the ego-vehicle. 
By training on real trajectories (\textit{e.g.}, from the nuScenes dataset) and then using the model as a simulator, 
DriveDreamer exemplifies Real2Sim: the resulting simulations carry over statistical patterns and "common-sense" physics from reality. 
Likewise, VISTA \cite{gao2024vista} uses neural rendering and learned behavior models to generate realistic virtual sensor outputs (camera feeds, LiDAR point clouds, etc.).
This realism is crucial for Sim2Real transfer – models or driving policies trained in these simulators will generalize better when deployed on actual vehicles. 
Another emerging concept is closed-loop generative simulation, where an autonomous vehicle's actions affect the simulation in real time, enabling interactive training. 
Traditional open-loop replay of real data does not capture how an AI driver's behavior might influence other agents (or vice versa). 
New simulators like DriveArena \cite{yang2024drivearena} and LimSim++ \cite{fu2024limsim} aim to address this by using generative agent models that respond to the ego vehicle's maneuvers in a responsive loop, rather than following a fixed script. 
Such systems let researchers test counterfactuals: \textit{e.g.}, "If our autonomous vehicle aggressively accelerates at an intersection, how would surrounding vehicles and pedestrians react?" – a question that generative agents can answer by simulating plausible responses. 
This is closely tied to the Real2Sim2Real generalization: 
the goal is for the closed-loop behavior in sim to be so realistic that an autonomous vehicle control policy developed there can be deployed with minimal fine-tuning in the real world 
(and conversely, that data from the real world can continuously update and improve the simulator). 

What's worse, generative models still struggle with valid edge-case physics. 
Digital twins can capture static environments very well (maps, road rules, etc.), but making sure that generated dynamic events (like a crash unfolding) obey real physics is non-trivial. 
There is active work on combining classical physics engines with generative AI – 
for example, using a physics simulator like CARLA \cite{dosovitskiy2017carla}, MetaDrive \cite{li2021metadrive}, or NVIDIA Drive Sim in the loop to validate that a generative model's outputs (trajectories, collisions) are feasible. 
In the near term, a practical approach is hybrid: use generative models to propose scenario variants, then filter or fine-tune them with physics-based checks and real-world data calibration. 
In the longer term, one can imagine city-scale digital twins continually updated with live data (traffic feeds, weather) and using generative AI to explore "what-if" scenarios (like emergency rerouting or infrastructure changes) – 
effectively becoming virtual testbeds for both policy and technology before deployment in the real city.
Achieving robust Real2Sim2Real loops will be a cornerstone for safe and efficient autonomous vehicle deployment, and it represents a convergence of transportation and AR (augmented reality)/VR (virtual reality) technology with AI.
\begin{tcolorbox}[colback=gray!5!white, colframe=black!50!gray]
    \begin{itemize}[leftmargin=0em]
        \item \fontsize{9}{10}\selectfont
              \textbf{Opportunities:}
              \textit{Future research should focus on reducing the high cost of building realistic digital twins by leveraging generative AI to reconstruct large-scale environments from 2D images or video. To further bridge the sim-to-real gap, advancements are needed in neural rendering and behavior modeling that better reflect real-world physics, especially in rare or edge-case scenarios. Additionally, hybrid approaches that integrate generative models with physics-based validation present a promising path to ensure physical plausibility without sacrificing diversity. Ultimately, the goal is to develop continuously updated, city-scale digital twins that serve as comprehensive virtual environments for training autonomous systems and validating new policies and technologies prior to real-world deployment.}
    \end{itemize}
\end{tcolorbox}

\subsection{Integration with Vehicle-to-Everything (V2X) Cooperative Systems}

\begin{wrapfigure}{L}{0.40\textwidth}
\includegraphics[width=\linewidth]{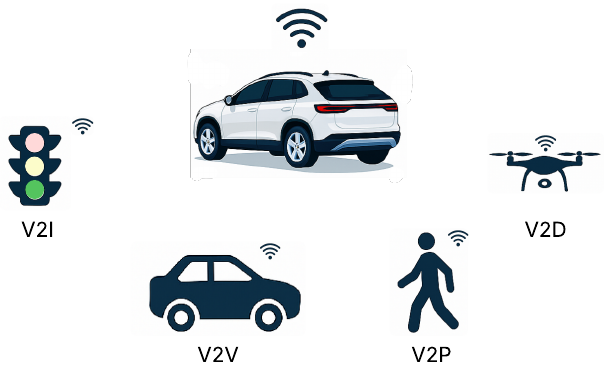}
    \caption{Generative AI-enhanced Vehicle-to-Everything (V2X) communication, enabling real-time coordination between autonomous vehicles, infrastructure, pedestrians, drones, and other road users through wireless connectivity.}
\end{wrapfigure}

As autonomous vehicles become increasingly connected, generative AI offers new opportunities to enhance Vehicle-to-Everything (V2X) cooperation, including Vehicle-to-Vehicle (V2V)~\cite{xu2023v2v4real,yu2025end,li2024comamba,song2024collaborative}, Vehicle-to-Infrastructure (V2I)~\cite{hu2024collaborative,xu2022v2x,xu2024v2x,wu2025v2x,you2024v2x,zimmer2024tumtraf, zimmer2024tumtraf}, Vehicle-to-Pedestrian (V2P)~\cite{gao2025langcoop}, and Vehicle-to-Drone (V2D) communication. Cooperation is widely seen as a force multiplier for safety and efficiency: indeed, some researchers argue that information exchange via V2X is the foundation for ethical driving and that autonomous vehicles must be cooperative to resolve multi-agent dilemmas safely \cite{sidorenko2023ethical}. 
Generative AI can contribute by coordinating and predicting multi-agent interactions~\cite{wei2021we} in ways that deterministic systems cannot. For example, a generative model could simulate the likely trajectories of other vehicles beyond line-of-sight, based on V2X messages about their intent or environment, effectively giving an autonomous vehicle a bird’s eye view of hidden hazards.
This relates to cooperative perception: through V2X, vehicles can share sensor data and warnings, allowing them to detect occluded or distant objects that their own sensors might miss \cite{huang2023v2x}. 
A diffusion or transformer-based model could take such shared data and generate a unified scene representation, filling in gaps with plausible detail (\textit{e.g.}, imagining the motion of a pedestrian that one car’s camera sees but another car around the corner can only infer). 
By fusing real observations and generative predictions, overall situational awareness is improved. 

Cooperative maneuvers are another area ripe for generative approaches. In current research, multi-agent reinforcement learning and game-theoretic models are used for vehicles negotiating merges or intersections. 
Generative models, especially those that can produce a distribution of possible behaviors, could enhance these by proposing creative solutions or negotiating strategies. 
For instance, two autonomous vehicles approaching a narrow bridge might implicitly “communicate” by sharing their internal generative forecasts: each car’s AI could simulate both itself and the other car in a variety of yielding/give-way decisions and agree on a plan that avoids collision. 
In essence, the vehicles would be performing a kind of coordinated counterfactual reasoning via generative world models. Early steps toward this vision can be seen in CMP \cite{wang2025cmp} and related frameworks that aim for unified V2X-integrated planning. 
A recent and more comprehensive development in this direction is Language-based cooperative driving \cite{gao2025langcoop}, which enables inter-vehicle negotiation through language-based message generation, enabling high communication efficiency and rich information exchange.
Its LLM-driven reasoning engine performs multi-turn deliberation based on self-observation, received messages, and recalled experience, facilitating safe and interpretable multi-agent cooperation at complex scenarios. 
This illustrates how generative models can not only synthesize shared V2X predictions but also actively coordinate future actions.

On the infrastructure side (V2I), generative AI could help traffic management centers communicate with autonomous vehicles more effectively. 
However, a significant challenge exists in connectivity and interoperability—autonomous vehicles are not naturally collaborative when equipped with different communication devices, algorithms, or when performing different tasks. 
This heterogeneity of agents remains a fundamental challenge, though approaches like central protocols have been proposed to address it \cite{gao2025stamp}. 
Despite these challenges, in the near term, we expect straightforward applications like natural language communication between vehicles and infrastructure. 
Imagine a vehicle that can parse a spoken or textual message from a roadside unit about an accident ahead and then generate a safe maneuver or route adjustment in response. 
This marries V2X data with the kind of language-based generative reasoning that large language models (LLMs) excel at \cite{gao2025langcoop}. 
Longer-term, more sophisticated collaborations become possible. A smart traffic light might broadcast a generative model's recommendation for approaching vehicles, for example, an adaptive speed profile to optimize traffic flow by simulating various possibilities. 
Conversely, an autonomous vehicle fleet could collectively generate an emergent traffic signal timing plan on the fly during abnormal congestion, essentially self-organizing the intersection control. 

The challenge of reliability and security in cooperative systems also remains unsolved. V2X channels have latency and packet loss: a generative model must account for uncertain or delayed information. 
Moreover, malicious actors could inject false data – a concern since an AI might hallucinate a very realistic but fake scenario if fed tampered V2X inputs. Robustness techniques (such as verifying consistency between an autonomous vehicle’s own sensors and the shared info) will be needed. 
Standards will also play a role: bodies like IEEE and 3GPP (for C-V2X 5G communication) may need to define new message types to convey AI-generated content (such as predicted occupancy grids or risk maps). 
International regulators are beginning to include connectivity in their automation frameworks; for example, the United Nations Economic Commission for Europe (UNECE) is updating its recommendations to account for connected ADS, and cooperative automation is seen reaching higher SAE levels safely \cite{unece29}. 
In the longer term, cooperative generative AI could extend beyond cars to include smart cities – envision traffic infrastructure, cars, drones, and even pedestrians’ smartphones all exchanging data and generative predictions to optimize mobility as a whole. Achieving this will require aligning many stakeholders (automakers, city planners, telecom providers), but the potential benefits in safety (\textit{e.g.} virtually zero blind crashes) and efficiency (platooning, smooth traffic) provide a strong incentive to explore this integration.

\begin{tcolorbox}[colback=gray!5!white, colframe=black!50!gray]
    \begin{itemize}[leftmargin=0em]
        \item \fontsize{9}{10}\selectfont
              \textbf{Opportunities:}
              \textit{Future research should focus on developing generative models that can effectively coordinate multi-agent interactions by simulating trajectories beyond line-of-sight and generating unified scene representations from shared sensor data. 
              Language-based cooperative driving frameworks merit further exploration to enable efficient inter-vehicle negotiation through natural language messaging. 
              Additionally, researchers should address the challenges of heterogeneous agent integration, reliability in communication channels, and security against malicious data injection. 
              Developing robust standards for AI-generated content transmission and creating frameworks that extend cooperation beyond vehicles to include infrastructure, drones, and pedestrians' devices would move the field toward truly integrated smart mobility ecosystems that optimize safety and efficiency.}
    \end{itemize}
\end{tcolorbox}

\subsection{Traffic Operation}

\begin{wrapfigure}{R}{0.50\textwidth}
    \includegraphics[width=\linewidth]{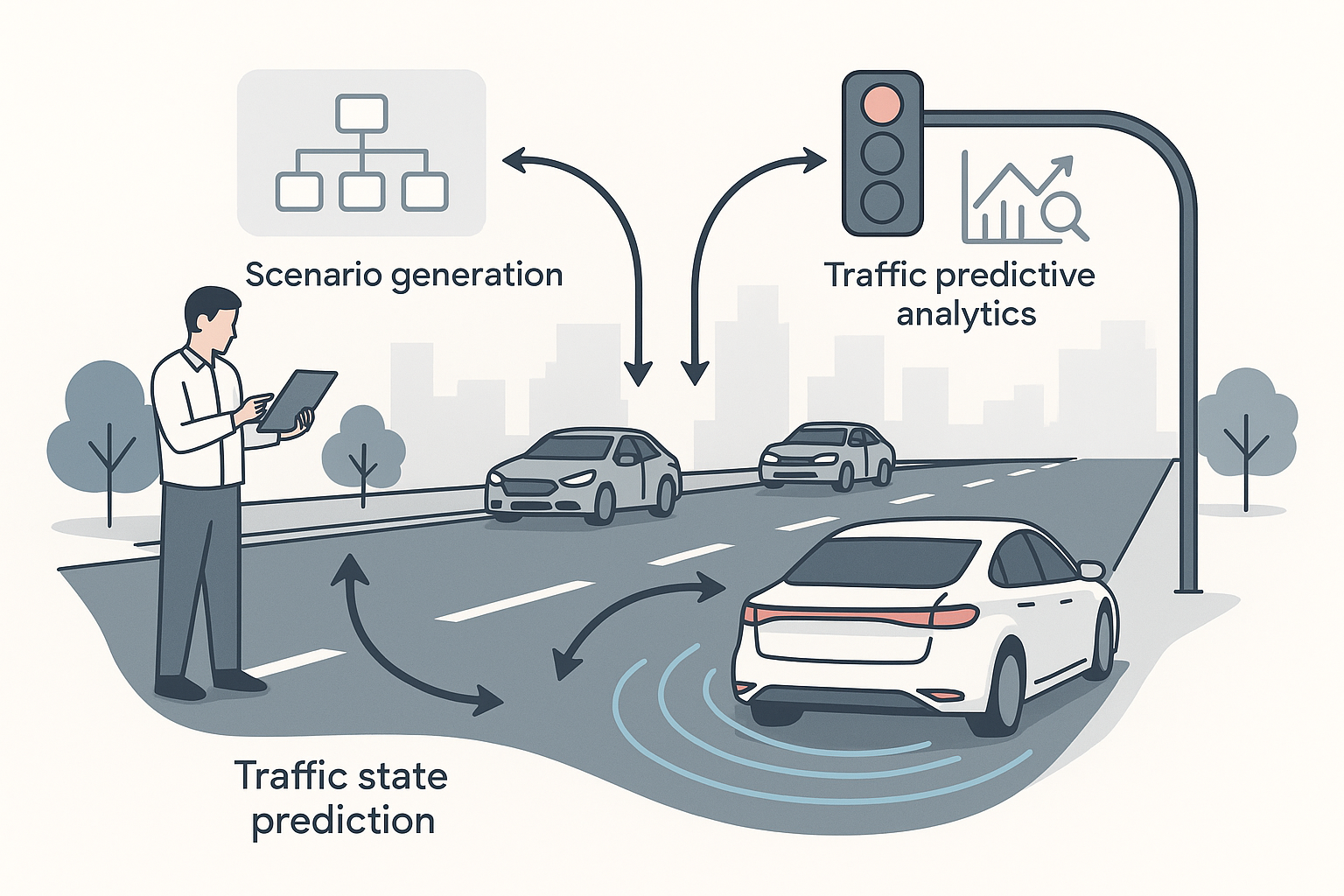}
    \caption{Generative AI in traffic operations enables scenario generation, predictive analytics, and traffic state forecasting to support adaptive control, optimize flow, and enhance safety in mixed-autonomy transportation systems.}
\end{wrapfigure}
Generative AI is transforming traffic operations for autonomous driving by enabling scenario generation and traffic predictive analytics to enhance safety and efficiency \cite{guo2024towards}. Furthermore, generative AI also allows for forecasting of dynamic traffic patterns and behaviors, providing the traffic managers with short-term predictions and supporting adaptive decision-making processes that optimize traffic flow, reduce congestion, and mitigate accident risks within autonomous transportation networks.

\paragraph{Traffic State Prediction}

As autonomous vehicles enter real-world transportation systems, traffic prediction becomes critical for transportation planning due to the emergence of mixed traffic environments \cite{zheng2023integrating, yan2023survey, lee2024congestion,mei2023uncertainty}. Autonomous vehicles and human-driven vehicles interact in complex ways, creating new, less predictable traffic patterns \cite{lee2023traffic}. Accurate forecasting is essential to manage lane allocation, signal control, and capacity planning, especially during the transition phase to full automation \cite{min2023deep, lee2021travel,wei2021recent,wei2019survey, yang2022toward}. However, conventional models struggle with AV-specific challenges such as varying automation behaviors, rare interactions, and a lack of representative historical data, especially in the early deployment phase \cite{lee2024automated}.

Generative models have advanced traffic prediction by modeling uncertainty, heterogeneity, and unobserved conditions. For example, Curb-GAN \cite{zhang2020curb} predicts traffic conditions before urban development plans are implemented. It uses conditional GANs with dynamic convolution and self-attention to learn spatio-temporal patterns. TrafficGAN \cite{zhang2019trafficgan} estimates traffic changes before construction starts by defining the problem of traffic estimation off-deployment. It applies dynamic filters to learn how traffic responds to demand and changes in the road network, outperforming baseline models. Similarly, GAN-based models, such as ForGAN \cite{koochali2019forgan}, D-GAN \cite{saxena2019dgan}, and SATP-GAN \cite{zhang2021satp}, improve probabilistic forecasting by capturing spatio-temporal patterns and external conditions. Jin et al. \cite{jin2022gan} proposed PL-WGAN, a short-term traffic speed prediction model for urban road networks. It uses Wasserstein GAN with GCN, RNN, and attention to capture spatio-temporal patterns and improve prediction accuracy. Furthermore, the ST-LLM \cite{liu2024spatial} and R2T-LLM \cite{guo2024responsible} frameworks demonstrate that LLMs can serve not only as predictive engines but also as interpretable decision support tools, addressing the problem of opacity of deep neural networks and supporting the deployment of responsible models.


\paragraph{Operation Performance Evaluation}

As autonomous driving technology enters the mainstream, accurately evaluating the performance of mobility systems becomes increasingly crucial \cite{garciaarca2018integrating, sariyer2024leveraging, lee2025iterative, lee2023equity}. With the growing complexity of AV-driven mobility environments, it is necessary to go beyond conventional approaches. In earlier stages, AI and data-driven methodologies were widely used in performance evaluation \cite{lee2024explainable, moslem2023systematic}. 

Recent advances in generative AI have significantly transformed performance monitoring by providing nuanced, personalized, and adaptable evaluation approaches. For example, Wang et al. \cite{wang2025systematic} conducted a systematic evaluation of generative models, introducing a novel graph-based metric specifically tailored for transportation network evaluation. Their results revealed considerable differences between synthetic and real data, highlighting the need for transportation-specific generative models. Jiang et al. \cite{jiang2024genai} introduced GenAI-Arena, an open evaluation platform that uses collective user feedback to provide robust evaluations of generative AI models. This user-driven evaluation addresses the limitations of traditional automatic metrics (\textit{e.g.}, FID, CLIP), offering more accurate reflections of real-world user satisfaction. Doshi et al. \cite{doshi2025generative} demonstrated generative AI's utility in strategic decision-making by aggregating evaluations from multiple large language models (LLMs). Their research indicated that while single generative evaluations might be inconsistent, aggregated assessments closely align with human expert judgments. Zhou et al. \cite{zhou2024alpharank} developed AlphaRank, a deep reinforcement learning-based method utilizing Monte Carlo simulations for ranking and selection problems. AlphaRank effectively manages trade-offs among mean, variance, and correlation, critical factors in evaluating autonomous driving performance. Zou et al. \cite{zou2025cognitive} proposed the Cognitive Tree framework, integrating Retrieval-Augmented Generation to prioritize critical real-time information dynamically.

\begin{tcolorbox}[colback=gray!5!white, colframe=black!50!gray]
    \begin{itemize}[leftmargin=0em]
        \item \fontsize{9}{10}\selectfont
              \textbf{Opportunities:}
              Generative AI enables more accurate prediction and realistic evaluation of mixed traffic environments by synthesizing diverse scenarios and capturing complex spatio-temporal dependencies, uncertainty, and external factors such as infrastructure changes and demand fluctuations. These capabilities support robust performance testing across critical dimensions—travel efficiency, accessibility, network reliability—and enhance proactive traffic management strategies like adaptive lane management, signal control, and capacity planning, essential for integrating autonomous vehicles effectively into dynamic transportation systems.
    \end{itemize}
\end{tcolorbox}

\subsection{Transportation Planning}


\begin{wrapfigure}{L}{0.50\textwidth}
    \includegraphics[width=\linewidth]{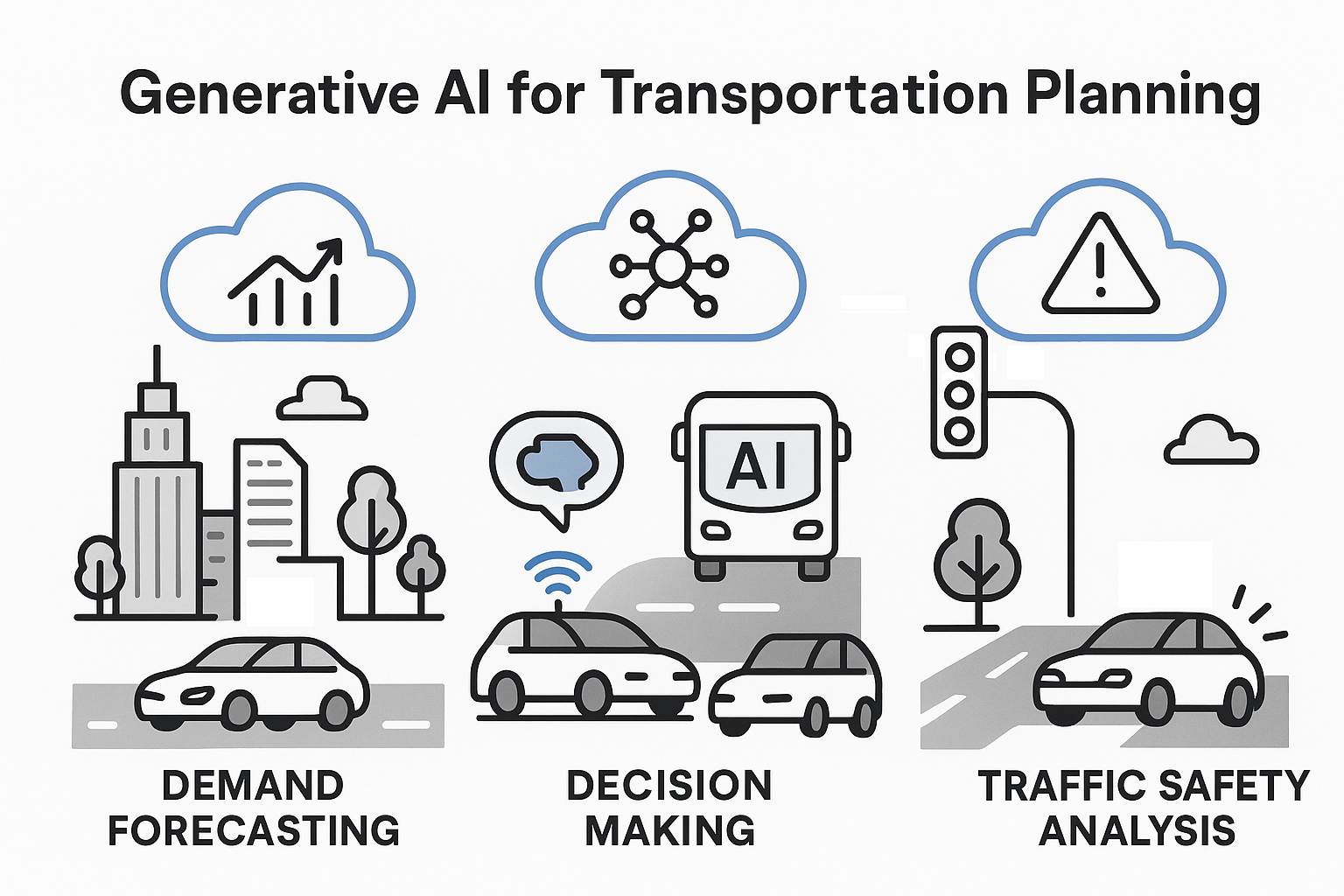}
    \caption{Generative AI improves demand forecasting, decision making, and traffic safety analysis.}
\end{wrapfigure}

Transportation planning is entering a new era shaped by autonomous driving and Generative AI. Their convergence is redefining how mobility systems are designed and managed, marking a paradigm shift in the transportation domain~\cite{da2025generative}. Traffic planners can leverage Generative AI’s data-driven insights and simulations to support the complex decision-making required for next-generation transportation networks.

\paragraph{Demand Forecasting}

One of the key concerns in transportation planning is that the introduction of new travel modes leads to changes in user behavior \cite{bruton2021introduction, yan2023survey, lee2022exploring, lee2021estimating, yao2019revisiting, wang2019unified}. The autonomous vehicle technology is not merely the addition of a new driving mode; it also fundamentally reshapes travel behavior by extending trip lengths, increasing travel flexibility, and activating latent demand. While overall trip frequency may not drastically increase, autonomous vehicles are expected to influence trip timing, chaining, and modal preferences \cite{dannemiller2021investigating, harb2018projecting, harb2022glimpse, farooq2018virtual, zmud2016consumer, salonen2019towards}. 

Generative AI addresses longstanding challenges in demand forecasting, particularly the scarcity of behavioral data and difficulty in modeling complex interactions. Zhang et al. \cite{zhang2024generative} proposed a semantic-aware framework for generative AI-enabled vehicular networks, applying deep reinforcement learning to optimize navigation and data transmission. Da et al. \cite{da2024openti} introduced Open-TI, a traffic intelligence framework powered by LLMs that handles full-cycle demand analysis and simulation through agent-to-agent communication. Zhang et al. \cite{zhang2024trafficgpt} presented a hybrid of LLMs and traffic foundation models that interactively process traffic data and aid decision-making. Movahedi and Choi  \cite{movahedi2024crossroads} showed that LLM-based adaptive signal controllers significantly improve urban flow efficiency, while Tang et al. \cite{tang2024llmcontrol} proposed a flexible traffic control architecture supporting autonomous and human-in-the-loop strategies. Kim et al. \cite{kim2024flexible} introduced a partially monotonic discrete choice model (DCM-LN) that incorporates domain knowledge while preserving the flexibility of machine learning, enhancing both accuracy and interpretability. In population synthesis, Kim et al. \cite{kim2023deep} also proposed a deep generative model that corrects structural and sampling zeros, producing more feasible and behaviorally consistent synthetic populations for agent-based models. \cite{rong2023goddag, zheng2021rebuilding} developed a graph neural network (GNN)-based generative model that uses domain adversarial training to predict origin-destination (OD) flows in cities with little or no historical mobility data.


\paragraph{Decision Making}

Generative AI is poised to transform decision-making at both the vehicle level and the transportation system level (urban planning and traffic management) \cite{zewe2023computers}. On the vehicle side, there is a shift toward viewing autonomous planning as a generative problem: rather than reacting with fixed rules, the autonomous vehicle imagines multiple future possibilities and then plans accordingly. A salient example is the concept of Generative End-to-End Driving. Traditional autonomous driving pipelines separated perception, prediction, and planning, which can ignore the coupled nature of these tasks. New approaches like GenAD (a 2024 framework) \cite{zheng2024genad} instead train a single model to generate the future trajectories of all agents (ego vehicle and others) in a structured latent space. By doing so, the model captures high-level interactions – for instance, how the ego vehicle’s lane change might influence a nearby driver to brake, which in turn affects the ego’s optimal plan a few seconds later. This joint prediction and planning through generative trajectory roll-outs has yielded more robust performance in simulations; GenAD demonstrated state-of-the-art vision-based planning on benchmarks by learning a latent “common sense” prior about feasible driving trajectories. In essence, the autonomous vehicle’s decision-making becomes imagination-driven: the better it can dream up what might happen next (within the bounds of realism), the better it can prepare and choose a safe action. 

Integration of LLMs and knowledge-based AI offers another dimension for decision-making. Large language models, with their vast encoded knowledge (including traffic rules or even human preferences), can be used to reason about unique situations. Early research shows that LLM-based agents, for instance, can simulate the behaviors of vehicles~\cite{da2024prompt}. By exploring a vast solution space, GenAI may identify infrastructure adjustments or traffic management policies that yield smoother, safer travel in a mixed AV–human environment~\cite{yao2024comal}. Imagine an autonomous car approaching an unusual construction setup; a language model could interpret a text description or dialogue with an advisory system (``Two lanes merge into one 500m ahead due to construction”) and help the autonomous vehicle decide to merge early, explaining that ``vehicles in the closing lane will need to merge, best to create space now.” In 2023, we saw early examples of this: Wayve’s LINGO-1/2 \cite{lingo2} could explain and justify driving decisions in natural language while driving in London, demonstrating a form of cognitive planning where the model articulates the reasoning (``reducing speed for the cyclist” as it actually does so). This not only aids transparency (which builds trust, as discussed later) but could improve decision quality by injecting high-level semantic context into low-level planning. 

From a societal perspective, generative AI-enabled transportation planning can lead to more resilient and responsive traffic systems. For example, generative models can be used to simulate evacuation plans for cities \cite{sevim2025simulation, lee2024leveraging}: in dynamically changing scenarios (like an earthquake damaging certain bridges), a generative traffic model could help decision-makers evaluate the best routing of vehicles and identify bottlenecks before they happen, effectively stress-testing city preparedness. Agencies such as departments of transportation might, in the future, maintain AI traffic twins of their cities, continuously running “what if” scenarios in the background. In the near term, however, the low-hanging fruit might be using generative AI tools to assist engineers in day-to-day tasks: we are already seeing products \cite{automotivetestingtechnologyinternationalTieupBetween} where one can query massive driving data logs with natural language (using generative AI) to find, say, all instances of illegal jaywalking in a dataset. This drastically cuts down the effort to gather decision-making corner cases for analysis. 

A remaining challenge is the validation and accountability of AI-driven decisions. Whether it is an autonomous vehicle making a split-second maneuver or a city deciding on a traffic scheme based on AI simulation, stakeholders will demand assurance that the decisions are sound. This is pushing research into interpretable generative models and probabilistic guarantees. For instance, an autonomous vehicle’s planning module might generate 1000 possible forecasts, but it should report not just the plan it chooses, but its confidence among the outcomes (addressing the question: What if the generative model is wrong?). At the city level, planners will need ways to interrogate the AI’s suggestion: “Why do you recommend narrowing this road?” – the AI should ideally answer in human terms (“Because in 90\% of simulations it reduced overall delay by 20\% without causing overflow to neighboring streets”). Such human-AI collaborative decision-making loops will become more common as trust in generative systems grows. In summary, generative AI is opening new frontiers in transportation decision making, enabling both micro-scale agility in vehicle behavior and macro-scale insights for urban mobility, but marrying these capabilities with the prudence and transparency required for public deployment remains an active area of development.


\paragraph{Traffic Safety Analysis}
Generative AI has emerged as a powerful tool for traffic safety analysis in autonomous driving, enabling an end-to-end pipeline that encompasses environmental sensing, behavior prediction, and risk evaluation. Its impact is particularly pronounced in active safety analysis, which aims to quantify potential traffic risks and proactively prevent crashes and incidents \cite{fan2024learning}. Unlike passive safety analysis, which relies on historical crash data and focuses on macroscopic traffic patterns~\cite{du2023safelight,du2023safety}, active safety analysis operates at a microscopic level. This approach directly considers specific traffic scenarios, driver behaviors, traffic signals, and vehicle dynamics to uncover the causal mechanisms behind traffic incidents rather than merely identifying correlations. Given the rarity and randomness of traffic events, an active safety approach is especially valuable in the realm of autonomous driving.

In the sensing phase, generative AI can improve data augmentation by producing realistic variations in weather, lighting, and road conditions based on inputs from aerial LiDAR scans, open-source mapping tools, and vehicle-mounted sensors (\textit{e.g.}, GPS, IMU and inclinometer data etc.). These synthetic variations bolster the robustness and generalizability of perception modules, which often struggle with out-of-distribution or rare events in real-world traffic \cite{ma2021multi,li2024adaptive, jiang2024cost, yang2024mitigating, yang2023cooperative, tian2025nuscenes}.

For behavior prediction and scenario generation, generative models learn intricate patterns of driver–vehicle interactions, including human-driven and autonomous vehicles operating in mixed traffic~\cite{wu2023graph, wu2024hypergraph}. By modeling stochastic and interactive behaviors, these systems can produce plausible trajectories and maneuvers that reflect realistic or even adversarial conditions. Such fidelity is especially crucial for identifying potential safety-critical situations that might otherwise be overlooked in deterministic simulations.

From a broader perspective, generative AI opens new opportunities for improving traffic safety by enabling the creation of synthetic risk maps at the city scale. Because traffic accidents are rare and real-world datasets often lack sufficient information for training, it is difficult to model safety-critical events using empirical data alone. By generating accident-prone scenarios and simulating high-risk locations, generative models can help identify hazardous areas and proactively inform autonomous driving systems of potential risks. Cai et al.~\cite{cai2020real} applied a deep convolutional generative adversarial network to balance imbalanced crash data and improve real-time crash prediction accuracy on expressways, outperforming traditional oversampling methods. Ding et al.~\cite{ding2022deep} proposed an augmented variational autoencoder to generate synthetic crash data for crash frequency models, effectively addressing excessive zero observations and enhancing model performance with heterogeneous data. Man et al.~\cite{man2022wasserstein} introduced a Wasserstein Generative Adversarial Network to handle extreme class imbalance in real-time crash risk prediction, achieving higher sensitivity and lower false alarm rates compared to conventional oversampling techniques. These synthetic insights can be used to support traffic management decisions, infrastructure planning, or onboard vehicle safety strategies, especially in regions where historical data is sparse or outdated~\cite{mei2023uncertainty, wei2020learning, mei2023reinforcement}.

Given the ambient traffic behavior, traffic infrastructure, and contextual background, surrogate safety measures (SSMs), such as time-to-collision (TTC) and deceleration rate to avoid collision (DRAC), are employed to quantify potential risks \cite{wang2021review, wang2024surrogate}. Essentially, each vehicle's motion can be described by ordinary differential equations (ODEs), and the evolution of relative distance along with other state variables can similarly be formulated as ODEs. Solving these equations provides an accurate estimation of TTC when the inter-vehicle gap falls below a predefined safety threshold. Although traditional SSMs often fail to incorporate high-fidelity vehicle dynamics and are typically limited to one-dimensional or piecewise one-dimensional analyses, recent work \cite{li2024beyond} introduces a generic approach to compute TTC regardless of the complexity of the vehicle dynamics or the analysis space. Extensions of this methodology are further demonstrated in \cite{zhang2024anticipatory, li2025adaptive, li2024disturbances, wu2025ai2active}, collectively advancing the field of traffic risk quantification. Besides the analytical solution, high-fidelity vehicle dynamics simulators \cite{tasora2016chrono, nvidia_physx} can also be applied to further assess the traffic safety.

Future investigations into GenAI-based traffic safety analysis for autonomous driving should pursue multiple key directions. For instance, incorporating advanced uncertainty quantification methods \cite{li2024autonomous, onal2024gaussian, schlauch2024informed} can yield calibrated risk assessments instead of overly confident single-value predictions, while integrating generative models into live digital twins \cite{zhang2025virtual, wang2021digital, wu2025digital, pumarola2021d, chen2024s} enables continuous, hardware-in-the-loop stress testing. A promising strategy is to merge data-driven generative frameworks with physics-based modeling \cite{giang2024conditional, djeumou2024one, tian2025physically, gan2025planning}, thereby accounting for the probabilistic behavior of drivers alongside the deterministic dynamics of vehicles. Moreover, adopting self-supervised learning \cite{salzmann2020trajectron++, tsuchiya2024online} and reinforcement learning \cite{shi2023physics, shi2021connected, shi2025predictive, yue2024hybrid} approaches can facilitate the ongoing refinement of synthetic scenarios, ensuring that synthetic data remains representative of evolving traffic conditions and emerging behavioral patterns. In parallel, adversarial or policy-informed scenario generators \cite{he2024trustworthy, xu2025diffscene, zhang2024chatscene, hao2023adversarial} can expose rare, high-impact edge cases. Crucially, extending these frameworks to incorporate end-to-end vehicle control optimization—through safe reinforcement learning policies and dynamics-aware control architectures—ensures that scenario-derived insights directly inform and enhance real-world control strategies for hazard mitigation \cite{li4940014nonlinear,li2024sequencing, zhang2023adaptive}. Finally, establishing uniform diversity and coverage benchmarks—together with harmonizing scenario outputs to accepted regulatory formats \cite{andreotti2020mathematical, chen2022generating}—will be critical for certification processes and widespread industry uptake.

\begin{tcolorbox}[colback=gray!5!white, colframe=black!50!gray]
    \begin{itemize}[leftmargin=0em]
        \item \fontsize{9}{10}\selectfont
              \textbf{Opportunities:}
              \textit{Generative AI enables autonomous vehicles to continuously generate and evaluate rare, high-risk scenarios. This is achieved by integrating robust uncertainty quantification with real-time digital twin simulations and hybrid data-driven and physics-based generators. Continuous self-supervised and reinforcement learning loops keep synthetic data aligned with evolving traffic patterns, while adversarial and policy-guided frameworks uncover critical edge cases. Unified diversity and coverage metrics, anchored to regulatory scenario definitions, further streamline certification processes and guide targeted infrastructure upgrades.}
    \end{itemize}
\end{tcolorbox}

\subsection{Economic Impacts of Autonomous Vehicles}

Autonomous driving will have significant socioeconomic impacts by transforming various aspects of daily life, including residential accessibility, job accessibility, and travel efficiency. Traditionally, transportation planners assess the economic value of mobility services within communities to inform decisions on budget distribution, infrastructure development, and expansion of services. Common evaluation criteria include changes in travel demand, accessibility, operational efficiency, and broader socioeconomic outcomes. However, because large-scale autonomous vehicle deployment has yet to occur, existing methods face limitations in empirically assessing the potential economic effects of autonomous vehicle integration.

For example, Metz \cite{metz2018developing} suggested that the use of autonomous vehicles in shared services could reduce the cost of taxi and public transportation operations. Shafiei et al.  \cite{shafiei2023impact} demonstrated that an increase in privately owned autonomous vehicles could have a negative impact on traffic congestion and empirically analyzed, through a case study in Melbourne, Australia, that a distance-based pricing scheme could mitigate these adverse effects.
Zhou et al. \cite{zhou2022review} pointed out that current research on machine learning-based longitudinal motion planning (mMP) for autonomous vehicles has focused primarily on safety, with insufficient attention to congestion mitigation, and emphasized the need to incorporate traffic efficiency objectives into future mMP training frameworks. Overtoom et al. \cite{overtoom2020assessing} noted that shared autonomous vehicles could create new bottlenecks due to their distinct driving behavior compared to conventional vehicles and suggested that infrastructure improvements, such as kiss-and-ride (K\&R) facilities, could alleviate these side effects.
Talebpour et al. \cite{talebpour2017investigating} found that operating dedicated lanes for autonomous vehicles could positively influence traffic flow and travel time reliability, particularly when the market penetration rate of autonomous vehicles exceeds 30–50 percentages. Rossi et al. \cite{rossi2018routing} analyzed the routing and redistribution of shared autonomous vehicles using a network flow approach and proposed a congestion-aware algorithm, showing that properly coordinated vehicle movements do not worsen traffic congestion.
Finally, Van den Berg and Verhoef \cite{van2016autonomous} used a dynamic equilibrium model to show that although autonomous vehicles may increase road capacity and lower the value of travel time (VOT) for users, changes in departure time behavior could impose negative externalities on existing users. Collectively, these studies suggest that autonomous vehicle technologies have the potential to generate significant social benefits.

In this context, generative AI provides a practical tool for simulating and evaluating the economic impacts of autonomous vehicle deployment prior to real-world implementation. By generating realistic traffic and land use scenarios, generative models can estimate how autonomous vehicle services might affect travel times and cost-efficiency at the community level. 

For example, generative AI can be used to simulate the projected outcomes of various autonomous vehicle deployment models, such as dedicated autonomous vehicle lanes, mixed operations with freight and passenger vehicles, or AV-based shuttle services. Xu et al. \cite{xu_generative_2023} proposed an architecture that synthesizes unlimited conditioned traffic and driving datasets using generative AI in the vehicular mixed reality Metaverse to generate driving scenarios. Similarly, Jia et al. \cite{jia2024dynamic} developed a dynamic test scenario generation method for autonomous vehicles based on conditional generative adversarial imitation learning, enabling the evaluation of autonomous vehicles’ ability to handle dynamic and interactive traffic environments. Tuncali et al. \cite{tuncali2018simulation} generated simulation-based adversarial tests for autonomous vehicles equipped with machine learning components to evaluate their robustness and performance under challenging conditions. In each case, it becomes possible to quantify the expected benefits in terms of travel time savings, improved access to economic centers (\textit{e.g.}, employment hubs, healthcare facilities), and potential gains in service efficiency relative to conventional systems.

This approach allows planners to assess the socioeconomic impacts of autonomous vehicle adoption. Even in the absence of empirical autonomous vehicle data, generative AI enables scenario-based forecasting under a variety of assumptions and conditions. In addition, it supports smarter public investment decisions and facilitates an efficient transition toward autonomous mobility systems.

\begin{tcolorbox}[colback=gray!5!white, colframe=black!50!gray]
    \begin{itemize}[leftmargin=0em]
        \item \fontsize{9}{10}\selectfont
              \textbf{Opportunities:}
              \textit{
             Generative AI offers significant future opportunities in transportation planning by enhancing prediction, simulation, and decision-making while building trust. It will enable advanced demand forecasting that captures AV-driven behavioral shifts using synthetic data and agent models. Proactive decision-making will emerge through 'imagination-driven' autonomous vehicles generating potential future trajectories and through system-level simulations optimizing traffic management and infrastructure under diverse conditions, including emergencies. Safety analysis will advance by continuously generating rare, high-risk scenarios—potentially within digital twins—integrating uncertainty quantification, hybrid data-physics models, and continuous learning loops (like RL and adversarial methods) to identify edge cases. Crucially, establishing interpretable models, effective human-AI collaboration, and standardized metrics tied to certification will be essential for validating these systems and guiding infrastructure improvements towards safer, more responsive transportation networks.} 
    \end{itemize}
\end{tcolorbox}

\subsection{Environmental Impacts of Autonomous Vehicles}

Autonomous driving and generative AI have complex environmental trade-offs. On one hand, there are clear potential benefits: autonomous vehicles can be optimally routed to reduce congestion and idling, they can drive more efficiently than humans (gentler acceleration, precise platooning, etc.), and generative data can reduce the need for fuel-burning test drives. For example, using a digital twin to test scenarios means fewer development vehicles running on test tracks or public roads, saving fuel and associated emissions. Mcity’s digital twin team highlighted that millions of miles can be tested virtually before a vehicle touches the real world, accelerating development while avoiding real-world mileage \cite{sun2025terasim}. Moreover, synthetic data generation might alleviate the need to drive fleets around just to gather corner-case data (which sometimes involves driving around hoping for rare events to occur). This could significantly cut down the vehicle miles traveled during the development phase of autonomous vehicles. 

Autonomous vehicles also dovetail with electrification. Many of the current robotaxis and test autonomous vehicles are electric vehicles (EVs). If generative AI leads to faster deployment of autonomous vehicles, and if those autonomous vehicles are predominantly EVs, there’s an environmental win in terms of reduced tailpipe emissions. Additionally, optimized traffic flow from widespread autonomous vehicle adoption could reduce overall emissions – studies have shown that smoothing stop-and-go traffic even slightly can have big fuel economy gains for everyone. Generative AI could be used by city planners to simulate and quantify these effects: \textit{e.g.}, creating a city-scale generative traffic model to compare emission levels with 0\%, 50\%, 100\% AV penetration. It could also help design eco-driving behaviors; for instance, a generative planner might prioritize energy efficiency in certain contexts (taking a route that is a bit longer but avoids steep hills to save battery in an EV, if time allows). 

On the other hand, the computational footprint of AI can be heavy. Training large generative models – whether they are vision models, world models, or language models for driving – requires substantial computational resources, often in energy-hungry data centers. There have been eye-opening estimates of the carbon footprint of AI: training a single big transformer-based model can emit on the order of hundreds of thousands of kilograms of $\text{CO}_2$ \cite{hao2019training}, equivalent to multiple lifetimes of driving a conventional car. Although these numbers are improving with efficiency and renewable energy, it is a factor to consider. If every automotive company trains its own massive driving model, the aggregate impact is non-trivial. Furthermore, once deployed, the onboard computers in autonomous vehicles run continuously and are far more power-demanding than a human brain. A 2023 MIT study \cite{zewe2023computers} warned that if we naively scale up autonomous vehicles, the energy used by their computation (sensors and CPUs/GPUs) could become a significant source of emissions, possibly outpacing the emissions we save by optimizing driving, unless hardware efficiency improves rapidly. They found that to keep computing emissions in check, given a growing autonomous vehicle fleet, we’d need to double computing efficiency roughly every 1.1 years, a pace even faster than historical Moore’s Law improvements. This calls for green AI practices: using energy-efficient algorithms, specialized hardware (like AV-specific chips optimized for neural networks), and training with renewable energy where possible. 

Another environmental angle is the effect of synthetic data on the need for real-world data collection. Collecting driving data can be energy-intensive (sending out cars to record sensor data). To the extent that generative models can create data, we save that energy. However, generating data on servers draws power too, so it is not free, but data center energy can be more easily offset or made renewable than gasoline burned on roads. It is a shift from distributed impact (lots of cars emitting) to concentrated impact (data centers). Policymakers might encourage autonomous vehicle developers to report the carbon footprint of their training and simulation efforts as part of a sustainability index. 

In the long term, autonomous driving could enable new services that have environmental benefits, such as dynamic ridesharing and robo-taxis, reducing the number of individually owned cars (thus reducing manufacturing emissions and land use for parking). Generative AI comes in by simulating these systems at scale: for example, generating travel demand patterns to see how a fleet of robo-taxis might satisfy mobility needs with far fewer vehicles than today’s private car ownership model. If successful, this can reduce the overall number of cars produced and resources consumed. On the flip side, easier travel (when a car drives itself, some might make trips they wouldn’t have before) could induce more travel demand, a known phenomenon called the rebound effect. Generative simulations can help study this: \textit{e.g.}, Uber and Lyft have used simulations to see how people might switch from public transit to autonomous vehicle ride-hailing if it is too convenient, which could increase congestion unless managed. 

An interesting twist is the environmental impact of infrastructure adaptation for autonomous vehicles. If we heavily instrument roads with sensors or V2X beacons to assist autonomous vehicles, that has its own energy/material footprint. Alternatively, if autonomous vehicles allow us to remove some infrastructure (like traffic signals in the far future, or streetlights if vehicles have night vision), there could be savings. These systemic effects are hard to predict, which is why city-scale generative modeling is valuable. In one optimistic scenario, autonomous vehicles plus generative urban planning might allow cities to reclaim land from parking lots for green spaces or urban forests (since autonomous pods could drop people off and go park themselves efficiently elsewhere, or not need parking if they’re in constant use). This could improve urban air quality and carbon sequestration locally. Such transitions may also benefit water quality and soil health by reducing surface runoff and land impermeability \cite{kwak2024deep}.

In summary, the net environmental impact of generative AI-powered autonomous vehicles will depend on how intelligently we deploy the technology. There is great potential for positive impact: smoother traffic, less idling, and fewer unnecessary miles driven (especially in testing) mean reduced emissions and energy use. But if we’re not careful, the computing emissions and potential increase in travel demand could offset those gains. The industry is aware of this balance. Efforts like OpenAI’s focus on efficiency, and automotive AI teams working on model compression and on-chip acceleration, are directly addressing the need to “green” the AI. One can imagine future autonomous vehicle marketing even highlighting energy-efficient AI as a feature (“Our self-driving AI runs on 50\% less energy than the competitor’s, giving you more range per charge”). Governments might also incentivize sharing of models to avoid redundant training – if multiple carmakers use a foundation model jointly (perhaps via a consortium), that could save the planet some duplicated training runs. Finally, as emphasized in a Communications of the ACM article, cutting the carbon footprint of AI is crucial as usage grows\cite{hao2019training}. For autonomous vehicles, this means innovation not just in how AI drives, but how AI is built and maintained, to ensure the autonomous revolution is also a green revolution.

\begin{tcolorbox}[colback=gray!5!white, colframe=black!50!gray]
    \begin{itemize}[leftmargin=0em]
        \item \fontsize{9}{10}\selectfont
              \textbf{Opportunities:}
              \textit{
              Future research opportunities include developing energy-efficient AI models and specialized hardware for autonomous vehicles to minimize computing emissions, creating standardized sustainability metrics (such as reporting the carbon footprint of AI training and simulations), and using generative AI to model the systemic environmental impacts of autonomous vehicle adoption at city scales. Further exploration is needed on how synthetic data can replace real-world testing without shifting emissions unsustainably to data centers, and how autonomous fleets might reshape urban infrastructure and mobility patterns to maximize environmental benefits while mitigating rebound effects. Collaborative approaches, like shared foundational AI models across companies and green AI training practices, also present critical areas for innovation.}
    \end{itemize}
\end{tcolorbox}

\subsection{Trustworthiness of Generative AI Models in Autonomous Driving}

Trustworthiness of AI requires AI models to be safe, accountable, fair, and ethical \citep{wing2021trustworthy}. Similarly, these attributes carry over to generative AI models. 
With the increasing role of generative AI in safety-critical driving tasks, the trustworthiness and safety assurance of these models have become paramount \cite{ huang2024position, wang2023decodingtrust, liu2023trustworthy,wang2024edit,wang2024sleepermark,ma2025position, huang2025trustworthiness,li2025secure,li2024dpu,li2025safeflow}. Generative models, especially large neural networks, are often black boxes that may produce plausible but incorrect outputs, a dangerous failure mode in driving. A core issue is uncertainty estimation: unlike traditional software, an AI might “improvise” in novel situations, so we need it to not only make a best guess but also know when it might be wrong. Recent research emphasizes quantifying the uncertainty of generative predictions (for example, predicting a distribution of trajectories rather than one deterministic path). Techniques like deep ensembles~\cite{lakshminarayanan2017simple, ganaie2022ensemble} and evidential neural nets~\cite{sensoy2018evidential, amini2020deep, ye2024uncertainty} are being applied to trajectory generators so that an autonomous vehicle can gauge when a situation falls outside its training distribution. In parallel, companies are developing runtime monitors. For example, Themis.AI \cite{themisaiUncertaintyAwareHuman} announced a “Risk-Aware Hallucination Detection” system to catch when any generative model (vision or language) starts producing dubious outputs. In the context of self-driving, this could mean an independent module checking if a generated scenario violates physical commonsense (say, a pedestrian appearing in two places at once in the model’s simulation) and flagging it.

\begin{wrapfigure}{R}{0.50\textwidth}
\includegraphics[width=\linewidth]{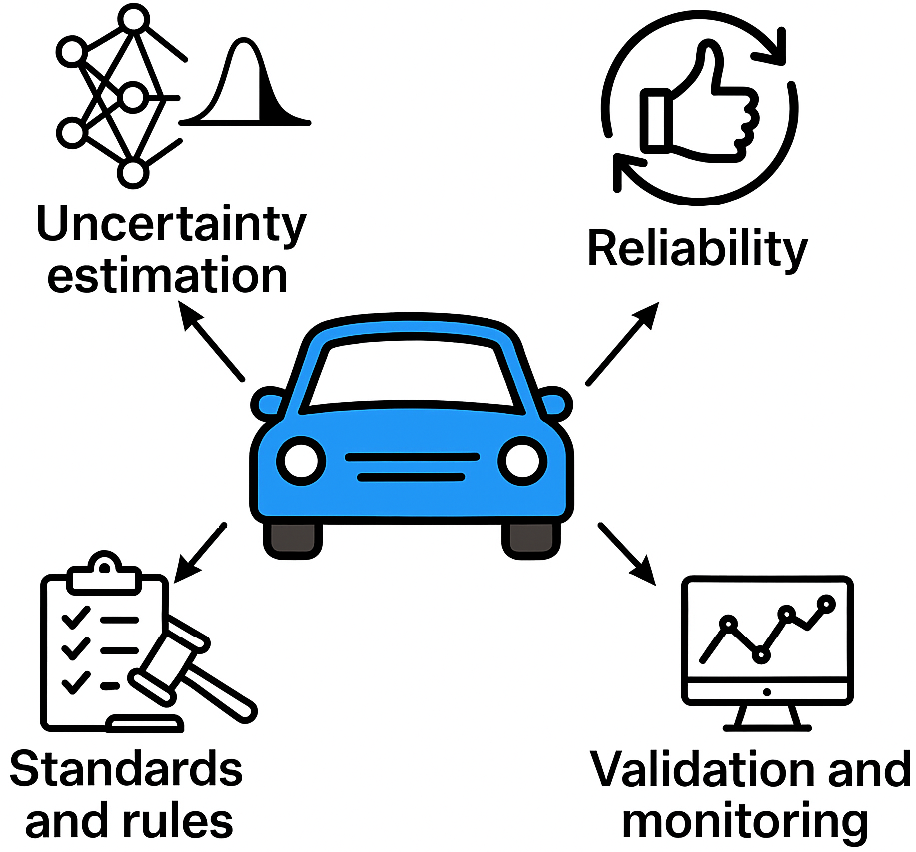} 
\caption{Trustworthy Autonomous Driving.}
\end{wrapfigure}

Another facet of trustworthiness is the integration with safety frameworks and standards. Traditional automotive systems undergo rigorous validation (ISO 26262 functional safety, etc.), and now, standards bodies are adapting these to AI. In 2023, the UNECE’s automotive working party began drafting guidelines on the use of AI in vehicles \cite{unece29}, aiming to recommend best practices so that AI components can be transparently evaluated. Similarly, ISO/PAS 8800 (road vehicles safety and AI) and UL 4600 (standard for autonomous product safety) are incorporating notions of runtime monitoring, fallback behaviors, and dataset management for AI. A likely outcome is that generative models will need to be paired with protective enveloping systems. For example, an autonomous vehicle planner might be generative, but an external rule-based “safety governor” monitors its suggested trajectories and will veto any that violate hard constraints (like leaving the roadway or exceeding safe deceleration limits). This concept of preventative and corrective safety layers is already present in designs like Waymo’s and Cruise’s stacks (they use multiple redundant systems). The challenge is ensuring the generative model and the rule-based safety net agree most of the time – otherwise the AI might propose maneuvers that get consistently blocked, which is inefficient. 

Validation of generative models poses new difficulties as well. Instead of just testing specific scenarios, one must test the model’s range of behaviors. This is where the earlier point about scenario generation loops back: ironically, we might use generative models (like scenario generators) to test other generative models (like driving policy networks) in simulation. This “AI vs AI” testing can flush out corner cases, but it is impossible to prove complete safety via testing alone because the space is so vast. Hence, researchers are exploring formal verification for neural networks. For simpler tasks like lane-keeping, some progress has been made verifying that a network will always keep within lane bounds given certain input ranges. For complex generative planners, formal methods are in infancy, but one can envision constraints (like conservation of momentum, collision avoidance) being enforced or checked during generation \cite{xing2024autotrust, xing2025re}. 

Trustworthiness also relates to consistency and reliability. A known issue with generative models is stochasticity: two runs might produce slightly different results. While diversity is good for exploration, in a deployed system, one usually wants predictability. Techniques such as seeded generation (to reproduce scenarios) or ensemble consensus (multiple generators agreeing on an outcome) could improve consistency. For instance, an autonomous vehicle could run two different generative models in parallel (perhaps one vision-based, one map-based) and only act when they largely agree on a plan; this is akin to N-version programming for AI \cite{ron2024galapagos}. 

\mpv{In the industry, Nvidia's Halos \cite{nvidia-halos}, launched in March 2025, is the pioneer solution to ensure the safety of drivers and other road users. Halos is a full-stack safety system designed to ensure the reliability of autonomous vehicles (AVs) from development to deployment. Halos integrates NVIDIA's hardware and software solutions, emphasizing AI-based, end-to-end AV stacks. It approaches the safety challenge from three levels: \ding{182} \textbf{Platform Safety}, to utilize safety-assessed systems-on-a-chip (SoCs) with built-in safety mechanisms, coupled with NVIDIA DriveOS, a safety-certified operating system, \ding{183} \textbf{Algorithmic Safety}, to incoporate libraries and APIs for safety data handling, enabling the filtering of undesirable behaviors and biases prior to training, and \ding{184} \textbf{Ecosystem Safety}, to provides diverse, unbiased datasets and automated safety evaluations, fostering continuous safety improvements and aiding in regulatory compliance. Additionally, the NVIDIA AI Systems Inspection Lab offers a platform for automakers and developers to verify the safe integration of their products, marking a significant step toward standardized AV safety practices. By addressing safety at multiple levels, Halos exemplifies a holistic approach to building trust in autonomous driving systems.}

Finally, building public trust is critical. This ties into transparency: an autonomous vehicle powered by generative AI should ideally be able to explain its decisions in human-understandable terms. As noted earlier, LLM-based explainers are being prototyped (like Nuro and Wayve’s systems that answer questions about actions in real-time \cite{understandingaiTransformerbasedNetworks}). If a car can say “I’m slowing down because I predict the car ahead will cut into my lane to avoid a parked vehicle,” that not only reassures passengers, it also provides a rationale that engineers and auditors can examine. In aviation, black-box AI is generally not allowed for core flight control; automotive safety regulators may similarly mandate a level of traceability, possibly through event logs that record what the AI thought would happen (its generated predictions) vs. what actually happened, to analyze any mishaps. In the near term, companies are already implementing extensive safety test suites for generative components. Cruise, for example, runs its driving AI through millions of simulated encounters nightly to measure disengagement rates, and Waymo has published methodologies for statistically significant safety performance evaluation (comparing miles per incident in sim vs real). We might soon see third-party auditing of AI models – analogous to crash testing, but for AI decision logic – where certified agencies run a suite of standardized scenario tests (some of them generatively created) and ensure the AI’s behavior falls within acceptable risk bounds. In conclusion, the promise of generative AI must be balanced with rigorous safety engineering. Combining empirical testing, theoretical analysis, and new regulations will be necessary. The encouraging news is that the industry is cognizant of this: safety and trustworthiness are front-and-center in autonomous vehicle discussions today, with generative AI developers increasingly publishing not just results, but also failure modes and uncertainty metrics. The coming years will likely bring standardized safety benchmarks for generative models and perhaps even real-world driving trials where an AI’s performance in rare scenarios is evaluated under regulatory supervision (much like crash tests). Achieving public trust will be the ultimate test, and it hinges on demonstrating that generative AI can be as safe and dependable as the components it aims to replace or augment \cite{shu2024generative}.

\begin{tcolorbox}[colback=gray!5!white, colframe=black!50!gray]
    \begin{itemize}[leftmargin=0em]
        \item \fontsize{9}{10}\selectfont
              \textbf{Opportunities:}
              \textit{Achieving trustworthy generative AI in autonomous driving necessitates coordinated progress across several critical dimensions. Key advancements involve developing robust runtime monitoring and safety governance to reliably detect and override outputs violating physical or logical constraints. Integrating formal verification methods, even partially, into generative planners offers provable safety assurances beyond empirical testing alone. Aligning generative AI components with emerging standards (\textit{e.g.}, ISO/PAS 8800) will streamline certification and accelerate public acceptance. Finally, embedding real-time explainable AI mechanisms to articulate model rationales is vital for building trust among users, engineers, and regulators. Collectively, these efforts aim for generative AI in autonomous vehicles that is systematically validated, transparently monitored, reliably audited, and broadly trusted for safe, scalable deployment.}
    \end{itemize}
\end{tcolorbox}

\subsection{Federated Generative AI in Autonomous Driving}
\label{sec:FedGenAI}
The demand for large-scale, diverse, and high-quality data in autonomous driving systems is growing rapidly. Traditional centralized training methods face challenges such as distributed data, privacy concerns, data security, and high costs, making these methods challenging. Federated learning (FL)~\cite{mcmahan2017communication, konevcny2016federated} offers a feasible solution to autonomous driving by enabling distributed training, where nodes collaboratively train a global model without sharing raw data~\cite{fantauzzo2022feddrive, song2023fedbevt, tian2022federated}. Meanwhile, GenAI provides new ways to enhance data acquisition and expand model capabilities. Generative models can synthesize high-quality perception data, traffic scenarios, and complex interactions, significantly alleviating the scarcity of real-world data. Combining FL and GenAI (FedGenAI)~\cite{liu2024integration} has the potential to break through current bottlenecks in autonomous driving AI development, specifically in the following areas:

\paragraph{Data Augmentation with Federated Collaboration}
In real-world applications, the data collected by autonomous vehicles exhibits non-independent and identically distributed (non-IID) characteristics~\cite{nguyen2022deep, song2022federated, wang2022federated}. Local datasets vary significantly due to differences in regions, weather, and traffic conditions. Current FL often struggles with performance degradation in these circumstances. By integrating GenAI, each vehicle can train a local generative model based on its real-world data to create diverse and complementary synthetic samples, expanding its own training set. During the FL process, the nodes can collaboratively train a cross-node generator, enabling privacy-friendly virtual data sharing that mitigates data scarcity and heterogeneity. This significantly enhances the generalization and robustness of the model in various environments. For example, vehicles operating mainly in sunny conditions can generate rainy or night driving scenarios to improve the cross-domain adaptability of the global model.

\paragraph{Personalized Models through Generative Adaptation}
As existing personalized FL works~\cite{pillutla2022federated, zhang2025mllm, song2023fedbevt}, a single global model often cannot accommodate all vehicle-specific variations, especially with significant differences in sensors, driving environments, or task requirements, such as cars versus trucks. With GenAI, vehicles can locally model their unique data distributions and use generated samples for local fine-tuning, achieving federated personalization. This improves the adaptability of the model to specific conditions without compromising local data privacy, allowing vehicles to dynamically optimize their perception and decision-making modules. 

\paragraph{Generative Inference for Communication-efficient V2X}
In cooperative intelligent transport systems (C-ITS), the V2X communication bandwidth is limited. Frequently transmitting deep models for FL can easily lead to communication bottlenecks~\cite{song2023v2x}. With the help of GenAI, servers may be able to extract more synthetic information from models that better describe distributed data, accelerating the convergence of FL.
In addition, using GenAI to generate communication-related information can improve the orchestration of FL. Reconstructing data related to the physical layer of the communication or traffic topology networks through GenAI, it becomes easier to estimate communication parameters such as the signal-to-noise ratio, latency, and throughput. This can help identify stragglers in the communication system and enable a more effective client selection in C-ITS.

\paragraph{Generative Scenarios for Validation and Deployment}
Autonomous systems must make decisions in extremely complex and dynamic environments, but collecting real-world data covering all edge cases is nearly impossible. GenAI can quickly synthesize diverse and physically plausible traffic scenarios to create simulation environments for training and validation. With FL, vehicles or roadside units can collaboratively train generative world models, producing simulated data under various traffic conditions for local training or testing of decision strategies. This not only accelerates strategy validation but also enhances system robustness against rare or dangerous events, such as sudden accidents or severe weather.

\begin{tcolorbox}[colback=gray!5!white, colframe=black!50!gray]
    \begin{itemize}[leftmargin=0em]
        \item \fontsize{9}{10}\selectfont
              \textbf{Opportunities:}
              \textit{Generative AI enhances federated learning by providing data augmentation, privacy-preserving synthetic data, and rare scenario simulation, improving model robustness and generalization. Meanwhile, federated learning empowers Generative AI by enabling distributed training across diverse and private datasets, increasing data diversity, protecting user privacy, and accelerating model personalization. Their integration unlocks new potential for building highly adaptive, scalable, and secure AI systems.}
    \end{itemize}
\end{tcolorbox}

\subsection{Deployment Challenge of Generative AI}

\begin{figure}[h!]
    \centering
    \includegraphics[width=0.8\linewidth]{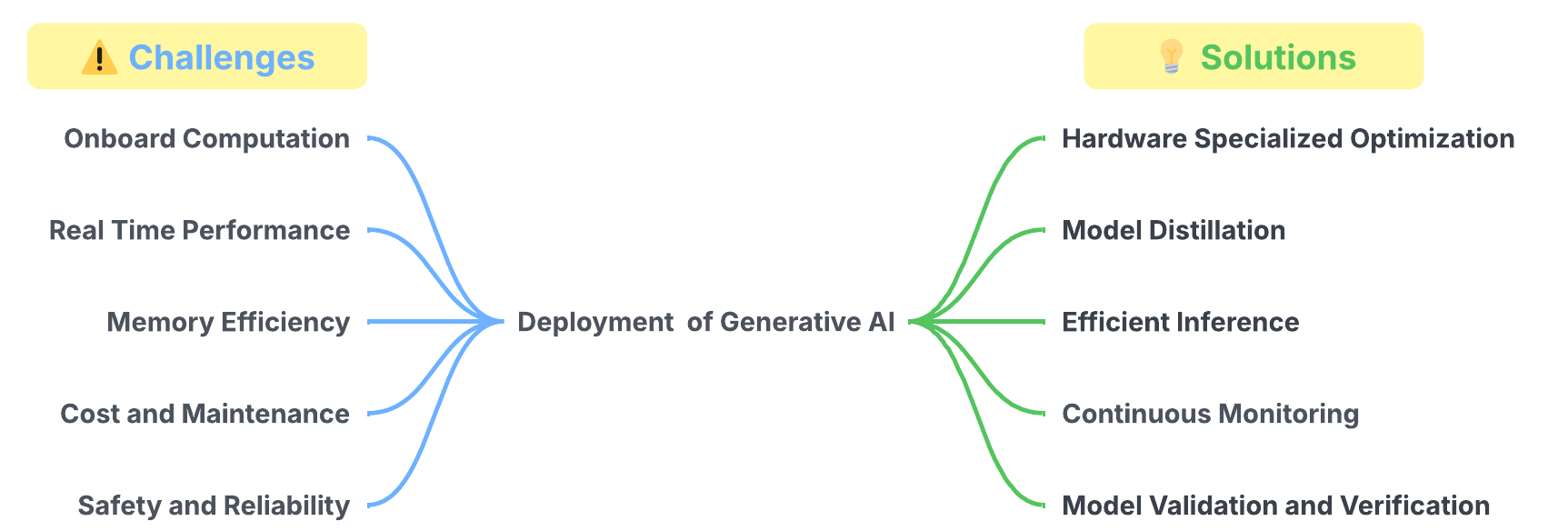}
    \caption{Challenges and potential solutions for Generative AI model deployment.}
\end{figure}

Deploying these new generative technologies on autonomous vehicles has also become an increasingly important research topic. Notable examples include Tesla's deployment of FSD (Full Self-Driving) beta utilizing sophisticated world models and vision-based systems, Waymo's extensive use of multimodal systems~\cite{hwang2024emma} in their autonomous taxi services, and Cruise's implementation of providing end-to-end driving policy from simulation~\cite{bewley2019learning}. Recent research has validated the potential of deploying LLMs~\cite{cui2024largelanguagemodelsautonomous,cui2024personalizedautonomousdrivinglarge} or VLMs~\cite{cui2024onboardvisionlanguagemodelspersonalized} on real vehicles in the testing field. However, the current deployment of generative AI is still facing several critical challenges.

The deployment of advanced AI models in autonomous vehicles faces significant computational resource limitations. Automotive hardware typically has lower computational capabilities compared to development systems, creating a substantial gap between model requirements and available resources~\cite{cui2024onboardvisionlanguagemodelspersonalized}. This constraint affects all aspects of system performance, from basic perception tasks to complex decision-making processes. The challenge is particularly obvious for large models like world models and multimodal LLMs, which require substantial GPU computing power that often exceeds typical automotive hardware capabilities. Moreover, some emerging models must be optimized for specialized automotive hardware accelerators, which often have different architectures from the current development GPUs. 

Model distillation, which transfers knowledge from large teacher networks, such as multimodal LLMs or generative AI models, to smaller student architectures, is a promising yet underexplored direction in the autonomous driving domain for addressing computational constraints~\cite{sreenivas2024llm}. Similarly, plug-and-play approaches offer another compelling avenue for integrating foundation model capabilities into on-device autonomous driving networks without adding additional computational complexity~\cite{ma_lampilot_2024,pan_vlp_2024}.   

Autonomous driving systems must meet extremely strict real-time processing requirements~\cite{cui_receive_2024}. Environmental representations must be updated within milliseconds, typically requiring latencies under 100 ms for full scene understanding and under 50 ms for critical decisions. This becomes particularly challenging when processing high-dimensional sensor data through complex models or when managing multiple concurrent processing streams. Additionally, it might take several seconds to get the reasoning results when the multimodal LLMs are used~\cite{cui2024onboardvisionlanguagemodelspersonalized}. Autonomous driving systems must maintain these time requirements consistently across varying operational conditions and computational loads.

Ensuring system safety and reliability requires extensive validation across a wide range of operational conditions. This includes testing system responses to edge cases, verifying behavior in degraded operating conditions~\cite{he2010single,zhu2024mwformer,10310160,li2024light}, and validating the effectiveness of fallback mechanisms. The challenge extends to maintaining clear audit trails of system decisions and ensuring alignment with different regulatory requirements. This is particularly critical for systems implementing AI alignment and multimodal LLMs, where behavior must consistently align with defined ethical guidelines while maintaining safety guarantees.

Memory efficiency is crucial given the limited RAM available in automotive computers. The challenge extends to managing memory bandwidth, particularly for multimodal systems that require significant data movement between different processing units. 

The deployment of these new models requires investment in hardware for safety and maintenance procedures. This includes the cost of validation and testing before real-world deployment, as well as maintenance to ensure consistent performance. The challenge extends to managing the life cycle of deployed systems, including calibration and potential hardware upgrades. Additionally, software systems will require continuous updates throughout the life cycle of autonomous vehicles. These updates may include model refinements, security patches, performance optimizations, and new feature deployments to ensure system stability and safety after each update.

\begin{tcolorbox}[colback=gray!5!white, colframe=black!50!gray]
    \begin{itemize}[leftmargin=0em]
        \item \fontsize{9}{10}\selectfont
              \textbf{Opportunities:}
              \textit{Model distillation and plug-and-play approaches represent promising yet underexplored directions for integrating advanced foundation model capabilities into on-device autonomous driving systems. These techniques can transfer knowledge from large, computationally intensive models to smaller, more efficient architectures suitable for automotive hardware without sacrificing critical capabilities. This approach addresses the fundamental gap between advanced AI model requirements and the limited computational resources available in vehicles.}
    \end{itemize}
\end{tcolorbox}

\subsection{Ethical Issues of Applying Generative AI to Autonomous Driving}
\label{ssec:ethical-issues}


The deployment of generative AI in autonomous driving raises profound ethical questions, ranging from classical dilemmas (like the trolley problem) to new issues of agency and bias. One major concern is decision-making in scenarios where harm is unavoidable. If an autonomous vehicle must choose between two catastrophic outcomes, how should a generative policy decide? To date, there is no single moral framework that comprehensively guides AI in such complex decisions, and oversimplified solutions can lead to morally questionable outcomes \cite{abcWhatMoral}. For example, a generative model might implicitly learn a bias (say, swerving one way because training data had more examples of cars doing so), which could translate to preferential harm, something society may deem unacceptable. Human drivers in emergencies react instinctively; an autonomous vehicle will react according to its programming or learned policy, which means we (designers and regulators) are effectively encoding a decision, whether we do so explicitly or not. This has led ethicists to argue for transparency in how these models are trained and what principles they follow. Some countries, like Germany, have early guidelines (\textit{e.g.} no decision shall be based on discriminating factors such as age or gender of potential victims), but encoding such rules in a neural model is non-trivial. 

Generative AI complicates this further because it can come up with solutions that a human might not consider. This creativity is usually positive (it might avoid an accident by taking an unconventional action), but it also means we need to anticipate and constrain the AI’s behavior ethically. Value alignment for generative models – ensuring the AI’s “intentions” align with human values \cite{mou2024individual} – is an open research area, heavily overlapping with the broader AI safety community. One approach is to incorporate explicit ethical reward signals or rules into the training process (for instance, penalizing any outcome where a pedestrian is hit, regardless of circumstance). Another approach is to have a secondary "ethics monitor" that evaluates the outcomes of the primary model’s plan and vetoes plans that cross moral red lines (similar to the safety governor idea, but checking ethical criteria). This might handle clear-cut cases (don’t hit people, period), but grey areas will persist. 

There is also the issue of agency and responsibility. If a generative AI makes a decision that leads to harm, who is accountable? This is already a legal and economic quandary for autonomous vehicles, but with AI that learns from data, it becomes even murkier: the manufacturer, the software developer, the data used to train, or the AI itself (in a metaphorical sense)? Society may be more willing to accept an accident caused by a human driver’s split-second error than one caused by an algorithm, even if statistically the algorithm causes fewer accidents. This is a known paradox in autonomous vehicle ethics: we expect autonomous vehicles to be much safer than humans to accept them, but any single failure gets magnified in public perception. Generative AI must therefore achieve a high bar. Answerability is key for accountability \cite{dubber2020oxford}. In other words, if a collision occurs, investigators will want to know who should be liable and responsible for this collision. 
Explainability and traceability of decisions is also important as to why the AI chose a certain action. As mentioned, recording the internal reasoning (even as high-level descriptions) can help assign responsibility or improve the system post-incident. 

Another ethical aspect is bias and fairness. Training data for generative models might underrepresent certain environments or behaviors – for example, if most driving data comes from urban areas, the model might perform poorly (or strangely) in rural or developing areas. This raises concerns of equitable deployment: will autonomous vehicles be safe for all communities or only those that match the training distribution? If an AI is more likely to “sacrifice” one type of road user over another because of subtle biases in data (imagine it learned to prioritize avoiding cars over bicycles, simply due to frequency of scenarios seen), that’s an unfair outcome. Ongoing research is looking at dataset balancing and using synthetic data generation as a remedy, intentionally generating more scenarios involving vulnerable road users, for instance, to teach the AI to protect them. Policymakers might even mandate that certain ethically significant scenarios (like interactions with pedestrians in crosswalks, or decisions in loss-of-control situations) are given extra weight in testing and training \cite{yang2022cooperative}. 

A novel proposition in the ethics literature is cooperative or collective ethics: EthicalV2X \cite{sidorenko2023ethical} argues that if vehicles communicate and cooperate, many ethical dilemmas can be mitigated or avoided. For example, instead of two cars independently making decisions (potentially to each sacrifice the other), they can cooperate via V2V and find a solution that reduces harm to both, effectively escaping the classical trolley setup through coordination. Generative AI can facilitate this by jointly simulating outcomes for all parties and finding Pareto-optimal solutions. Of course, this requires all actors (vehicles) to be autonomous and communicative, which is a future scenario not applicable to mixed traffic today. 

We must also consider privacy and data ethics as part of the ethical challenges. Generative models often require huge amounts of data (videos of city streets, driver behavior logs, etc.). Using these data raises questions: Are we properly anonymizing pedestrians and other drivers? Could a generative model inadvertently “re-identify” someone or expose private information? For instance, if a model is trained on video that includes someone’s license plate, and then it generates a remarkably similar scene, is that a privacy breach? Techniques like differential privacy could be employed during training to minimize memorization of specific details. There’s also an emerging concept of data governance for autonomous vehicles: ensuring that the way data is collected (often by vehicles driving through communities) and then used for model training adheres to consent and privacy laws. Europe’s GDPR and the forthcoming AI Act are touching on this, classifying AI in mobility as high-risk and requiring documentation of training data provenance and bias assessments. 

\begin{tcolorbox}[colback=gray!5!white, colframe=black!50!gray]
    \begin{itemize}[leftmargin=0em]
        \item \fontsize{9}{10}\selectfont
              \textbf{Opportunities:}
              \textit{The application of generative AI to autonomous driving opens new ethical opportunities in areas such as value alignment, bias mitigation, and accountability. Research is needed to design transparent ethical frameworks, integrate real-time "ethics monitors" into decision loops, and create explainable systems that can trace AI reasoning. Generative models can also be leveraged to synthesize fair, privacy-preserving datasets and simulate rare or cooperative scenarios (e.g., Ethical V2X communication) to minimize harm. Collaboration with regulators will be crucial to set standards for safety, fairness, and data governance, ensuring that generative AI supports inclusive, responsible, and community-centered mobility.}
    \end{itemize}
\end{tcolorbox}

\subsection{On the Human-AI Collaborations}

Rather than replacing humans, generative AI in autonomous driving opens possibilities for collaboration between humans and AI, both in the design/testing phase and in real-time operation. One evident avenue is through human-in-the-loop simulation and design. Engineers can work hand-in-hand with generative models to craft scenarios or improve the AI’s behavior. For example, an autonomous vehicle developer might notice the model performs poorly in a certain scenario; using a human-in-the-loop approach, they could tweak parameters or provide a few example demonstrations, and have a generative scenario engine produce dozens of variations of that scenario for retraining. This kind of interactive training (similar to reinforcement learning from human feedback, RLHF) could significantly improve model performance on edge cases. We are essentially steering the generative model using human insight. Early research has shown that combining human feedback with uncertainty estimation can be powerful. One study reported that integrating human interventions based on an AI’s uncertainty led to a 16× reduction in collision rate in simulation \cite{themisaiUncertaintyAwareHuman}. That suggests a future where autonomous vehicles might ask for help when uncertain: the car could alert a remote human supervisor or even a passenger in complex situations (at least at lower automation levels) and query guidance. 

In the development cycle, AI-assisted tooling is burgeoning. Natural-language-based query systems (as mentioned with the dSPACE example) enable engineers to sift through petabytes of driving data quickly. This accelerates debugging and scenario discovery. Moreover, generative AI (like code generation models) can assist in writing simulation scripts, configuring experiments, or even generating test cases automatically. We can think of this as the autonomous vehicle equivalent of pair-programming: the engineer specifies high-level goals (“create a nighttime scenario with heavy rain and a jaywalking pedestrian in front of a left-turning car”) and the generative system produces the simulation setup. This lowers the barrier to testing creative scenarios. 

During real-world operation, human-AI collaboration manifests in new ways inside and around the vehicle. A striking example is the conversational capability demonstrated in Wayve’s LINGO \cite{marcu2023lingoqa} – where a passenger (or pedestrian) could literally talk to the car and get meaningful responses. This turns the vehicle into an interactive agent rather than a mute machine. Generative AI, especially language and image generation, can facilitate these new interaction modalities. The car might flash a synthesized human-like eye on a screen or a text banner to signal intent. These are areas being explored in HMI (Human-Machine Interface) research for autonomous vehicles \cite{wang2026drive}. 

Another form of human-AI partnership is shared control. For partial automation (Level 2/3 systems), generative models could work with human drivers to enhance safety. For instance, an AI co-pilot might monitor the environment and generate gentle corrective inputs or warnings if it foresees a hazard that the human hasn’t reacted to. This goes beyond current ADAS, because a generative co-pilot could be more proactive, almost like an instructor with a brake pedal on the passenger side. If the human is driving erratically (drowsy or distracted), the AI could even take over briefly or suggest a break, having “imagined” the likely outcome of continued inattention (this touches on driver monitoring systems and mental state detection too). Conversely, a human driver could overrule or guide an AI suggestion: maybe the car planned an efficient but aggressive maneuver the user is uncomfortable with; a quick voice command “take it easy” might prompt the generative planner to bias toward a more conservative driving style. Such real-time personalization is an exciting opportunity where the autonomous vehicle adapts to individual passenger preferences or stress levels. In the long run, your autonomous car might know that you’re a nervous rider and thus it generates smoother, slower-driving scenarios when you’re on board, versus when it is doing a logistics run by itself, where it might drive more boldly within safe limits. 

\begin{tcolorbox}[colback=gray!5!white, colframe=black!50!gray]
    \begin{itemize}[leftmargin=0em]
        \item \fontsize{9}{10}\selectfont
              \textbf{Opportunities:}
              \textit{Generative AI in autonomous driving fosters powerful human-AI collaboration across development and real-world operation. In development, humans can guide generative models through interactive simulation and training, improving performance on rare or challenging scenarios. AI-assisted tools accelerate debugging, scenario creation, and test case generation. In real-time driving, generative AI enables new human-machine interaction modes, from conversational interfaces to shared control systems where AI acts as a proactive co-pilot, personalizing behavior based on human input and preferences. Ultimately, human-AI partnerships promise safer, more adaptable, and more user-centered autonomous vehicles.}
    \end{itemize}
\end{tcolorbox}

\subsection{Broader Implications for Urban Studies and Geography}

Advances in autonomous driving, coupled with powerful generative AI techniques, are opening new frontiers in urban studies and geography, beyond merely augmenting transportation research. Generative AI enables the possibility of generating diverse environmental settings and places across multiple scales and modalities. 
For instance, leveraging GAN- and diffusion-based architectures can generate high-quality 2D street view images \cite{zhuang2024hearing}, 3D point clouds \cite{luo2021diffusion}, and even 3D buildings \cite{wei2023buildiff}. By parameterizing specific spatial attributes such as land use objects and building typology, researchers can deploy these generative models to simulate a variety of urban morphologies \cite{zhou2024controlcity}. 
With the ability to generate and synthesize different environmental settings using GenAI, the real world might be transformed as a virtual “laboratory” or a “world simulator” \cite{zhu2024sora}, wherein social behaviors and physical environmental processes might be rigorously tested, and directly benefit autonomous vehicle research itself. Synthetic driving environments populated with realistic variations in topography, spatial relationships, and environmental settings allow an autonomous system to be tested in rare or extreme conditions. 

The large-scale deployment of autonomous vehicles may fundamentally redefine urban data acquisition. Outfitted with high-resolution cameras, multi-beam LiDAR scanners, and environmental sensors (e.g., air-quality sensors, acoustic arrays), each vehicle continuously samples its surroundings, generating rich, geotagged streams of multimodal observations \cite{zhang2023autonomous}. Unlike stationary monitoring stations with sparse spatial coverage, these mobile fleets could capture fine-grained spatio-temporal variations in microclimate, noise levels, traffic flows \cite{anjomshoaa2018city}, and infrastructure conditions at an unprecedented resolution, a new form of “Volunteered Geographic Information” in the era of GenAI \cite{goodchild2007citizens}. Such high‐frequency, longitudinal observations could not only capture dynamic patterns of geographic phenomena and human activities, but also empower the next generation GIS platform \cite{li2025giscience}. 

While the integration of GenAI and autonomous vehicles unlocks significant opportunities for urban and geographic areas, it simultaneously necessitates the development of urban governance and policy frameworks to address potential risks \cite{ye2025artificial}. In particular, two critical concerns require immediate attention, including the management of hallucinations in generative outputs and the protection of (geo)privacy \cite{kang2024artificial}. These are important aspects to enhancing human trust and ensuring the responsible deployment of GenAI in autonomous vehicles and more broadly.

A foremost challenge is the phenomenon of hallucination inherent in generative models. Beyond technical solutions for addressing hallucination, more attention should be paid to developing governance frameworks and ethical guidelines. The responsibility and accountability for hallucination-related harms should be discussed \cite{hacker2023regulating, kang2025human}. Key open questions include: If a hallucinated dataset or simulation influences public policy with adverse effects, who should take responsibility? Should there be standards for traceability, enabling the provenance of AI-generated content to be audited and linked to specific decision processes? How can policy workflows be designed to track, document, and mitigate hallucination risks that may occur in model training and their real-world applications across various environmental settings?

As GenAI and autonomous vehicles reshape geospatial data collection methods and urban environmental monitoring, safeguarding (geo)privacy will be another important issue. Several critical questions may arise about protecting individual rights of geographic locations, data ownership, and democratic oversight \cite{kim2021people}. To address geoprivacy concerns, both technical innovation and regulatory governance are necessary. On the technical side, privacy-preserving methods offer promising solutions. For instance, synthetic data generation methods can generate realistic but non-identifiable information that maintains key statistical properties of the original data while minimizing the risk of re-identification \cite{rao2020lstm}. Federated learning methods allow decentralized model training across distributed data sources without transmitting raw data \cite{rao2021privacy}. However, technical methods alone are insufficient to ensure trust or legitimacy, and it is important to establish governance frameworks for the ethical collection, use, and dissemination of geographic data. 
Furthermore, it is crucial to actively involve stakeholders, including local communities, civil society organizations, and policymakers \cite{du2024artificial}. Such participatory processes ensure that AI-driven urban technologies serve the broader interests and prioritize human values. 

In sum, the emergence of GenAI and autonomous vehicles represents a potential paradigm shift in simulating and sensor urban environments, and brings new opportunities as challenges for urban governance. It is necessary to embrace new technological advancements and to develop cooperative, and participatory policy frameworks. Through the integration of interdisciplinary research, the fields of urban studies and geography could help shape a more ethical and resilient future for AI-integrated cities.

\begin{tcolorbox}[colback=gray!5!white, colframe=black!50!gray]
    \begin{itemize}[leftmargin=0em]
        \item \fontsize{9}{10}\selectfont
              \textbf{Opportunities:}
              Future research should focus on using generative AI to create realistic environments across diverse place settings for urban and geographic studies, advancing methods for high-resolution, autonomous vehicle-based environmental data collection, and developing governance frameworks to manage hallucinations and protect geoprivacy when developing and using GenAI. 
              Key actions include establishing standards for traceability and accountability in AI-generated content, designing guidelines for synthetic data collection and use, and actively engaging stakeholders to prioritize human values to enhance trust, transparency, and resilience to integrate GenAI into urban systems.
    \end{itemize}
\end{tcolorbox}

\subsection{Drones, UAVs, and the Low-Altitude Economy}

Many concepts in generative AI for driving extend naturally to autonomous drones and low-altitude air mobility. Drones face analogous challenges: they must navigate dynamic environments, avoid collisions, and coordinate with other agents (including ground vehicles and aerial vehicles). 
Generative models can assist by predicting complex 3D trajectories and environmental factors for drones. For instance, recent research has applied generative modeling to micro-weather patterns (like wind gusts in urban canyons) to improve UAV flight safety \cite{shah2025generative}. By learning to generate realistic wind fields and turbulence scenarios, drones can be trained to handle conditions that are too dangerous to test in real life, enhancing reliability in the face of weather uncertainties. Similarly, the concept of real2sim2real is being explored for urban air mobility: \textit{e.g.} using GANs to expand limited datasets of drone flight logs with synthetic data for rare events like GPS outages or emergency landings. This mirrors what’s done for cars (augmenting long-tail events) but in three dimensions. 

In the emerging low-altitude economy, which includes delivery drones, autonomous air taxis, and surveillance UAVs, generative AI could play a role in traffic management and collision avoidance. We can imagine a future “sky traffic control” AI that uses generative models to simulate thousands of drone flight paths over a city, identifying conflict points and dynamically routing drones in real-time to prevent congestion in popular altitudes or locations. 
Drones may also communicate with ground vehicles (a delivery drone might signal an autonomous car to clear a driveway for a landing). Generative models that span both aerial and ground domains could coordinate such interactions, essentially treating the entire urban environment as one integrated system. 
This may lead to air-ground cooperative generative models, where, for example, a drone delivering medical supplies during heavy traffic could generate an optimal meetup spot with an autonomous ground courier, balancing air and road travel time. 

\begin{wrapfigure}{L}{0.50\textwidth}
\includegraphics[width=\linewidth]{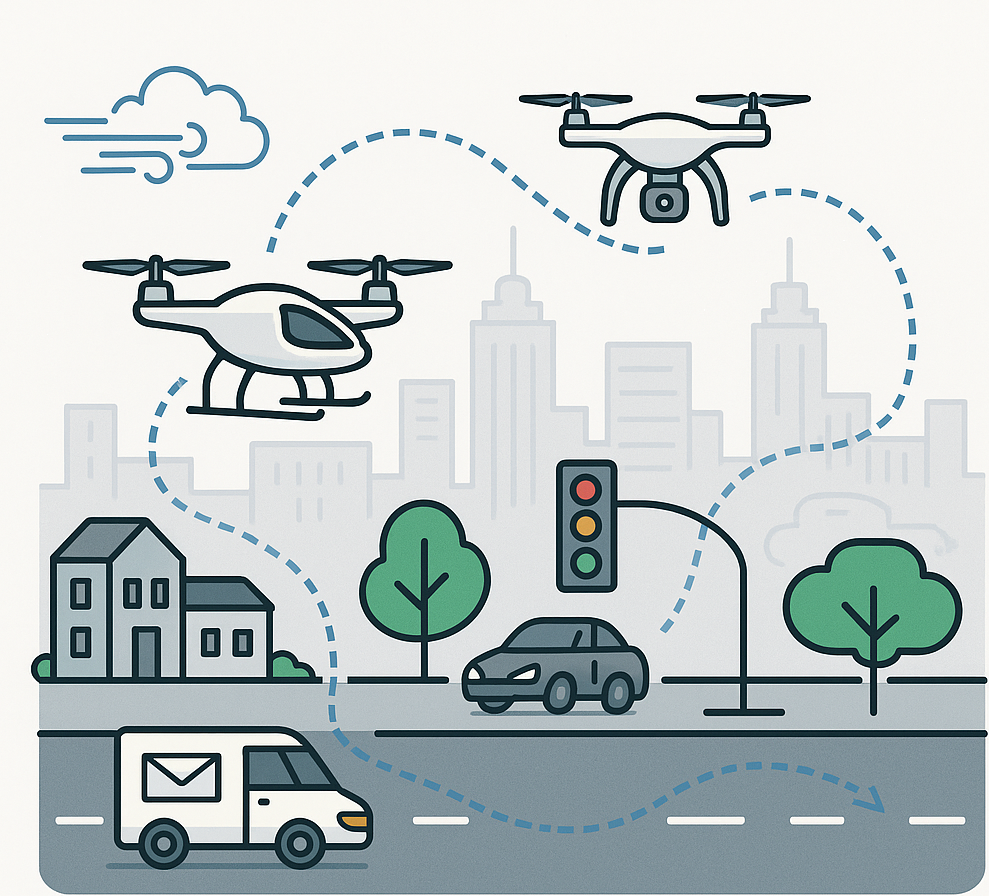} 
\caption{In the future, autonomous vehicles will extend to drones.}
\end{wrapfigure}

From a safety standpoint, regulators (like the FAA and NASA in the U.S.) are already testing unmanned traffic management (UTM) systems \cite{kopardekar2016unmanned}. We anticipate that simulation-driven certification will be important for drones, too. Just as self-driving cars undergo millions of simulated miles, drone developers will use generative world models to simulate bird encounters, powerline interference, payload failures, etc. One interesting research direction is using generative AI to simulate rare emergency scenarios for drones, for instance, how to safely glide and land a delivery drone after a motor failure in various urban landscapes. Another is 3D environment generation: creating rich 3D city models (buildings, trees, electromagnetic interference maps) where drones can be trained. Companies might leverage existing map data and use generative algorithms to add plausible details (like spontaneously generating new construction sites or crane operations in the virtual city) to test drone responses. 

The low-altitude economy also raises unique public acceptance issues, akin to what autonomous vehicles face. Noise, privacy, and airspace crowding are concerns. Generative AI can help address some of these by optimizing drone routes for minimal noise (\textit{e.g.} generating flight paths that avoid hovering over sensitive areas like schools or hospitals whenever possible) and by simulating the impact of large-scale drone deployment on communities (for instance, generating the soundscape of a neighborhood with 50 drone overflights per hour to gauge acceptability). In essence, before hundreds of drones fill the skies, we can use generative simulation to explore social and environmental impacts and inform policymaking (much like city traffic simulations do for ground vehicles). 

In the near term, we will likely see drone-specific generative models for perception and planning, similar to what we have in cars, such as vision transformers that generate probable trajectories for other aircraft or birds, giving drones an early warning system \cite{gao2025airv2x}. Long-term, the boundary between ground and air transportation AI might blur: a unified logistics AI could decide how to split deliveries between trucks and drones by simulating the entire multimodal delivery process generatively, optimizing speed, cost, and carbon footprint. The convergence of autonomous cars and drones under the umbrella of generative AI will push us toward a truly heterogeneous autonomous transportation network.

\begin{tcolorbox}[colback=gray!5!white, colframe=black!50!gray]
    \begin{itemize}[leftmargin=0em]
        \item \fontsize{9}{10}\selectfont
              \textbf{Opportunities:}
              \textit{Researchers should focus on developing generative models specifically tuned for predicting complex 3D trajectories in various environments, enhancing drone reliability in unpredictable conditions. Further work is needed on integrated air-ground cooperative models that can optimize interactions between aerial and ground vehicles in shared spaces.}
    \end{itemize}
\end{tcolorbox}

\subsection{Toward Health and Well-being-Aware Autonomous Mobility}


Autonomous driving is increasingly viewed not only a safety innovation but also as a technology to facilitate health and well-being. One important health outcome is the reduction in traffic fatalities by eliminating accidents caused by human error. According to a survey by the U.S. Department of Transportation, the reduction could be as high as 94\% in the United States \cite{singh2015critical}. Globally, this could result in saving up to 10 million lives per decade \cite{fleetwood2017public}. With the integration of generative artificial intelligence (GenAI), the capabilities and benefits of autonomous vehicles can extend to many dimensions beyond navigation and safety, such as personalized health support, emotional engagement, and healthcare accessibility. As mobility becomes increasingly entwined with health equity and digital inclusion, GenAI offers the possibility of transforming autonomous vehicles into active agents of healthcare that support individuals and communities. 


GenAI can support individual health in and out of the autonomous vehicles by improving driving comfort and behavior. Inside autonomous vehicles, advanced monitoring systems can track physiological symptoms, such as fatigue and illness, through cameras and biosensors \cite{visconti2025innovative}. GenAI assistant can then interpret these signals and respond appropriately in real time, for example, by initiating a health check dialogue, adjusting the speed of an autonomous vehicle safely, or even contacting emergency services when medical risks are detected. Emotionally adaptive in-car interfaces are also being empowered by generative AI. Experimental “empathic vehicles” leverage generative models to sense a passenger’s mood, via facial expressions, vocal tone, or wearable data, and adjust the interior environment, such as lighting or music, to enhance comfort \cite{khattak2023quantifying}. Beyond moment-to-moment emotion recognition, GenAI can personalize interaction based on longer-term psychological profiles. For instance, personality-informed models may adapt tone, pacing, or content in ways that reduce anxiety or earn trust, particularly for riders who score high on neuroticism or exhibit low confidence in driving \cite{bai2025study}. These psychologically attuned interventions are especially valuable in high-stress contexts such as nighttime driving or traffic congestion. Meanwhile, outside autonomous vehicles,  comfort is shaped by macroscopic factors such as traffic flow, road curvature, stop frequency, and jerk dynamics during turns or braking. Recent studies have demonstrated how road-level information can be integrated into a multi-head attention model to predict discomfort and inform global path planning \cite{chen2025predicting}. By synthesizing these external signals with passenger health status captured internally, generative AI can enable truly health-aware trajectory planning, which can personalize routes not only for speed or safety, but also for minimizing physiological and psychological discomfort, particularly for individuals experiencing “automatophobia”, a condition marked by phobic symptoms when using autonomous cars \cite{meinlschmidt2023anticipated}. This convergence of inside and outside sensing, mediated through GenAI, marks a shift toward proactive, health-centered mobility.


GenAI autonomous driving creates new opportunities to serve vulnerable groups and advance public health goals. Mobility is a well-recognized determinant of health: when older adults stop driving, their reduced mobility often leads to social isolation and worse health outcomes \cite{siegfried2021older}. By embedding GenAI into assistant systems, self-driving cars could provide older adults and people with disabilities new mobility options, helping them maintain independence and access medical care and social activities \cite{othman2022exploring}. These AI-driven assistants can accommodate a wide range of special needs by delivering health-related reminders, guiding users to medical facilities, or providing simplified, multimodal interfaces designed for specific populations with cognitive or visual impairments. Such functionality not only improves accessibility but also redefines the vehicle as a supportive healthcare partner. Indeed, researchers have emphasized that well-designed autonomous services can actively “promote wellness” by reconnecting users with essential services. Scaling such GenAI-enabled services equitably across communities will be crucial to maximizing their long-term public health impact.


The integration of generative AI into autonomous vehicles is especially transformative in emergency and high-stakes healthcare scenarios. One critical application prototype is the autonomous mobile clinic (AMC), defined as a self-navigating, AI-enabled vehicle equipped to deliver on-site medical services \cite{liu2022autonomous}. These systems represent a convergence of AI, telemedicine, and mobile diagnostics, offering not just rapid transportation but also real-time care in underserved or high-risk environments \cite{huang2024health}. Unlike traditional ambulances focused solely on speed and transport, AMCs can serve as integrated care touchpoints: performing diagnostics, consulting with remote doctors, and generating patient-specific care plans powered by GenAI agents. Studies have shown that such mobile systems significantly improve patient satisfaction, reduce delays, and provide a scalable model for equitable access, especially in low-resource or post-disaster areas. Critically,  when paired with Internet of Things (IoT)-connected equipment (\textit{e.g.} smart wheelchairs) \cite{hou2024autonomous}, GenAI can enable these mobile clinics to deliver dynamic, patient-centered interactions, including adaptive explanations of procedures, multilingual health counseling, and context-aware routing based on patient conditions. This positions AMCs not only as a tool for acute care, but as a lifeline for longitudinal, community-embedded healthcare. As healthcare systems globally seek to reduce fragmentation and expand integrated care, GenAI AMCs could help overcome logistical barriers while offering high-satisfaction, cost-effective health deliveries.

Looking ahead, safety, trust, and fairness will be critical principles in the design of generative AI systems embedded in autonomous vehicles \cite{ehsani2024advancing}. Models used for in-vehicle medical dialogue, triage support, or behavior-sensitive interaction must be rigorously validated to ensure reliability and user well-being \cite{lawrence2024opportunities}. At the same time, institutional and regulatory frameworks must evolve to support the application of GenAI in mobile health beyond traditional clinical settings. With careful governance and human oversight, health-aware autonomous vehicles, whether used in moments of crisis or day-to-day health support, will become an integral component of future equitable and responsive healthcare ecosystems.

\begin{tcolorbox}[colback=gray!5!white, colframe=black!50!gray]
    \begin{itemize}[leftmargin=0em]
        \item \fontsize{9}{10}\selectfont
              \textbf{Opportunities:}
              \textit{Generative AI advances the health-aware capabilities of autonomous vehicles by improving physiological and emotional monitoring, personalizing ride experiences, and facilitating timely interventions at the point of care. At scale, these systems offer new opportunities to expand healthcare access, strengthen public health resilience, and promote equitable, patient-centered mobility solutions.}
    \end{itemize}
\end{tcolorbox}

\subsection{Generative Autonomous Systems for Disaster Management}

Disaster scenarios pose extreme challenges due to their unpredictability, dynamism, and scale, causing widespread damage to critical infrastructure and vital communication networks~\cite{Comfort2005}.
Autonomous systems (\textit{e.g.}, UAVs, ground robots, autonomous vehicles) are invaluable for tasks like aerial surveillance, search and rescue (SAR), and infrastructure monitoring, as they can rapidly survey large, dangerous areas.
However, current autonomous systems face limitations in disaster zones.
Events such as the recent 2025 California wildfires~\cite{ca-wildfire} and the earlier Texas power crisis due to the 2021 Winter Storm Uri~\cite{texas-winter-storm-wiki} clearly demonstrated existing technological gaps. Thick smoke during the wildfires severely disrupted UAVs' vision and LiDAR capabilities, while extensive ice and snow during Winter Storm critically impaired sensor performance in ground-based autonomous vehicles. These real-world challenges underscore the urgency of enhancing autonomous perception systems for these extreme, 'black swan' scenarios that occur beyond normal weather conditions~\cite{Gazzea2023,Lyu2023}.

On the one hand, GenAI can potentially bridge critical gaps by synthesizing realistic, diverse datasets of rare or hazardous conditions that are often underrepresented in traditional data collection. Synthetic scenarios that replicate dense smoke, severe flooding, structural debris, or icy road conditions allow autonomous systems to train and refine their perception modules more comprehensively~\cite{Abbaspour2024}. Recent studies indicate that perception systems trained with such synthetic data achieve notably improved detection reliability and accuracy under adverse conditions, directly translating to improved operational safety in unpredictable disaster environments.
On the other hand, generative AI also enhances the decision-making capabilities of autonomous systems through large language models~\cite{Lei2025}. These models interpret high-level, human-defined missions, translate complex disaster response objectives into actionable tasks, and coordinate multiple robotic units dynamically. Multimodal extensions of LLMs further integrate and synthesize diverse data sources—including sensor streams, satellite imagery, and textual updates—into coherent situational reports, providing critical intelligence to emergency management teams during rapidly changing disaster scenarios.

A particularly promising application of GenAI is its support of evacuation planning~\cite{moreno2024generative}, a crucial component of disaster management closely related to autonomous driving contexts. Generative models can simulate extensive, realistic evacuation scenarios, enabling planners and autonomous vehicle operators to proactively analyze various contingencies, identify optimal evacuation routes, anticipate congestion points, and strategically position emergency resources. For instance, during wildfire~\cite{boroujeni2024comprehensive} or flood events~\cite{do2023generalizing}, generative scenario simulations can test numerous evacuation strategies under evolving conditions, thereby significantly enhancing the effectiveness and safety of evacuation operations in real crises. Such proactive evacuation modeling and analysis represent a direct, impactful benefit derived from GenAI technologies, bridging disaster management with intelligent transportation system advancements.

However, critical challenges persist. Bridging the simulation-to-reality gap remains complex, especially for these rare scenarios~\cite{li2023climatenerf}; simulated environments must sufficiently mirror real-world unpredictability and variability. Furthermore, existing evaluation metrics for general  (such as FID or LPIPS) often fall short of accurately assessing the relevance and practical utility of synthetic disaster data due to their unique characteristics. Moreover, effective deployment also demands computationally efficient models compatible with the limited resources of edge devices such as drones and robotic platforms. Lastly, similar to issues we mentioned in~\cref{ssec:ethical-issues}, ethical guidelines, privacy safeguards, and robust governance frameworks are essential to ensure transparency, accountability, and fairness in AI-supported decision-making processes.
These aspects require researchers and practitioners to responsibly taken care of in future research.

\begin{tcolorbox}[colback=gray!5!white, colframe=black!50!gray]
    \begin{itemize}[leftmargin=0em]
        \item \fontsize{9}{10}\selectfont
              \textbf{Opportunities:}
              \textit{Looking forward, integrating generative simulation, advanced perception modeling, and intelligent coordination tools could significantly shift disaster management from reactive response toward anticipatory preparedness and enhanced societal resilience. By drawing insights from fields such as autonomous driving and intelligent transportation systems, GenAI-driven solutions can substantially improve the capacity of communities to respond to and recover from complex, large-scale disaster events.}
    \end{itemize}
\end{tcolorbox}

\subsection{Potential Negative Societal Impacts}
\label{ssec:potential-negative-impact}

GenAI-powered autonomous systems promise profound benefits across many domains, as aforementioned in previous sections.
For example, it can dramatically expand access: self-driving vehicles can serve passengers who cannot drive (such as the elderly or disabled), effectively providing ``door-to-door'' mobility that was previous impossible~\cite{litman2017autonomous}
By coordinating vehicles in real time, GenAI systems can optimize traffic efficiency. Connected autonomous vehicles can form platoons or adopt optimal spacing, reducing bottlenecks and increasing roadway throughput. In safety, removing human error could greatly cut crashes: U.S. data show that ~94\% of traffic fatalities involve human factors, and one analysis found autonomous vehicles could dramatically cut deaths from drunk driving (30\% of fatalities), speeding (22\%), and fixed-object collisions (17.5\%)~\cite{maccarthy2024evolving}.
Environmentally, GenAI can reduce emissions by favoring electric vehicles and more efficient driving. Autonomous cars avoid inefficient behaviors (like needless idling or circling for parking) by taking optimal routes, smoothing speed profiles, and quickly decelerating for red lights~\cite{onat2023rebound}.
In practice, autonomous vehicles can lessen traffic congestion and thus lower fuel burn per trip, while also enabling a faster shift to low-carbon powertrains.
Early deployments already hint at public interest: companies like Waymo now offer hundreds of thousands of paid robotaxi rides per week in major cities~\cite{genai-help-vehicle},  and autonomous delivery drones and ground robots (also aided by GenAI perception) are being piloted for last-mile logistics.
These innovations hold promise for smarter urban planning, as rich autonomous vehicle data can inform adaptive traffic controls and city design. In disaster management, AI-powered drones extend response capabilities: recent reports note that autonomous drones provide vital reconnaissance and early hazard detection in emergencies~\cite{zhao2019uav}.
Overall, GenAI could transform automated vehicles and transportation into a safer, more sustainable, and more inclusive system.
Nevertheless, the potential deployment of these technologies also necessitates acknowledging significant societal risks, including hurdles related to public perception, regulatory uncertainty, the potential for increased economic inequality, and complex human factors.

\paragraph{Public Perception and Trust Issues}

Yet these benefits are contingent on broad societal acceptance and trust, which are far from guaranteed. Surveys show drivers remain wary: only ~13\% of U.S. motorists say they would trust riding in a self-driving car (only slightly up from 9\% a year earlier), while a majority (~60\%) report fear of autonomous vehicles~\cite{driving-trust}.
A 2023 J.D. Power study similarly found U.S. confidence in autonomous vehicles scored just 37 out of 100~\cite{driving-trust-2}.
Such skepticism is fueled by very public incidents of hostility. 
For example, in January 2025, a crowd in Los Angeles violently attacked an empty Waymo robotaxi-smashing windows, tearing off a door, and kicking the vehicle, in an apparent “street takeover” scenario~\cite{group-vandalize}.
These and other reported attacks (including attempts to burn test autonomous vehicles in Texas) illustrate that segments of the public may actively resist shared self-driving cars and trucks~\cite{crowd-shatters}.

Such incidents inevitably undermine public trust, complicating the deployment timeline and procedures for autonomous vehicles. While industry emphasizes the rarity of hostile events and reaffirms commitments to safety and investigation, these occurrences highlight the critical need for proactive community engagement and effective trust-building strategies. Achieving widespread acceptance necessitates a multi-pronged approach centered on transparency regarding capabilities, limitations, and incident reporting; robust public education initiatives using clear, accessible language to demystify the technology; and demonstrable safety through rigorous testing protocols and potentially shared safety standards. As analysts note, regulatory environments emphasizing safety and open communication are crucial for fostering public confidence. Furthermore, facilitating human-robot coexistence in shared spaces may benefit from clear design cues—such as external displays or intuitive lighting signals indicating autonomous operation and intent—and ensuring autonomous vehicles operate with predictable and considerate behavior, such as reliably yielding to pedestrians and avoiding abrupt maneuvers, which aligns with expectations of social agents. Ultimately, securing a broad social license to operate, built on proven reliability and public trust, is paramount; without it, public apprehension and resistance could significantly impede the adoption of GenAI-powered autonomous driving, jeopardizing the realization of its profound potential for enhanced safety and efficiency.

\paragraph{Legal Responsibility and Regulatory Challenges}

When generative AI causes harm, explainability and treaceability come first, but who is legally liable is crucial to apportion loss and risk among the involved parties. 
Current traffic laws and insurance models are based on human drivers and do not neatly accommodate a driverless vehicle.
Policymakers are grappling with this gap: as one analysis observes, regulators ``have not put in place an effective regulatory system that assures the public that safety concerns have been adequately addressed,'' and fundamental questions like the assignment of liability remain unsettled~\cite{medsker2024acm}.
In most jurisdictions, if an autonomous car causes a crash, it is unclear whether fault lies with the manufacturer’s AI, the owner, or another party. Some propose shifting liability onto manufacturers (except in cases of gross owner negligence)~\cite{gherardini2024impact}, but no uniform standard exists yet. Recent years have seen a growing literature in understanding how to design efficient economic liability framework when autonomous vehicles are involved in crashes, especially when they interact with human drivers in mixed autonomy \cite{chatterjee13,smith17,shavell20,geistfeld17,di20,chen2023legal,dawid24}. However, there are many open questions in tort law and criminal law. Legal reform is urgently needed, not only for autonomous vehicles, but also for broader digital products that rely on algorithmic decision making resulting from AI and generative AI.
Insurance frameworks must similarly adapt, potentially moving from driver-centric policies to product-liability models or new usage-based premiums.

Beyond high-level liability, integrating autonomous agents into human societal systems reveals fundamental points of friction. Critical interactions, such as how law enforcement or first responders should engage with a driverless vehicle during traffic stops or post-accident scenarios, currently lack established protocols designed for non-human actors, creating operational uncertainty and novel risks. This specific gap mirrors a broader regulatory fragmentation: a complex patchwork of disparate, often lagging, state and national laws governs autonomous vehicle testing, deployment, and safety requirements. Such regulatory disarray not only impedes innovation and complicates cross-jurisdictional operation but can also undermine public confidence through perceived inconsistencies in safety oversight. Establishing robust, harmonized legal and regulatory frameworks is, therefore, more than a bureaucratic step; it is a societal imperative for defining responsible automation. These frameworks must encompass clear definitions of automation levels, rigorous safety validation procedures, updated traffic laws for mixed autonomy, and equitable liability principles. As experts caution, the persistent lag in achieving global policy alignment casts a shadow of uncertainty, potentially stifling the progress of GenAI-driven autonomous systems and raising profound questions about our collective readiness to govern these transformative technologies responsibly.

\paragraph{Economic Inequality due to Technopoly}
A third concern is that GenAI-enabled autonomous driving could deepen economic divides both within societies and between them. The technology is being driven largely by a handful of wealthy nations and tech giants–Google’s Waymo, Apple, Tesla, and major automotive consortia~\cite{market-share}–raising fears of “technopoly” concentration. Critics caution that, as with smartphones, dominant firms may lock consumers into proprietary autonomous vehicle platforms and extract most of the value. Moreover, developing regions with less capital for new infrastructure (such as smart highways and high-bandwidth networks) may reap few of the benefits of autonomous vehicles in the near term, potentially widening the gap between rich and poor countries~\cite{sanaullah2017autonomous,zali2022autonomous}. In wealthier cities, early adopters may enjoy safe driverless taxis and high-tech transit, while rural or poorer areas lag behind. As mentioned in~\cite{gherardini2024impact}, access to self-driving cars could become a luxury good, reinforcing social stratification: “private cars… represent a ‘status symbol’ that differentiates classes of people on the base of the economic position,” and autonomous vehicles may similarly be accessible only to those who can afford them. Other factors (gender, age, and technology literacy) further skew who can benefit from generative autonomy~\cite{chen2024distributional}. Hence, without policies to ensure equitable access (\textit{e.g.}, subsidized transit fleets or shared community autonomous vehicle programs), the promised mobility gains could mainly accrue to privileged groups, exacerbating inequality.

On the employment front, autonomous vehicles threaten to displace large segments of the workforce in transportation. Millions drive for a living worldwide: in the United States alone, driving occupations (truck drivers, delivery drivers, taxi/Uber drivers, bus drivers, etc.) account for a substantial share of jobs. A major report estimated that a rapid shift to fully autonomous vehicles could eliminate over 4 million driving jobs in the U.S.~\cite{beede2017employment}, hitting those with less education hardest. Globally, whole industries are preparing for the loss of conventional driving roles. In ride-hailing specifically, even Uber’s CEO has openly predicted that in 10–20 years, “autonomous vehicles… will take over the same routes humans drive today”~\cite{uber-ceo}. These disruptions could outpace the economy’s ability to retrain and absorb displaced workers, leading to social and political stress. Communities dependent on trucking or taxi employment would need significant support to transition, for example, by training workers for new roles in EV maintenance, fleet management, or remote vehicle supervision~\cite{groshen2019preparing}. The rise of GenAI for autonomy thus raises fundamental questions about the future of work. Without proactive measures (such as education programs or phased adoption timelines), the technology could deliver productivity gains unevenly, enriching the tech investors while leaving many former drivers behind.

\begin{tcolorbox}[colback=gray!5!white, colframe=black!50!gray]
    \begin{itemize}[leftmargin=0em]
        \item \fontsize{9}{10}\selectfont
              \textbf{Opportunities:}
              \textit{Significant gaps remain in fostering public trust, ensuring equitable access, and addressing socioeconomic disruptions. Future research should prioritize enhancing transparency and public education to build widespread societal acceptance, while simultaneously developing robust regulatory and liability frameworks to govern AI-driven transportation. Additionally, proactive strategies are required to mitigate economic inequality resulting from workforce displacement, emphasizing retraining and creating new employment opportunities in emerging AV-related sectors.}
    \end{itemize}
\end{tcolorbox}

\clearpage

\section{Concluding Remarks}
\label{sec:conclusion}

Generative foundation models are at the forefront of redefining autonomous driving, heralding an era where data-driven innovations substantially enhance vehicle autonomy across perception, planning, simulation, and evaluation. This survey has systematically synthesized the transformative potential of generative architectures, including VAEs, GANs, Diffusion Models, NeRFs, 3DGS, and multimodal LLMs, highlighting their roles in synthesizing realistic sensor data, predicting intricate traffic scenarios, and enabling human-aligned decision-making.

Beyond simply improving autonomous driving capabilities, generative models fundamentally alter the way autonomous systems are conceptualized, developed, and validated. They empower engineers and researchers to address the critical challenge of the ``long tail'' of rare, hazardous, and unpredictable scenarios through high-fidelity synthetic data generation and sophisticated digital twins. This transformation extends beyond technical realms, fostering a future where autonomous vehicles actively enhance urban mobility, promote environmental sustainability, and substantially reduce traffic-related fatalities, thereby profoundly benefiting humanity and society.

However, the path towards fully integrated, reliable, and ethically sound autonomous mobility still presents significant hurdles. Key challenges include the computational efficiency required for real-time deployment, ensuring robust generalization in diverse and dynamic conditions, and developing standardized benchmarks for rigorous, safety-critical evaluation. In addition to technical challenges, the successful societal adoption of generative autonomous systems hinges on establishing transparent, accountable frameworks that inspire and uphold public confidence. The rise of collaborative human–AI models, such as the humans-as-sensors paradigm and mixed-expert systems, signals a shift toward more interpretable, context-aware, and adaptive autonomous technologies. These emerging strategies emphasize the importance of human-centered AI by incorporating human insight and real-world situational feedback directly into system operations, thereby strengthening safety, responsiveness, and public receptivity. Similar to developments in Geospatial Artificial Intelligence, where ethical imperatives such as privacy protection and mitigation of algorithmic bias are central, the advancement of autonomous mobility must be grounded in responsible innovation that aligns with broader societal principles and values~\cite{ye2025human}.

Looking ahead, generative AI in autonomous driving is to advance through the convergence of sophisticated language understanding, environmental perception, and physics-grounded reasoning, enabled by multimodal foundation models. These integrated paradigms, guided by human-centered design and rigorous ethical standards, hold the potential to deliver unprecedented levels of autonomy that align with societal values and enhance human welfare ~\cite{ye2023toward}.
Yet, the broader integration of generative AI into complex decision-making systems, such as those governing autonomous vehicles, raises pressing ethical and governance challenges. As AI increasingly influences high-stakes, real-time decisions, concerns about algorithmic bias, lack of transparency, accountability gaps, and privacy risks become more acute. Without careful design and oversight, these technologies risk reinforcing systemic inequalities, obscuring the logic behind critical decisions, and diminishing public trust. These issues are not unique to mobility but reflect broader patterns in AI-driven decision-making across domains~\cite{sanchez2025ethical}.

To navigate this landscape responsibly, future systems must prioritize transparency in algorithmic processes, incorporate diverse and representative data, and support mechanisms for continuous human oversight. Public engagement, ethical auditing, and inclusive governance structures are essential to ensure that AI systems reflect collective values and serve all communities equitably.

In the long term, the rise of generative AI compels a fundamental rethinking of how we design and govern intelligent systems. It challenges conventional notions of autonomy and demands new frameworks that balance innovation with ethical responsibility. As generative models become embedded in everyday decision-making, from transportation to healthcare to finance, societies must foster adaptive policies, interdisciplinary research, and public education initiatives to build resilience and trust. Ultimately, the successful integration of generative AI into autonomous driving, and beyond, will depend on our collective ability to align technological progress with the principles of fairness, accountability, and human dignity.

\bibliographystyle{unsrt}
\clearpage
\bibliography{bibliography}

\end{document}